\documentclass[letterpaper]{article} %

\usepackage{aaai2026}
\usepackage{times}  %
\usepackage{helvet}  %
\usepackage{courier}  %
\usepackage[hyphens]{url}  %
\usepackage{graphicx} %
\usepackage{natbib}  %
\usepackage{caption} %
\usepackage{algorithm}
\usepackage{algorithmic}

\usepackage{newfloat}
\usepackage{listings}
\DeclareCaptionStyle{ruled}{labelfont=normalfont,labelsep=colon,strut=off} %
\floatstyle{ruled}
\newfloat{listing}{tb}{lst}{}
\floatname{listing}{Listing}
\title{Quantifying and Improving Adaptivity in Conformal Prediction through Input Transformations}
\author{
    Sooyong Jang\textsuperscript{\rm 1},
    Insup Lee\textsuperscript{\rm 1}
}
\affiliations{
    \textsuperscript{\rm 1}
PRECISE Center, University of Pennsylvania, USA\\
    \{sooyong, lee\}@seas.upenn.edu
}

\usepackage{booktabs}       %
\usepackage{amsfonts}       %
\usepackage{nicefrac}       %
\usepackage{microtype}      %
\usepackage{xcolor}         %

\usepackage{amsthm}
\usepackage{amsmath}
\usepackage{multirow,multicol}
\usepackage{caption}
\usepackage{mathtools}
\usepackage{comment}

\usepackage{subcaption}
\usepackage{placeins}

\newtheoremstyle{dotafter}%
  {}%
  {}%
  {\itshape}%
  {}%
  {\bfseries}%
  {.}%
  { }%
  {}%

\theoremstyle{dotafter}
\newtheorem{property}{Property}

\def\Prob{\mathbb{P}}

\def\As{\mathcal{A}}
\def\Is{\mathcal{I}}

\def\Ns{\mathcal{N}}
\def\Ps{\mathcal{P}}

\begin{document}

\maketitle

\begin{abstract}
Conformal prediction constructs a set of labels instead of a single point prediction, while providing a probabilistic coverage guarantee.
Beyond the coverage guarantee, adaptiveness to example difficulty is an important property.
It means that the method should produce larger prediction sets for more difficult examples, and smaller ones for easier examples.
Existing evaluation methods for adaptiveness typically analyze coverage rate violation or average set size across bins of examples grouped by difficulty.
However, these approaches often suffer from imbalanced binning, which can lead to inaccurate estimates of coverage or set size.
To address this issue, we propose a binning method that leverages input transformations to sort examples by difficulty, followed by uniform-mass binning.
Building on this binning, we introduce two metrics to better evaluate adaptiveness. 
These metrics provide more reliable estimates of coverage rate violation and average set size due to balanced binning, leading to more accurate adaptivity assessment.
Through experiments, we demonstrate that our proposed metric correlates more strongly with the desired adaptiveness property compared to existing ones.
Furthermore, motivated by our findings, we propose a new adaptive prediction set algorithm that groups examples by estimated difficulty and applies group-conditional conformal prediction. 
This allows us to determine appropriate thresholds for each group.
Experimental results on both (a) an Image Classification (ImageNet) (b) a medical task (visual acuity prediction)
show that our method outperforms existing approaches according to the new metrics.
\end{abstract}

\begin{links}
    \link{Code}{https://github.com/sooyongj/t-aps}
    \link{Extended version}{https://arxiv.org/pdf/2511.11472}
\end{links}

\section{Introduction}
\label{sec:introduction}

Researchers have used conformal prediction to ensure the reliability of machine learning predictions.
Unlike traditional models that produce a single point prediction, conformal prediction outputs a set of labels.
This set ensures that it contains the true label with high probability.
Beyond this coverage guarantee, adaptivity to example difficulty is a crucial property \citep{seedat2023improving}.
It enables the method to produce smaller prediction sets for easier instances while allocating larger sets to more challenging ones.

A key challenge in developing adaptive prediction sets is the precise evaluation of adaptiveness as existing standard metrics have limitations.
A common approach involves comparing the coverage rate violation or the average set size across subgroups of examples, where the subgroups are constructed based on example difficulty.
Typically, difficulty is inferred either from the prediction set size or from the rank of the ground truth label within the softmax output.
In particular, previous works have measured the worst-case coverage rate violation across bins based on set size (\emph{e.g.,} SSCV \citep{angelopoulos2021uncertainty}, ESCV \citep{huang2024conformal}), or have compared changes in coverage rate violation \citep{angelopoulos2021uncertainty} or relative average set size changes \citep{huang2024conformal} across bins based on the ground truth rank.

However, these methods have a notable weakness: the estimation of the coverage rate violation or average set size within each bin can be unreliable due to the imbalance in the number of examples across bins.
If a bin contains few examples, the estimation becomes highly biased.
Conversely, if a bin contains the majority of examples, the coverage rate violation will be close to zero, since a prediction set algorithm is designed to closely match the target coverage rate.

A straightforward way to mitigate the issue of uneven number of examples in each bin is to apply the uniform-mass binning \citep{kumar2019verified, gupta2020distribution}, where each bin contains approximately the same number of examples.
However, it is challenging to apply uniform-mass binning with the ground truth rank (one well-known method of estimating example difficulty \citep{angelopoulos2021uncertainty, huang2024conformal}), due to the long-tailed nature of its distribution.
Most examples have a high rank (corresponding to a high softmax value for the ground truth), while only a few examples have a low rank.
For example, a classifier with 80~\% accuracy assigns rank one to the correct label in 80~\% of cases.
This skew makes uniform-mass binning difficult to apply, as several bins may end up containing only examples with rank one, resulting in homogeneous groups that do not meaningfully reflect varying difficulty.

Therefore, we propose an alternative sorting method based on a different difficulty estimation.
Specifically, we estimate the difficulty $D(x) \in [0,1]$ (or equivalently, its ease as $E(x) = 1 - D(x)$) by measuring the variability in the model's predictions after applying small perturbations to the input.
For example, if we add a small noise $\delta$ to an input $x$, an easy example would result in similar predictions ($f(x) \approx f(x+\delta)$).
We empirically demonstrate that this difficulty-based sorting helps form bins that mitigate the aforementioned issue.

Building on this idea, we propose the two improved metrics
to better evaluate the adaptiveness according to the difficulty.
Our metrics first construct bins by leveraging the difficulty estimated through input transformations.
Then, they compute (1) the worst-case coverage rate violation similar to SSCV and ESCV and (2) the relationship between average set size and the average difficulty (ground truth label ranks) across the bins.
The first metric assesses if a prediction set algorithm maintains coverage guarantees across difficulty level.
The second metric evaluates whether the algorithm constructs prediction set with appropriate size according to the difficulty.
While the first metric is straightforward, as it analyzes the coverage rate violation across bins (consistent with prior work (SSCV and ESCV)), the second metric requires additional analysis to determine its validity as an adaptivity metric.
We define the desired adaptiveness property and show that our proposed metric achieves the highest statistical correlation with the property satisfaction rate on the ImageNet \citep{russakovsky2015imagenet}.

Since this experiment shows that we can estimate difficulty using input transformations, we further propose a new adaptive prediction set algorithm by extending this idea.
The key extension is to group examples based on estimated difficulty and to separately apply the prediction set algorithm within each group.
This separate treatment allows us to learn different thresholds for each group, leading to more appropriate set sizes according to difficulty.
To ensure coverage guarantee, we build on ideas from group-conditional conformal prediction \citep{vovk2003mondrian, jung2023batch,gibbs2025conformal,jin2025confidence}, where coverage is enforced not only marginally but also within each predefined subgroup.

We evaluate our proposed prediction set algorithm on 
two tasks: an image classification task and a visual acuity prediction task.
Our experiments demonstrate that the proposed method outperforms existing baselines.
Specifically, it achieves lower worst coverage rate violations and establishes a closer relationship between average set size and average example difficulty.

Our contributions are as follows:
\begin{itemize}
    \item We propose two metrics for evaluating adaptiveness, which provide better assessment compared to existing metrics (Section \ref{sec:metric}).
    \item We introduce an adaptive conformal prediction set algorithm that leverages transformation-based binning and group-conditional conformal prediction (Section \ref{sec:algorithm}).
    \item We empirically demonstrate 
    that our proposed method outperforms other baselines across nine base classifiers on ImageNet (Section \ref{sec:image}) 
    and four base classifiers on visual acuity prediction (Section \ref{sec:va}).
\end{itemize}
\section{Related Work}
\label{sec:related_work}

\paragraph{Adaptiveness Evaluation Metric.}
Several metrics exist to evaluate adaptivity.
SSCV \citep{angelopoulos2021uncertainty} and ESCV \citep{huang2024conformal} assess coverage rate violations across different bins based on prediction set sizes, measuring whether the algorithms maintain the target coverage rate across varying levels of difficulty.
Metrics ``Deficit'' and ``Excess''~\citep{navratil2020uncertainty, seedat2023improving}) quantify how tightly the prediction sets are aligned with the ground truth labels.
Additionally, previous studies employ empirical analyses, such as analyzing coverage rate violations and average set sizes stratified by ground truth rank \citep{angelopoulos2021uncertainty, huang2024conformal}.
Section~\ref{sec:metric} provides a more detailed analysis of these metrics.

\paragraph{Adaptive prediction set.}
Multiple prior works use softmax scores 
to construct adaptive prediction sets, allowing set sizes to vary according to example difficulty \citep{sadinle2019least, romano2020classification, angelopoulos2021uncertainty, huang2024conformal}.
They incorporate
softmax outputs into their non-conformity score functions, leveraging either individual softmax scores \citep{sadinle2019least} or the full softmax score distribution over labels \citep{romano2020classification, angelopoulos2021uncertainty, huang2024conformal} to assess difficulty.

\paragraph{Transformation and uncertainty quantification.}
Earlier studies show that transformations help quantify uncertainty.
\citet{ayhan2018test} compute the interquartile range (IQR) of the top softmax scores after applying input transformations, using it as a measure of uncertainty to identify challenging cases in a simulated clinical workflow.
This idea further extends to confidence estimation (calibration) \citep{bahat2020classification,jang2021improving, rizve2022towards}.
\citet{bahat2020classification} use the mean of top softmax scores for perturbed images as confidence, while 
\citet{jang2021improving} leverage lossy label-invariant transformations to group examples for confidence calibration.
Similarly, \citet{rizve2022towards} quantify uncertainty by measuring the variance of the top softmax scores across perturbed images, using this value for temperature scaling.
In addition, \citet{tao2024consistency} use prediction consistency for uncertainty estimation.
Although \citet{guo2025sample} do not use transformations, they estimate uncertainty based on the logit gap.

\paragraph{Group-conditional conformal prediction.}
Beyond marginal coverage guarantee, a growing body of research focuses on group-conditional coverage guarantee.
This setting aims to ensure that prediction sets achieve the target coverage rate within predefined subgroups of examples.
For example, in disease prediction, the model should maintain consistent coverage across different demographic groups, such as race.
Various 
researchers propose algorithms for
group-conditional conformal prediction \citep{vovk2003mondrian,jung2023batch,gibbs2025conformal,jin2025confidence}, with different properties.
These differences include whether the groups can overlap and whether the grouping is based soley on features or other factors.
\section{Transformation based metric}
\label{sec:metric}

This section describes metrics for evaluating adaptivity. 
\citet{navratil2020uncertainty, seedat2023improving} introduce ``Deficit" and ``Excess" \citep{navratil2020uncertainty, seedat2023improving} for regression problems.
Specifically, Deficit quantifies the number of classes required to include the ground truth, whereas Excess assess the number of classes that can be removed while maintaining coverage, both following softmax score order.
SSCV \citep{angelopoulos2021uncertainty}, a widely used metrics, quantifies worst-case deviation by comparing the target coverage rate with the empirical coverage rate across different size strata.
Similarly, ESCV \citep{huang2024conformal} relates closely to SSCV but considers individual prediction set sizes rather than size ranges.
Additionally, two empirical analyses (Rank-CV, Rank-SS) utilize ground truth rank-based binning.
The ground truth rank refers to the position of the true label within the sorted softmax scores.
Using this binning, Rank-CV compares coverage violation across bins to assess consistency, while Rank-SS examines whether average set sizes increase with rising average ground truth rank.

However, these metrics have certain limitations.
Deficit and Excess are biased toward one side; for example, prediction sets containing all labels achieve zero Deficit.
Both SSCV and ESCV use set size-based binning, which can yield inaccurate metric when bins have few examples.
Rank-CV and Rank-SS share SSCV and ESCV's limitation due to their similar binning strategy.

We focus on a common limitation of SSCV, ESCV, and the two empirical analysis methods: all rely on binning examples.
Such methods may yield inaccurate estimates when bins are imbalanced.
To address this, we propose using a uniform-mass binning, which ensures each bin contains approximately the same number of examples.
However, applying uniform-mass binning based on ground truth label rank is challenging, due to its long-tailed distributions, \emph{i.e.,} most examples have a rank of one.
Thus, a more suitable difficulty sorting method is required to ensure that the average difficulty across bins varies smoothly.

\begin{figure}[!bt]
    \centering
    \includegraphics[width=0.81\linewidth]{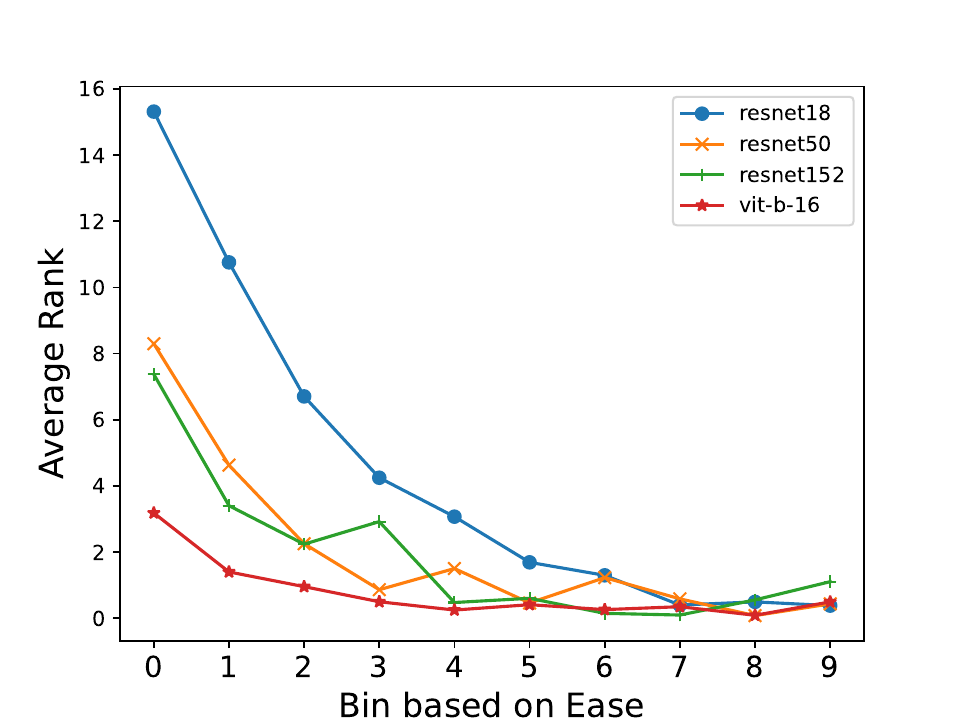}
    \captionof{figure}{Ease vs. Average Rank.}
    \label{fig:ease_vs_avgrank}
\end{figure}

We leverage a method to estimate example difficulty by utilizing input transformations.
At a high level, if the output of a prediction model $f$ remains unchanged after applying an input transformation, such as adding small noise to an example $x$, the example $x$ can be considered easy for the model.
Accordingly, we define the ease of example $x$ with respect to the model $f$, denoted as $E: \mathcal{X} \to [0,1]$, formally:

\begin{equation}
\label{eqn:similarity}
E(x;f) = \frac{1}{L} \sum_{l=1}^L \text{sim}(f(x), f(t_l(x)),
\end{equation} 
where $\text{sim}(x,y)$ denotes a similarity between $x$ and $y$, such as the cosine similarity, and $t_l(x)$ represents a transformation (\emph{e.g.,} $t_l(x) = x + \delta_l$, where $\delta_l \sim  \Ns(0, \sigma^2)).$

Regarding the input transformations, we use Gaussian noise for simplicity; other transformations (\textit{e.g.,} color jitters) work if severity is controlled --- excessive change mostly alters predictions. 
Adversarial perturbations are less suitable, but may work with robust models.

To verify the effectiveness of this sorting's method, we plot the average ground truth rank within each bin on ImageNet dataset using the four different base classifiers.
As shown in Figure \ref{fig:ease_vs_avgrank}, for all base classifiers, the average rank generally decreases as the ease increases, confirming the validity of our sorting strategy for rank-based binning.

\begin{figure}[!bt]
    \centering
    \includegraphics[width=\linewidth]{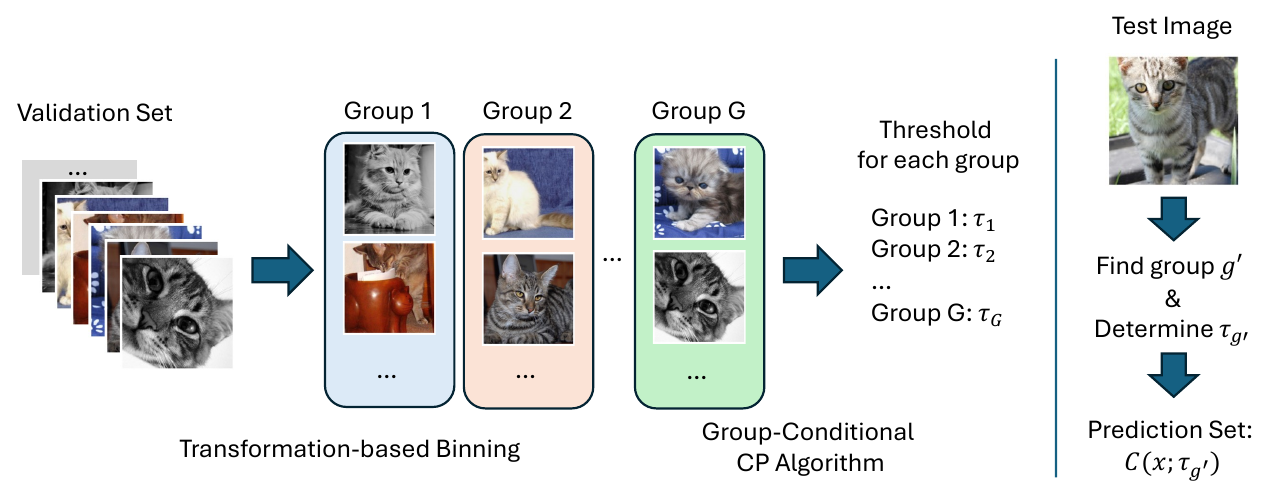}
    \caption{Algorithm Process. The left part illustrates the calibration process, while the right part shows the construction of prediction sets for test images.}
    \label{fig:algorithm Process}
\end{figure}

We now describe how examples are binned using $E(x;f)$.
We partition the range of ease scores $[0, 1]$ into $B$ bins using uniform-mass binning.
Formally, let $a:\{1,..., N\} \to \{1, ..., N\}$ be an indexing function that sorts examples such that $E(x_{a(1)}) \le E(x_{a(2)}) \le \cdots \le E(x_{a(N)})$.
Then, the set of indices for bin $b$ is defined as:
\begin{multline}
\label{eqn:binning}
\Is_b = \left\{a(i) \middle| \left\lfloor \frac{(b-1)N}{B} \right\rfloor < i \le \left\lfloor \frac{bN}{B} \right\rfloor \right\} \\ \text{for } b=1,2,\dots, B.
\end{multline}
This process is summarized in Algorithm \ref{alg:binning} in Appendix.

Based on this transformation-based binning, we define two evaluation metrics. 
(1) Transformation-Based binning Coverage Violation (T-CV): the worst-case coverage rate violation across bins, and (2) Transformation-Based binning Set Size relationship (T-SS): the relationship between the average set size and the difficulty across bins.
The two metrics are formally defined as:
\begin{equation}
\label{eq:first_metric}
    \textbf{T-CV} =  \max_{b\in [B]} \left|\frac{\sum_{i\in\Is_b}\mathbb{I}(y_i \in C(x_i))}{|\Is_b|} - (1-\alpha)\right|
\end{equation}
\begin{equation}
\label{eq:second_metric}
\textbf{T-SS} = R^2(\{r(\Is_1), \cdots, r(\Is_B)\}, \{s(\Is_1), \cdots, s(\Is_B) \}),
\end{equation}
where $R^2$ is the (signed) R-squared value (Coefficient of Determination) and $r(\Is_b), s(\Is_b)$ are the average rank and average set size for the group of examples $\Is_b$, respectively.
Here, $R^2(X,Y)$ = $\text{sign}(a) \times \text{max}\left(0, 1 - \frac{\sum(y-\hat{y})^2}{\sum(y-\bar{y})^2}\right)$, where $a$ is the coefficient for a linear regressor, $y, \hat{y}, \bar{y}$ are the true value, predicted value, and mean true value, respectively.
Thus, T-SS measures how closely the average rank and average set size of each bin are linearly related, with the sign indicating the direction of their relationship (positive or negative correlation).

\begin{table}
\centering
\scriptsize
\begin{tabular}{c cccc c}
\toprule
Rank & Defict & Excess & SSCV & ESCV & T-SS\\
\midrule
1 & 12 & 0 & 0 & 0 & 24\\
2 & 23 & 0 & 1 & 0 & 12\\
3 & 0 & 9 & 16 & 11 & 0\\
4 & 1 & 12 & 11 & 12 & 0\\
5 & 0 & 15 & 8 & 13 & 0\\
\midrule
Avg. & 1.72 & 4.17 & 3.58 & 4.07 & \textbf{1.33}\\
\bottomrule
\end{tabular}
\caption{Metric comparison result. We rank each metric across thirty six cases, based on nine models and four algorithms, with its frequency and average rank reported.}
\label{tab:metric_comparison}
\end{table}

While T-CV directly evaluates coverage rate violation, similar to prior metrics (SSCV and ESCV), T-SS focuses on the different aspect, average set size.
Therefore, the effectiveness of T-SS needs to be further validated as an adaptivity metric.
To validate T-SS, we define a desired property (Property \ref{pro:adap_metric}) that adaptive prediction sets should satisfy and analyze the correlation between each metric and the degree of this property being met.
The next subsection describes the experimental setup and presents the results.

\begin{property}
\label{pro:adap_metric}
(Difficulty and Set Size Relationship). For any two subsets $g_1, g_2$ of size $m$, let $d_1, d_2$ be their average difficulty and $s_1,s_2$ be their average prediction set size. 
Then, a desirable adaptivity property requires that one of the following holds:
\begin{align*}
d_1 \le d_2 \implies s_1 \le s_2 \quad  \text{ or } \quad d_1 > d_2 \implies s_1 > s_2.
\end{align*}
In words, groups of harder examples should correspond to larger average prediction set size.
\end{property}

\paragraph{Experiment.}
We conduct an experiment to compare T-SS (Equation \ref{eq:second_metric}) with existing metrics,
focusing on their ability to estimate the satisfaction rate of the desired property.
We perform $R=1000$ trials, and in each trial, we sample $M=2000$ examples from ImageNet validation set.
For each $M$ examples, we estimate the property satisfaction rate by sampling $m=100$ examples $T=10000$ times and checking whether the property holds.
We also compute each metric on the same $M$ examples.
Finally, we calculate the Spearman correlation \citep{ca468a70-0be4-389a-b0b9-5dd1ff52b33f} between the property satisfaction rates and the corresponding metric values across the $R$ trials. 

We use nine base classifiers (ResNet18, ResNet50, ResNet152 \citep{he2016deep}, ViT-B-16, ViT-L-16, ViT-H-14 \citep{dosovitskiy2020image}, EfficientNet-V2-M, EfficientNet-V2-L \citep{tan2021efficientnetv2}, Swin-V2-B \citep{liu2022swin}) and four prediction set algorithms (LAC \citep{sadinle2019least}, APS \citep{romano2020classification}, RAPS \citep{angelopoulos2021uncertainty}, SAPS \citep{huang2024conformal}) on ImageNet.

We rank the five metrics based on the Spearman correlation coefficients and summarize the result in Table \ref{tab:metric_comparison}. 
We provide the full results, including p-values in Table \ref{tab:apdx_full_metric_result} in the Appendix.
As shown in Table \ref{tab:metric_comparison}, T-SS outperforms existing metrics, achieving the best rank in most cases.
Moreover, the correlation coefficients for T-SS are statistically significant based on the corresponding p-values. 
This implies that T-SS is strongly correlated with the property satisfaction rate, indicating its effectiveness in evaluating adaptivity.

\section{Prediction Set Algorithm}
\label{sec:algorithm}
In this section, we describe our algorithm, which is based on the transformation-based binning introduced in Section \ref{sec:metric}.

\begin{table*}[!bt]
\scriptsize
\centering
\begin{tabular}{c c cccccc| cccccc}
\toprule
    \multirow{2}{*}{Base Model} & \multirow{2}{*}{Acc.} & \multicolumn{6}{c}{T-CV} & \multicolumn{6}{c}{T-SS}\\   
    & & LAC & APS & RAPS & SAPS & O-LAC & O-SAPS & LAC & APS & RAPS & SAPS & O-LAC & O-SAPS \\
\midrule
\multicolumn{14}{c}{$\alpha=0.10$}\\
\midrule
ResNet18 & 69.89 & 0.192 & \textbf{0.040} & 0.055 & 0.255 & 0.060 & \underline{0.053} & 0.869 & \underline{0.924} & 0.921 & 0.827 & \textbf{0.926} & 0.912\\
ResNet50 & 80.81 & 0.323 & 0.148 & 0.212 & 0.290 & \underline{0.068} & \textbf{0.057} & 0.512 & 0.210 & 0.617 & 0.620 & \textbf{0.835} & \underline{0.807}\\
ResNet152 & 82.26 & 0.243 & 0.060 & 0.273 & 0.260 & \underline{0.052} & \textbf{0.047} & 0.674 & 0.756 & 0.614 & 0.728 & \textbf{0.783} & \underline{0.769}\\
ViT-B-16 & 85.22 & 0.297 & 0.065 & \textbf{0.045} & 0.055 & \underline{0.050} & \textbf{0.045} & 0.636 & \textbf{0.814} & \underline{0.810} & 0.743 & 0.797 & 0.755\\
ViT-L-16 & 88.20 & 0.463 & 0.075 & 0.180 & \textbf{0.040} & \underline{0.042} & \textbf{0.040} & 0.296 & \textbf{0.823} & 0.558 & 0.760 & \underline{0.793} & 0.776\\
ViT-H-14 & 88.53 & 0.430 & \underline{0.043} & 0.160 & 0.050 & \underline{0.043} & \textbf{0.040} & 0.287 & 0.611 & 0.495 & 0.619 & \textbf{0.627} & \underline{0.623}\\
Efficientnet-V2-M & 85.11 & 0.258 & 0.050 & 0.230 & 0.253 & \underline{0.045} & \textbf{0.040} & 0.517 & 0.408 & 0.368 & 0.554 & \textbf{0.624} & \underline{0.595}\\
Efficientnet-V2-L & 85.68 & 0.290 & \underline{0.050} & 0.190 & 0.057 & \underline{0.050} & \textbf{0.045} & 0.473 & 0.507 & 0.479 & 0.538 & \textbf{0.557} & \underline{0.553}\\
Swin-V2-B & 83.93 & 0.243 & \underline{0.050} & \underline{0.050} & 0.075 & \textbf{0.040} & \textbf{0.040} & 0.627 & \textbf{0.703} & 0.686 & 0.670 & \underline{0.694} & \underline{0.694}\\
    \midrule
    Average & - & 0.304 & 0.065 & 0.155 & 0.148 & \underline{0.050} & \textbf{0.045} & 0.543 & 0.640 & 0.616 & 0.673 & \textbf{0.737} & \underline{0.720}\\
    Avg. Rank from CD & - & 5.44 & 2.44 & 3.89 & 4.11 & 1.33 & \textbf{1.11} & 5.33 & 2.67 & 3.56 & 3.56 & \textbf{1.11} & \underline{1.67}\\
\midrule
\multicolumn{14}{c}{$\alpha=0.05$}\\
\midrule
ResNet18 & 69.89 & 0.112 & \textbf{0.033} & 0.045 & 0.165 & \textbf{0.033} & \underline{0.035} & 0.912 & \underline{0.926} & 0.920 & 0.828 & \textbf{0.930} & 0.910\\
ResNet50 & 80.81 & 0.153 & 0.110 & 0.165 & 0.100 & \textbf{0.035} & \underline{0.040} & 0.680 & 0.095 & 0.474 & 0.636 & \textbf{0.828} & \underline{0.807}\\
ResNet152 & 82.26 & 0.110 & \underline{0.033} & 0.163 & 0.107 & \textbf{0.030} & \underline{0.033} & 0.763 & 0.689 & 0.679 & 0.727 & \textbf{0.791} & \underline{0.769}\\
ViT-B-16 & 85.22 & 0.107 & 0.043 & \underline{0.033} & 0.053 & \textbf{0.030} & \underline{0.033} & 0.700 & \textbf{0.802} & \underline{0.770} & 0.725 & \textbf{0.802} & 0.745\\
ViT-L-16 & 88.20 & 0.143 & \underline{0.040} & 0.090 & 0.058 & \textbf{0.030} & \textbf{0.030} & 0.690 & \underline{0.803} & 0.711 & 0.743 & \textbf{0.813} & 0.776\\
ViT-H-14 & 88.53 & 0.115 & \underline{0.030} & \textbf{0.028} & 0.048 & \underline{0.030} & \textbf{0.028} & 0.593 & 0.542 & 0.585 & 0.613 & \textbf{0.621} & \underline{0.619}\\
Efficientnet-V2-M & 85.11 & 0.110 & \textbf{0.030} & 0.145 & 0.112 & \underline{0.030} & \textbf{0.028} & 0.558 & 0.402 & 0.504 & 0.554 & \textbf{0.627} & \underline{0.598}\\
Efficientnet-V2-L & 85.68 & 0.110 & \underline{0.035} & 0.115 & 0.080 & \textbf{0.033} & \textbf{0.033} & 0.536 & 0.478 & 0.491 & 0.536 & \textbf{0.570} & \underline{0.552}\\
Swin-V2-B & 83.93 & 0.107 & 0.035 & 0.050 & 0.055 & \textbf{0.028} & \underline{0.030} & 0.666 & 0.690 & 0.678 & 0.670 & \textbf{0.705} & \underline{0.695}\\
\midrule
Average & - & 0.119 & 0.043 & 0.093 & 0.086 & \textbf{0.031} & \underline{0.032} & 0.678 & 0.603 & 0.646 & 0.670 & \textbf{0.743} & \underline{0.719}\\
Avg. Rank from CD & - & 5.00 & \underline{1.78} & 4.11 & 4.33 & \textbf{1.00} & \textbf{1.00} & 3.78 & 3.67 & 4.22 & 3.67 & \textbf{1.00} & \underline{1.89}\\
\bottomrule
\end{tabular}
\caption{Our Metrics. The second-to-last row shows the average value over different models, and the last row shows the average rank based on Critical Difference (CD) analysis. $B=50$.}
\label{tab:alg_results_our_metrics}
\end{table*}

\subsection{Algorithm overview}
As illustrated in Section \ref{sec:metric}, transformation-based binning effectively groups examples by similar difficulty.
Building on this, we argue that applying conformal prediction separately within each group enables the algorithm to determine more appropriate thresholds, leading to improved adaptivity.
To ensure coverage guarantees in this group-wise setting, we adopt an idea from group-conditional conformal prediction.
 
The group conditional coverage guarantee was proposed to address the limitations of marginal coverage.
Instead of the marginal coverage guarantee,
the group-conditional coverage guarantee ensures:
\begin{equation}
    \label{eqn:group_coverage}
    \Prob_{(x_i,y_i) \sim \Ps}[y_i \in C(x_i) | G(x_i,y_i) = g] \ge 1-\alpha, \quad \forall g \in [H],
\end{equation}
where $\Ps$ is the data distribution, $C(x_i)$ is the prediction set,
$G(x_i,y_i)$
is the group assignment function, $\alpha$ is the target miscoverage rate, and $H$ is the number of groups.

We define the group assignment function $G(x_i,y_i) \coloneq b \text{ such that } i \in \Is_b$, and apply a group-conditional conformal prediction algorithm based on this assignment.

\subsection{Algorithm detail}
The algorithm is described in Figure \ref{fig:algorithm Process} and Algorithm \ref{alg:algorithm} (Appendix).
At a high level, we first group examples based on their estimated difficulty using transformation-based binning (Left part in Figure \ref{fig:algorithm Process}) 
and then apply a group-conditional conformal prediction algorithm to obtain different conformal predictors for each group (Middle).
Given a test data point $x$, we identify its corresponding bin, and use the threshold assigned to that bin to construct the prediction set (Right).

Multiple group-conditional conformal algorithms \citep{vovk2003mondrian,jung2023batch,ding2023class,gibbs2025conformal,jin2025confidence} exist for $\As$.
These algorithms differ in aspects such as whether the group overlapping is allowed \citep{jung2023batch,gibbs2025conformal,jin2025confidence} and whether the group assignment depends solely based on the feature $x$ or on both the feature $x$ \citep{jung2023batch,gibbs2025conformal,jin2025confidence} and label $y$ \citep{vovk2003mondrian,ding2023class}.
However, in our setting, Mondrian Conformal Prediction (MCP) \citep{vovk2003mondrian} is sufficient.
The approach forms groups and applies conformal prediction individually within each group.
Regarding the randomness in the grouping function, separating the calibration sets for binning and for CP preserves group-conditional coverage, whereas using a single calibration set suffices for marginal coverage.

\section{Application to Image Classification}
\label{sec:image}

We conduct an experiment to compare our proposed algorithm with other adaptive prediction set algorithms.
\paragraph{Dataset and set-up.}
We use ImageNet \citep{russakovsky2015imagenet} in this section.
The original dataset consists of training, validation and test set, but we use only the validation set.

We repeat each experiment $R=100$ times and we report the median value of each metric as the final result.
In each repetition, we randomly sample a validation set $D_{val}$ of size $n_{val}=20,000$ and a test set $D_{test}$ of size $n_{test} = 20,000$ from the original validation set.
For algorithms requiring hyperparameter tuning (including ours), we evenly split $D_{val}$ into a calibration set $D_{cal}$ and a tuning set $D_{tune}$. 
Algorithms that do not require hyperparameter tuning use the entire $D_{val}$.
This setup ensures that all methods use the same amount of validation data.

\paragraph{Baselines and our algorithm.}
We compare our method against four adaptive prediction set algorithms: LAC \citep{sadinle2019least}, APS \citep{romano2020classification}, RAPS \citep{angelopoulos2021uncertainty}, SAPS \citep{huang2024conformal}.
We implement these baselines using TorchCP library \citep{huang2024torchcp}.
For our method, we use the non-conformity score functions of LAC and SAPS as the base conformal prediction algorithms, and denote the resulting method as O-LAC and O-SAPS, respectively.

Some algorithms, including ours, require tuning hyperparameters, such as the number of bins and the transformations types in our method, the regularization weight in RAPS, and the ranking weight in SAPS.
While RAPS uses SSCV and average set size for tuning, we instead use our proposed metric (T-SS) for hyperparameter selection to better align with our goal of improving adaptivity.

For all methods, we set target miscoverage rate $\alpha \in \{0.10, 0.05\}$. 

\paragraph{Metrics.}
We primarily use the two proposed metrics (T-CV and T-SS with the number of bins $B=50$) to compare the adaptivity of prediction set algorithms.
We also compute the empirical coverage rate and average set size to evaluate the coverage guarantee and the efficiency of the algorithms, and include them (Table \ref{tab:apdx_alg_results_cov_avgss}) in the Appendix.
Further results using existing evaluation metrics are presented in Tables \ref{tab:apdx_alg_results_deficit_excess} and \ref{tab:apdx_alg_results_sscv_escv} in the Appendix. 
We also report the performance of our metric with an alternative bin count ($B \in \{ 100, 30\}$), provided in the Appendix (Tables \ref{tab:apdx_alg_results_our_metrics_B100} and \ref{tab:apdx_alg_results_our_metrics_B30}).

\subsection{Results}

Table \ref{tab:apdx_alg_results_cov_avgss} reports the coverage rate and average set size.
Overall, all algorithms achieve empirical coverage rates close to the target coverage rate ($1-\alpha\in \{0.90, 0.95\}$).
In terms of average set size, LAC produces the smallest sets on average, while APS yields the largest sets, with the others showing similar sizes.
Specifically, binning in our algorithm reduces the samples per bin, making threshold estimation harder and slightly increasing size. However, ours shows comparable sizes to others except LAC (worse adaptivity one).

Table \ref{tab:alg_results_our_metrics} presents the result using our proposed metrics, T-CV and T-SS.
The `Average' row indicates the mean values of the respective metrics, while `Avg. Rank from CD' represents the average rank of each algorithm derived from the Critical Difference (CD) analysis.
For the CD analysis, we first perform Friedman Test, which indicates statistically significant differences among the algorithms (all p-values are significant, $< 0.001$).
We then construct CD diagrams, shown in the Appendix (Figures \ref{fig:apdx_cd_tcv_alpha0.05} - \ref{fig:apdx_cd_tss_alpha0.20}), 
and rank algorithms based on the CD results, assigning the same rank to algorithms that are not statistically different.
Finally, we compute the average rank of each algorithm
to compare their relative performance.

Based on the metric values
and the average ranks from the CD analysis,
we analyze the adaptiveness performance of individual algorithms in detail.
First, O-SAPS achieves the smallest T-CV value, with O-LAC achieving the second-best performance.
This is noteworthy as there is a general trend where the average set size correlates with T-CV: algorithms with smaller average set sizes (\emph{e.g.}, LAC) tend to have higher T-CV values, while those with larger set sizes (\emph{e.g.}, APS) have lower T-CV value.
Given that APS produces the largest average set size, it is expected that APS achieves a low T-CV value.
In this context, it is notable that our algorithms record the low T-CV values while maintaining not overly large average set sizes unlike APS.

Regarding T-SS, our two algorithms achieve the best (O-LAC) and the second best (O-SAPS), indicating that the prediction set sizes respond adaptively to example difficulty.
Although APS generally performs well in T-SS, it shows a notable exception on ResNet50, where its exceptionally large average set size leads to a weaker relationship between set size and difficulty.

In summary, our methods better adapt the prediction set size according to the difficulty (as indicated by T-SS), while maintaining the coverage rates close to the target level across different difficulty levels (T-CV).

\begin{figure*}[!bt]
    \centering
    \begin{subfigure}{0.38\linewidth}
        \includegraphics[width=\linewidth]{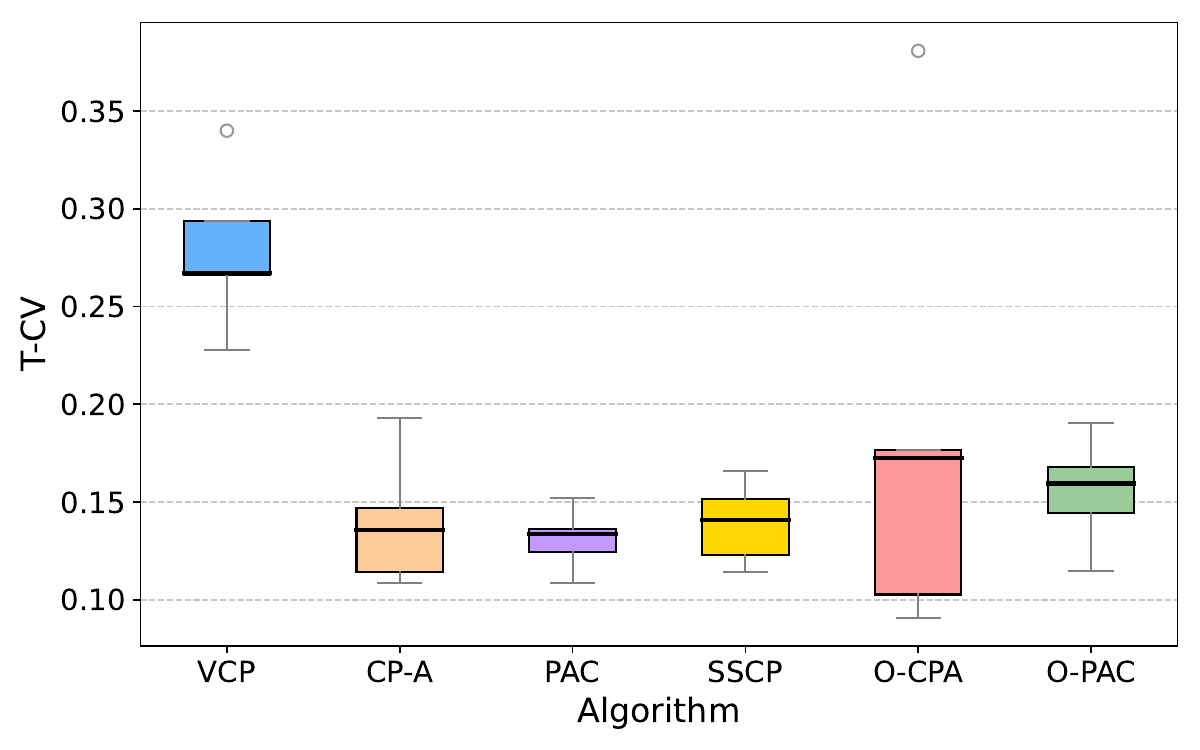}
        \caption{T-CV}
    \end{subfigure}%
    \begin{subfigure}{0.38\linewidth}
        \includegraphics[width=\linewidth]{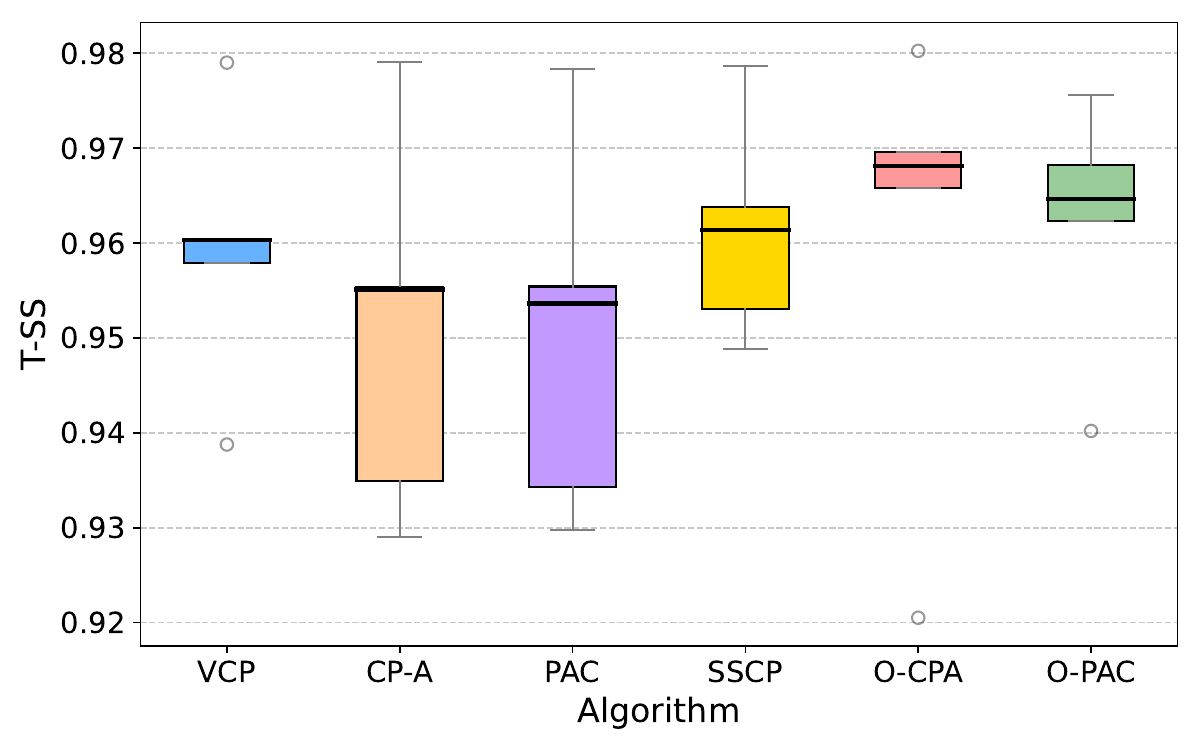}
        \caption{T-SS}
    \end{subfigure}
    \caption{Visual Acuity Prediction Result. EfficientNet-V2-S. $\alpha=0.30$}
    \label{fig:va_eff_alpha0.30}
\end{figure*}

\paragraph{Limitations.}
One limitation of our approach is that as the base classifier becomes more accurate, the T-SS score tends to decrease.
For example, O-LAC achieves a T-SS of 0.926 with ResNet18 and 0.627 with ViT-H-14, even though ViT-H-14 attains a significantly higher classification accuracy (69.89\% vs. 88.20\%).
We hypothesize that this phenomenon is due to the distribution of the ground truth rank.
T-SS measures the relationship between the average ground truth rank and the average set size across bins. 
A more accurate classifier tends to have more examples with the ground truth rank of one.
This leads to many bins having similar average ranks, which reduces variability of difficulty and limits maximum T-SS score.
In the extreme case of a base classifier achieving 100\% accuracy, all bins would have an average rank of one, making it impossible to compute T-SS meaningfully.
For a similar reason, our metrics become less effective when the target miscoverage rate $\alpha$ is high, \emph{e.g.,} $\alpha \in \{0.15, 0.20\}$. 
Please refer to the results provided in Appendix (Table \ref{tab:apdx_alg_results_our_metrics}).
A higher $\alpha$ allows more miscoverage cases, leading to smaller prediction sets overall, including for more difficult examples.
This reduces the contrast in set sizes across different difficulty levels, making it harder to accurately assess the relationship between example difficulty and prediction set size --- the core aspect that T-SS aims to measure.

Despite these limitations, T-SS remains useful for comparing different prediction set algorithms under a fixed base classifier and the reasonable target miscoverage rate considering the safety. 
Under these conditions, all methods operate under the same conditions, ensuring a fair and meaningful evaluation.
\begin{figure}[!bt]
    \centering
    \includegraphics[width=0.8\linewidth]{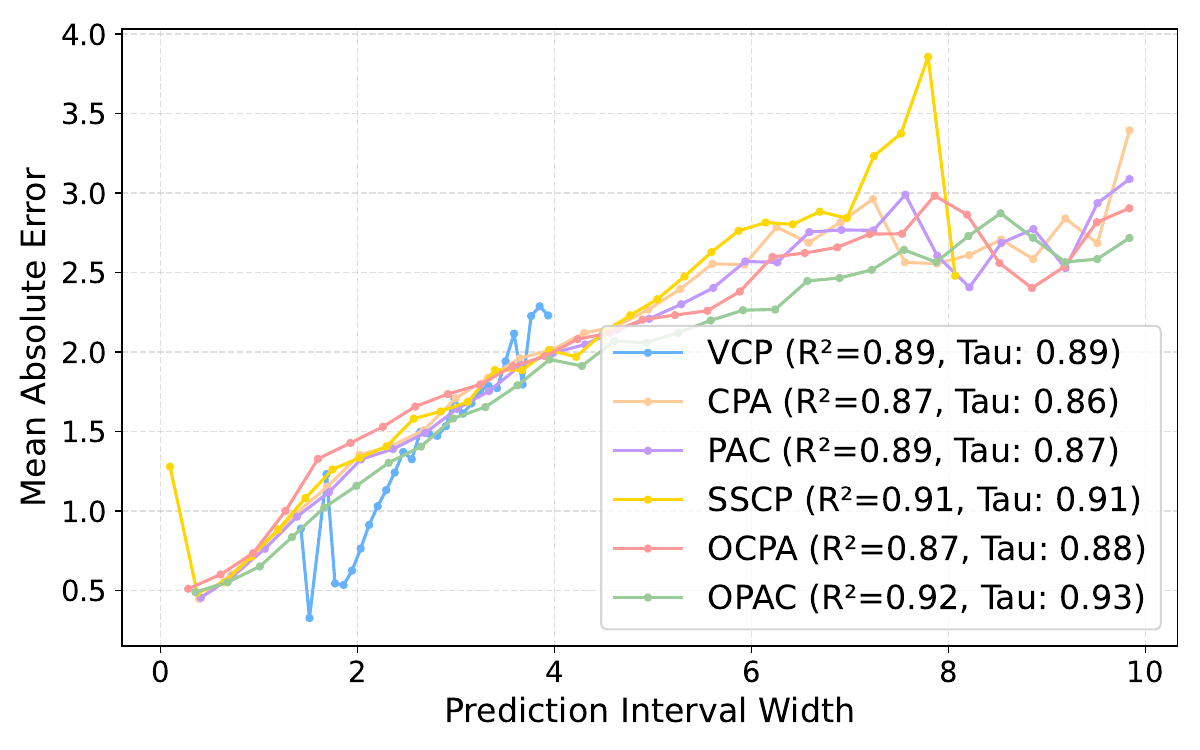}
    \caption{Prediction Interval Width and Prediction Error. EfficientNet-V2-S. $\alpha=0.30$.}
    \label{fig:va_rel}
\end{figure}
\section{Application to Visual Acuity Prediction}
\label{sec:va}

We apply our adaptive prediction set algorithm to a medical task as a practical application.
Specifically, we consider visual acuity (VA) prediction, a clinically relevant task studied in prior works \cite{kim2022deep,paul2023accuracy,pmlr-v259-jang25a}.
Given a retinal fundus image, $x_i$, the goal is to predict the visual acuity $y_i$.
The task can be formulated in various ways; here, we adopt the same formulation as in \cite{pmlr-v259-jang25a}.
In particular, the task is to train a model $f(x)$ for Gaussian distribution over visual acuity (\emph{i.e.,} a regression task), and then apply conformal prediction to obtain prediction intervals with a coverage guarantee.

We use the dataset from \cite{kim2022deep} and the trained models from \cite{pmlr-v259-jang25a}, and apply our algorithms to them.
Because our algorithm is originally designed for classification, we modify it to handle regression.
First, we replace cosine similarity with a new similarity metric defined as, $
\text{similarity}(y_1, y_2) = \exp{\left(-|y_1-y_2|\right)}.
$
Second, instead of using the prediction set size, we evaluate our method using the width of prediction interval.

\subsection{Experiment set-up}

\paragraph{Dataset.}
We use the same dataset as in \cite{kim2022deep, pmlr-v259-jang25a}. 
For a detailed description of the dataset, please refer to \cite{pmlr-v259-jang25a}.

\paragraph{Baselines.}
We implement four baselines:
\begin{itemize}
    \item \textbf{CP} and \textbf{CP-A} are the vanilla CP approaches using different non-conformity score functions: $s_{\text{CP}}(x,y) = |f_{\mu}(x) - y|$ for CP, and $s_{\text{CP-A}}(x,y) = \frac{|f_{\mu}(x) - y|}{f_{\sigma}(x)}$ for CP-A.
    In other words, CP-A is the extension of CP that incorporates the estimated standard deviation, allowing for prediction interval with varying width.
    \item \textbf{PAC} is the PAC prediction interval algorithm used in the paper \cite{pmlr-v259-jang25a}.
    \item \textbf{SSCP} \cite{seedat2023improving} leverages the loss from a self-supervised learning task loss to estimate uncertainty, which is then used to compute non-conformity score.
\end{itemize}

We also implement two versions of the adaptive prediction set algorithm, O-CPA and O-PAC which are based on CP-A and PAC, respectively.
We use the same model from \cite{pmlr-v259-jang25a}, and apply our two versions to each base model. 

\subsection{Results}
Following the set-up in \cite{pmlr-v259-jang25a}, we repeat each experiment with five different random seeds.
All experiments use a target miscoverage rate $\alpha \in \{0.20, 0.30, 0.40\}$ and for PAC-based methods, we use a significance level of $\delta=0.00001$.
Figure~\ref{fig:va_eff_alpha0.30} shows results with EfficientNet-V2-S at miscoverage rate $\alpha=0.30$. Full results appear in the Appendix (Tables \ref{tab:res_va_adaptivity}-\ref{tab:res_va_general}, Figures \ref{fig:va_adaptivity_alpha0.20}-\ref{fig:va_adaptivity_alpha0.40}).

As shown in Figure~\ref{fig:va_eff_alpha0.30}, our methods have similar T-CV values compared to other baselines except VCP, and outperform all baselines in terms of T-SS.
Our methods tend to generate conservative intervals when the number of examples is small due to the group-coverage algorithm.
This results in an over-coverage rate as shown in Table \ref{tab:res_va_general}, and a higher T-CV --- the over-coverage increases the T-CV value as well.
In terms of T-SS, ours achieves higher values, indicating a stronger linear relationships between average error and average interval width, which implies better adaptivity.

We further analyze how effectively each algorithm captures adaptivity by comparing the relationship between prediction interval width and prediction error.
The common assumption is that wider intervals are indicative of higher prediction errors, as they reflect greater uncertainty in the base model's predictions.
To quantify this relationship, we plot the average prediction interval widths against the mean absolute error.
We compute the $R^2$ value and Kendall's Tau \citep{kendall1938new} to measure the strength of the relationships between interval width and prediction error.
The results for EfficientNet-V2-S with desired miscoverage level of $\alpha=0.30$ are presented in Figure \ref{fig:va_rel}.
Our approach achieves higher $R^2$ and tau values, indicating a stronger positive correlation between interval width and prediction error.
This suggests that the prediction intervals constructed by our method more accurately reflect the model's uncertainty, enabling clinicians to utilize interval width as a meaningful measure of uncertainty in predictions.
We also note that our methods consistently outperform other algorithms across different base models and target miscoverage levels (Figure \ref{fig:va_rel_all_models} in Appendix).
Furthermore, for narrow intervals (which are clinically useful), our methods tend to maintain a positive correlation between interval width and prediction error.
\section{Conclusion}
\label{sec:conclusion}

We identify that existing metrics for evaluating the adaptiveness of conformal prediction sets face challenges primarily due to imbalanced binning.
To address this issue, we propose a transformation-based binning method and introduced two new metrics leveraging this approach.
We demonstrate that these metrics effectively evaluate the adaptivity; in particular, one of them is effective at measuring the extent to which the desired property for adaptiveness is satisfied.
Building on this foundation, we propose an adaptive prediction set algorithm that combines the transformation-based binning with group-conditional conformal prediction.
Through experiments on 
two tasks,
we show that our algorithm outperforms existing baselines in terms of the proposed metrics.

Our method builds on the theoretical coverage guarantees of conformal prediction (CP) and group-conditional CP. 
Extending the theory toward expected set size \citep{dhillon2024expected,huang2024conformal} or establishing tighter bounds connecting the number of samples per bin to calibration \citep{kumar2019verified,gupta2020distribution} would be a valuable direction for future work.
\section*{Acknowledgements}
This work was supported by NIH 1R01EY037101 and ARO MURI W911NF-20-1-0080.
We gratefully acknowledge Professor Jin Hyun Kim and Professor Yong-Seop Han for providing the visual acuity dataset.

\bibliography{references}

\appendix
\onecolumn
\section{Algorithms}
The section includes the two algorithms (Algorithm \ref{alg:binning} \& \ref{alg:algorithm}) which are described in the main text.
In Algorithm \ref{alg:binning}, we further denote $Q_b \coloneqq (l_b, u_b)$ as the bin edges of $\Is_b$, so that the binning can equivalently be expressed as:
\begin{equation}
\label{eqn:binning_edges}
\Is_b = \left\{j \middle| l_b \leq E(x_j) < u_b \right\}, \quad \text{for } b=1,2,\dots, B.
\end{equation}

\begin{algorithm}[!htb]
\caption{Transformation-based Binning (T-Binning)}
\label{alg:binning}
\textbf{Input:} Dataset $D$, Predictor $f$, Number of transformations $L$, Number of bins $B$\\
\textbf{Output:} Set of indices for binned data, $\Is_{1,\dots,B}$, Set of bin edges $Q_{1, \dots,B}$
\begin{algorithmic}[1]
\STATE \textbf{Function} \texttt{TBinning}($D$, $f$, $L$, $B$)

\FOR{each example $x \in D$}
    \STATE Obtain the prediction for the original input: $f(x)$
    \STATE Compute predictions for the transformed inputs: $f(t_l(x))$ for $l=1,\dots, L$
    \STATE Evaluate the example difficulty score $E(x; f)$ using Equation~\ref{eqn:similarity}
\ENDFOR
\STATE Sort the examples by $E(x; f)$
\STATE Partition the sorted examples into $B$ bins to obtain $\Is_b$ and $Q_b$ (Equations \ref{eqn:binning} and \ref{eqn:binning_edges})
\RETURN $\Is_{1,\dots,B}$, $Q_{1, \dots,B}$
\end{algorithmic}
\end{algorithm}

\begin{algorithm}[!h]
\caption{Prediction Set Algorithm}
\label{alg:algorithm}
\textbf{Input:} Validation dataset $D_{val}$, Predictor $f$, Number of transformations $L$, Group conditional conformal prediction algorithm $\As$\\
\textbf{Output:} Prediction set $C(x)$
\begin{algorithmic}[1]
\STATE Split $D_{val}$ into $D_{cal}$ and $D_{tune}$
\STATE Find the best hyperparameter using $D_{cal}$ and $D_{tune}$
\STATE $\Is_{1,\dots,B^*}, Q_{1,\dots,B^*} \leftarrow$ \texttt{TBinning}($D, f, L, B^*$) \hfill $\triangleright$ Algorithm \ref{alg:binning}
\STATE $h_{1,...,B^*} \leftarrow$ Apply $\As$ with $\Is_b$
\STATE Find $b$ for $x$ using $Q_{1, \dots, B^*}$
\STATE Compute $C(x)$ using $h_b$
\RETURN $C(x)$
\end{algorithmic}
\end{algorithm}

\section {Additional Experimental Details}
In this section, we describe additional experimental details mentioned in the main text.
We begin with how we select hyperparameters for the algorithms.

\paragraph{Hyperparameter Tuning.}
RAPS, SAPS, and our proposed methods involve hyperparameters, and we need to select the appropriate ones.
For RAPS and SAPS, We follow the hyperparmeter selection from their original papers.
RAPS selects $k^*$ based on the ground truth ranks in validation data, and chooses the best regularization weight $\lambda \in \{0.001, 0.01, 0.1, 0.2, 0.5\}$ based on the SSCV metric.
For SAPS, we select the best weight $w\in \{0, 0.02\} \cup \{0.05, 0.10, 0.15, \dots, 0.65\}$ based on SSCV.

For our algorithms, we tune the number of bins $B \in \{10,20,30\}$ and  the transformation severity, controlled by the standard deviation $\sigma\in\{0.03, 0.05, 0.07,0.10\}$ of Gaussian Noise.
In the case of O-SAPS, we additionally tune the weight parameter $w$ using the same candidate values as SAPS.

\section{Additional Experimental Results}
We include additional experimental results in this section.

\subsection{Metric Comparison Experiment}
Table \ref{tab:apdx_full_metric_result} presents the full correlation coefficients and corresponding p-values from the metric validation experiment in Section \ref{sec:metric}.
Overall, T-SS shows the highest correlation with statistically significant p-values ($<$ .001).
While the Deficit metric also shows statistically significant correlations, T-SS generally achieves stronger correlation coefficients.
Another notable pattern is that Deficit consistently exhibits a strong negative correlation, whereas other metrics tend to show either positive correlation or negative correlation with lower magnitude.
This indicates that a lower Deficit value corresponds to a higher property satisfaction rate. 
A smaller Deficit implies that the prediction set size is closer to the ground-truth rank, without yet including the ground-truth. 
Consequently, groups with higher difficulty (\emph{i.e.,} a larger average ground-truth rank) tend to have larger prediction sets, increasing the likelihood that the property holds.

\begin{table}[!htb]
\scriptsize
    \centering
    \begin{tabular}{cc ccccc ccccc}
    \toprule
    \multirow{2}{*}{Model} & \multirow{2}{*}{ Algorithm} &  \multicolumn{5}{c}{Coefficient} & \multicolumn{5}{c}{P-value}\\
    & & Deficit & Excess & SSCV & ESCV & T-SS & Deficit & Excess & SSCV & ESCV & T-SS\\
    \midrule
    \multirow{4}{*}{ResNet18} & LAC & -0.4548 & -0.0364 & 0.0902 & -0.0179 & \textbf{0.6105} & $<$ .001 & 0.250 & 0.004 & 0.572 & $<$ .001\\
    & APSR & -0.5539 & -0.0482 & 0.0185 & -0.0018 & \textbf{0.6364} & $<$ .001 & 0.127 & 0.558 & 0.954 & $<$ .001 \\
    & RAPS & -0.4134 & -0.1253 & -0.1605 & 0.1123 & \textbf{0.5987} & $<$ .001 & $<$ .001 & $<$ .001 & $<$ .001 & $<$ .001 \\
    & SAPS & -0.4696 & -0.0766 & 0.1409 & 0.0322 & \textbf{0.6000} & $<$ .001 & 0.015 & $<$ .001 & 0.309 & $<$ .001 \\
\midrule
    \multirow{4}{*}{ResNet50} & LAC & \textbf{-0.6589} & -0.0134 & 0.0598 & 0.0553 & 0.6438 & $<$ .001 & 0.672 & 0.059 & 0.081 & $<$ .001\\
    & APSR & -0.0750 & -0.1357 & -0.1654 & 0.0527 & \textbf{0.2707} & 0.018 & $<$ .001 & $<$ .001 & 0.096 & $<$ .001\\
    & RAPS & -0.6128 & 0.0118 & 0.0041 & -0.0110 & \textbf{0.7369} & $<$ .001 & 0.708 & 0.896 & 0.727 & $<$ .001\\
    & SAPS & -0.6263 & -0.0092 & 0.0567 & 0.0585 & \textbf{0.7024} & $<$ .001 & 0.772 & 0.073 & 0.064 & $<$ .001\\
 \midrule
    \multirow{4}{*}{ResNet152} & LAC & -0.6088 & -0.0097 & 0.0861 & -0.0210 & \textbf{0.7094} & $<$ .001 & 0.760 & 0.006 & 0.508 & $<$ .001\\
    & APSR & -0.6855 & -0.0081 & -0.0403 & 0.0110 & \textbf{0.6998} & $<$ .001 & 0.799 & 0.203 & 0.728 & $<$ .001\\
    & RAPS &-0.5995 & -0.0148 & -0.0814 & 0.1679 & \textbf{0.7427} & $<$ .001 & 0.640 & 0.010 & $<$ .001 & $<$ .001\\
    & SAPS & -0.6164 & -0.0244 & 0.0161 & 0.0237 & \textbf{0.7512} & $<$ .001 & 0.440 & 0.612 & 0.454 & $<$ .001\\
\midrule
    \multirow{4}{*}{ViT-B-16} & LAC  & -0.7785 & -0.0644 & 0.0134 & -0.0299 & \textbf{0.8322} & $<$ .001 & 0.042 & 0.672 & 0.345 & $<$ .001\\
    & APSR & \textbf{-0.8262} & 0.0399 & -0.0147 & 0.0069 & 0.7289 & $<$ .001 & 0.208 & 0.643 & 0.828 & $<$ .001\\
    & RAPS & -0.7982 & -0.0314 & -0.0292 & 0.1092 & \textbf{0.8266} & $<$ .001 & 0.321 & 0.357 & 0.001 & $<$ .001\\
    & SAPS & -0.7866 & -0.0192 & 0.0584 & 0.0568 & \textbf{0.8621} & $<$ .001 & 0.545 & 0.065 & 0.073 & $<$ .001\\
\midrule
    \multirow{4}{*}{ViT-L-16} & LAC  & \textbf{-0.6257} & -0.1174 & -0.0763 & -0.0310 & 0.6068 & $<$ .001 & 0.545 & 0.065 & 0.073 & $<$ .001\\
    & APSR & \textbf{-0.8678} & 0.0867 & - & -0.1325 & 0.7199 & $<$ .001 & 0.006 & NaN & $<$ .001 & $<$ .001\\
    & RAPS & -0.8406 & -0.0218 & 0.0501 & 0.0394 & \textbf{0.8607} & $<$ .001 & 0.490 & 0.113 & 0.213 & $<$ .001\\
    & SAPS & -0.8360 & 0.0000 & 0.0757 & 0.0342 & \textbf{0.8621} & $<$ .001 & 0.999 & 0.017 & 0.279 & $<$ .001\\
\midrule
    \multirow{4}{*}{ViT-H-14} & LAC  & -0.4780 & -0.1454 & -0.1201 & -0.1110 & \textbf{0.5334}  & $<$ .001 & $<$ .001 & $<$ .001 & $<$ .001 & $<$ .001\\
    & APSR & \textbf{-0.6691} & 0.0814 & 0.0373 & -0.1938 & 0.6266 & $<$ .001 & $<$ .001 & 0.010 & 0.239 & $<$ .001\\
    & RAPS & \textbf{-0.8280} & -0.0606 & 0.0144 & 0.1026 & 0.7429 & $<$ .001 & 0.055 & 0.649 & 0.001 & $<$ .001\\
    & SAPS & -0.8151 & -0.0161 & 0.0211 & -0.0778 & \textbf{0.8341} & $<$ .001 & 0.611 & 0.506 & 0.014 & $<$ .001\\
\midrule
    \multirow{4}{*}{EfficientNet-V2-M} & LAC  & -0.7897 & -0.0031 & 0.0220 & 0.0159 & \textbf{0.8166} & $<$ .001 & 0.923 & 0.486 & 0.615 & $<$ .001\\
    & APSR & -0.3751 & -0.1815 & -0.1919 & 0.1155 & \textbf{0.4840} & $<$ .001 & $<$ .001 & $<$ .001 & $<$ .001 & $<$ .001\\
    & RAPS & -0.7671 & -0.0237 & 0.0351 & 0.0025 & \textbf{0.8588} & $<$ .001 & 0.455 & 0.268 & 0.938 & $<$ .001\\
    & SAPS & -0.7842 & 0.0436 & 0.0605 & -0.0014 & \textbf{0.8608} & $<$ .001 & 0.168 & 0.056 & 0.965 & $<$ .001\\
\midrule
    \multirow{4}{*}{EfficientNet-V2-L} & LAC  & \textbf{-0.8246} & -0.0784 & 0.0504 & -0.0501 & 0.5278 & $<$ .001 & 0.013 & 0.111 & 0.113 & $<$ .001\\
    & APSR & \textbf{-0.7922} & 0.0361 & 0.0141 & -0.0670 & 0.6021 & $<$ .001 & 0.254 & 0.656 & 0.034 & $<$ .001\\
    & RAPS & \textbf{-0.8206} & 0.0206 & 0.0290 & 0.0125 & 0.6181 & $<$ .001 & 0.254 & 0.656 & 0.034 & $<$ .001\\
    & SAPS & \textbf{-0.8210} & -0.0133 & 0.0266 & 0.0229 & 0.6330 & $<$ .001 & 0.675 & 0.400 & 0.470 & $<$ .001\\
\midrule
\multirow{4}{*}{Swin-V2-B} & LAC  & \textbf{-0.7264} & -0.0045 & 0.0051 & 0.0398 & 0.7064 & $<$ .001 & 0.888 & 0.871 & 0.208 & $<$ .001\\
& APSR & \textbf{-0.7189} & -0.0016 & -0.1974 & 0.0873 & 0.7114 & $<$ .001 & 0.959 & $<$ .001 & 0.006 & $<$ .001\\
& RAPS & -0.7219 & 0.0004 & 0.0474 & 0.1344 & \textbf{0.7837} & $<$ .001 & 0.990 & 0.134 & $<$ .001 & $<$ .001\\
& SAPS & -0.7329 & 0.0196 & -0.0333 & 0.0489 & \textbf{0.7511} & $<$ .001 & 0.536 & 0.292 & 0.123 & $<$ .001\\
    \bottomrule
    \end{tabular}
    \caption{Correlation Coefficient and P-value. B=50. Bold numbers represent the best correlation coefficients.}
    \label{tab:apdx_full_metric_result}
\end{table}

\subsection{Hyperparameter Sensitivity}
\begin{figure*}[!h]
    \centering
    \begin{subfigure}{0.32\linewidth}
        \includegraphics[width=\linewidth]{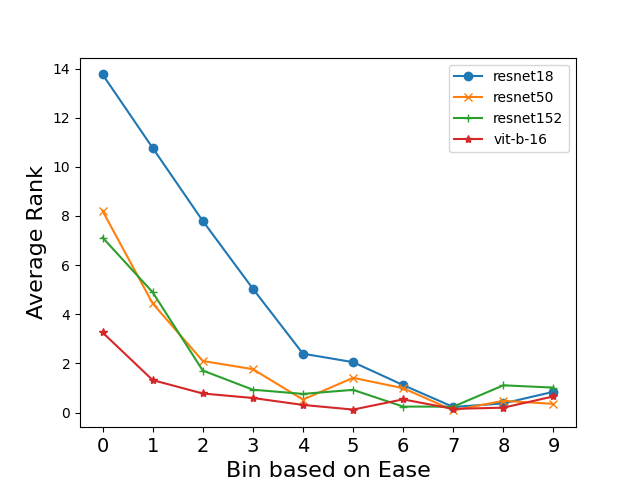}
        \caption{Brightness}
    \end{subfigure}%
    \begin{subfigure}{0.32\linewidth}
        \includegraphics[width=\linewidth]{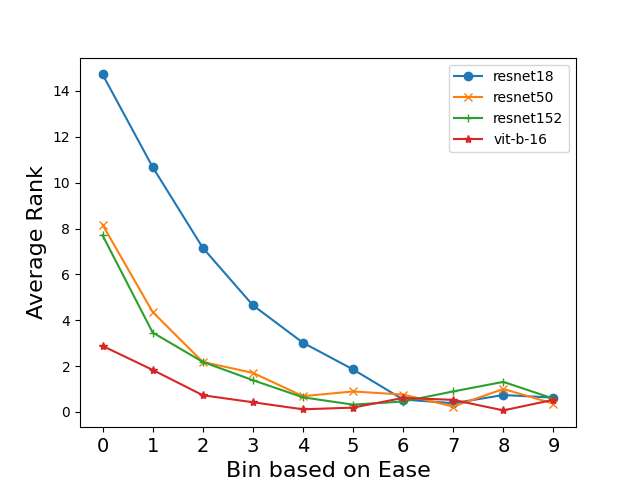}
        \caption{Contrast}
    \end{subfigure}%
    \begin{subfigure}{0.32\linewidth}
        \includegraphics[width=\linewidth]{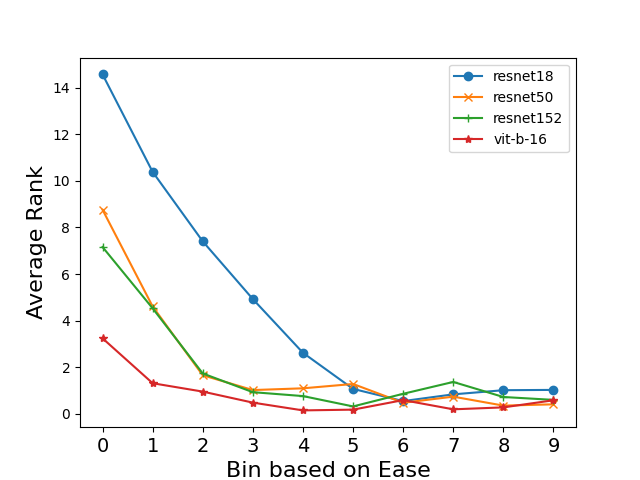}
        \caption{Saturation}
    \end{subfigure}%
    \caption{Different Input Transformations.}
    \label{fig:apdx_diff_trans}
\end{figure*}

\begin{figure}
    \centering
    \includegraphics[width=0.5\linewidth]{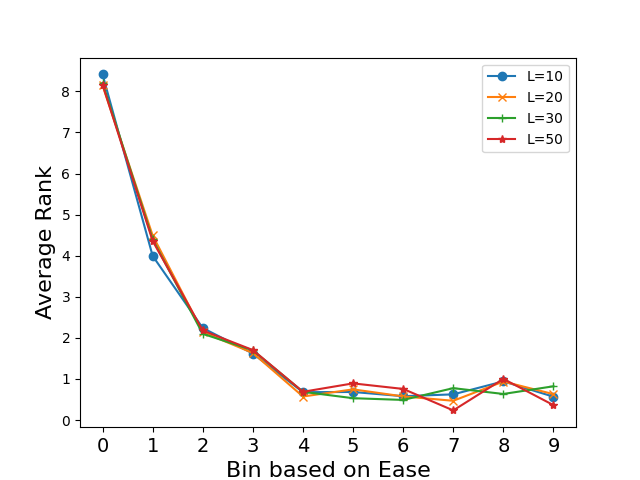}
    \caption{Different Number of Input Transformation.}
    \label{fig:apdx_diff_num_trans}
\end{figure}

We perform two supplementary experiments to analyze the sensitivity of two key hyperparameters: the input transformation type and the number of input transformations.
First, we incorporate three additional transformations (brightness, contrast, and saturation) into the image classification experiment and present the results in Figure \ref{fig:apdx_diff_trans}.
Next, we vary the number of transformations (10, 20, 30, and 50) and report the corresponding results in Figure \ref{fig:apdx_diff_num_trans}.
All experiments show that our approach is less sensitive to both the type and the number of transformations.

\subsection{Image Classification Experiments - Algorithm Comparisons}
Table \ref{tab:apdx_alg_results_cov_avgss} and \ref{tab:apdx_alg_results_our_metrics} present results for the two target miscoverage rates, $\alpha\in\{0.20, 0.15\}$, as the main text already includes the results for the other settings.
Table \ref{tab:apdx_alg_results_deficit_excess} and \ref{tab:apdx_alg_results_sscv_escv}, on the other hand, report results for all four target miscoverage rates, $\alpha\in\{0.20,0.15,0.10,0.05\}$

In Table \ref{tab:apdx_alg_results_cov_avgss}, all algorithms mostly show the coverage rate close to the target levels.
Among them, LAC consistently achieves the smallest average set size, while APS yields the largest.
Table \ref{tab:apdx_alg_results_our_metrics} compares the algorithms using our proposed metrics, showing a similar pattern with the results reported in Table \ref{tab:alg_results_our_metrics} in the main text.
One notable observation is that for higher values of $\alpha$, our approaches (O-LAC and O-SAPS) do not consistently outperform other methods when using vision transformer architectures.
We hypothesize that this is related to the high accuracy of these models, which leads to two unfavorable effects under high target miscoverage rates.
First, most examples have a ground truth rank of one, due to the model's high accuracy.
Second, higher values of $\alpha$ result in smaller prediction sets being constructed by our approaches (see Table \ref{tab:apdx_alg_results_cov_avgss}). 
These two factors reduce the variation in average ranks and average set sizes across bins, making it difficult to achieve high T-SS scores.

Table \ref{tab:apdx_alg_results_deficit_excess} and \ref{tab:apdx_alg_results_sscv_escv} evaluate the algorithms using existing metrics: Deficit, Excess, SSCV, and ESCV.
Deficit and Excess are closely related to average set size.
The smallest set size (LAC) results in the best Excess and the worst Deficit, while the largest set size (APS) shows the opposite trend.
SSCV and ESCV exhibit more varied pattern, as they depend on the the number of examples within each bin.

Lastly, we also evaluated our metrics using a different number of bins.
The results, shown in Table \ref{tab:apdx_alg_results_our_metrics_B100}, exhibit a pattern similar to that observed with $B=50$.

\begin{table}[!htb]
\scriptsize
\centering
\begin{tabular}{c c cccccc| cccccc}
\toprule
\multirow{2}{*}{Base Model} & \multirow{2}{*}{Acc.} & \multicolumn{6}{c}{Coverage Rate} & \multicolumn{6}{c}{Average Set Size}\\   
& & LAC & APS & RAPS & SAPS & O-LAC & O-SAPS & LAC & APS & RAPS & SAPS & O-LAC & O-SAPS \\
\midrule
\multicolumn{14}{c}{$\alpha=0.20$}\\
\midrule
ResNet18 & 69.89 & 0.799 & 0.799 & 0.798 & 0.801 & 0.799 & 0.803 & \textbf{1.50} & 5.93 & 5.02 & \underline{2.71} & 3.07 & 3.77\\
ResNet50 & 80.81 & 0.798 & 0.798 & 0.798 & 0.799 & 0.803 & 0.801 & \textbf{0.98} & 10.88 & 2.95 & \underline{1.33} & 1.52 & 1.99\\
ResNet152 & 82.26 & 0.798 & 0.803 & 0.803 & 0.803 & 0.801 & 0.805 & \textbf{0.92} & 8.31 & 2.07 & 1.56 & \underline{1.40} & 2.72\\
ViT-B-16 & 85.22 & 0.798 & 0.803 & 0.802 & 0.803 & 0.795 & 0.805 & \textbf{0.88} & 2.02 & 1.66 & 1.38 & \underline{1.14} & 1.21\\
ViT-L-16 & 88.20 & 0.799 & 0.802 & 0.801 & 0.802 & 0.798 & 0.804 & \textbf{0.85} & 2.87 & 1.33 & \underline{1.02} & \underline{1.02} & 1.07\\
ViT-H-14 & 88.53 & 0.800 & 0.801 & 0.801 & 0.801 & 0.793 & 0.804 & \textbf{0.85} & 2.00 & 1.21 & \underline{0.99} & 1.00 & 1.04\\
EfficientNetV2-M & 85.11 & 0.795 & 0.801 & 0.801 & 0.803 & 0.803 & 0.805 & \textbf{0.88} & 5.57 & 1.76 & 1.34 & \underline{1.19} & 1.29\\
EfficientNetV2-L & 85.68 & 0.797 & 0.800 & 0.800 & 0.800 & 0.802 & 0.803 & \textbf{0.87} & 4.69 & 1.63 & \underline{1.09} & \underline{1.10} & 1.17\\
Swin-V2-B & 83.93 & 0.800 & 0.799 & 0.800 & 0.801 & 0.802 & 0.805 & \textbf{0.90} & 2.09 & 1.73 & 1.38 & \underline{1.25} & 1.33\\
\midrule
\multicolumn{14}{c}{$\alpha=0.15$}\\
\midrule
ResNet18 & 69.89 & 0.850 & 0.847 & 0.848 & 0.850 & 0.851 & 0.852 & \textbf{2.15} & 8.47 & 6.74 & \underline{3.18} & 3.25 & 3.18\\
ResNet50 & 80.81 & 0.848 & 0.847 & 0.848 & 0.848 & 0.851 & 0.850 & \textbf{1.15} & 18.47 & 1.68 & \underline{1.54} & 1.87 & 3.78\\
ResNet152 & 82.26 & 0.850 & 0.853 & 0.853 & 0.853 & 0.851 & 0.856 & \textbf{1.07} & 14.37 & 2.69 & 1.77 & \underline{1.72} & 3.42\\
ViT-B-16 & 85.22 & 0.851 & 0.851 & 0.851 & 0.852 & 0.846 & 0.854 & \textbf{0.99} & 3.03 & 1.56 & 1.56 & \underline{1.31} & 1.40\\
ViT-L-16 & 88.20 & 0.852 & 0.850 & 0.851 & 0.852 & 0.848 & 0.854 & \textbf{0.93} & 4.70 & 1.55 & 1.26 & \underline{1.15} & 1.21\\
ViT-H-14 & 88.53 & 0.849 & 0.850 & 0.851 & 0.852 & 0.846 & 0.855 & \textbf{0.92} & 3.09 & 1.39 & 1.24 & \underline{1.12} & 1.17\\
EfficientNetV2-M & 85.11 & 0.849 & 0.850 & 0.852 & 0.853 & 0.854 & 0.855 & \textbf{0.99} & 9.36 & 2.22 & 1.53 & \underline{1.39} & 1.58\\
EfficientNetV2-L & 85.68 & 0.847 & 0.851 & 0.851 & 0.851 & 0.853 & 0.853 & \textbf{0.97} & 8.55 & 2.03 & 1.57 & \underline{1.26} & 1.37\\
Swin-V2-B & 83.93 & 0.849 & 0.849 & 0.849 & 0.852 & 0.851 & 0.854 & \textbf{1.02} & 3.45 & 1.66 & 1.56 & \underline{1.48} & 1.64\\
\midrule
\multicolumn{14}{c}{$\alpha=0.10$}\\
\midrule
ResNet18 & 69.89 & 0.900 & 0.900 & 0.898 & 0.899 & 0.901 & 0.904 & \textbf{3.66} & 13.82 & 9.50 & \underline{5.03} & 7.15 & 9.60\\
ResNet50 & 80.81 & 0.897 & 0.898 & 0.899 & 0.898 & 0.899 & 0.901 & \textbf{1.47} & 34.62 & 2.68 & \underline{1.89} & 2.62 & 5.35\\
ResNet152 & 82.26 & 0.901 & 0.900 & 0.902 & 0.901 & 0.900 & 0.904 & \textbf{1.35} & 26.82 & 1.82 & \underline{1.64} & 2.40 & 4.50\\
ViT-B-16 & 85.22 & 0.901 & 0.900 & 0.901 & 0.901 & 0.895 & 0.902 & \textbf{1.16} & 5.73 & 1.91 & 1.78 & \underline{1.58} & 1.78\\
ViT-L-16 & 88.20 & 0.901 & 0.900 & 0.900 & 0.901 & 0.901 & 0.903 & \textbf{1.05} & 9.20 & \underline{1.29} & 1.41 & 1.33 & 1.45\\
ViT-H-14 & 88.53 & 0.898 & 0.902 & 0.901 & 0.902 & 0.899 & 0.905 & \textbf{1.03} & 6.08 & \underline{1.26} & 1.39 & 1.30 & 1.40\\
EfficientNetV2-M & 85.11 & 0.902 & 0.900 & 0.905 & 0.903 & 0.902 & 0.906 & \textbf{1.18} & 17.52 & 1.60 & \underline{1.40} & 1.73 & 2.37\\
EfficientNetV2-L & 85.68 & 0.898 & 0.900 & 0.899 & 0.901 & 0.903 & 0.904 & \textbf{1.13} & 17.48 & \underline{1.46} & 1.59 & 1.51 & 1.78\\
Swin-V2-B & 83.93 & 0.902 & 0.898 & 0.900 & 0.902 & 0.899 & 0.902 & \textbf{1.24} & 7.63 & 2.08 & \underline{1.76} & 1.90 & 2.21\\
\midrule
\multicolumn{14}{c}{$\alpha=0.05$}\\
\midrule
ResNet18 & 69.89 & 0.948 & 0.950 & 0.950 & 0.950 & 0.951 & 0.952 & \textbf{9.14} & 29.63 & 16.15 & \underline{14.71} & 16.35 & 19.79\\
ResNet50 & 80.81 &  0.950 & 0.950 & 0.949 & 0.950 & 0.951 & 0.953 & \textbf{2.59} & 80.43 & 5.08 & 5.20 & \underline{4.89} & 9.65\\
ResNet152 & 82.26 & 0.952 & 0.950 & 0.953 & 0.952 & 0.950 & 0.953 & \textbf{2.27} & 62.79 & 4.26 & \underline{2.96} & 4.43 & 7.38\\
ViT-B-16 & 85.22 &  0.949 & 0.950 & 0.949 & 0.950 & 0.949 & 0.952 & \textbf{1.62} & 16.55 & 2.63 & \underline{2.08} & 2.43 & 2.94\\
ViT-L-16 & 88.20 & 0.951 & 0.951 & 0.950 & 0.951 & 0.954 & 0.955 & \textbf{1.34} & 23.75 & 1.75 & \underline{1.65} & 1.84 & 2.28\\
ViT-H-14 & 88.53 &  0.952 & 0.953 & 0.952 & 0.952 & 0.952 & 0.955 & \textbf{1.31} & 18.98 & 2.40 & \underline{1.62} & 1.79 & 2.06\\
EfficientNetV2-M & 85.11 & 0.953 & 0.951 & 0.954 & 0.953 & 0.952 & 0.955 & \textbf{1.76} & 38.56 & 2.85 & \underline{2.31} & 2.84 & 5.26\\
EfficientNetV2-L & 85.68 & 0.952 & 0.950 & 0.952 & 0.951 & 0.954 & 0.954 & \textbf{1.57} & 45.77 & 2.45 & \underline{1.97} & 2.32 & 3.20\\
Swin-V2-B & 83.93 & 0.952 & 0.948 & 0.950 & 0.949 & 0.952 & 0.953 & \textbf{1.88} & 23.89 & 3.59 & \underline{2.97} & 3.20 & 4.28\\
\bottomrule
\end{tabular}
\caption{Coverage Rate and Average Set Size. $\alpha \in \{0.20,0.15.0.10,0.05\}$. Bold numbers indicate the best (\emph{i.e.,} smallest) average set size, while underlined numbers represent the second-best. All algorithms achieve empirical coverage rates close to the target coverage rate ($1-\alpha\in \{0.80,0.85,0.90, 0.95\}$).
In terms of average set size, LAC produces the smallest sets on average, while APS yields the largest sets, with the others showing similar sizes.}
\label{tab:apdx_alg_results_cov_avgss}
\end{table}

\begin{table}[!bt]
\scriptsize
\centering
\begin{tabular}{c c cccccc| cccccc}
\toprule
\multirow{2}{*}{Base Model} & \multirow{2}{*}{Acc.} & \multicolumn{6}{c}{T-CV} & \multicolumn{6}{c}{T-SS}\\   
& & LAC & APS & RAPS & SAPS & O-LAC & O-SAPS & LAC & APS & RAPS & SAPS & O-LAC & O-SAPS \\
\midrule
\multicolumn{14}{c}{$\alpha=0.20$}\\
\midrule
ResNet18 & 69.89 & 0.352 & \textbf{0.053} & \underline{0.060} & 0.140 & 0.070 & 0.070 & 0.634 & 0.916 & \underline{0.919} & 0.873 & \textbf{0.921} & 0.918\\
ResNet50 & 80.81 & 0.573 & 0.182 & 0.170 & 0.227 & \underline{0.117} & \textbf{0.070} & -0.647 & 0.379 & 0.713 & 0.670 & \textbf{0.828} & \underline{0.798}\\
ResNet152 & 82.26 & 0.613 & 0.100 & \textbf{0.060} & \underline{0.062} & 0.092 & 0.075 & -0.707 & \textbf{0.794} & \underline{0.793} & 0.744 & 0.766 & 0.776\\
ViT-B-16 & 85.22 & 0.770 & 0.087 & 0.075 & 0.100 & \underline{0.065} & \textbf{0.060} & -0.723 & \underline{0.808} & \textbf{0.818} & 0.762 & 0.759 & 0.757\\
ViT-L-16 & 88.20 & 0.785 & 0.115 & 0.082 & 0.095 & \underline{0.062} & \textbf{0.053} & -0.711 & \underline{0.813} & \textbf{0.826} & 0.766 & 0.766 & 0.774\\
ViT-H-14 & 88.53 & 0.790 & \underline{0.055} & \textbf{0.053} & 0.105 & 0.060 & \underline{0.053} & -0.581 & \underline{0.620} & \textbf{0.636} & \underline{0.620} & 0.605 & \underline{0.620}\\
EfficientNetV2-M & 85.11 & 0.685 & 0.073 & \textbf{0.055} & 0.068 & 0.063 & \underline{0.058} & -0.600 & 0.422 & \underline{0.605} & 0.604 & \textbf{0.624} & \underline{0.604}\\
EfficientNetV2-L & 85.68 & 0.593 & 0.063 & \textbf{0.057} & 0.125 & \underline{0.063} & \textbf{0.057} & -0.522 & 0.537 & \textbf{0.565} & 0.540 & 0.539 & \underline{0.548}\\
Swin-V2-B & 83.93 & 0.708 & 0.067 & 0.070 & 0.068 & \underline{0.060} & \textbf{0.057} & -0.681 & 0.657 & 0.676 & 0.686 & \underline{0.691} & \textbf{0.694}\\
\midrule
Average & -  & 0.652 & 0.088 & 0.076 & 0.110 & \underline{0.072} & \textbf{0.061} & -0.504 & 0.661 & \textbf{0.728} & 0.696 & \underline{0.722} & 0.721\\
Avg. Rank from CD & - & 6.00 & 2.67 & 1.56 & 3.67 & \underline{1.44} & \textbf{1.33} & 6.00 & 2.89 & \textbf{1.56} & 3.33 & \underline{2.00} & 2.11\\
\midrule
\multicolumn{14}{c}{$\alpha=0.15$}\\
\midrule
ResNet18 & 69.89 & 0.275 & \textbf{0.052} & \underline{0.058} & 0.170 & 0.060 & 0.063 & 0.790 & 0.919 & \textbf{0.928} & 0.856 & \underline{0.926} & 0.916\\
ResNet50 & 80.81 & 0.457 & 0.173 & 0.290 & 0.255 & \underline{0.100} & \textbf{0.065} & 0.079 & 0.301 & 0.401 & 0.651 & \textbf{0.831} & \underline{0.800}\\
ResNet152 & 82.26 & 0.440 & 0.090 & \textbf{0.052} & \underline{0.060} & 0.075 & 0.065 & 0.217 & \underline{0.780} & \textbf{0.787} & 0.743 & 0.775 & 0.774\\
ViT-B-16 & 85.22 & 0.570 & 0.070 & \underline{0.057} & 0.080 & \textbf{0.055} & \textbf{0.055} & -0.575 & \underline{0.811} & \textbf{0.817} & 0.756 & 0.777 & 0.751\\
ViT-L-16 & 88.20 & 0.742 & 0.095 & 0.070 & 0.060 & \underline{0.053}& \textbf{0.048} & -0.760 & \underline{0.821} & \textbf{0.828} & 0.771 & 0.777 & 0.770\\
ViT-H-14 & 88.53 & 0.757 & \textbf{0.050} & \textbf{0.050} & \underline{0.065} & \textbf{0.050} & \textbf{0.050} & -0.586 & \underline{0.625} & \textbf{0.644} & 0.620 & 0.620 & 0.622\\
EfficientNetV2-M & 85.11 & 0.532 & 0.060 & \underline{0.055} & \textbf{0.047} & \underline{0.055} & \textbf{0.047} & -0.470 & 0.411 & 0.561 & 0.593 & \textbf{0.626} & \underline{0.601}\\
EfficientNetV2-L & 85.68 & 0.455 & 0.060 & \underline{0.052} & 0.065 & 0.058 & \textbf{0.050} & -0.443 & 0.528 & \textbf{0.563} & 0.540 & \underline{0.554} & 0.550\\
Swin-V2-B & 83.93 & 0.525 & 0.058 & 0.058 & \underline{0.053} & \underline{0.053} & \textbf{0.050} & -0.002 & 0.678 & 0.689 & 0.679 & \underline{0.689} & \textbf{0.696}\\
\midrule
Average & - & 0.528 & 0.079 & 0.082 & 0.095 & \underline{0.062} & \textbf{0.055} & -0.194 & 0.653 & 0.691 & 0.690 & \textbf{0.731} & \underline{0.720}\\
Avg. Rank from CD & - & 6.00 & 2.89 & 2.00 & 3.00 & \underline{1.89} & \textbf{1.11} & 6.00 & 2.89 & \textbf{1.44} & 3.22 & \underline{1.56} & 2.11\\
\bottomrule
\end{tabular}
\caption{Our Metrics (T-CV, T-SS). $\alpha \in \{0.20,0.15\}$. Bold values denote the best case, while the underlined values represent the second-best. LAC mostly shows the negative T-SS values where the base classifier already satisfies the target coverage rate.}
\label{tab:apdx_alg_results_our_metrics}
\end{table}

\begin{table}[!bt]
\scriptsize
\centering
\begin{tabular}{c c cccccc| cccccc}
\toprule
\multirow{2}{*}{Base Model} & \multirow{2}{*}{Acc.} & \multicolumn{6}{c}{Deficit} & \multicolumn{6}{c}{Excess}\\   
& & LAC & APS & RAPS & SAPS & O-LAC & O-SAPS & LAC & APS & RAPS & SAPS & O-LAC & O-SAPS \\
\midrule
\multicolumn{14}{c}{$\alpha=0.20$}\\
\midrule
ResNet18 & 69.89 & 4.18 & \textbf{2.91} & \underline{3.08} & 3.73 & 3.61 & 3.51 & \textbf{0.26} & 3.41 & 2.64 & \underline{1.01} & 1.25 & 1.81\\
ResNet50 & 80.81 & 2.10 & \textbf{1.59} & \underline{1.76} & 1.97 & 1.90 & 1.80 & \textbf{0.04} & 9.43 & 1.67 & \underline{0.25} & 0.37 & 0.78\\
ResNet152 & 82.26 & 1.99 & \textbf{1.52} & \underline{1.69} & 1.77 & 1.79 & 1.66 & \textbf{0.00} & 6.92 & 0.85 & 0.43 & \underline{0.28} & 1.46\\
ViT-B-16 & 85.22 & 0.92 & \textbf{0.70} & \underline{0.72} & 0.75 & 0.80 & 0.77 & \textbf{0.00} & 0.92 & 0.58 & 0.34 & \underline{0.15} & 0.19\\
ViT-L-16 & 88.20 & 0.61 & \textbf{0.45} & \underline{0.47} & 0.51 & 0.52 & 0.50 & \textbf{0.00} & 1.86 & 0.34 & \underline{0.07} & 0.09 & 0.11\\
ViT-H-14 & 88.53 & 0.57 & \textbf{0.43} & \underline{0.45} & 0.48 & 0.50 & 0.47 & \textbf{0.00} & 1.02 & 0.24 & \underline{0.06} & 0.07 & 0.10\\
EfficientNetV2-M & 85.11 & 1.22 & \textbf{0.97} & \underline{1.03} & 1.09 & 1.08 & 1.51 & \textbf{0.00} & 4.44 & 0.69 & 0.31 & \underline{0.17} & 0.26\\
EfficientNetV2-L & 85.68 & 1.19 & \textbf{0.94} & \underline{1.01} & 1.07 & 1.09 & 1.04 & \textbf{0.00} & 3.56 & \underline{0.57} & 0.10 & 0.11 & 0.17\\
Swin-V2-B & 83.93 &1.48 & \textbf{1.23} & \underline{1.25} & 1.31 & 1.33 & 1.31 & \textbf{0.00} & 0.95 & 0.60 & 0.32 & \underline{0.20} & 0.27\\
\midrule
\multicolumn{14}{c}{$\alpha=0.15$}\\
\midrule
ResNet18 & 69.89 & 3.91 & \textbf{2.52} & \underline{2.74} & 3.56 & 3.25 & 3.18 & \textbf{0.64} & 5.56 & 4.04 & \underline{1.36} & 2.18 & 3.11\\
ResNet50 & 80.81 & 2.00 & \textbf{1.43} & 1.91 & 1.89 & 1.78 & \underline{1.68} & \textbf{0.11} & 16.87 & 0.54 & \underline{0.39} & 0.60 & 2.39\\
ResNet152 & 82.26 & 1.88 & \textbf{1.38} & 1.60 & 1.70 & 1.68 & \underline{1.57} & \textbf{0.05} & 12.83 & 1.38 & 0.57 & \underline{0.50} & 2.07\\
ViT-B-16 & 85.22 & 0.82 & \textbf{0.60} & \underline{0.67} & 0.69 & 0.73 & 0.70 & \textbf{0.01} & 1.84 & 0.43 & 0.45 & \underline{0.24} & 0.31\\
ViT-L-16 & 88.20 & 0.53 & \textbf{0.38} & \underline{0.41} & 0.44 & 0.45 & 0.44 & \textbf{0.00} & 3.62 & 0.51 & 0.24 & \underline{0.15} & 0.19\\
ViT-H-14 & 88.53 & 0.50 & \textbf{0.36} & \underline{0.39} & 0.40 & 0.43 & 0.41 & \textbf{0.00} & 2.04 & 0.36 & 0.23 & \underline{0.13} & 0.16\\
EfficientNetV2-M & 85.11 & 1.13 & \textbf{0.87} & \underline{0.98} & 1.00 & 1.00 & \underline{0.98} & \textbf{0.01} & 8.12 & 1.08 & 0.43 & \underline{0.29} & 0.47\\
EfficientNetV2-L & 85.68 & 1.11 & \textbf{0.84} & \underline{0.94} & 0.99 & 1.01 & 0.98 & \textbf{0.01} & 7.32 & 0.90 & 0.48 & \underline{0.20} & 0.30\\
Swin-V2-B & 83.93 &1.38 & \textbf{1.12} & \underline{1.20} & 1.24 & 1.24 & 1.22 & \textbf{0.02} & 2.20 & 0.49 & 0.43 & \underline{0.34} & 0.50\\
\midrule
\multicolumn{14}{c}{$\alpha=0.10$}\\
\midrule
ResNet18 & 69.89 & 3.45 & \textbf{2.00} & 2.40 & 3.56 & 2.72 & \underline{2.70} & \textbf{1.68} & 10.40 & 6.41 & \underline{3.16} & 4.46 & 6.83\\
ResNet50 & 80.81 & 1.87 & \textbf{1.23} & 1.78 & 1.82 & 1.61 & \underline{1.55} & \textbf{0.30} & 32.80 & 1.39 & \underline{0.67}& 1.19 & 3.83\\
ResNet152 & 82.26 & 1.75 & \textbf{1.21} & 1.73 & 1.71 & 1.53 & \underline{1.45} & \textbf{0.19} & 25.12 & 0.65 & \underline{0.47} & 1.02 & 3.04\\
ViT-B-16 & 85.22 & 0.72 & \textbf{0.49} & 0.59 & 0.62 & 0.64 & \underline{0.62} & \textbf{0.09} & 4.43 & 0.70 & 0.60 & \underline{0.42} & 0.61\\
ViT-L-16 & 88.20 & 0.44 & \textbf{0.31} & 0.40 & 0.38 & 0.38 & \underline{0.37} & \textbf{0.03} & 8.04 & \underline{0.23} & 0.33 & 0.26 & 0.37\\
ViT-H-14 & 88.53 & 0.41 & \textbf{0.28} & 0.37 & 0.35 & 0.36 & \underline{0.34} & \textbf{0.02} & 4.94 & \underline{0.20} & 0.31 & 0.24 & 0.33\\
EfficientNetV2-M & 85.11 & 1.02 & \textbf{0.75} & 0.99 & 1.00 & 0.91 & \underline{0.88} & \textbf{0.10} & 16.16 & \underline{0.49} & 0.30 & 0.54 & 1.16\\
EfficientNetV2-L & 85.68 & 1.01 & \textbf{0.73} & 0.97 & 0.93 & 0.92 & \underline{0.88} & \textbf{0.07} & 16.15 & 0.37 & 0.45 & \underline{0.36} & 0.63\\
Swin-V2-B & 83.93 & 1.27 & \textbf{0.99} & 1.15 & 1.20 & 1.13 & \underline{1.12} & \textbf{0.13} & 6.24 & 0.83 & \underline{0.57} & 0.65 & 0.97\\
\midrule
\multicolumn{14}{c}{$\alpha=0.05$}\\
\midrule
ResNet18 & 69.89 & 2.50 & \textbf{1.25} & \underline{1.85} & 2.85 & 1.83 & 2.00 & \textbf{6.20} & 25.45 & 12.73 & \underline{12.12} & 12.72 & 16.36\\
ResNet50 & 80.81  & 1.63 & \textbf{0.91} & 1.61 & 1.55 & 1.33 & \underline{1.32} & \textbf{1.18} & 78.32 & 3.63 & 3.67 & \underline{3.19} & 7.93\\
ResNet152 & 82.26 &  1.53 & \textbf{0.97} & 1.54 & 1.52 & 1.30 & \underline{1.29} & \textbf{0.90} & 60.90 & 2.90 & \underline{1.59} & 2.84 & 5.77\\
ViT-B-16 & 85.22  & 0.59 & \textbf{0.36} & \underline{0.50} & 0.54 & 0.51 & \underline{0.50} & \textbf{0.42} & 15.12 & 1.33 & \underline{0.85} & 1.14 & 1.67\\
ViT-L-16 & 88.20  & 0.34 & \textbf{0.22} & 0.32 & 0.32 & 0.30 & \underline{0.29} & \textbf{0.23} & 22.51 & 0.61 & \underline{0.51} & 0.67 & 1.11\\
ViT-H-14 & 88.53  & 0.31 & \textbf{0.19} & \underline{0.25} & 0.28 & 0.27 & 0.26 & \textbf{0.20} & 17.74 & 1.23 & \underline{0.48} & 0.64 & 0.91\\
EfficientNetV2-M  & 85.11 & 0.88 & \textbf{0.59} & 0.87 & 0.87 & 0.77 & \underline{0.75} & \textbf{0.53} & 37.05 & 1.61 & \underline{1.08} & 1.51 & 3.90\\
EfficientNetV2-L  & 85.68 & 0.88 & \textbf{0.57} & 0.86 & 0.86 & 0.81 & \underline{0.79} & \textbf{0.39} & 44.25 & 1.25 & \underline{0.76} & 1.06 & 1.94\\
Swin-V2-B & 83.93 &  1.10 & \textbf{0.79} & 1.03 & 1.05 & \underline{0.97} & \underline{0.97} & \textbf{0.61} & 22.30 & 2.22 & \underline{1.64} & 1.78 & 2.87\\
\bottomrule
\end{tabular}
\caption{Deficit and Excess. $\alpha \in \{0.20, 0.15,0.10,0.05\}$. Bold numbers indicate the best result, and underlined numbers indicate the second-best}
\label{tab:apdx_alg_results_deficit_excess}
\end{table}

\begin{table}[!bt]
\scriptsize
\centering
\begin{tabular}{c c cccccc| cccccc}
\toprule
\multirow{2}{*}{Base Model} & \multirow{2}{*}{Acc.} & \multicolumn{6}{c}{SSCV} & \multicolumn{6}{c}{ESCV}\\   
& & LAC & APS & RAPS & SAPS & O-LAC & O-SAPS & LAC & APS & RAPS & SAPS & O-LAC & O-SAPS \\
\midrule
\multicolumn{14}{c}{$\alpha=0.20$}\\
\midrule
ResNet18 & 69.89 & 0.062 & 0.081 & \textbf{0.050} & \underline{0.053} & 0.083 & 0.067 & 0.800 & 0.800 & 0.800 & \underline{0.342} & 0.467 & \textbf{0.214}\\
ResNet50 & 80.81 & 0.097 & 0.198 & 0.111 & \textbf{0.012} & \underline{0.063} & 0.129 & \textbf{0.105} & 0.200 & 0.200 & \underline{0.155} & 0.200 & 0.200\\
ResNet152 & 82.26 & 0.135 & 0.192 & \textbf{0.083} & \underline{0.084} & 0.185 & 0.148 & \textbf{0.135} & \underline{0.200} & \underline{0.200} & 0.234 & 0.300 & \underline{0.200}\\
ViT-B-16 & 85.22 & \textbf{0.002} & 0.200 & 0.200 & 0.095 & \underline{0.069} & 0.085 & \textbf{0.110} & 0.800 & 0.300 & 0.316 & \underline{0.200} & 0.216\\
ViT-L-16 & 88.20 & \textbf{0.002} & 0.200 & 0.145 & \underline{0.009} & 0.079 & 0.088 & 0.141 & 0.300 & 0.200 & \textbf{0.129} & 0.200 & \underline{0.136}\\
ViT-H-14 & 88.53 & \textbf{0.001} & 0.200 & 0.115 & 0.011 & 0.087 & \underline{0.086} & 0.146 & 0.200 & 0.800 & \underline{0.142} & 0.200 & \textbf{0.139}\\
EfficientNetV2-M & 85.11 & 0.200 & 0.196 & 0.117 & 0.099 & \textbf{0.073} & \underline{0.075} & \underline{0.200} & 0.300 & \underline{0.200} & 0.300 & 0.300 & \textbf{0.138}\\
EfficientNetV2-L & 85.68 & \textbf{0.003} & 0.191 & 0.127 & \underline{0.022} & 0.052 & 0.127 & \textbf{0.117} & 0.800 & 0.200 & \underline{0.135} & 0.200 & 0.173\\
Swin-V2-B & 83.93 & \textbf{0.001} & 0.135 & 0.200 & 0.076 & 0.076 & \underline{0.050} & \textbf{0.089} & 0.800 & 0.800 & 0.300 & 0.300 & \underline{0.175}\\
\midrule
\multicolumn{14}{c}{$\alpha=0.15$}\\
\midrule
ResNet18 & 69.89 & 0.141 & 0.062 & 0.064 & 0.055 & \textbf{0.053} & \textbf{0.053} & 0.850 & 0.850 & 0.850 & \underline{0.382} & 0.850 & \textbf{0.189}\\
ResNet50 & 80.81 & \underline{0.025} & 0.146 & 0.051 & \textbf{0.013} & 0.052 & 0.111 & \textbf{0.039} & 0.150 & \underline{0.080} & 0.141 & 0.176 & 0.150\\
ResNet152 & 82.26 & \textbf{0.007} & 0.142 & 0.065 & \underline{0.053} & 0.133 & 0.108 & \textbf{0.047} & 0.350 & \underline{0.183} & 0.244 & 0.850 & \underline{0.183}\\
ViT-B-16 & 85.22 & 0.073 & 0.129 & \textbf{0.035} & 0.057 & 0.052 & \underline{0.041} & \textbf{0.073} & 0.850 & 0.350 & 0.250 & \underline{0.150} & 0.220\\
ViT-L-16 & 88.20 & 0.150 & 0.150 & 0.111 & 0.070 & \textbf{0.054} & \underline{0.061} & 0.150 & 0.150 & 0.150 & \underline{0.123} & 0.150 & \textbf{0.100}\\
ViT-H-14 & 88.53 & \textbf{0.002} & 0.150 & 0.090 & 0.068 & \underline{0.056} & \underline{0.056} & \textbf{0.071} & 0.350 & 0.850 & 0.350 & 0.150 & \underline{0.139}\\
EfficientNetV2-M & 85.11 & \underline{0.086} & 0.146 & 0.090 & \underline{0.067} & \textbf{0.056} & \underline{0.67} & \textbf{0.086} & 0.350 & \underline{0.150} & 0.267 & 0.350 & 0.158\\
EfficientNetV2-L & 85.68 & 0.084 & 0.145 & 0.090 & 0.079 & \textbf{0.047} & \underline{0.069} & \textbf{0.084} & 0.350 & 0.150 & \underline{0.133} & 0.850 & 0.150\\
Swin-V2-B & 83.93 & 0.052 & 0.117 & \underline{0.034} & 0.041 & 0.090 & \textbf{0.033} & \textbf{0.052} & 0.850 & 0.350 & 0.264 & 0.450 & \underline{0.162}\\
\midrule
\multicolumn{14}{c}{$\alpha=0.10$}\\
\midrule
ResNet18 & 69.89 & 0.218 & 0.046 & 0.044 & \textbf{0.002} & 0.042 & \underline{0.036} & 0.900 & 0.900 & 0.900 & \underline{0.357} & 0.900 & \textbf{0.241}\\
ResNet50 & 80.81 & 0.062 & 0.118 & 0.085 & \textbf{0.032} & \underline{0.037} & 0.111 & \textbf{0.100} & \textbf{0.100} & 0.290 & \underline{0.126} & 0.400 & 0.186\\
ResNet152 & 82.26 & 0.088 & 0.093 & \textbf{0.024} & \underline{0.034} & 0.039 & 0.064 & \textbf{0.150} & 0.233 & 0.398 & 0.224 & 0.471 & \underline{0.161}\\
ViT-B-16 & 85.22 & 0.100 & 0.097 & \textbf{0.015} & \underline{0.021} & 0.100 & 0.021 & \textbf{0.100} & 0.233 & 0.400 & 0.500 & 0.275 & \underline{0.186}\\
ViT-L-16 & 88.20 & 0.014 & 0.100 & \textbf{0.017} & 0.033 & 0.043 & \underline{0.030} & \textbf{0.100} & \textbf{0.100} & 0.298 & \underline{0.146} & 0.150 & \textbf{0.100}\\
ViT-H-14 & 88.53 & 0.033 & 0.097 & \textbf{0.017} & 0.031 & 0.100 & \underline{0.028} & \textbf{0.033} & 0.400 & 0.264 & 0.400 & \underline{0.100} & 0.122\\
EfficientNetV2-M & 85.11 & 0.100 & 0.095 & \textbf{0.006} & \underline{0.021} & 0.051 & 0.057 & \textbf{0.100} & \underline{0.150} & 0.355 & 0.297 & 0.900 & 0.131\\
EfficientNetV2-L & 85.68 & \underline{0.035} & 0.096 & \textbf{0.010} & 0.037 & 0.090 & 0.045 & \textbf{0.049} & 0.400 & 0.336 & 0.211 & 0.275 & \underline{0.132}\\
Swin-V2-B & 83.93 & 0.092 & 0.090 & \underline{0.025} & \textbf{0.015} & 0.059 & 0.028 & \textbf{0.107} & 0.900 & 0.400 & \underline{0.271} & 0.700 & 0.189\\
\midrule
\multicolumn{14}{c}{$\alpha=0.05$}\\
\midrule
ResNet18 & 69.89 & 0.059 & 0.027 & 0.035 & \textbf{0.000} & 0.024 & \underline{0.014} & 0.950 & 0.950 & 0.950 & \underline{0.279} & 0.950 & \textbf{0.200}\\
ResNet50 & 80.81 & 0.200 & 0.135 & 0.091 & \underline{0.089} & \textbf{0.016} & 0.118 & 0.298 & \textbf{0.050} & 0.162 & 0.258 & 0.950 & \underline{0.093}\\
ResNet152 & 82.26 & 0.228 & 0.047 & \underline{0.023} & 0.041 & \textbf{0.021} & 0.030 & 0.950 & \textbf{0.200} & \underline{0.322} & 0.329 & 0.950 & \textbf{0.200}\\
ViT-B-16 & 85.22 &  0.103 & 0.050 & \underline{0.019} & 0.021 & 0.022 & \textbf{0.018} & 0.236 & \textbf{0.061} & 0.950 & 0.350 & 0.450 & \underline{0.150}\\
ViT-L-16 & 88.20 &  0.147 & 0.049 & 0.104 & 0.030 & \textbf{0.019} & \underline{0.024} & 0.175 & \underline{0.093} & 0.950 & 0.236 & 0.200 & \textbf{0.075}\\
ViT-H-14 & 88.53 &  0.096 & 0.050 & 0.021 & 0.019 & \underline{0.018} & \textbf{0.017} & 0.283 & \textbf{0.061} & 0.950 & 0.258 & 0.950 & \underline{0.114}\\
EfficientNetV2-M & 85.11 & 0.117 & 0.047 & 0.079 & 0.061 & \textbf{0.022} & \underline{0.034} & 0.283 & \textbf{0.200} & \underline{0.206} & 0.254 & 0.950 & 0.283\\
EfficientNetV2-L & 85.68 & 0.120 & 0.048 & 0.102 & 0.046 & \underline{0.041} & \textbf{0.029} & 0.950 & 0.283 & 0.250 & \underline{0.236} & 0.950 & \textbf{0.200}\\
Swin-V2-B & 83.93 & 0.155 & 0.045 & \underline{0.017} & 0.036 & 0.028 & \textbf{0.012} & \underline{0.325} & 0.950 & 0.450 & 0.379 & 0.700 & \textbf{0.172}\\
\bottomrule
\end{tabular}
\caption{SSCV and ESCV. $\alpha\in\{0.20, 0.15,0.10,0.05\}$. Bold numbers indicate the best result, and underlined numbers indicate the second-best}
\label{tab:apdx_alg_results_sscv_escv}
\end{table}

\begin{table}[!bt]
\scriptsize
\centering
\begin{tabular}{c c cccccc| cccccc}
\toprule
\multirow{2}{*}{Base Model} & \multirow{2}{*}{Acc.} & \multicolumn{6}{c}{T-CV} & \multicolumn{6}{c}{T-SS}\\   
& & LAC & APS & RAPS & SAPS & O-LAC & O-SAPS & LAC & APS & RAPS & SAPS & O-LAC & O-SAPS \\
\midrule
\multicolumn{14}{c}{$\alpha=0.20$}\\
\midrule
ResNet18 & 69.89 & 0.375 & 0.085 & \textbf{0.100} & 0.160 & \underline{0.105} & \underline{0.105} & 0.594 & 0.850 & 0.852 & 0.814 & \textbf{0.864} & \underline{0.855}\\
ResNet50 & 80.81 & 0.590 & 0.200 & 0.190 & 0.245 & \underline{0.130} & \textbf{0.090} & -0.561 & 0.318 & 0.614 & 0.575 & \textbf{0.712} & \underline{0.687}\\
ResNet152 & 82.26 & 0.635 & 0.120 & \textbf{0.085} & \textbf{0.085} & 0.120 & \underline{0.095} & -0.565 & 0.610 & \textbf{0.630} & 0.601 & \underline{0.619} & 0.624\\
ViT-B-16 & 85.22 &  0.790 & 0.100 & \underline{0.095} & 0.115 & \underline{0.095} & \textbf{0.085} & -0.569 & \underline{0.598} & \textbf{0.619} & 0.594 & 0.576 & 0.589\\
ViT-L-16 & 88.20 &  0.790 & 0.120 & 0.095 & 0.110 & \underline{0.085} & \textbf{0.080} & -0.558 & \underline{0.624} & \textbf{0.657} & 0.611 & 0.610 & 0.617\\
ViT-H-14 & 88.53 &  0.795 & \textbf{0.080} & \textbf{0.080} & 0.135 & \underline{0.090} & \textbf{0.080} & -0.420 & 0.426 & \textbf{0.452} & 0.441 & 0.433 & \underline{0.443}\\
EfficientNetV2-M & 85.11 & 0.695 & \underline{0.090} & \textbf{0.080} & 0.095 & \underline{0.090} & \textbf{0.080} & -0.433 & 0.274 & 0.425 & 0.437 & \textbf{0.442} & \underline{0.439}\\
EfficientNetV2-L & 85.68 & 0.605 & \underline{0.090} & \textbf{0.085} & 0.160 & \underline{0.090} & \textbf{0.085} & -0.399 & 0.388 & \textbf{0.423} & 0.409 & \underline{0.417} & 0.412\\
Swin-V2-B & 83.93 &  0.740 & 0.090 & 0.095 & 0.090 & \underline{0.085} & \textbf{0.080} & -0.487 & 0.431 & 0.454 & \underline{0.483} & \textbf{0.490} & 0.487\\
\midrule
Average & - & 0.668 & 0.108 & 0.101 & 0.133 & \underline{0.099} & \textbf{0.087} & -0.378 & 0.502 & \underline{0.570} & 0.552 & \textbf{0.574} & \underline{0.573}\\
Avg. Rank from CD & - & 6.00 & 1.78 & \underline{1.33} & 3.00 & 1.67 & \textbf{1.11} & 6.00 & 3.33 & \underline{1.67} & 2.56 & \underline{1.67} & \textbf{1.44}\\
\midrule
\multicolumn{14}{c}{$\alpha=0.15$}\\
\midrule
ResNet18 & 69.89 & 0.285 & \textbf{0.080} & 0.100 & 0.180 & 0.100 & \underline{0.090} & 0.737 & \underline{0.854} & \underline{0.865} & 0.801 & \textbf{0.866} & 0.855\\
ResNet50 & 80.81 & 0.485 & 0.190 & 0.305 & 0.270 & \underline{0.110} & \textbf{0.085} & 0.059 & 0.249 & 0.341 & 0.558 & \textbf{0.711} & \underline{0.686}\\
ResNet152 & 82.26 & 0.455 & 0.100 & \textbf{0.075} & \underline{0.080} & 0.100 & \underline{0.080} & 0.172 & 0.610 & \underline{0.627} & 0.601 & \textbf{0.628} & 0.624\\
ViT-B-16 & 85.22 &  0.580 & 0.085 & \underline{0.080} & 0.095 & 0.085 & \textbf{0.070} & -0.426 & \underline{0.613} & \textbf{0.624} & 0.588 & 0.592 & 0.583\\
ViT-L-16 & 88.20 & 0.750 & 0.105 & 0.080 & 0.080 & \underline{0.075} & \textbf{0.070} & -0.605 & \underline{0.638} & \textbf{0.664} & 0.610 & 0.623 & 0.620\\
ViT-H-14 & 88.53 &  0.770 & \textbf{0.070} & \underline{0.075} & 0.080 & \underline{0.075} & \underline{0.075} & -0.418 & 0.428 & \textbf{0.456} & 0.444 & \underline{0.448} & 0.446\\
EfficientNetV2-M & 85.11 & 0.550 & \underline{0.075} & 0.080 & \textbf{0.070} & \underline{0.075} & \textbf{0.070} & -0.304 & 0.281 & 0.396 & 0.430 & \textbf{0.440} & \underline{0.433}\\
EfficientNetV2-L & 85.68 & 0.475 & \underline{0.080} & \textbf{0.075} & 0.090 & \underline{0.080} & \textbf{0.075} & -0.327 & 0.383 & 0.420 & 0.409 & \textbf{0.427} & \underline{0.421}\\
Swin-V2-B & 83.93 & 0.535 & \underline{0.080} & 0.085 & 0.080 & \underline{0.075} & \textbf{0.070} & -0.001 & 0.430 & 0.476 & \underline{0.478} & \textbf{0.489} & \textbf{0.489}\\
\midrule
Average & -  & 0.543 & 0.096 & 0.106 & 0.114 & \underline{0.086} & \textbf{0.076} & -0.124 & 0.498 & 0.541 & 0.547 & \textbf{0.580} & \underline{0.573}\\
Avg. Rank from CD & - & 6.00 & 2.44 & 2.11 & 2.78 & \underline{2.00} & \textbf{1.11} & 6.00& 4.00 & \underline{1.89} & 3.33 & \textbf{1.33} & \underline{1.89}\\
\midrule
\multicolumn{14}{c}{$\alpha=0.10$}\\
\midrule
ResNet18 & 69.89 & 0.215 & \textbf{0.060} & 0.095 & 0.275 & 0.090 & \underline{0.075} & 0.813 & \underline{0.860} & 0.856 & 0.771 & \textbf{0.865} & 0.853\\
ResNet50 & 80.81 & 0.335 & 0.165 & 0.240 & 0.305 & \underline{0.080} & \textbf{0.075} & 0.436 & 0.171 & 0.534 & 0.534 & \textbf{0.715} & \underline{0.693}\\
ResNet152 & 82.26 & 0.270 & \underline{0.070} & 0.285 & 0.280 & \underline{0.070} & \textbf{0.065} & 0.554 & 0.593 & 0.491 & 0.585 & \textbf{0.637} & \underline{0.621}\\
ViT-B-16 & 85.22 &  0.305 & 0.075 & \underline{0.065} & 0.070 & 0.070 & \textbf{0.060} & 0.499 & \textbf{0.621} & \textbf{0.621} & 0.577 & \underline{0.608} & 0.581\\
ViT-L-16 & 88.20 &  0.500 & \underline{0.080} & 0.200 & \textbf{0.060} & \textbf{0.060} & \textbf{0.060} & 0.209 & \textbf{0.650} & 0.442 & 0.604 & \underline{0.639} & 0.619\\
ViT-H-14 & 88.53 &  0.455 & \textbf{0.055} & 0.170 & 0.065 & 0.070 & \underline{0.060} & 0.200 & 0.418 & 0.354 & 0.444 & \textbf{0.450} & \underline{0.447}\\
EfficientNetV2-M & 85.11 & 0.270 & \textbf{0.060} & 0.240 & 0.265 & \underline{0.065} & \textbf{0.060} & 0.385 & 0.282 & 0.263 & 0.399 & \textbf{0.443} & \underline{0.423}\\
EfficientNetV2-L & 85.68 & 0.330 & \textbf{0.065} & 0.215 & \underline{0.070} & \textbf{0.065} & \textbf{0.065} & 0.378 & 0.366 & 0.369 & 0.411 & \textbf{0.429} & \underline{0.428}\\
Swin-V2-B & 83.93 &  0.260 & \underline{0.070} & 0.075 & 0.095 & \textbf{0.060} & \textbf{0.060} & 0.440 & 0.452 & 0.461 & 0.469 & \textbf{0.494} & \underline{0.484}\\
\midrule
Average & -  & 0.327 & 0.078 & 0.176 & 0.165 & \underline{0.069} & \textbf{0.064} & 0.435 & 0.490 & 0.488 & 0.533 & \textbf{0.587} & \underline{0.572}\\
Avg. Rank from CD & - & 5.44 & 2.00 & 4.11 & 3.33 & \underline{1.44} & \textbf{1.11} & 5.11 & 3.44 & 3.67 & 3.22 & \textbf{1.00} & \underline{1.78}\\
\midrule
\multicolumn{14}{c}{$\alpha=0.05$}\\
\midrule
ResNet18 & 69.89 & 0.140 & \textbf{0.045} & 0.065 & 0.200 & \underline{0.050} & \textbf{0.045} & 0.854 & \underline{0.864} & 0.857 & 0.774 & \textbf{0.871} & 0.852\\
ResNet50 & 80.81 & 0.180 & 0.125 & 0.185 & 0.115 & \textbf{0.045} & \textbf{0.050} & 0.591 & 0.077 & 0.413 & 0.548 & \textbf{0.711} & \underline{0.692}\\
ResNet152 & 82.26 & 0.125 & \underline{0.050} & 0.175 & 0.125 & \textbf{0.045} & \textbf{0.045} & \underline{0.621} & 0.544 & 0.534 & 0.593 & \textbf{0.640} & 0.619\\
ViT-B-16 & 85.22 &  0.120 & \underline{0.050} & \underline{0.050} & 0.065 & \textbf{0.045} & \textbf{0.045} & 0.542 & \textbf{0.620} & 0.592 & 0.563 & \underline{0.611} & 0.580\\
ViT-L-16 & 88.20 &  0.180 & \underline{0.050} & 0.105 & 0.070 & \textbf{0.040} & \textbf{0.040} & 0.544 & \underline{0.640} & 0.564 & 0.587 & \textbf{0.659} & 0.623\\
ViT-H-14 & 88.53 &  0.130 & \textbf{0.040} & \textbf{0.040} & 0.060 & \underline{0.045} & \textbf{0.040} & 0.424 & 0.372 & 0.415 & 0.438 & \underline{0.441} & \textbf{0.443}\\
EfficientNetV2-M & 85.11 & 0.120 & \underline{0.045} & 0.155 & 0.125 & \underline{0.045} & \textbf{0.040} & 0.406 & 0.277 & 0.357 & 0.411 & \textbf{0.443} & \underline{0.422}\\
EfficientNetV2-L & 85.68 & 0.120 & \textbf{0.045} & 0.125 & \underline{0.090} & \textbf{0.045} & \textbf{0.045} & 0.416 & 0.349 & 0.374 & 0.411 & \textbf{0.427} & \underline{0.421}\\
Swin-V2-B & 83.93 & 0.125 & 0.055 & 0.070 & 0.065 & \textbf{0.040} & \underline{0.045} & 0.468 & 0.457 & 0.459 & 0.469 & \textbf{0.500} & \underline{0.487}\\
\midrule
Average & -  & 0.138 & \underline{0.056} & 0.108 & 0.102 & \textbf{0.044} & \textbf{0.044} & 0.541 & 0.467 & 0.507 & 0.533 & \textbf{0.589} & \underline{0.571}\\
Avg. Rank from CD & - & 5.00 & \underline{1.89} & 4.33 & 4.33 & \textbf{1.11} & \textbf{1.11} & 3.44 & 4.22 & 4.11 & 3.56 & \textbf{1.00} & \underline{1.89ß}\\
\bottomrule
\end{tabular}
\caption{Our Metrics (T-CV, T-SS). $\alpha \in \{0.20,0.15,0.10,0.05\}$. $B=100$. Bold values denote the best case, while the underlined values represent the second-best. LAC mostly shows the negative T-SS values where the base classifier already satisfies the target coverage rate.}
\label{tab:apdx_alg_results_our_metrics_B100}
\end{table}

\begin{table}[!bt]
\scriptsize
\centering
\begin{tabular}{c c cccccc| cccccc}
\toprule
\multirow{2}{*}{Base Model} & \multirow{2}{*}{Acc.} & \multicolumn{6}{c}{T-CV} & \multicolumn{6}{c}{T-SS}\\   
& & LAC & APS & RAPS & SAPS & O-LAC & O-SAPS & LAC & APS & RAPS & SAPS & O-LAC & O-SAPS \\
\midrule
\multicolumn{14}{c}{$\alpha=0.20$}\\
\midrule
ResNet18 & 69.89 & 0.338 & \textbf{0.037} & \underline{0.044} & 0.116 & 0.064 & 0.057 & 0.670 & \underline{0.953} & \textbf{0.957} & 0.910 & 0.952 & 0.952\\
ResNet50 & 80.81 & 0.566 & 0.163 & 0.149 & 0.213 & \underline{0.103} & \textbf{0.058} & -0.705 & 0.428 & 0.774 & 0.721 & \textbf{0.892} & \underline{0.859}\\
ResNet152 & 82.26 & 0.612 & 0.092 & \textbf{0.047} & \underline{0.056} & 0.082 & 0.068 & -0.753 & \textbf{0.849} & \textbf{0.849} & 0.805 & 0.817 & \underline{0.834}\\
ViT-B-16 & 85.22 &  0.764 & 0.060 & \textbf{0.053} & 0.100 & \underline{0.055} & \textbf{0.052} & -0.808 & \underline{0.879} & \textbf{0.886} & 0.843 & 0.837 & 0.836t\\
ViT-L-16 & 88.20 & 0.780 & 0.096 & 0.066 & 0.083 & \underline{0.053} & \textbf{0.043} & -0.781 & \underline{0.872} & \textbf{0.882} & 0.840 & 0.841 & 0.839\\
ViT-H-14 & 88.53 & 0.779 & 0.045 & \textbf{0.041} & 0.094 & 0.049 & \underline{0.043} & -0.716 & 0.745 & \textbf{0.753} & 0.747 & 0.736 & \underline{0.749}\\
EfficientNetV2-M & 85.11 & 0.675 & 0.068 & \textbf{0.044} & \underline{0.052} & \underline{0.052} & \textbf{0.044} & -0.721 & 0.492 & 0.705 & 0.716 & \textbf{0.757} & \underline{0.720}\\
EfficientNetV2-L & 85.68 & 0.576 & 0.052 & \textbf{0.042} & 0.103 & 0.052 & \underline{0.043} & -0.698 & \textbf{0.774} & \underline{0.764} & 0.725 & 0.716 & 0.732\\
Swin-V2-B & 83.93 &  0.695 & 0.054 & 0.059 & 0.059 & \underline{0.053} & \textbf{0.050} & -0.823 & 0.814 & \underline{0.819} & 0.816 & \textbf{0.823} & \textbf{0.823}\\
\midrule
Average & - & 0.697 & 0.067 & \underline{0.050} & 0.078 & 0.057 & \textbf{0.049} & -0.757 & 0.775 & \textbf{0.808} & 0.785 & \underline{0.790} & \underline{0.790}\\
Avg. Rank from CD & - & 6.00 & 2.89 & 1.89 & 4.00 & \underline{1.78} & \textbf{1.33} & 6.00 & 2.44 & \textbf{1.33} & 3.33 & 2.00 & \underline{1.89}\\
\midrule
\multicolumn{14}{c}{$\alpha=0.15$}\\
\midrule
ResNet18 & 69.89 & 0.251 & \textbf{0.035} & \underline{0.043} & 0.141 & 0.053 & 0.051 & 0.826 & \underline{0.958} & \textbf{0.963} & 0.892 & 0.955 & 0.951\\
ResNet50 & 80.81 & 0.446 & 0.151 & 0.276 & 0.240 & \underline{0.089} & \textbf{0.051} & 0.082 & 0.337 & 0.434 & 0.702 & \textbf{0.895} & \underline{0.865}\\
ResNet152 & 82.26 & 0.425 & 0.084 & \textbf{0.044} & \underline{0.046} & 0.067 & 0.059 & 0.283 & \underline{0.838} & \textbf{0.844} & 0.802 & 0.829 & 0.835\\
ViT-B-16 & 85.22 & 0.547 & 0.051 & \textbf{0.045} & 0.077 & 0.050 & \underline{0.048} & -0.611 & \underline{0.880} & \textbf{0.886} & 0.835 & 0.852 & 0.833\\
ViT-L-16 & 88.20 & 0.715 & 0.078 & 0.054 & 0.051 & \underline{0.044} & \textbf{0.037} & -0.837 & \underline{0.881} & \textbf{0.888} & 0.843 & 0.848 & 0.842\\
ViT-H-14 & 88.53 & 0.721 & 0.043 & \underline{0.042} & 0.059 & 0.045 & \textbf{0.039} & -0.707 & 0.745 & \textbf{0.759} & \underline{0.749} & 0.743 & \underline{0.749}\\
EfficientNetV2-M & 85.11 & 0.526 & 0.054 & 0.043 & \textbf{0.036} & 0.044 & \underline{0.038} & -0.512 & 0.486 & 0.672 & 0.710 & \textbf{0.751} & \underline{0.18}\\
EfficientNetV2-L & 85.68 & 0.433 & \underline{0.049} & \textbf{0.039} & 0.051 & 0.052 & \underline{0.039} & -0.641 & \underline{0.756} & \textbf{0.760} & 0.722 & 0.727 & 0.733\\
Swin-V2-B & 83.93 & 0.497 & 0.047 & 0.049 & 0.048 & \underline{0.045} & \textbf{0.042} & -0.008 & \textbf{0.834} & \underline{0.832} & 0.806 & 0.821 & 0.823\\
\midrule
Average & -  & 0.507 & 0.066 & 0.071 & 0.083 & \underline{0.054} & \textbf{0.042} & -0.236 & 0.746 & 0.782 & 0.785 & \textbf{0.825} & \underline{0.817}\\
Avg. Rank from CD & - & 6.00 & 2.78 & 2.00 & 3.11 & \underline{1.89} & \textbf{1.33} & 6.00 & 2.44 & \textbf{1.67} & 3.56 & \underline{2.11} & 2.44\\
\midrule
\multicolumn{14}{c}{$\alpha=0.10$}\\
\midrule
ResNet18 & 69.89 & 0.173 & \textbf{0.029} & 0.042 & 0.236 & \underline{0.039} & 0.041 & 0.902 & \textbf{0.962} & \underline{0.960} & 0.865 & 0.958 & 0.948\\
ResNet50 & 80.81 & 0.304 & 0.126 & 0.199 & 0.275 & \underline{0.058} & \textbf{0.041} & 0.553 & 0.231 & 0.666 & 0.671 & \textbf{0.901} & \underline{0.872}\\
ResNet152 & 82.26 & 0.234 & 0.056 & 0.262 & 0.251 & \underline{0.044} & \textbf{0.041} & 0.733 & 0.813 & 0.658 & 0.788 & \textbf{0.835} & \underline{0.832}\\
ViT-B-16 & 85.22 & 0.286 & 0.053 & \textbf{0.037} & 0.047 & 0.041 & \underline{0.039} & 0.710 & \underline{0.877} & \textbf{0.880} & 0.824 & 0.869 & 0.838\\
ViT-L-16 & 88.20 & 0.408 & 0.063 & 0.164 & \textbf{0.031} & \textbf{0.031} & \underline{0.032} & 0.340 & \textbf{0.886} & 0.608 & 0.829 & \underline{0.862} & 0.851\\
ViT-H-14 & 88.53 & 0.402 & \underline{0.035} & 0.143 & 0.047 & \textbf{0.034} & \textbf{0.034} & 0.446 & 0.726 & 0.604 & \underline{0.749} & \textbf{0.755}& 0.747\\
EfficientNet-V2-M & 85.11 & 0.251 & 0.042 & 0.218 & 0.239 & \underline{0.037} & \textbf{0.032} & 0.632 & 0.480 & 0.438 & 0.672 & \textbf{0.740} & \underline{0.708}\\
EfficientNet-V2-L & 85.68 & 0.263 & \underline{0.041} & 0.164 & 0.052 & \underline{0.041} & \textbf{0.035} & 0.662 & 0.718 & 0.636 & 0.724 & \underline{0.733} & \textbf{0.740}\\
Swin-V2-B & 83.93 & 0.239 & 0.044 & 0.046 & 0.071 & \textbf{0.033} & \underline{0.034} & 0.738 & \textbf{0.852} & \underline{0.829} & 0.787 & 0.823 & 0.823\\
\midrule
Average & -  & 0.284 & 0.054 & 0.142 & 0.139 & \underline{0.040} & \textbf{0.037} & 0.635 & 0.727 & 0.698 & 0.768 & \textbf{0.831} & \underline{0.818}\\
Avg. Rank from CD & - & 5.56 & \underline{2.44} & 3.89 & 4.11 & \textbf{1.11} & \textbf{1.11} & 5.33 & 2.56 & 3.89 & 3.44 & \textbf{1.22} & \underline{1.67}\\
\midrule
\multicolumn{14}{c}{$\alpha=0.05$}\\
\midrule
ResNet18 & 69.89 & 0.086 & \textbf{0.024} & 0.032 & 0.152 & \underline{0.025} & \underline{0.025} & 0.946 & \textbf{0.966} & 0.955 & 0.864 & \underline{0.964} & 0.946\\
ResNet50 & 80.81 & 0.139 & \underline{0.088} & 0.151 & 0.090 & \textbf{0.029} & \textbf{0.029} & 0.731 & 0.107 & 0.516 & 0.683 & \textbf{0.896} & \underline{0.871}\\
ResNet152 & 82.26 & 0.100 & 0.027 & 0.144 & 0.096 & \textbf{0.022} & \underline{0.029} & 0.818 & 0.750 & 0.739 & 0.791 & \textbf{0.847} & \underline{0.829}\\
ViT-B-16 & 85.22 & 0.112 & 0.038 & \underline{0.023} & 0.052 & \textbf{0.022} & 0.025 & 0.773 & \underline{0.871} & 0.839 & 0.799 & \textbf{0.873} & 0.827\\
ViT-L-16 & 88.20 & 0.126 & 0.037 & 0.093 & 0.054 & \textbf{0.023} & \underline{0.026} & 0.754 & \underline{0.874} & 0.780 & 0.810 & \textbf{0.883} & 0.853\\
ViT-H-14 & 88.53 & 0.094 & \underline{0.026} & \textbf{0.023} & 0.041 & \textbf{0.023} & \textbf{0.023} & 0.719 & 0.644 & 0.691 & \underline{0.739} & \underline{0.739} & \textbf{0.743}\\
EfficientNetV2-M & 85.11 & 0.099 & \underline{0.026} & 0.142 & 0.111 & \underline{0.026} & \textbf{0.022} & 0.675 & 0.473 & 0.607 & 0.670 & \textbf{0.769} & \underline{0.712}\\
EfficientNetV2-L & 85.68 & 0.106 & \textbf{0.026} & 0.109 & 0.076 & \underline{0.027} & \textbf{0.026} & 0.708 & 0.671 & 0.660 & 0.720 & \textbf{0.736} & \underline{0.734}\\
Swin-V2-B & 83.93 & 0.103 & 0.032 & 0.046 & 0.043 & \textbf{0.023} & \underline{0.025} & 0.788 & \underline{0.829} & 0.809 & 0.789 & \textbf{0.841} & 0.819\\
\midrule
Average & - & 0.107 & 0.036 & 0.085 & 0.079 & \textbf{0.024} & \underline{0.026} & 0.768 & 0.687 & 0.733 & 0.763 & \textbf{0.835} & \underline{0.815}\\
Avg. Rank from CD & - & 4.78 & \underline{2.22} & 4.11 & 4.44 & \textbf{1.11} & \textbf{1.11} & 3.78 & 3.67 & 4.33 & 3.89 & \textbf{1.00} & \underline{1.67}\\
\bottomrule
\end{tabular}
\caption{Our Metrics (T-CV, T-SS). $\alpha \in \{0.20,0.15,0.10,0.05\}$. $B=30$. Bold values denote the best case, while the underlined values represent the second-best. LAC mostly shows the negative T-SS values where the base classifier already satisfies the target coverage rate.}
\label{tab:apdx_alg_results_our_metrics_B30}
\end{table}

\FloatBarrier

\subsection{Image Classification Experiments - Critical Difference Diagrams}
In this section, we present the critical difference diagrams discussed in Section \ref{sec:algorithm}.
Additionally, we provide the critical difference diagrams for different values of $B\in \{30,100\}$.
From these diagrams, we compute the average rank of each algorithm and include the results accordingly.

\begin{figure}[!hb]
    \centering
    \begin{subfigure}[b]{0.45\textwidth}
        \includegraphics[width=\textwidth]{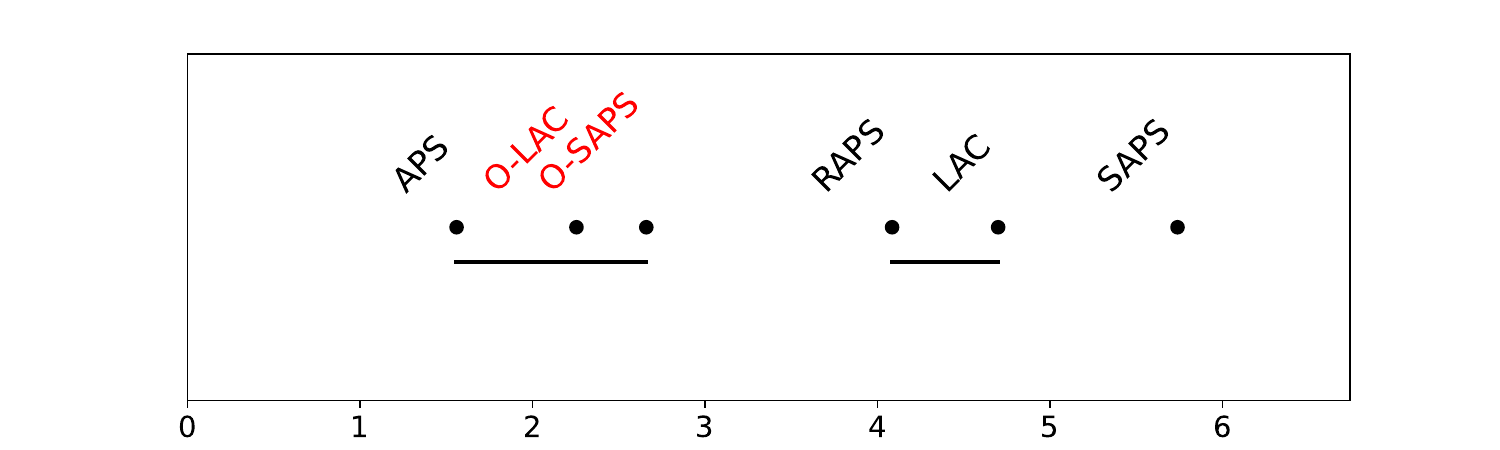}
        \caption{ResNet18}
    \end{subfigure}
    \begin{subfigure}[b]{0.45\textwidth}
        \includegraphics[width=\textwidth]{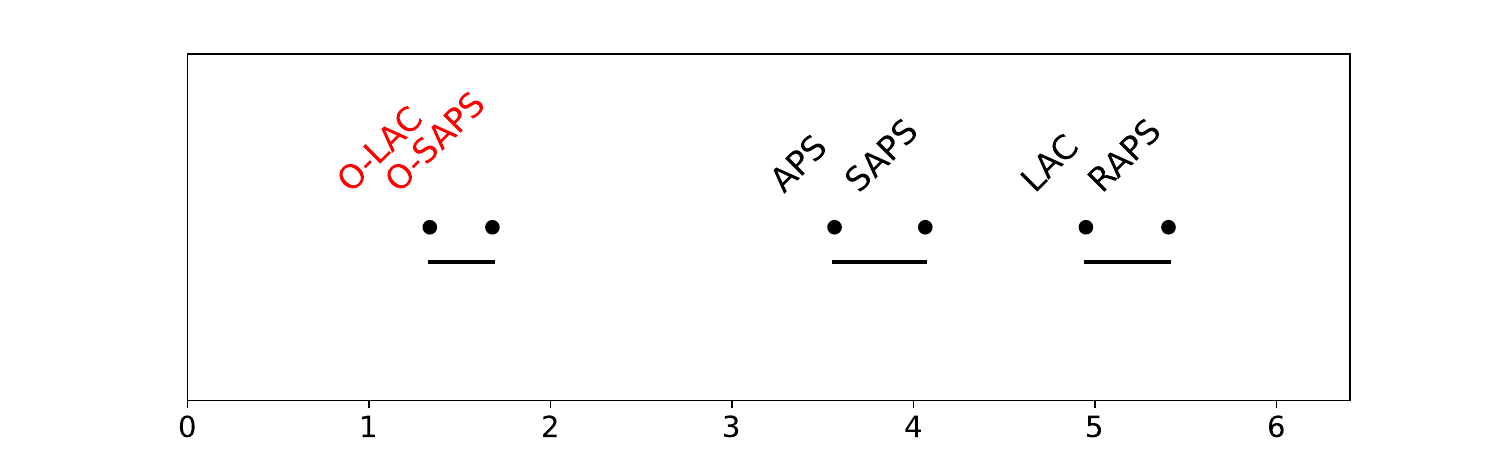}
        \caption{ResNet50}
    \end{subfigure}
    \begin{subfigure}[b]{0.45\textwidth}
        \includegraphics[width=\textwidth]{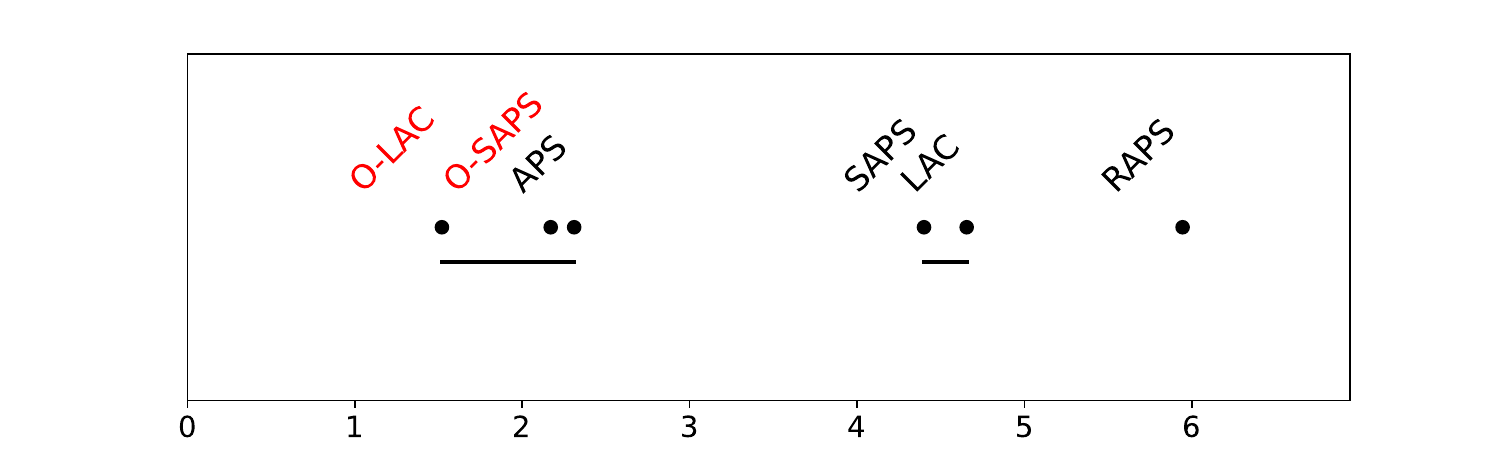}
        \caption{ResNet152}
    \end{subfigure}
    \begin{subfigure}[b]{0.45\textwidth}
        \includegraphics[width=\textwidth]{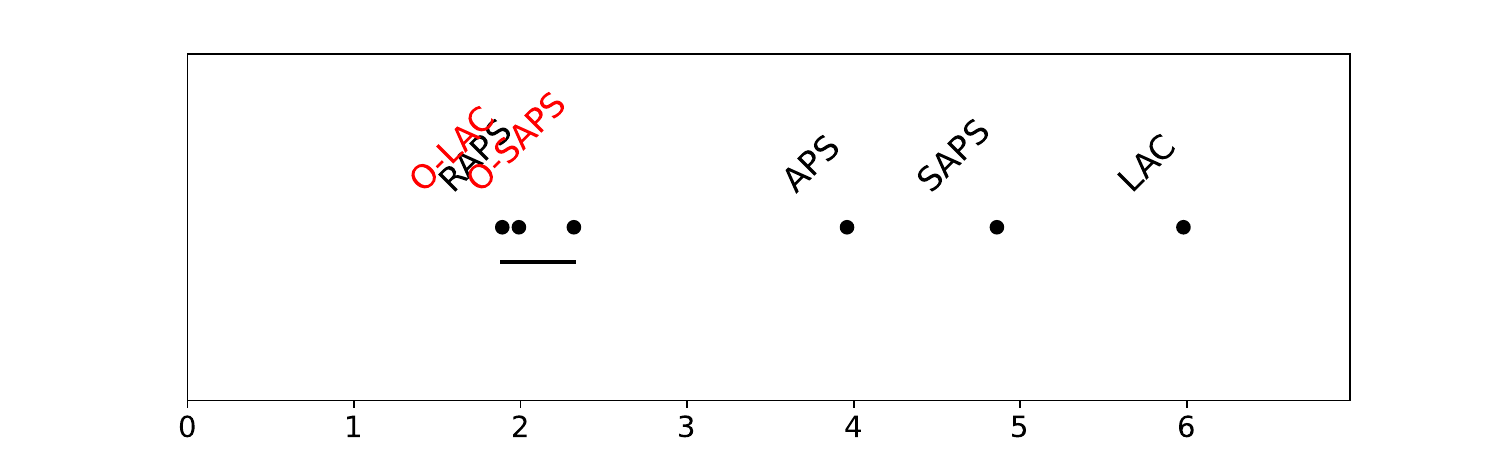}
        \caption{ViT-B-16}
    \end{subfigure}
    \begin{subfigure}[b]{0.45\textwidth}
        \includegraphics[width=\textwidth]{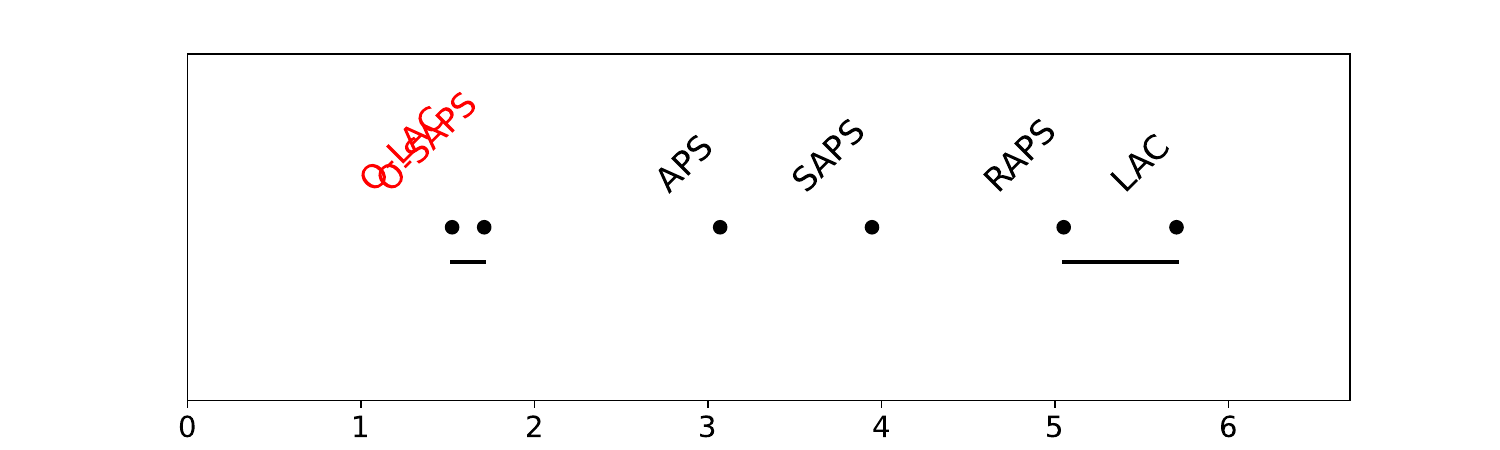}
        \caption{ViT-L-16}
    \end{subfigure}
    \begin{subfigure}[b]{0.45\textwidth}
        \includegraphics[width=\textwidth]{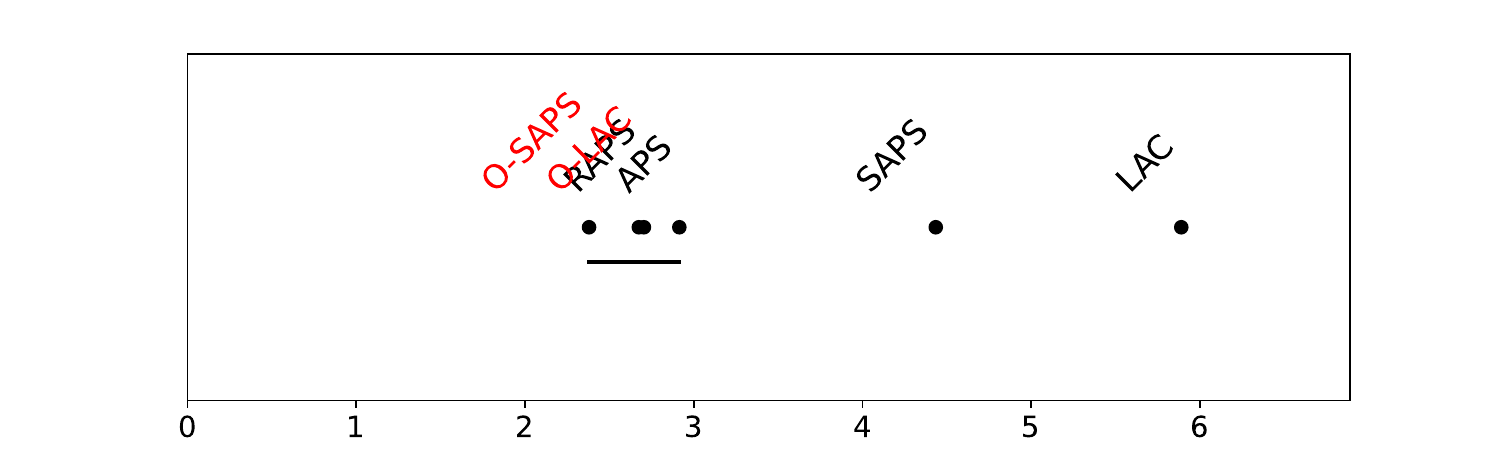}
        \caption{ViT-H-14}
    \end{subfigure}
    \begin{subfigure}[b]{0.45\textwidth}
        \includegraphics[width=\textwidth]{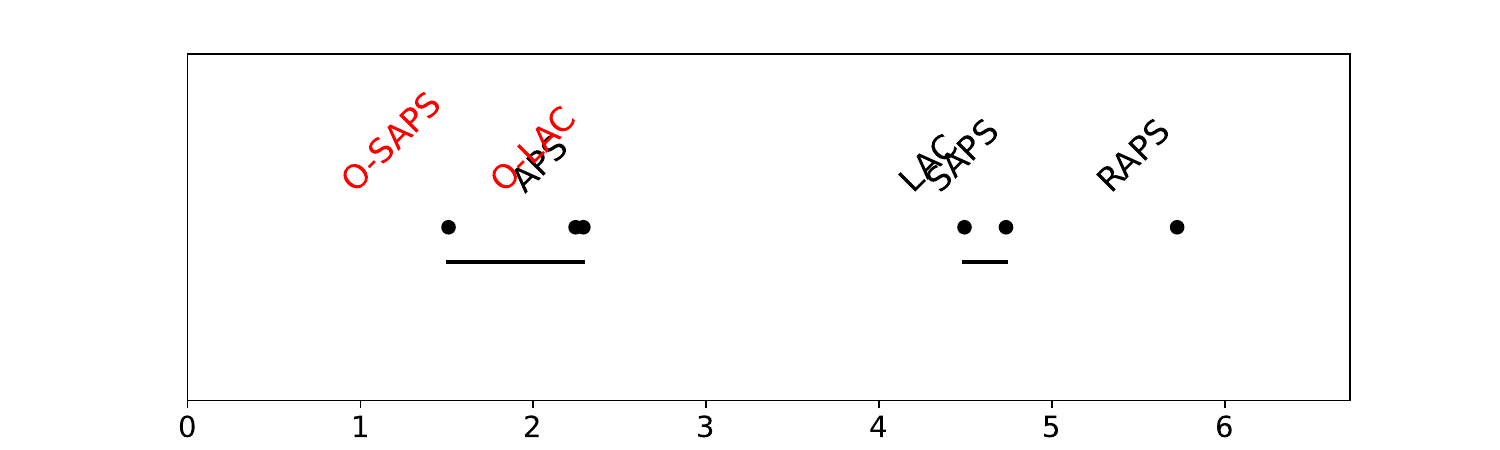}
        \caption{EfficientNet-V2-M}
    \end{subfigure}
    \begin{subfigure}[b]{0.45\textwidth}
        \includegraphics[width=\textwidth]{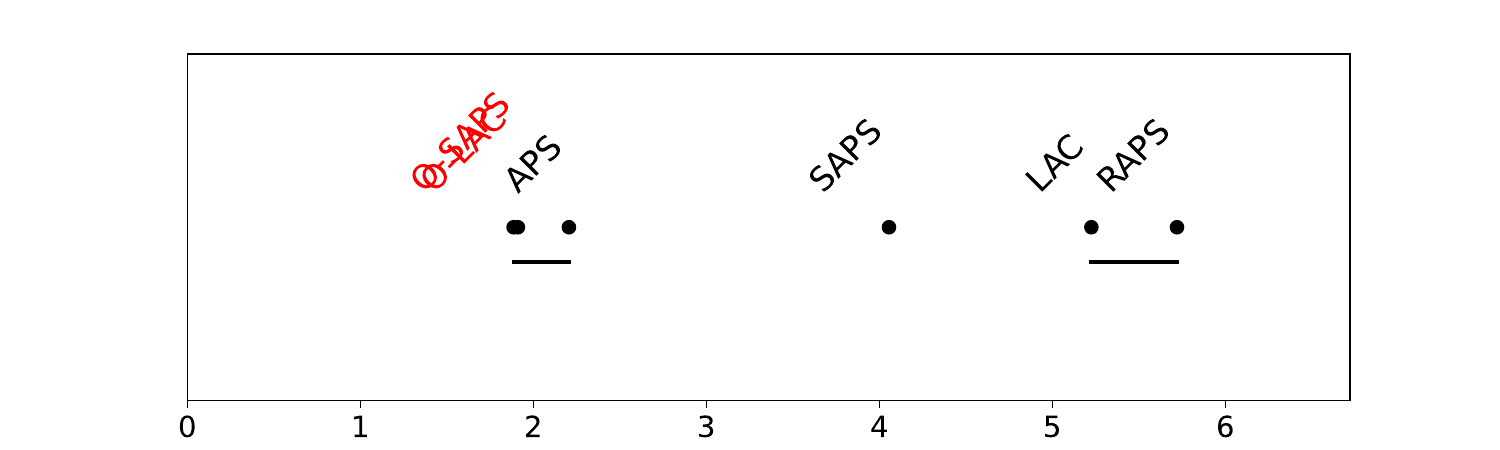}
        \caption{EfficientNet-V2-L}
    \end{subfigure}
    \begin{subfigure}[b]{0.45\textwidth}
        \includegraphics[width=\textwidth]{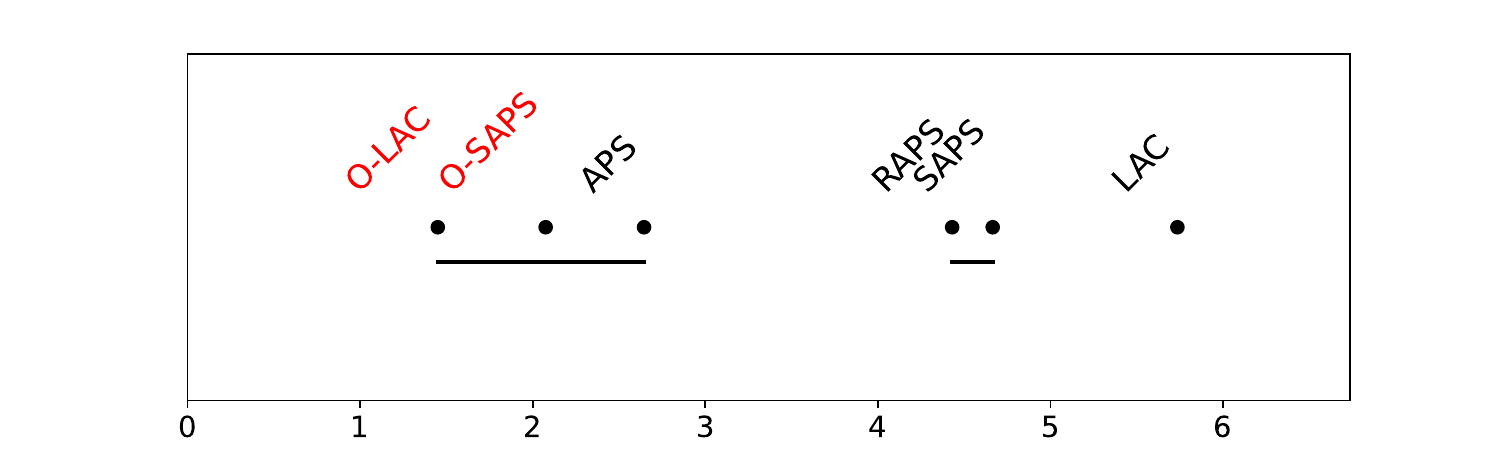}
        \caption{Swin-V2-B}
    \end{subfigure}
    \caption{Critical Difference Diagrams. T-CV, $\alpha=0.05, B=50$. The rank analysis based on these figures is summarized as `Avg. Rank from CD' in Table~\ref{tab:alg_results_our_metrics} in the main text.}
    \label{fig:apdx_cd_tcv_alpha0.05}
\end{figure}

\begin{figure}[!bt]
    \centering
    \begin{subfigure}[b]{0.45\textwidth}
        \includegraphics[width=\textwidth]{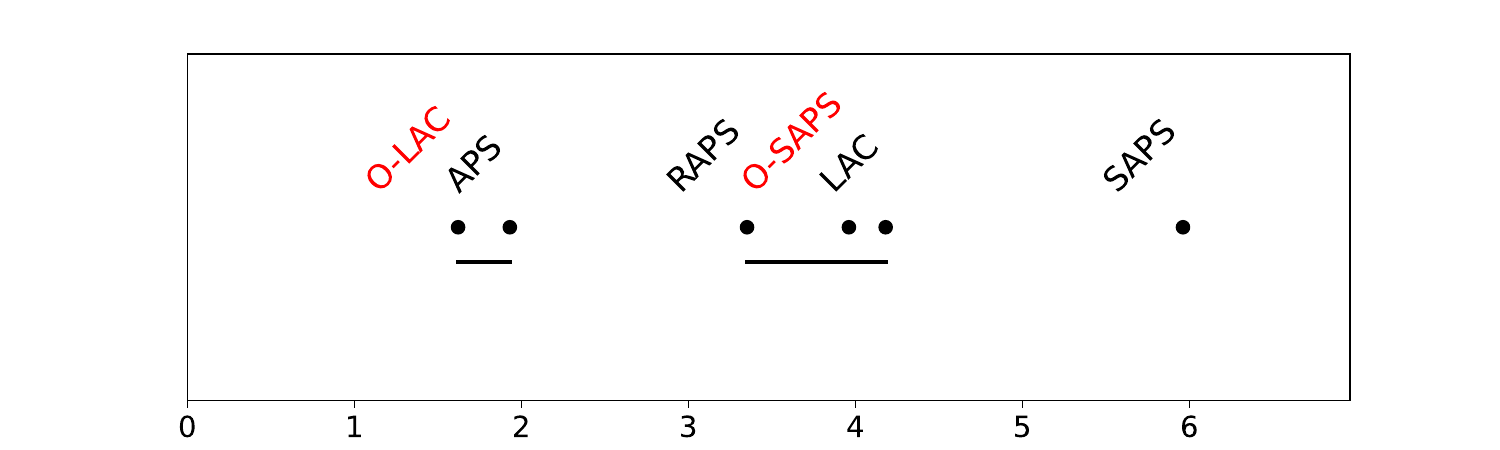}
        \caption{ResNet18}
    \end{subfigure}
    \begin{subfigure}[b]{0.45\textwidth}
        \includegraphics[width=\textwidth]{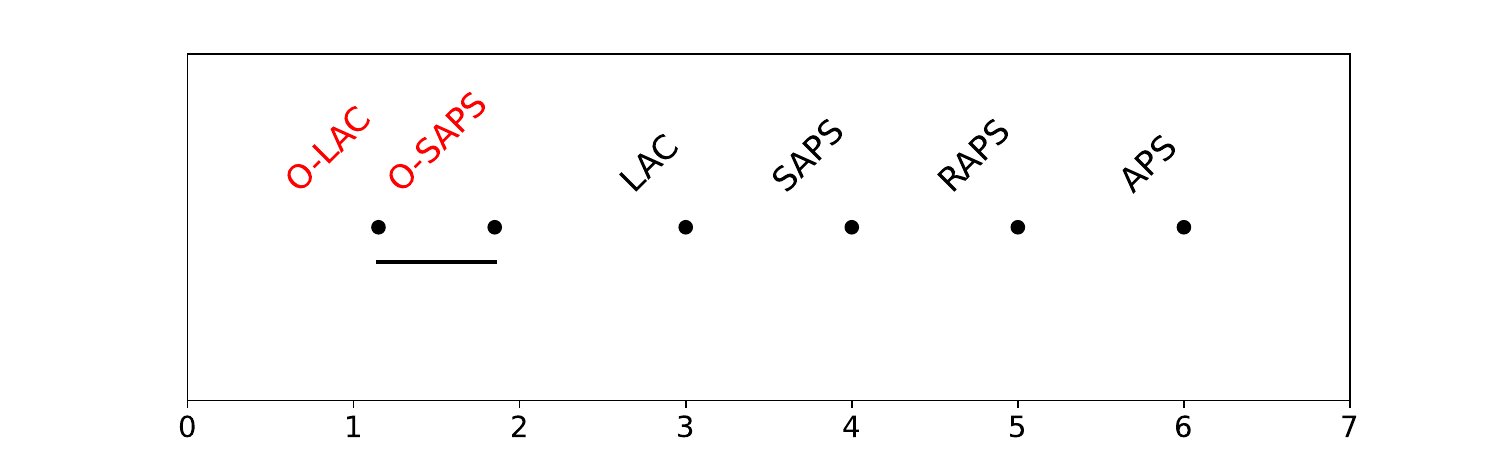}
        \caption{ResNet50}
    \end{subfigure}
    \begin{subfigure}[b]{0.45\textwidth}
        \includegraphics[width=\textwidth]{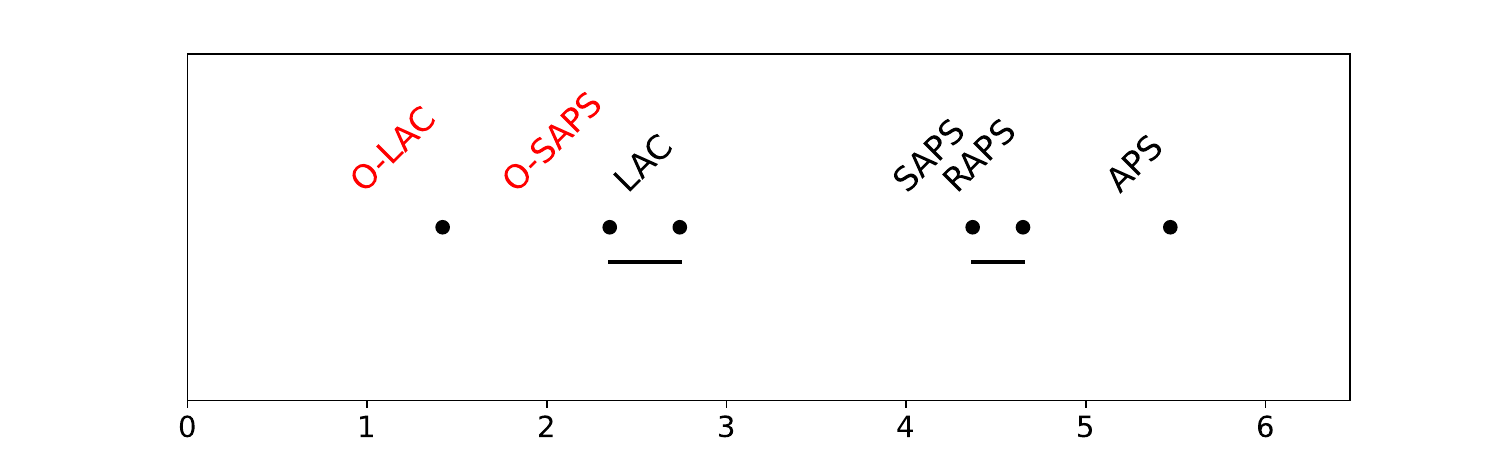}
        \caption{ResNet152}
    \end{subfigure}
    \begin{subfigure}[b]{0.45\textwidth}
        \includegraphics[width=\textwidth]{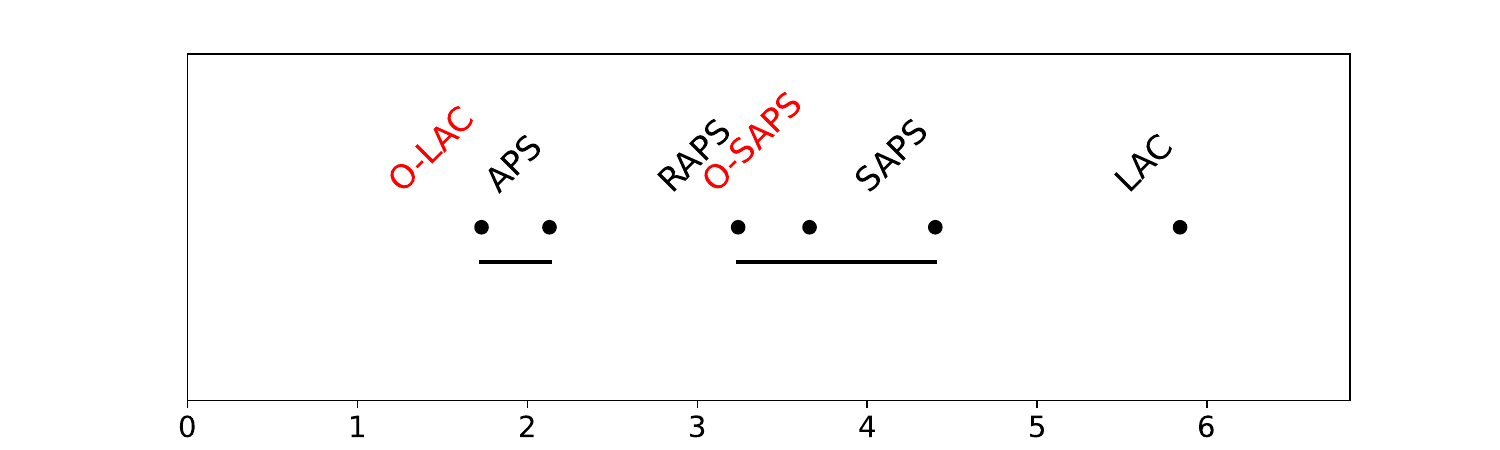}
        \caption{ViT-B-16}
    \end{subfigure}
    \begin{subfigure}[b]{0.45\textwidth}
        \includegraphics[width=\textwidth]{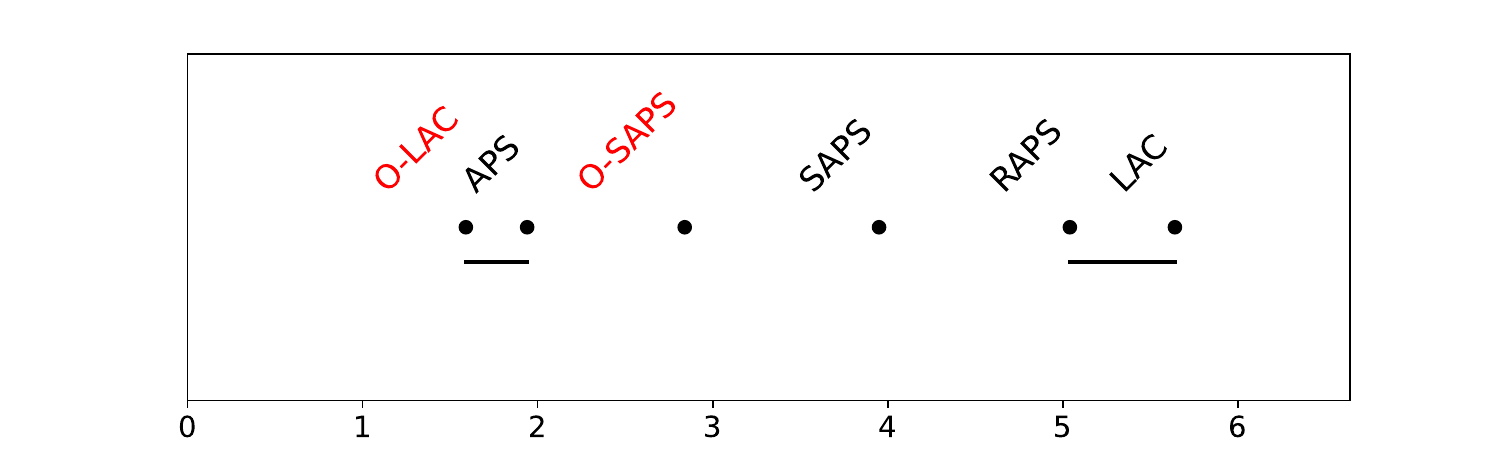}
        \caption{ViT-L-16}
    \end{subfigure}
    \begin{subfigure}[b]{0.45\textwidth}
        \includegraphics[width=\textwidth]{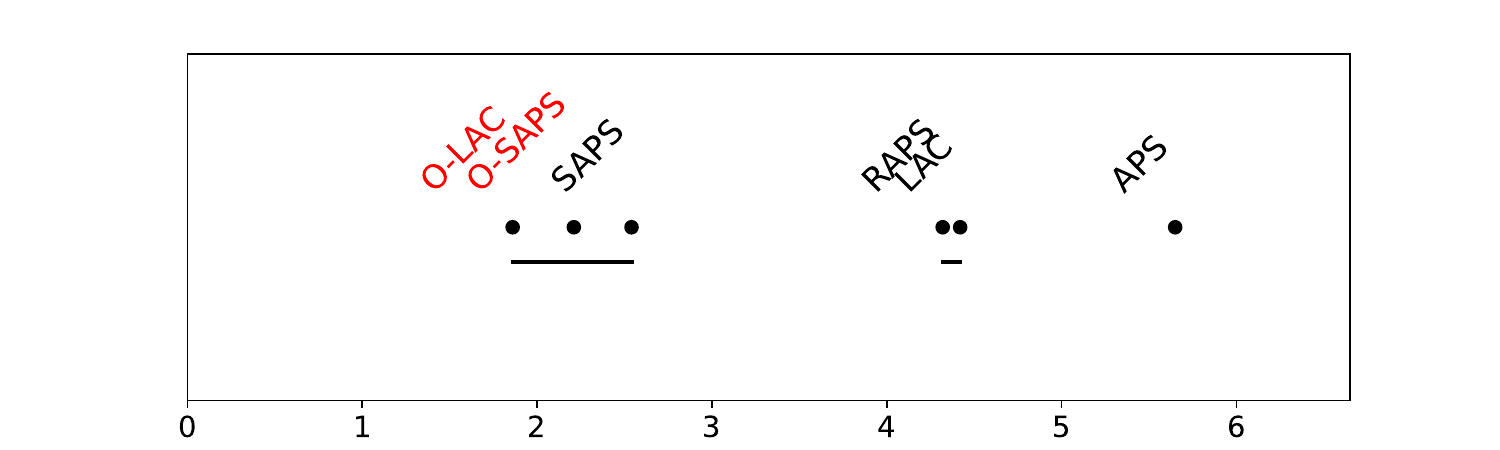}
        \caption{ViT-H-14}
    \end{subfigure}
    \begin{subfigure}[b]{0.45\textwidth}
        \includegraphics[width=\textwidth]{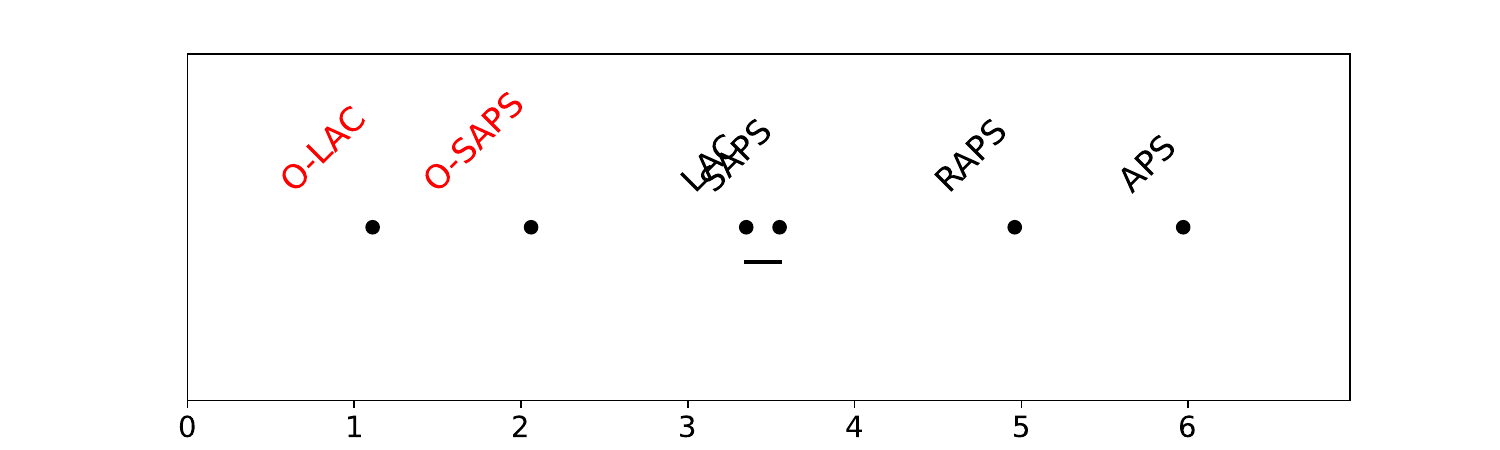}
        \caption{EfficientNet-V2-M}
    \end{subfigure}
    \begin{subfigure}[b]{0.45\textwidth}
        \includegraphics[width=\textwidth]{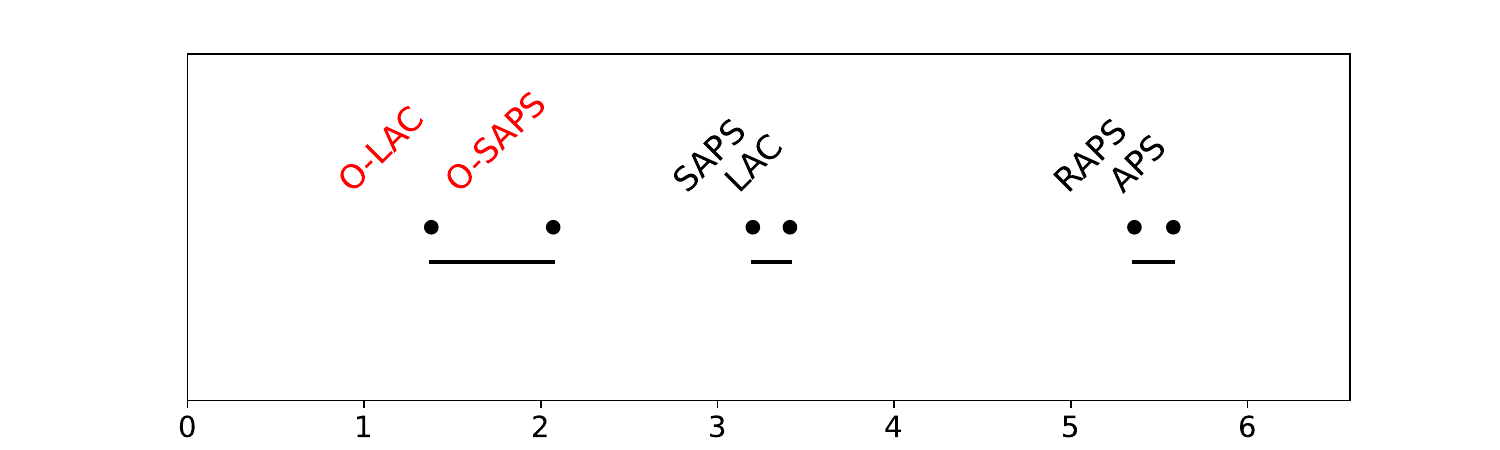}
        \caption{EfficientNet-V2-L}
    \end{subfigure}
    \begin{subfigure}[b]{0.45\textwidth}
        \includegraphics[width=\textwidth]{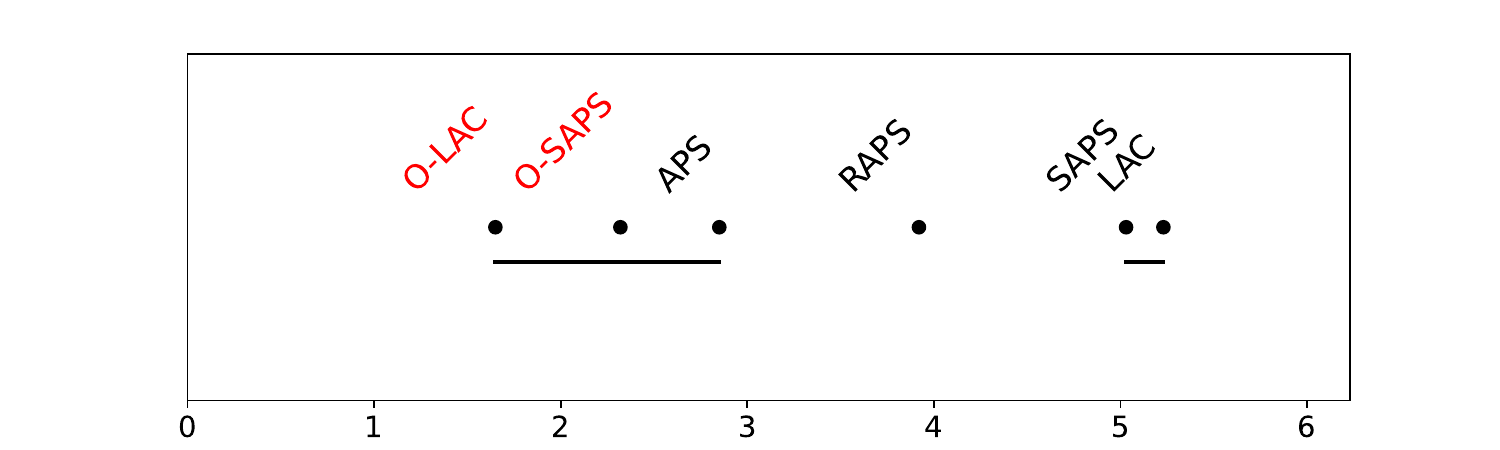}
        \caption{Swin-V2-B}
    \end{subfigure}
    \caption{Critical Difference Diagrams. T-SS, $\alpha=0.05, B=50$. The rank analysis based on these figures is summarized as `Avg. Rank from CD' in Table~\ref{tab:alg_results_our_metrics} in the main text.}
    \label{fig:apdx_cd_tss_alpha0.05}
\end{figure}

\begin{figure}[!bt]
    \centering
    \begin{subfigure}[b]{0.45\textwidth}
        \includegraphics[width=\textwidth]{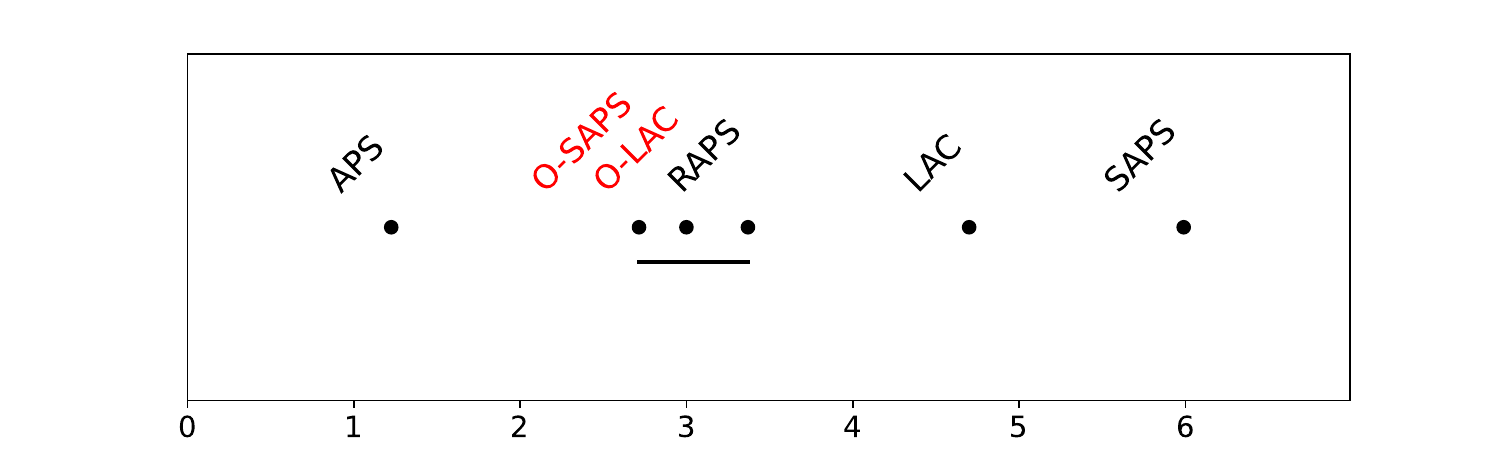}
        \caption{ResNet18}
    \end{subfigure}
    \begin{subfigure}[b]{0.45\textwidth}
        \includegraphics[width=\textwidth]{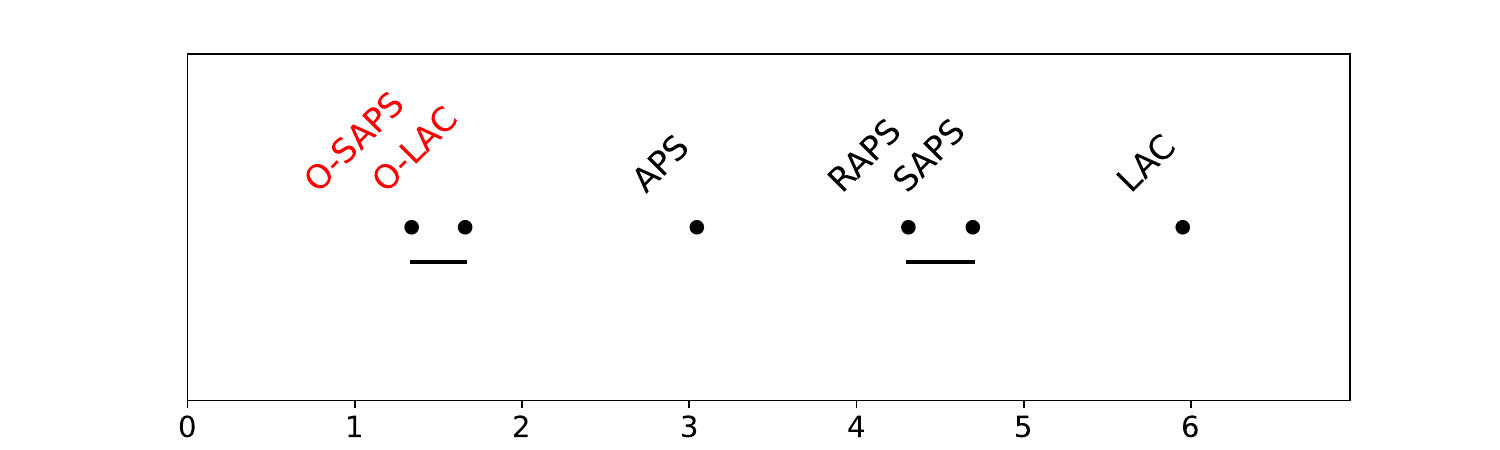}
        \caption{ResNet50}
    \end{subfigure}
    \begin{subfigure}[b]{0.45\textwidth}
        \includegraphics[width=\textwidth]{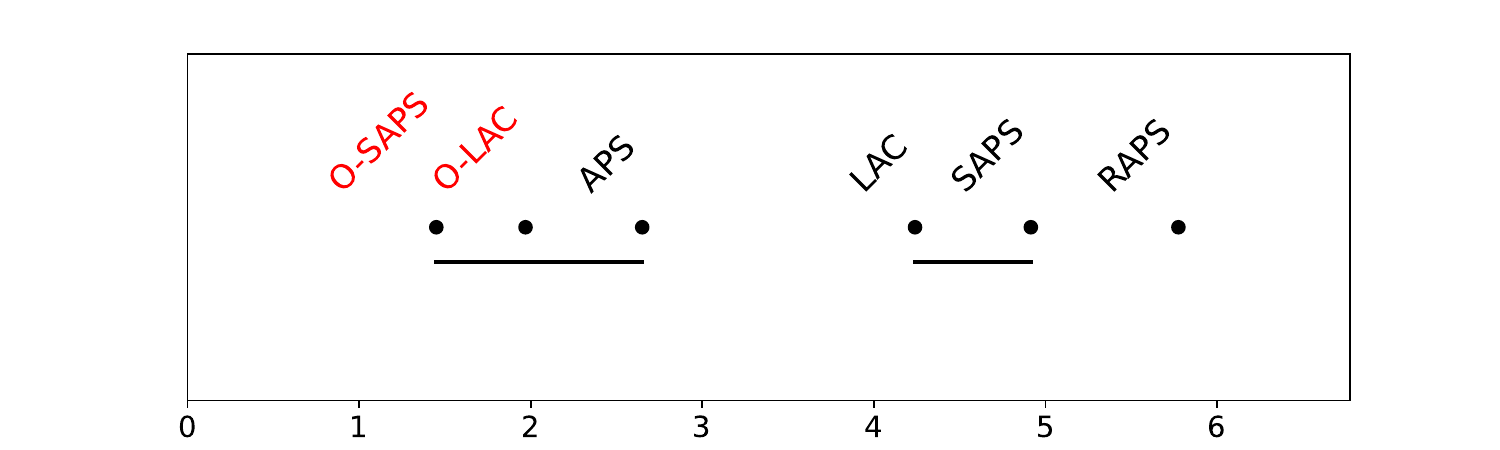}
        \caption{ResNet152}
    \end{subfigure}
    \begin{subfigure}[b]{0.45\textwidth}
        \includegraphics[width=\textwidth]{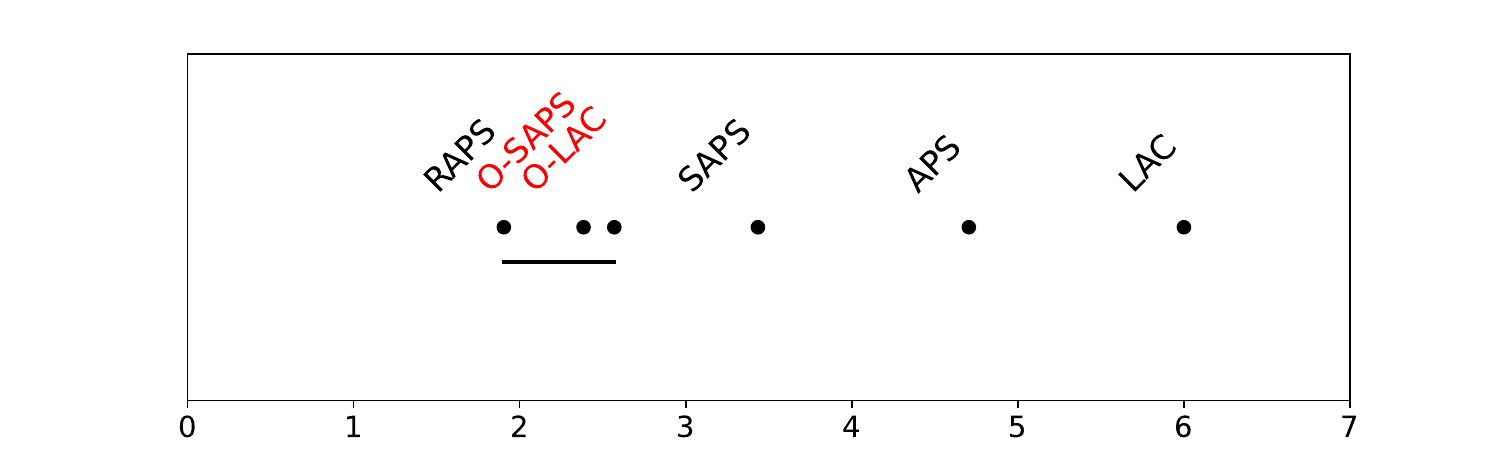}
        \caption{ViT-B-16}
    \end{subfigure}
    \begin{subfigure}[b]{0.45\textwidth}
        \includegraphics[width=\textwidth]{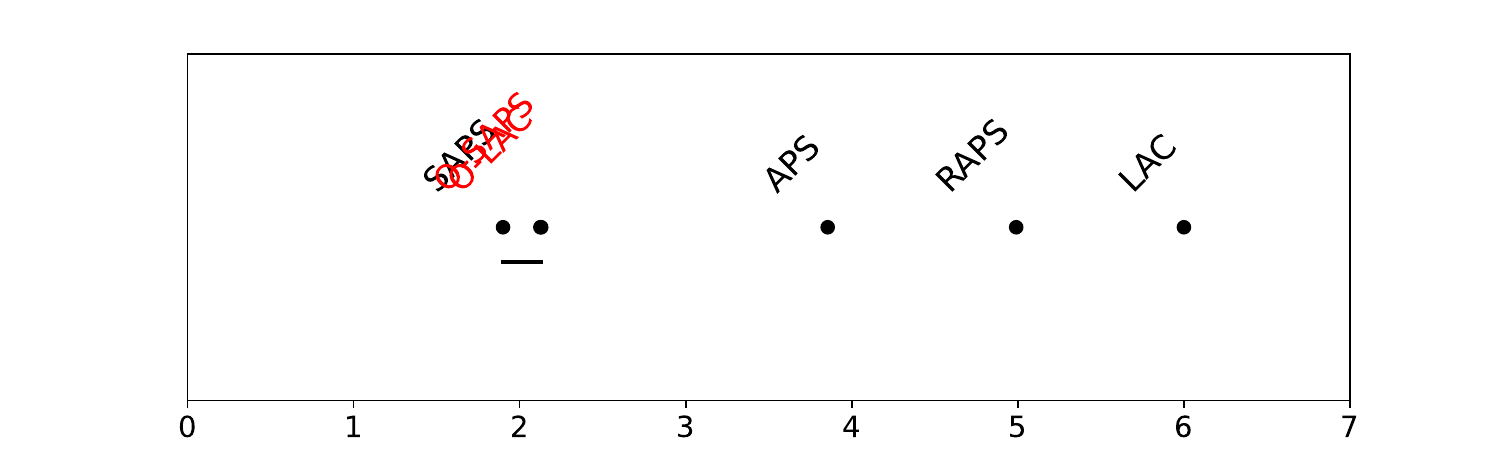}
        \caption{ViT-L-16}
    \end{subfigure}
    \begin{subfigure}[b]{0.45\textwidth}
        \includegraphics[width=\textwidth]{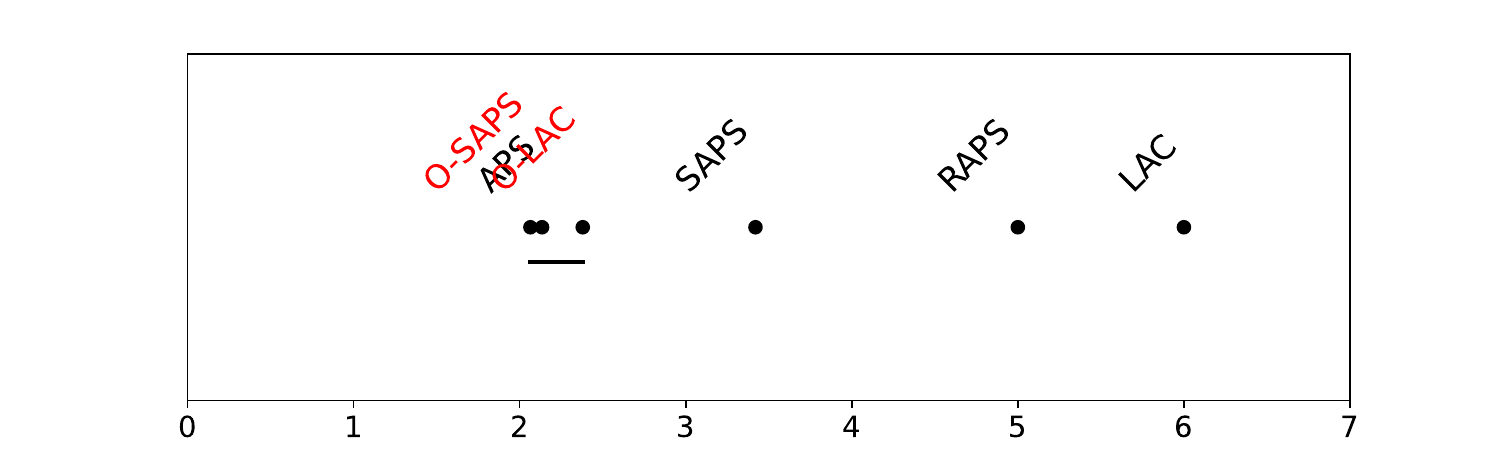}
        \caption{ViT-H-14}
    \end{subfigure}
    \begin{subfigure}[b]{0.45\textwidth}
        \includegraphics[width=\textwidth]{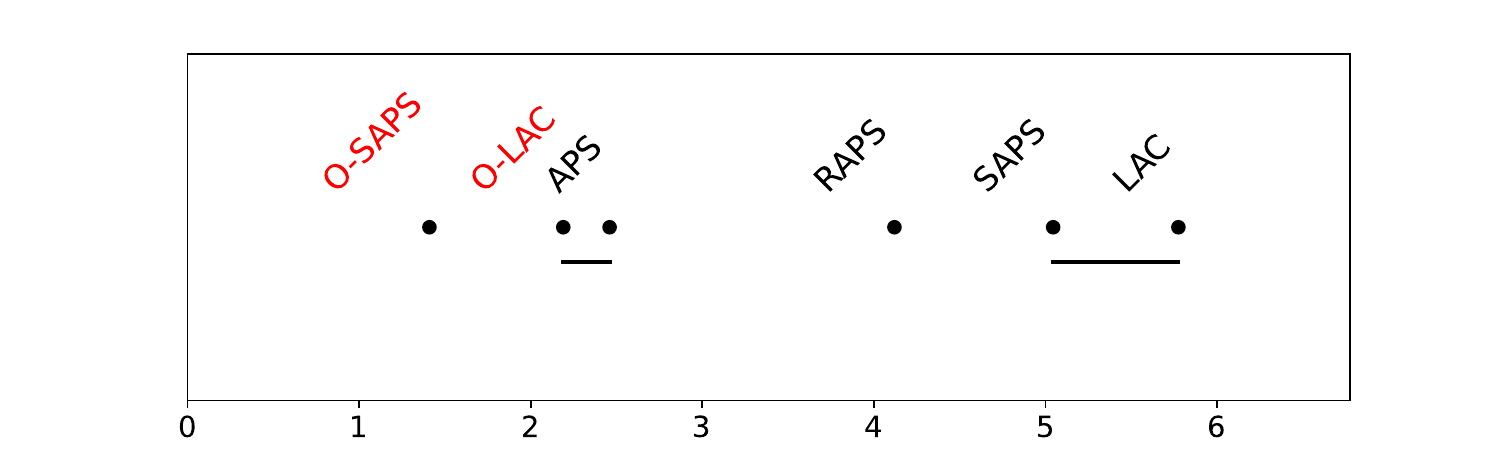}
        \caption{EfficientNet-V2-M}
    \end{subfigure}
    \begin{subfigure}[b]{0.45\textwidth}
        \includegraphics[width=\textwidth]{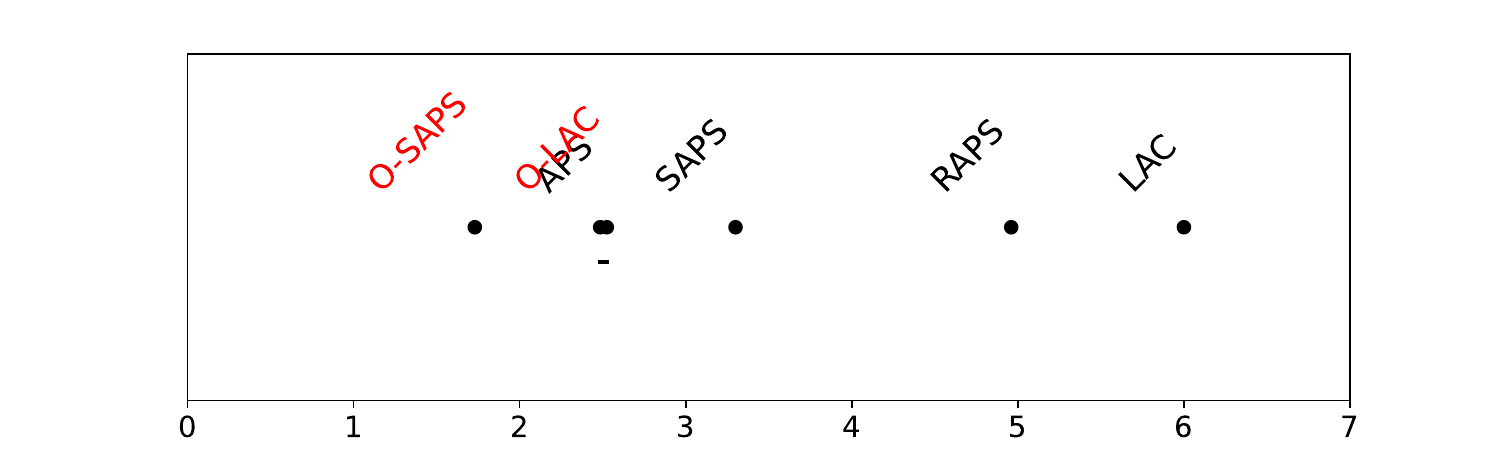}
        \caption{EfficientNet-V2-L}
    \end{subfigure}
    \begin{subfigure}[b]{0.45\textwidth}
        \includegraphics[width=\textwidth]{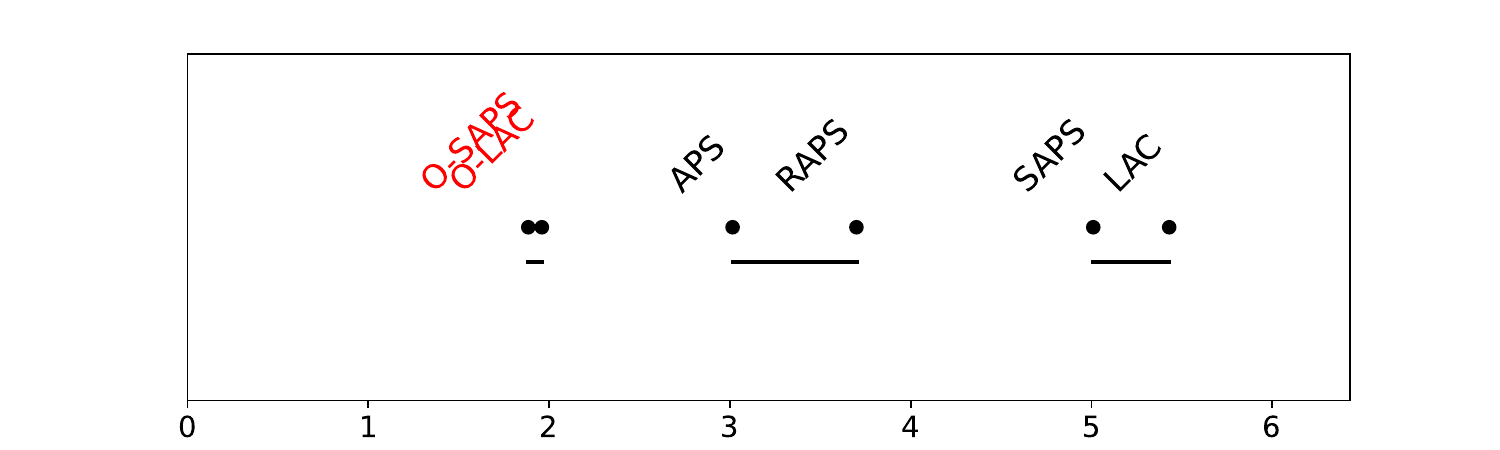}
        \caption{Swin-V2-B}
    \end{subfigure}
    \caption{Critical Difference Diagrams. T-CV, $\alpha=0.10, B=50$. The rank analysis based on these figures is summarized as `Avg. Rank from CD' in Table~\ref{tab:alg_results_our_metrics} in the main text.}
    \label{fig:apdx_cd_tcv_alpha0.10}
\end{figure}

\begin{figure}[!bt]
    \centering
    \begin{subfigure}[b]{0.45\textwidth}
        \includegraphics[width=\textwidth]{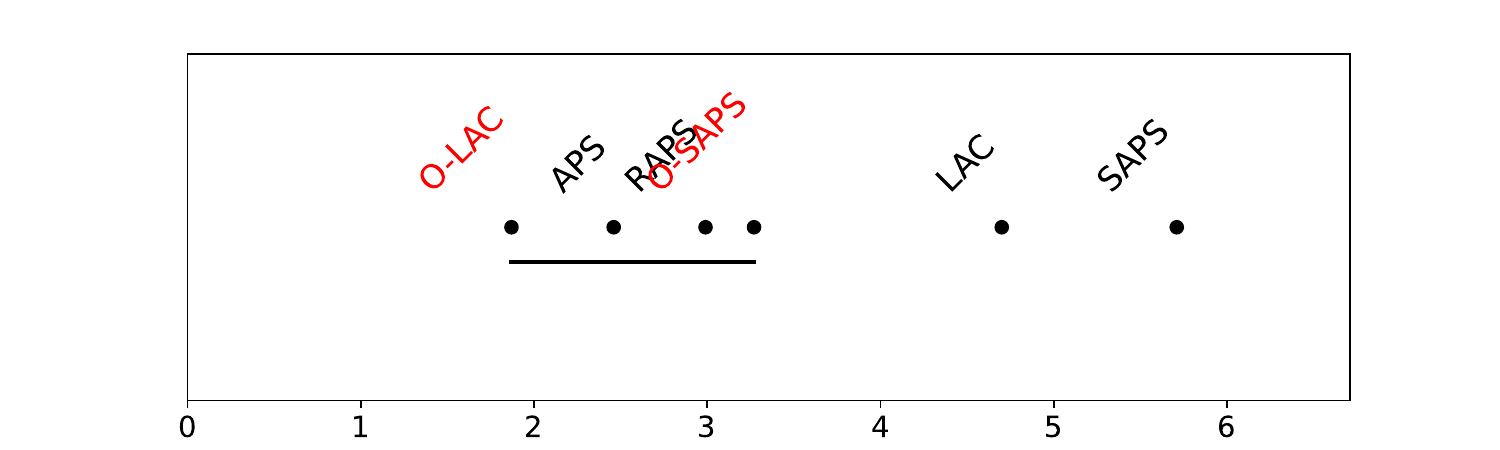}
        \caption{ResNet18}
    \end{subfigure}
    \begin{subfigure}[b]{0.45\textwidth}
        \includegraphics[width=\textwidth]{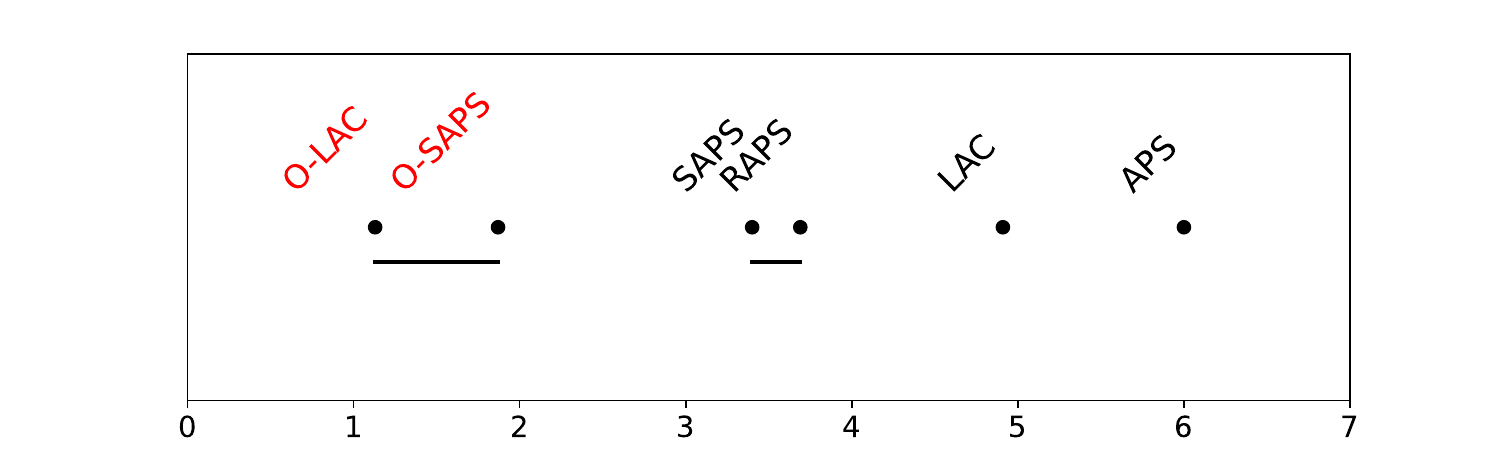}
        \caption{ResNet50}
    \end{subfigure}
    \begin{subfigure}[b]{0.45\textwidth}
        \includegraphics[width=\textwidth]{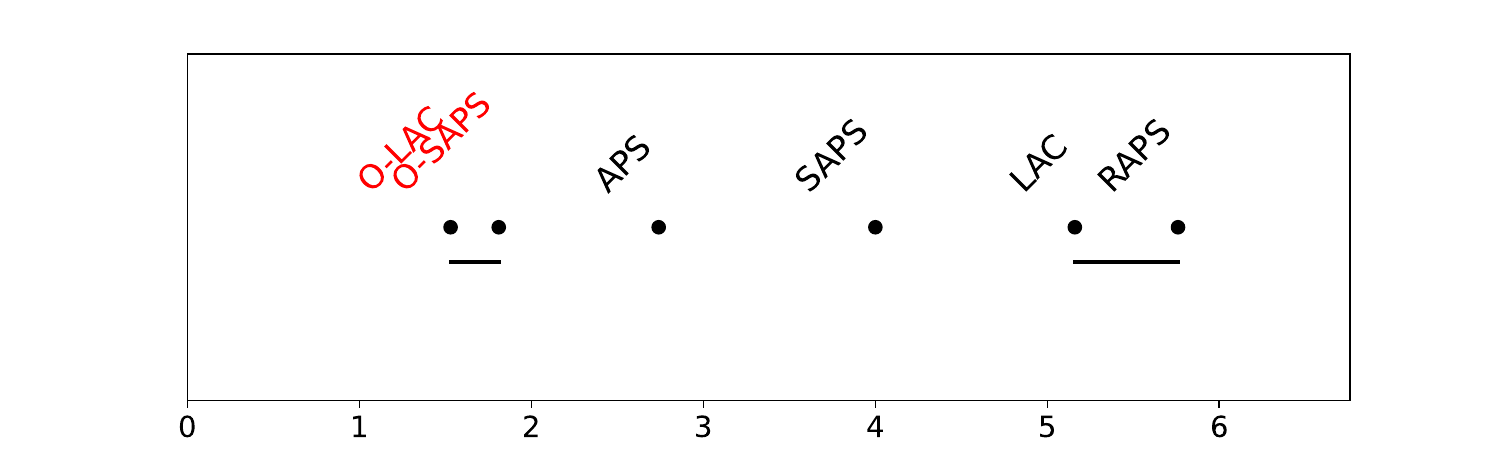}
        \caption{ResNet152}
    \end{subfigure}
    \begin{subfigure}[b]{0.45\textwidth}
        \includegraphics[width=\textwidth]{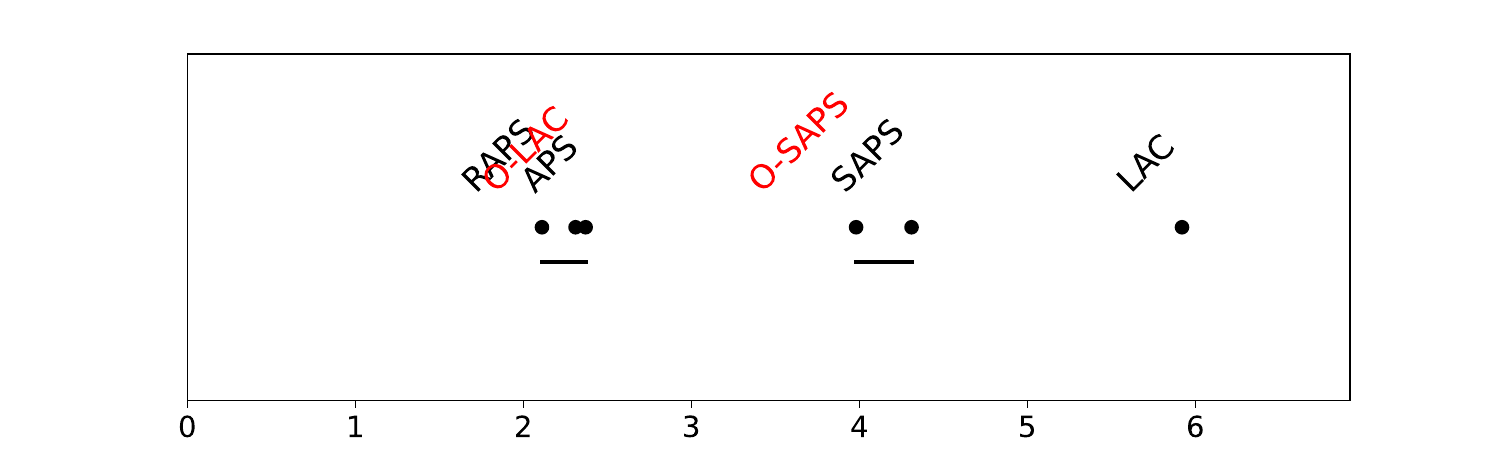}
        \caption{ViT-B-16}
    \end{subfigure}
    \begin{subfigure}[b]{0.45\textwidth}
        \includegraphics[width=\textwidth]{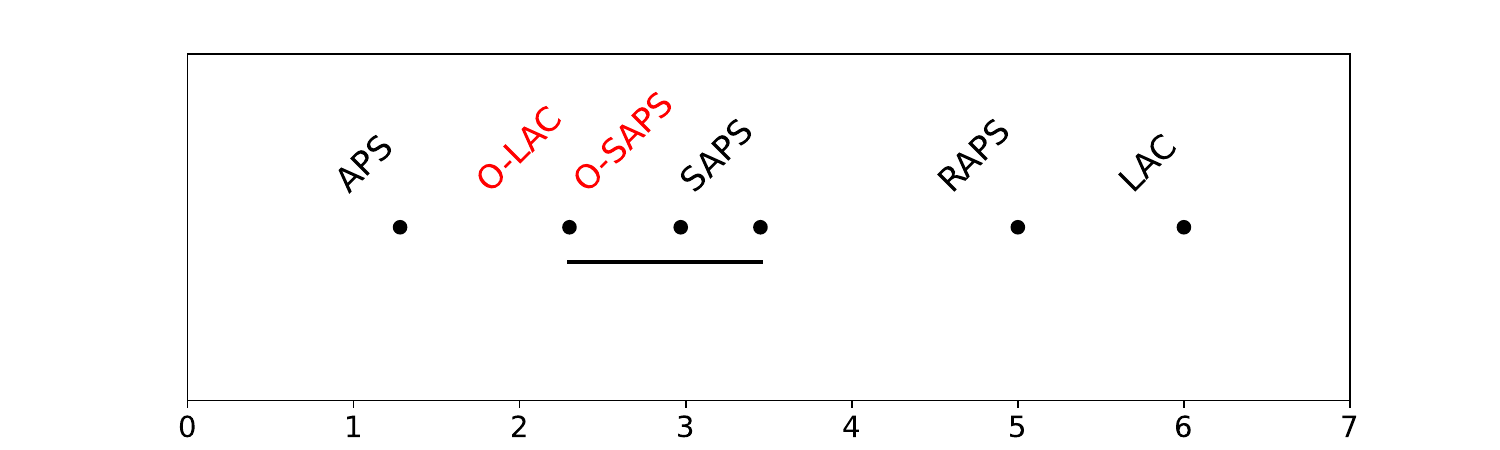}
        \caption{ViT-L-16}
    \end{subfigure}
    \begin{subfigure}[b]{0.45\textwidth}
        \includegraphics[width=\textwidth]{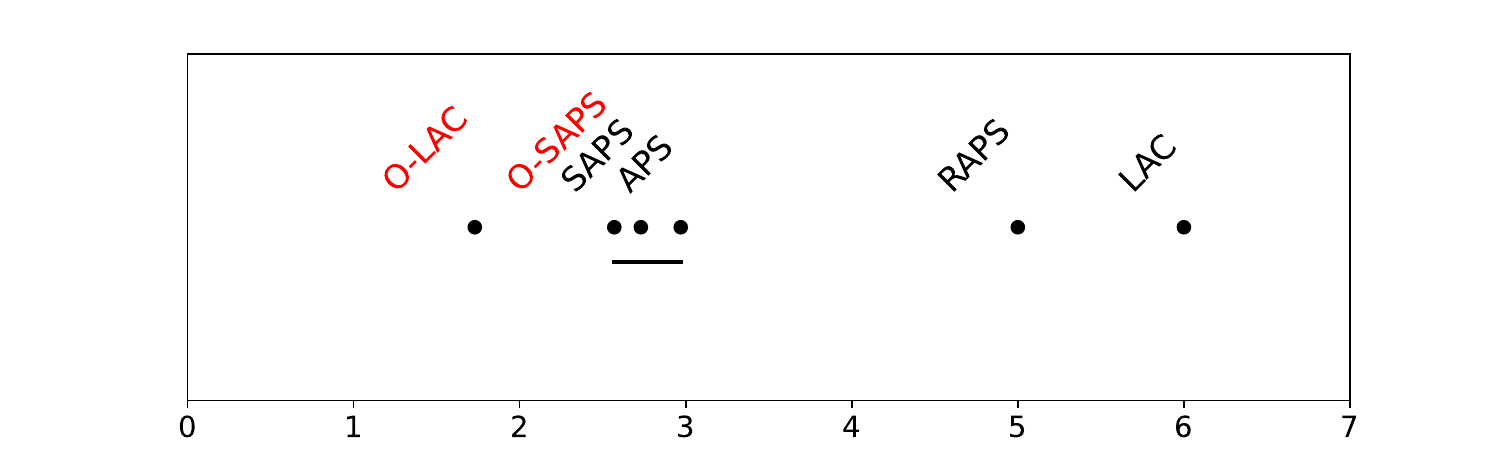}
        \caption{ViT-H-14}
    \end{subfigure}
    \begin{subfigure}[b]{0.45\textwidth}
        \includegraphics[width=\textwidth]{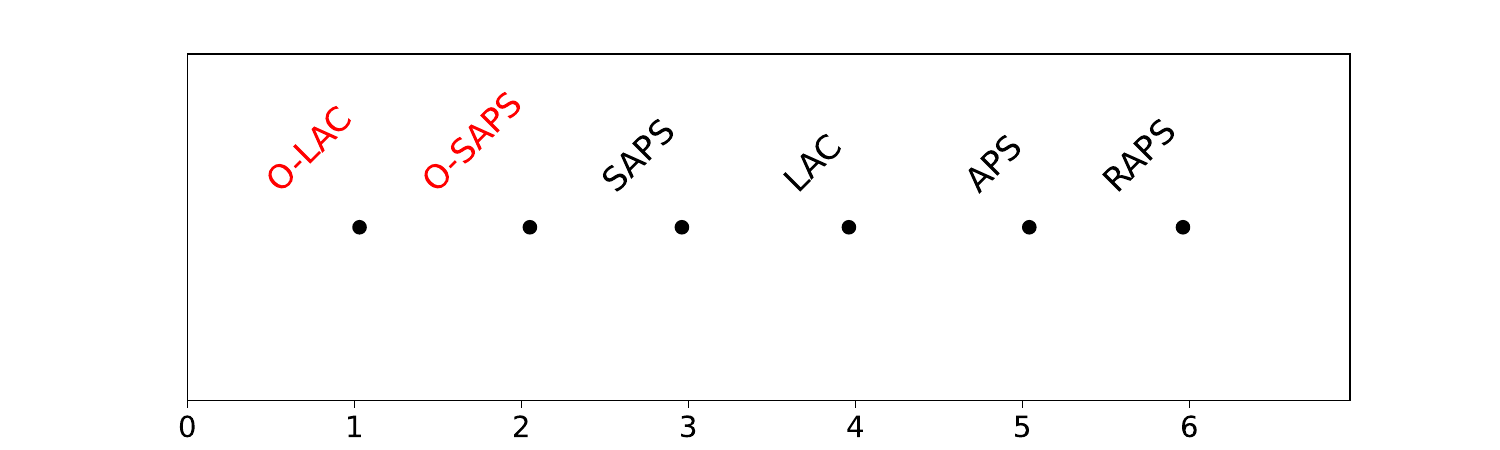}
        \caption{EfficientNet-V2-M}
    \end{subfigure}
    \begin{subfigure}[b]{0.45\textwidth}
        \includegraphics[width=\textwidth]{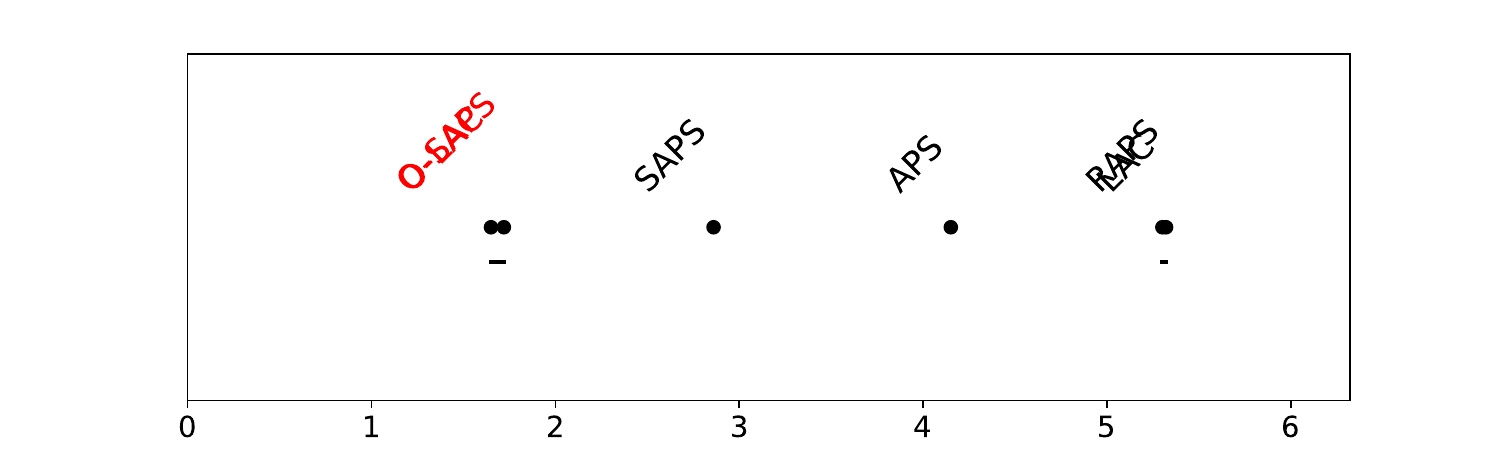}
        \caption{EfficientNet-V2-L}
    \end{subfigure}
    \begin{subfigure}[b]{0.45\textwidth}
        \includegraphics[width=\textwidth]{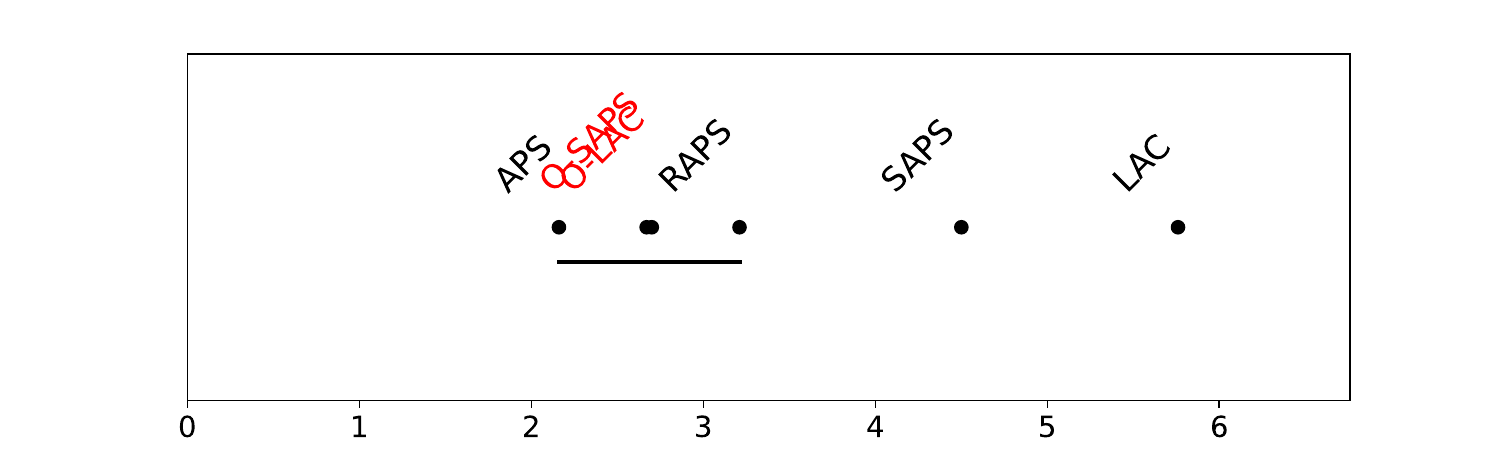}
        \caption{Swin-V2-B}
    \end{subfigure}
    \caption{Critical Difference Diagrams. T-SS, $\alpha=0.10, B=50$. The rank analysis based on these figures is summarized as `Avg. Rank from CD' in Table~\ref{tab:alg_results_our_metrics} in the main text.}
    \label{fig:apdx_cd_tss_alpha0.10}
\end{figure}

\begin{figure}[!bt]
    \centering
    \begin{subfigure}[b]{0.45\textwidth}
        \includegraphics[width=\textwidth]{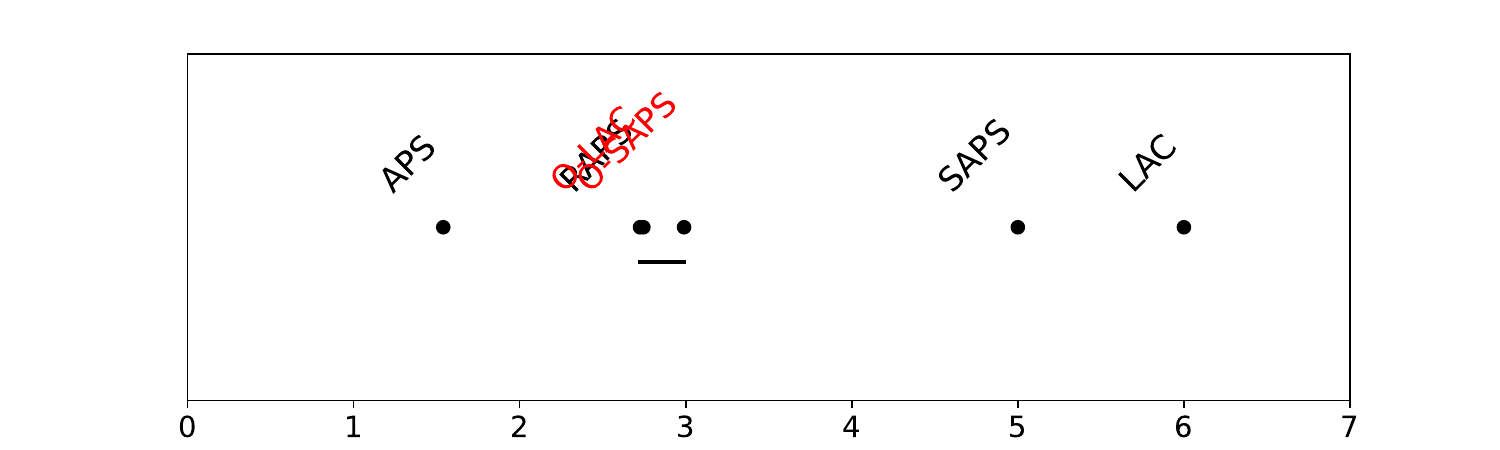}
        \caption{ResNet18}
    \end{subfigure}
    \begin{subfigure}[b]{0.45\textwidth}
        \includegraphics[width=\textwidth]{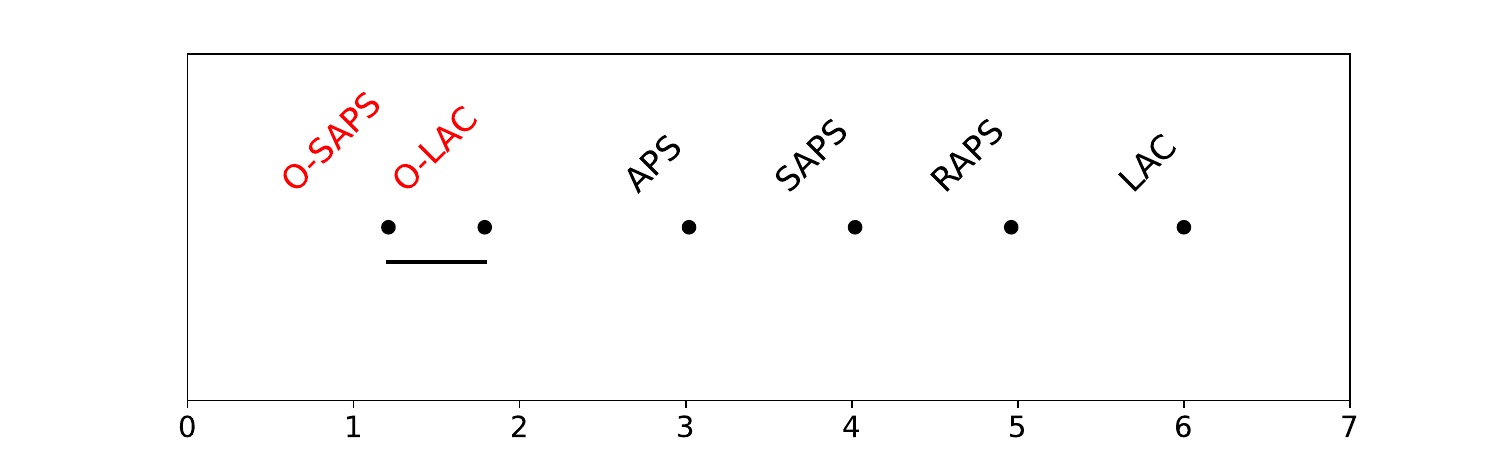}
        \caption{ResNet50}
    \end{subfigure}
    \begin{subfigure}[b]{0.45\textwidth}
        \includegraphics[width=\textwidth]{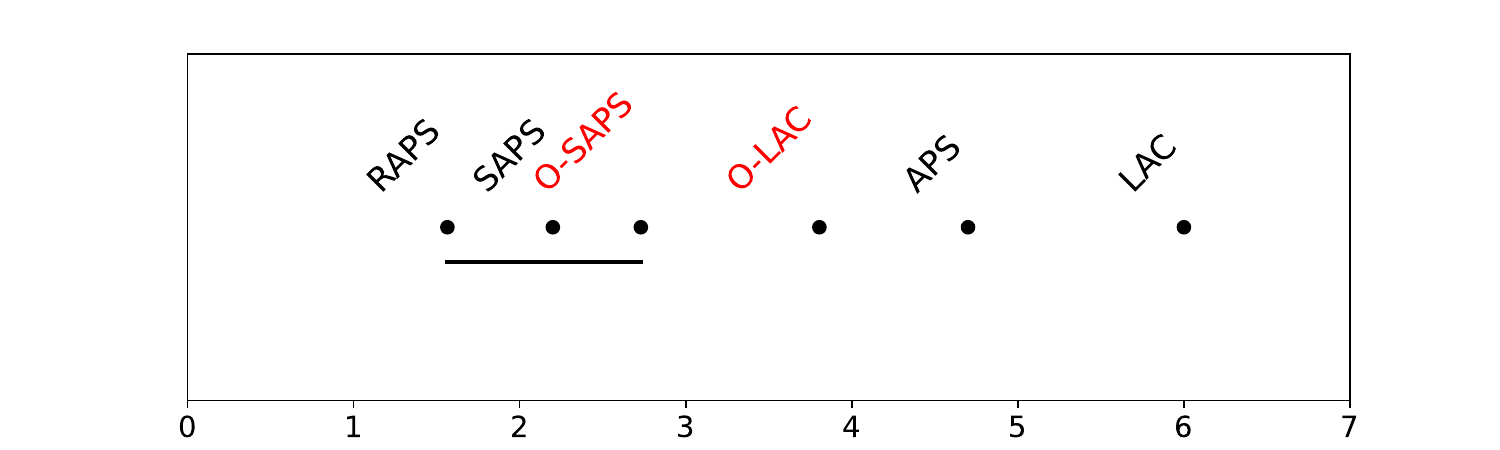}
        \caption{ResNet152}
    \end{subfigure}
    \begin{subfigure}[b]{0.45\textwidth}
        \includegraphics[width=\textwidth]{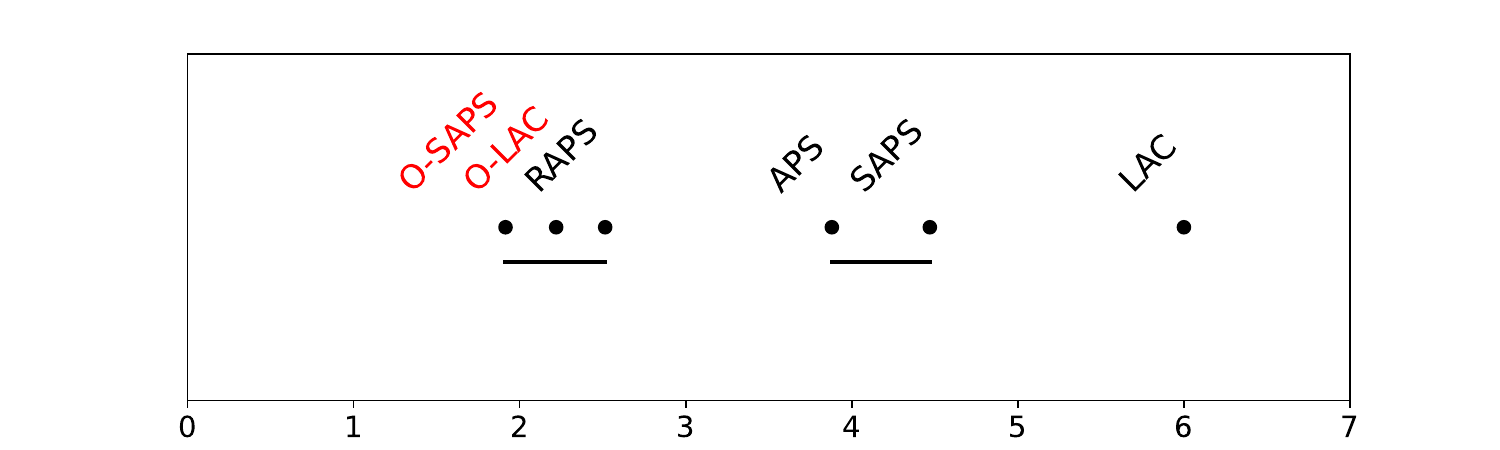}
        \caption{ViT-B-16}
    \end{subfigure}
    \begin{subfigure}[b]{0.45\textwidth}
        \includegraphics[width=\textwidth]{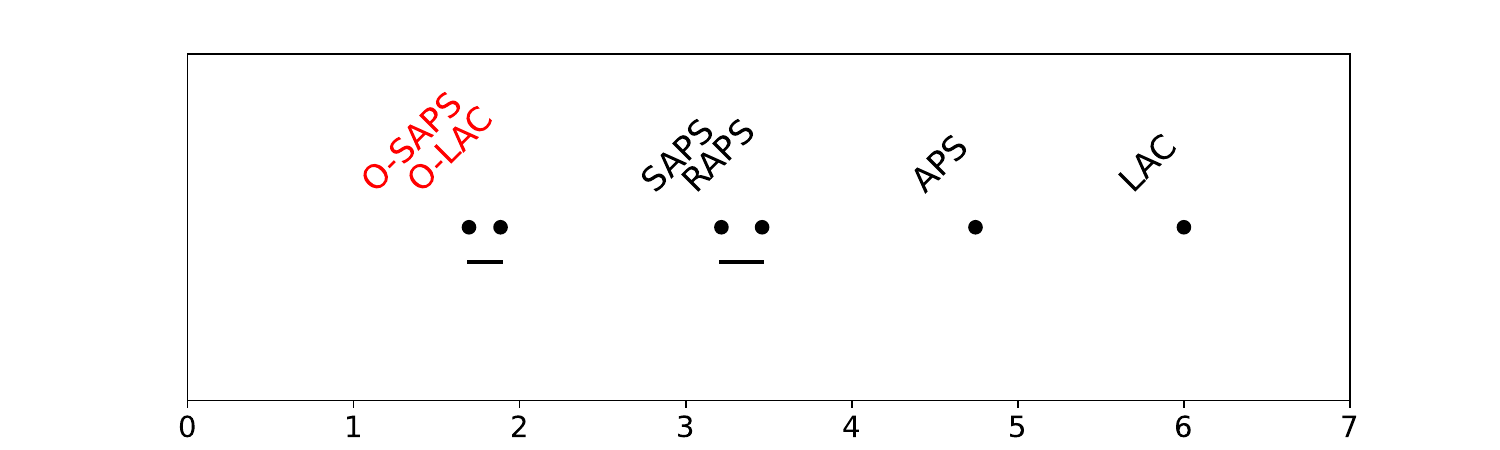}
        \caption{ViT-L-16}
    \end{subfigure}
    \begin{subfigure}[b]{0.45\textwidth}
        \includegraphics[width=\textwidth]{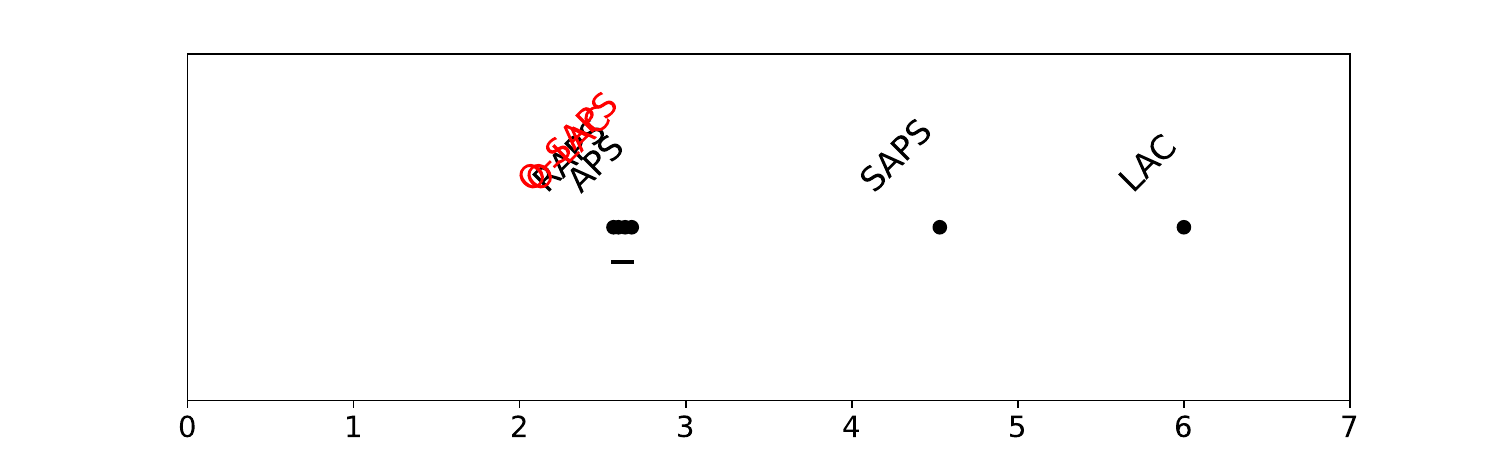}
        \caption{ViT-H-14}
    \end{subfigure}
    \begin{subfigure}[b]{0.45\textwidth}
        \includegraphics[width=\textwidth]{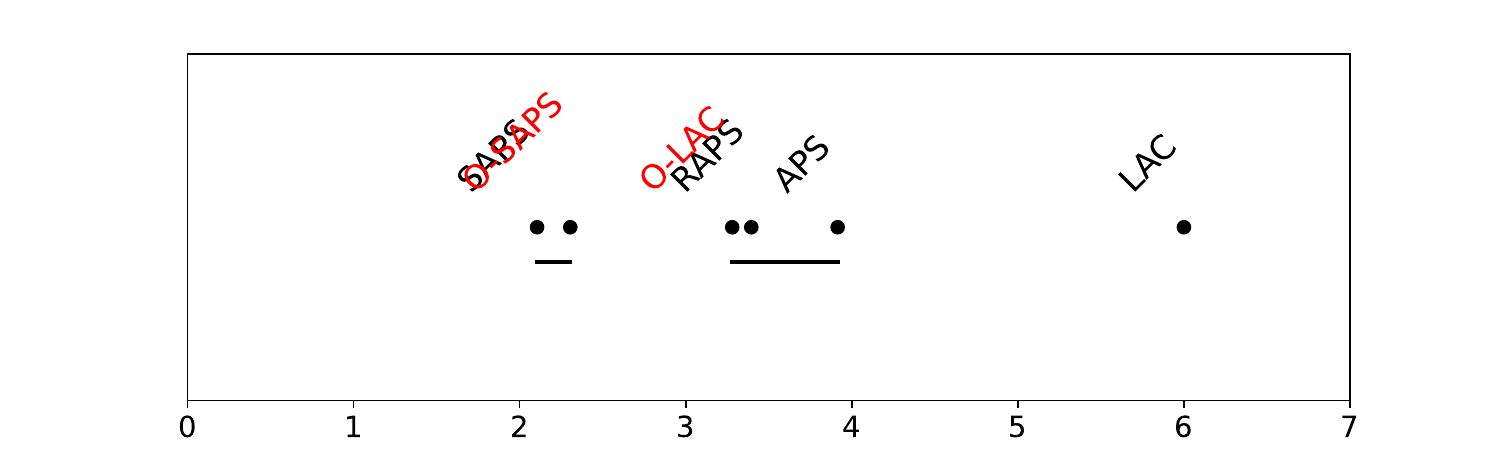}
        \caption{EfficientNet-V2-M}
    \end{subfigure}
    \begin{subfigure}[b]{0.45\textwidth}
        \includegraphics[width=\textwidth]{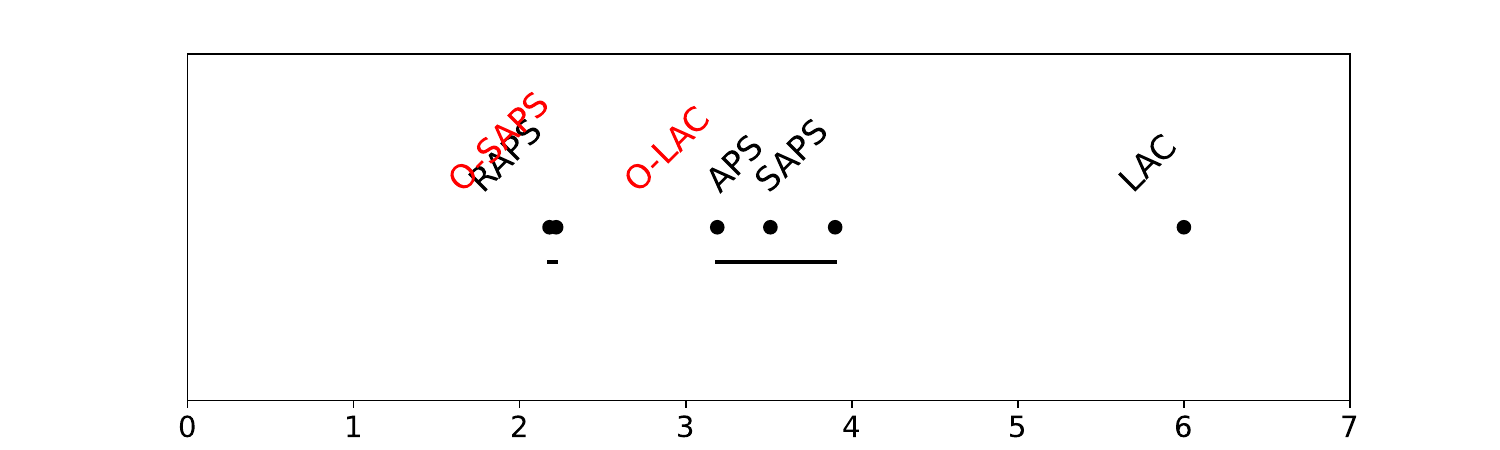}
        \caption{EfficientNet-V2-L}
    \end{subfigure}
    \begin{subfigure}[b]{0.45\textwidth}
        \includegraphics[width=\textwidth]{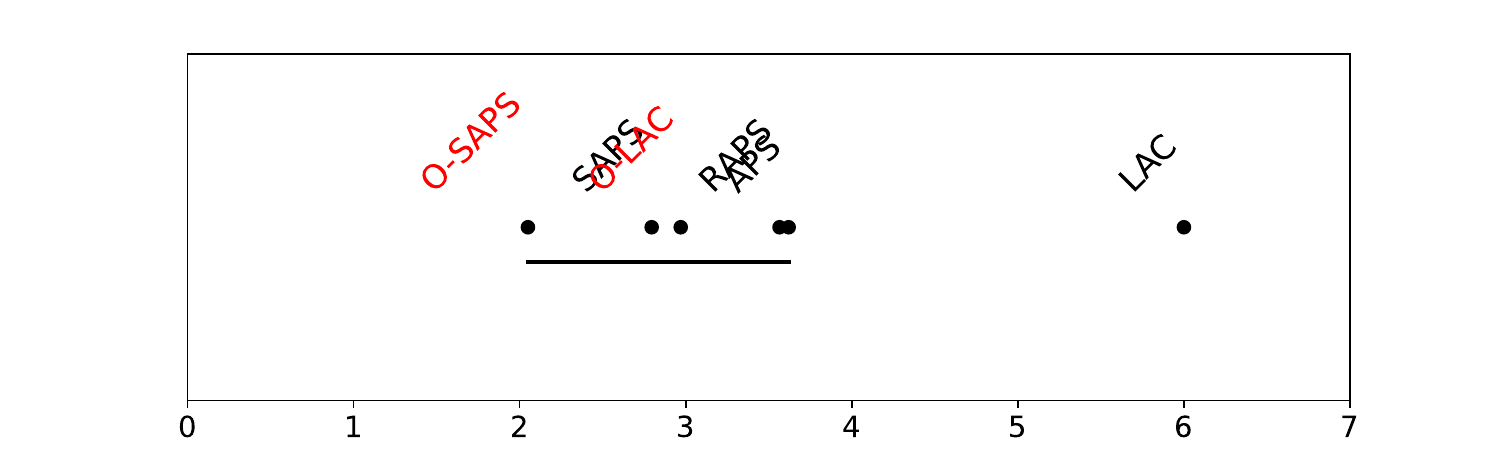}
        \caption{Swin-V2-B}
    \end{subfigure}
    \caption{Critical Difference Diagrams. T-CV, $\alpha=0.15, B=50$. The rank analysis based on these figures is summarized as `Avg. Rank from CD' in Table~\ref{tab:apdx_alg_results_our_metrics} in the main text.}
    \label{fig:apdx_cd_tcv_alpha0.15}
\end{figure}

\begin{figure}[!bt]
    \centering
    \begin{subfigure}[b]{0.45\textwidth}
        \includegraphics[width=\textwidth]{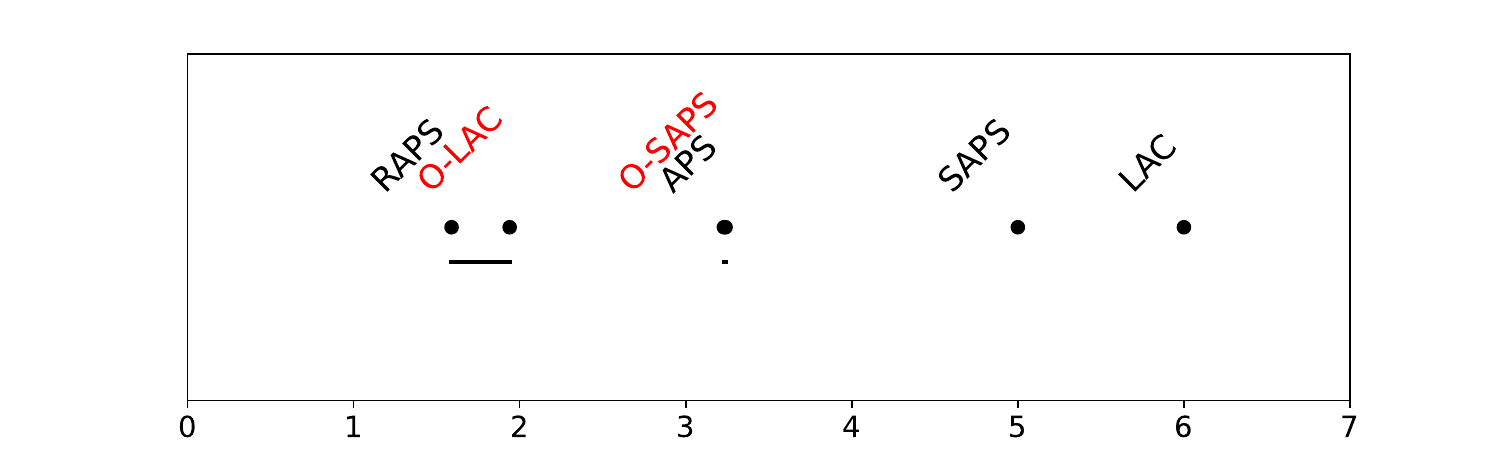}
        \caption{ResNet18}
    \end{subfigure}
    \begin{subfigure}[b]{0.45\textwidth}
        \includegraphics[width=\textwidth]{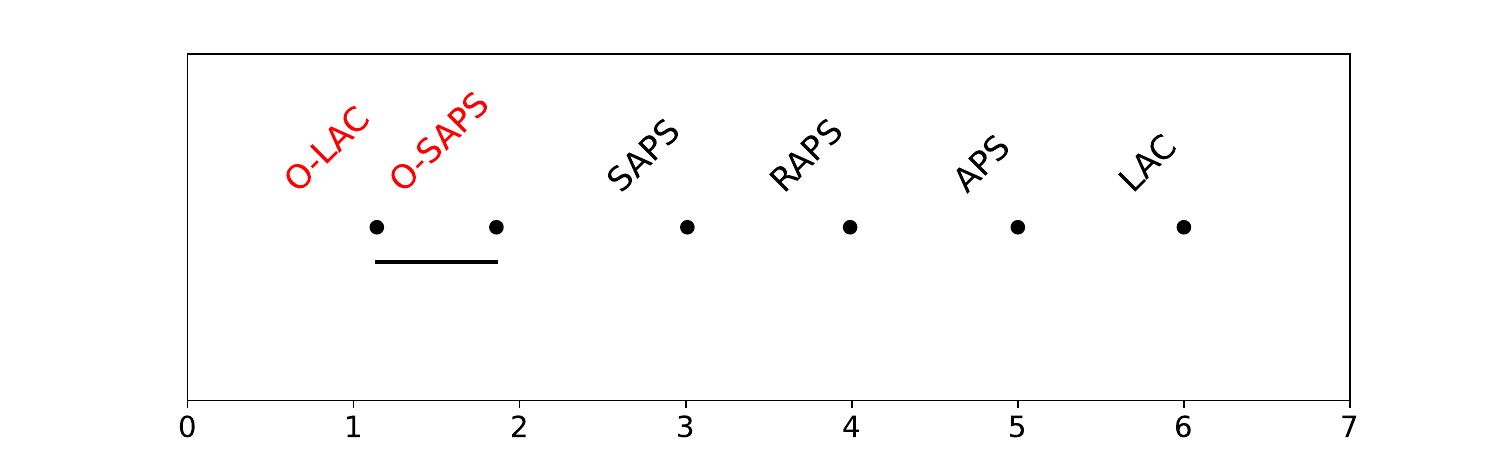}
        \caption{ResNet50}
    \end{subfigure}
    \begin{subfigure}[b]{0.45\textwidth}
        \includegraphics[width=\textwidth]{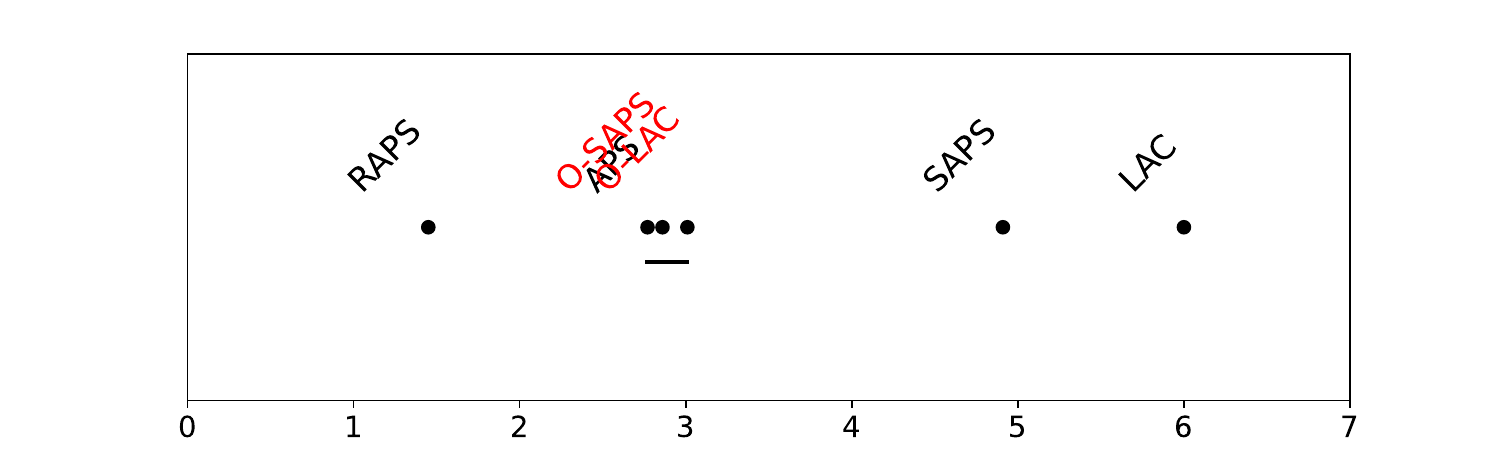}
        \caption{ResNet152}
    \end{subfigure}
    \begin{subfigure}[b]{0.45\textwidth}
        \includegraphics[width=\textwidth]{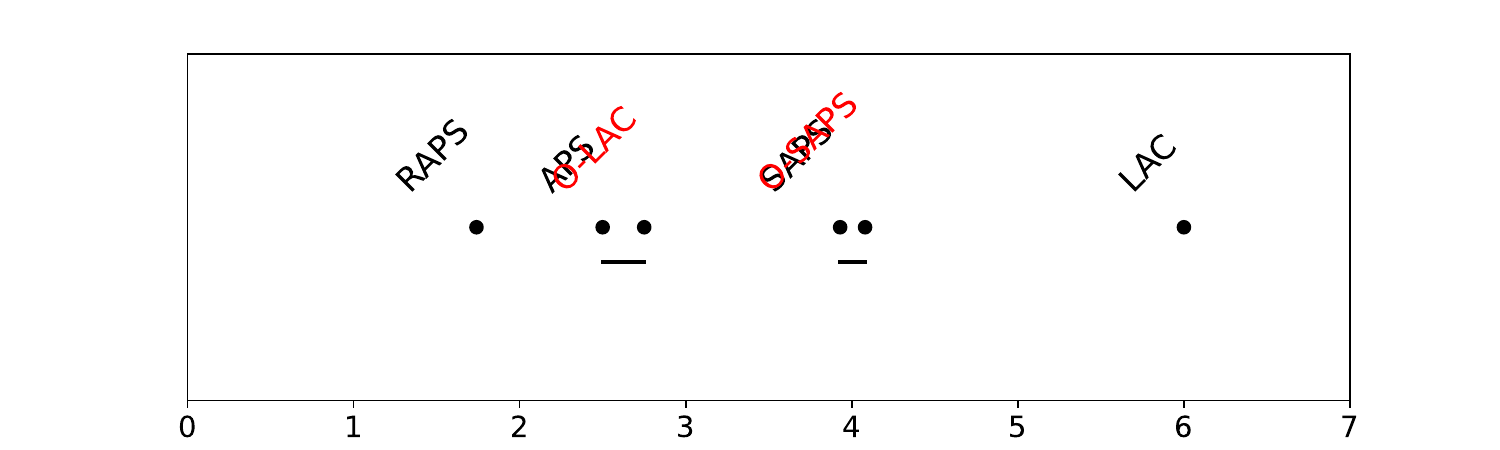}
        \caption{ViT-B-16}
    \end{subfigure}
    \begin{subfigure}[b]{0.45\textwidth}
        \includegraphics[width=\textwidth]{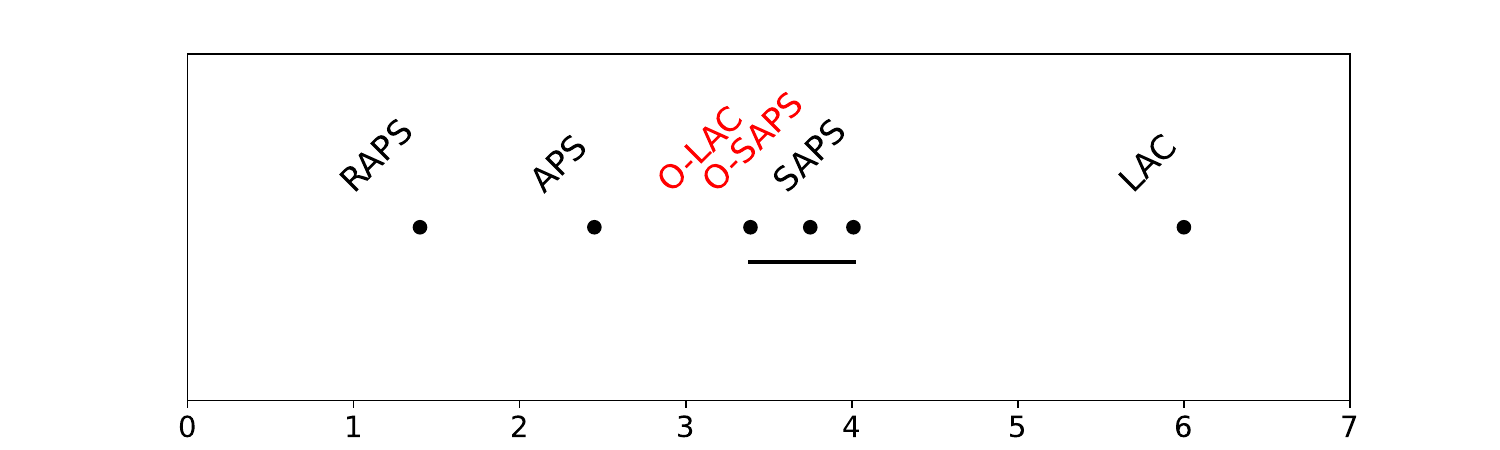}
        \caption{ViT-L-16}
    \end{subfigure}
    \begin{subfigure}[b]{0.45\textwidth}
        \includegraphics[width=\textwidth]{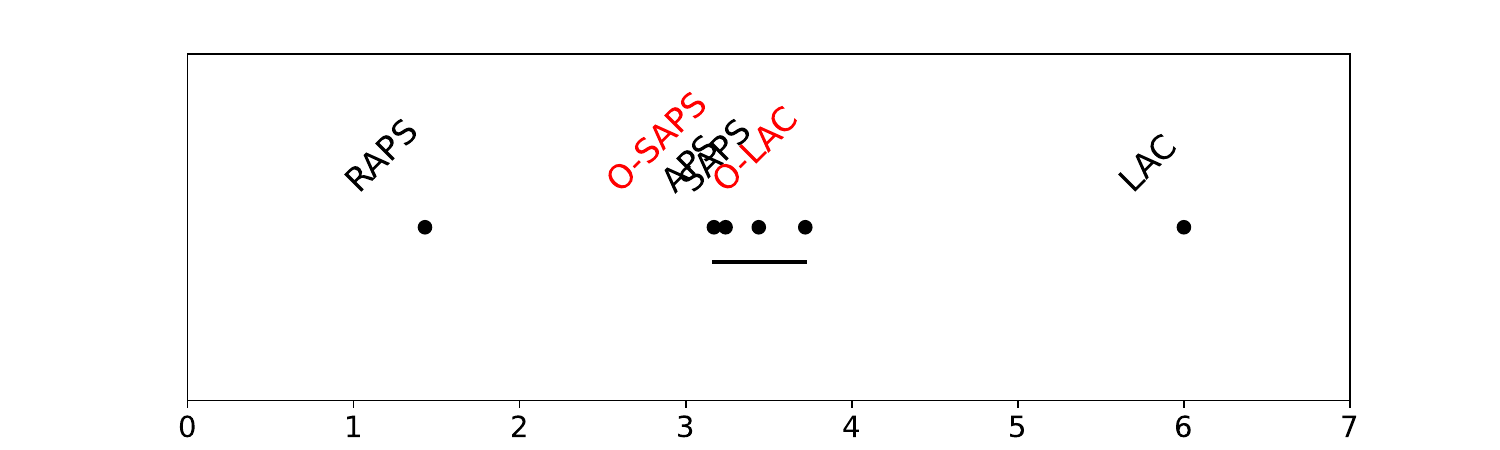}
        \caption{ViT-H-14}
    \end{subfigure}
    \begin{subfigure}[b]{0.45\textwidth}
        \includegraphics[width=\textwidth]{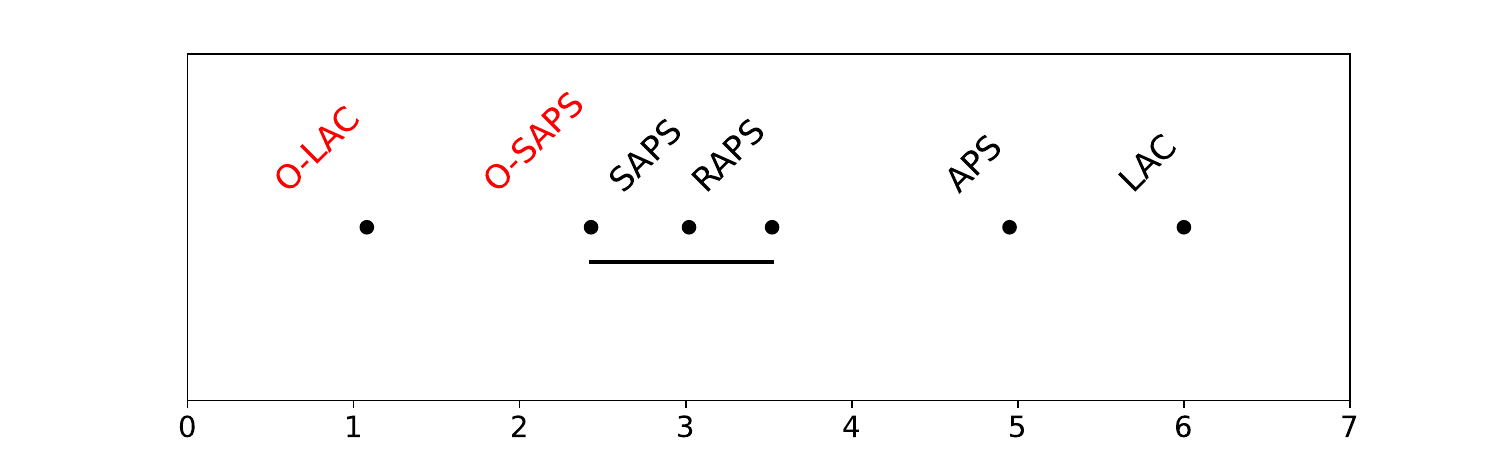}
        \caption{EfficientNet-V2-M}
    \end{subfigure}
    \begin{subfigure}[b]{0.45\textwidth}
        \includegraphics[width=\textwidth]{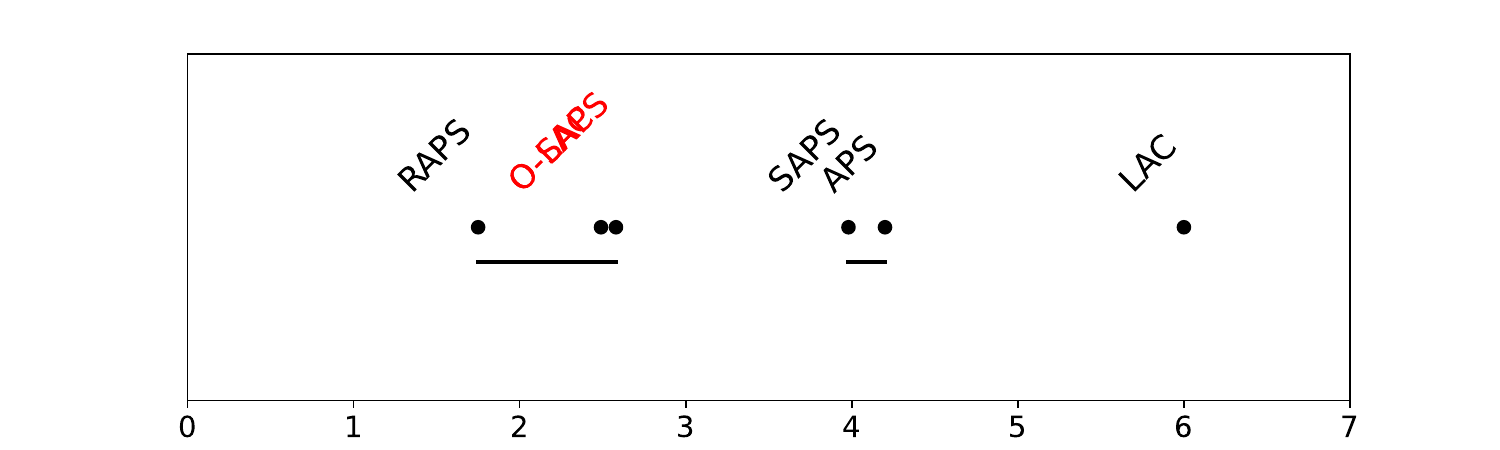}
        \caption{EfficientNet-V2-L}
    \end{subfigure}
    \begin{subfigure}[b]{0.45\textwidth}
        \includegraphics[width=\textwidth]{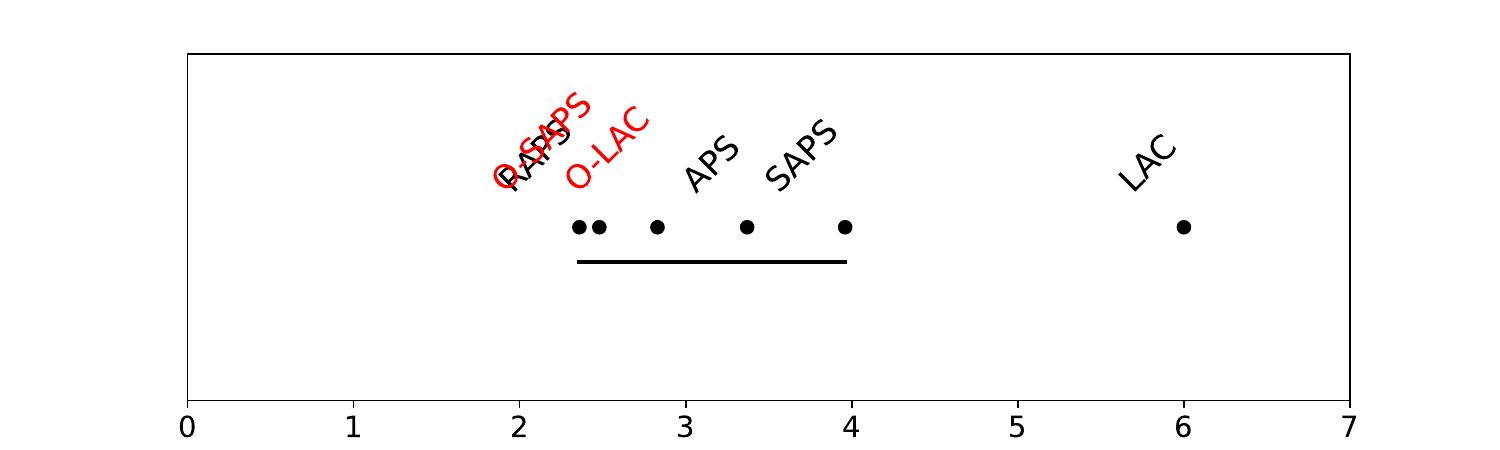}
        \caption{Swin-V2-B}
    \end{subfigure}
    \caption{Critical Difference Diagrams. T-SS, $\alpha=0.15, B=50$. The rank analysis based on these figures is summarized as `Avg. Rank from CD' in Table~\ref{tab:apdx_alg_results_our_metrics} in the main text.}
    \label{fig:apdx_cd_tss_alpha0.15}
\end{figure}

\begin{figure}[!bt]
    \centering
    \begin{subfigure}[b]{0.45\textwidth}
        \includegraphics[width=\textwidth]{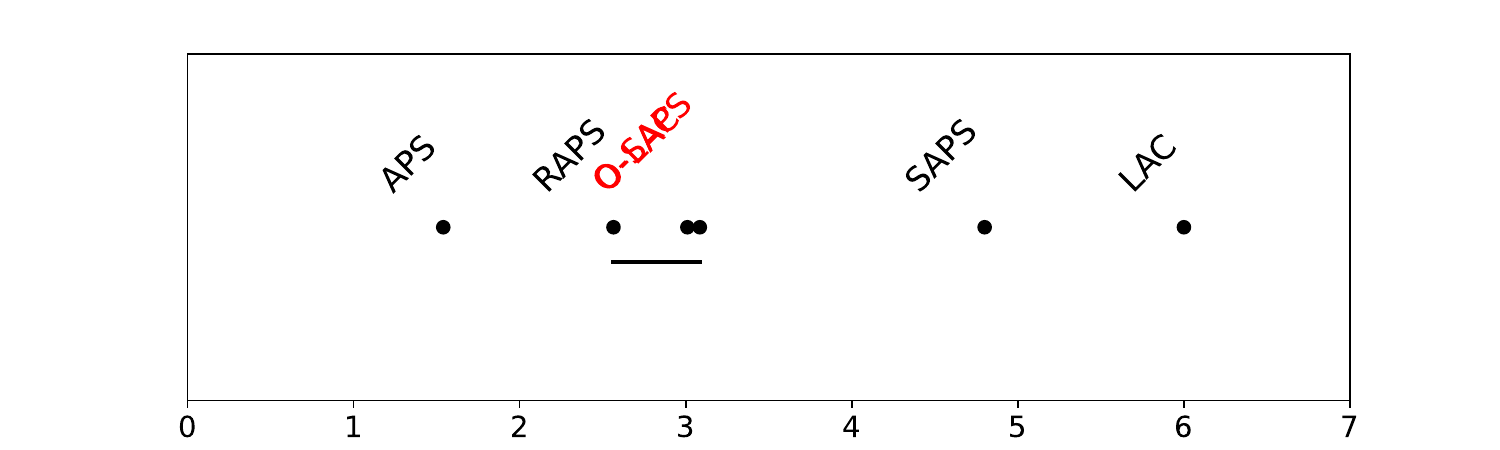}
        \caption{ResNet18}
    \end{subfigure}
    \begin{subfigure}[b]{0.45\textwidth}
        \includegraphics[width=\textwidth]{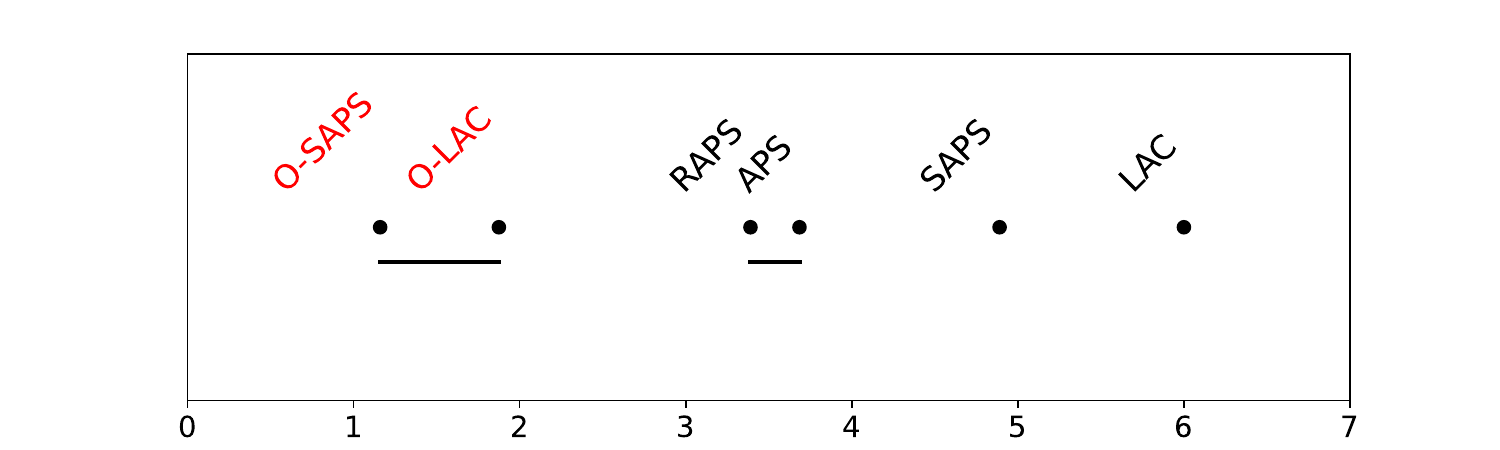}
        \caption{ResNet50}
    \end{subfigure}
    \begin{subfigure}[b]{0.45\textwidth}
        \includegraphics[width=\textwidth]{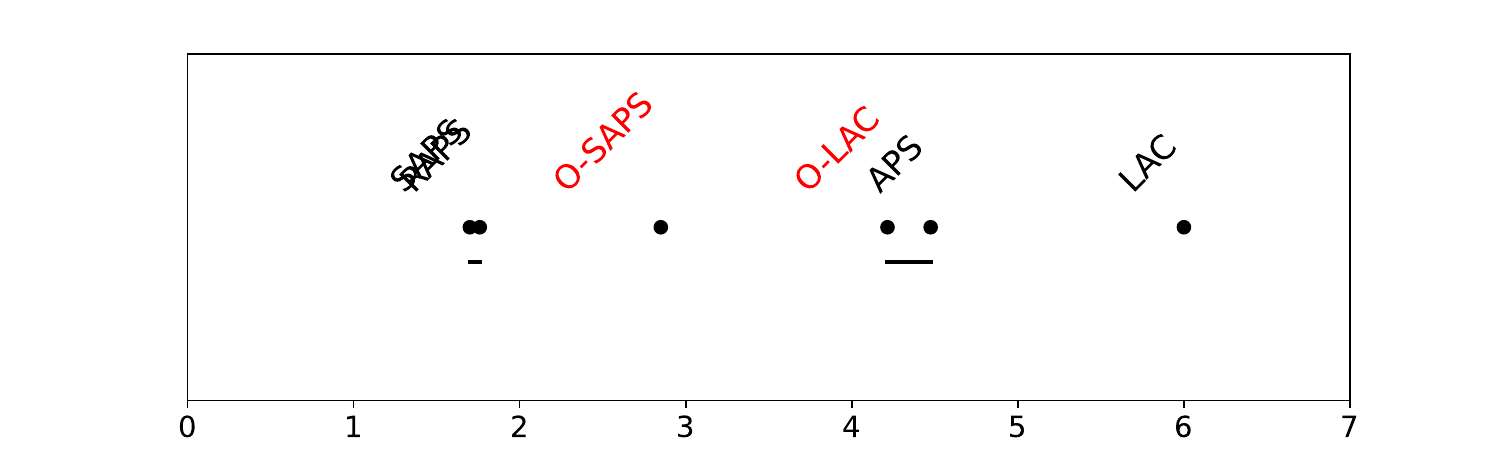}
        \caption{ResNet152}
    \end{subfigure}
    \begin{subfigure}[b]{0.45\textwidth}
        \includegraphics[width=\textwidth]{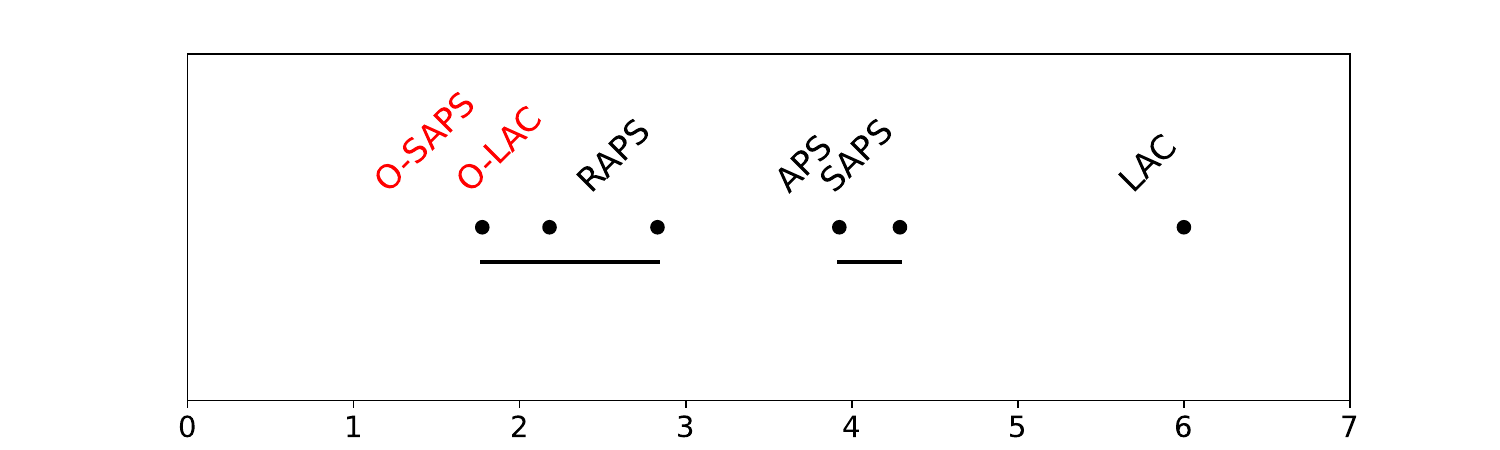}
        \caption{ViT-B-16}
    \end{subfigure}
    \begin{subfigure}[b]{0.45\textwidth}
        \includegraphics[width=\textwidth]{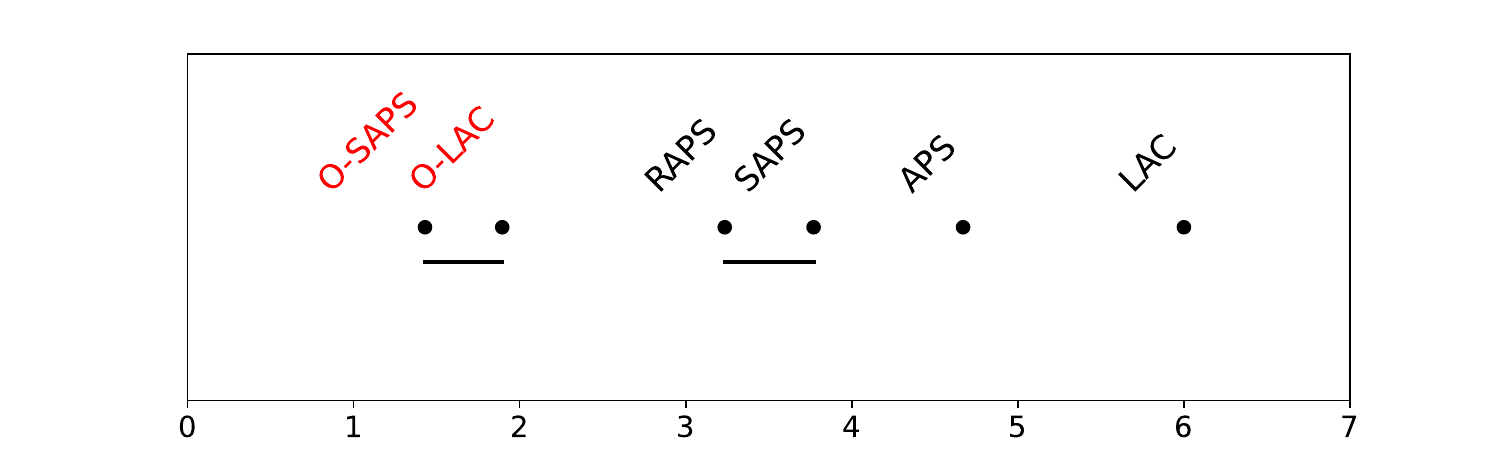}
        \caption{ViT-L-16}
    \end{subfigure}
    \begin{subfigure}[b]{0.45\textwidth}
        \includegraphics[width=\textwidth]{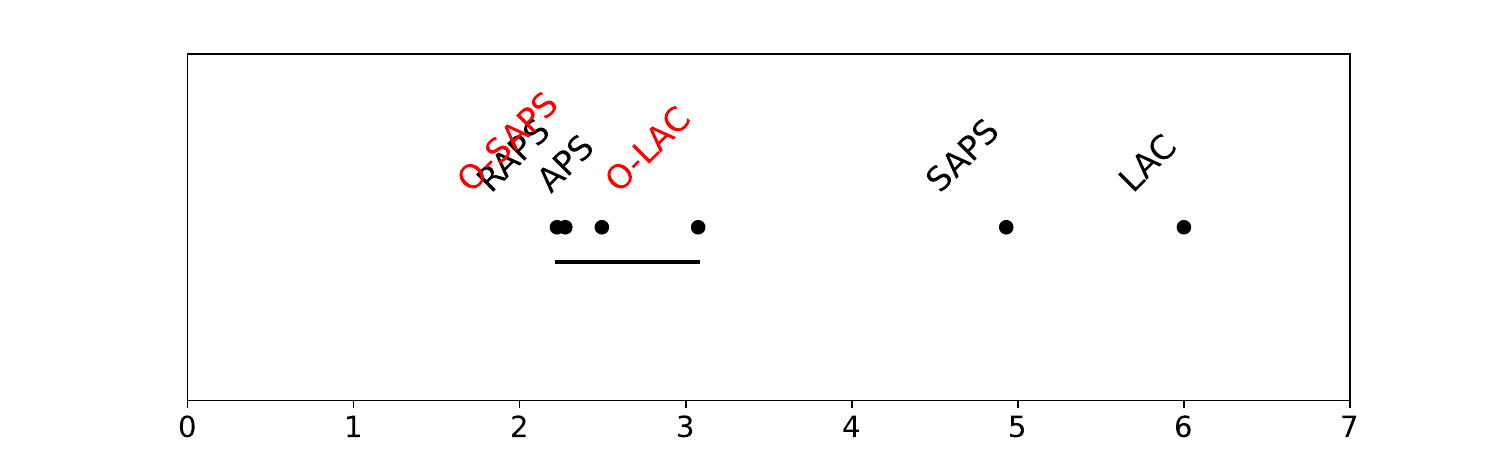}
        \caption{ViT-H-14}
    \end{subfigure}
    \begin{subfigure}[b]{0.45\textwidth}
        \includegraphics[width=\textwidth]{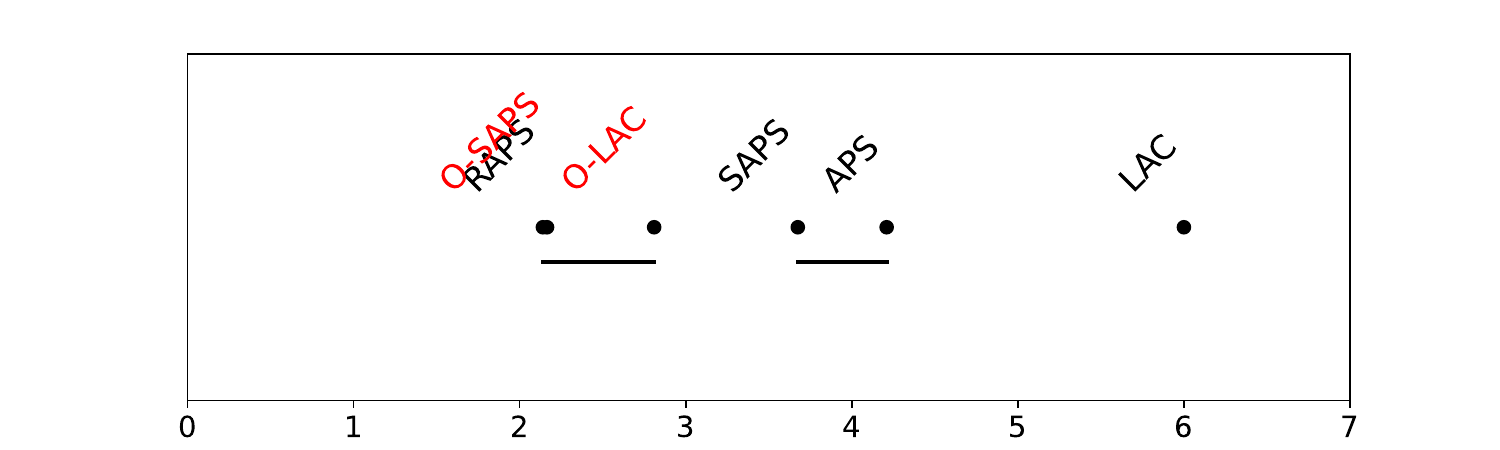}
        \caption{EfficientNet-V2-M}
    \end{subfigure}
    \begin{subfigure}[b]{0.45\textwidth}
        \includegraphics[width=\textwidth]{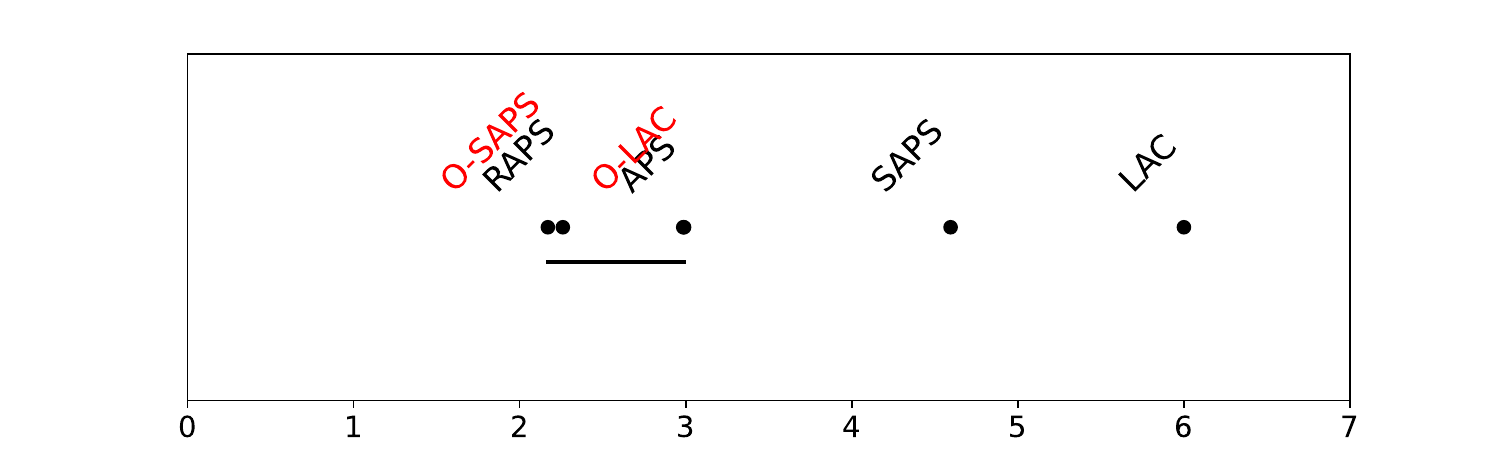}
        \caption{EfficientNet-V2-L}
    \end{subfigure}
    \begin{subfigure}[b]{0.45\textwidth}
        \includegraphics[width=\textwidth]{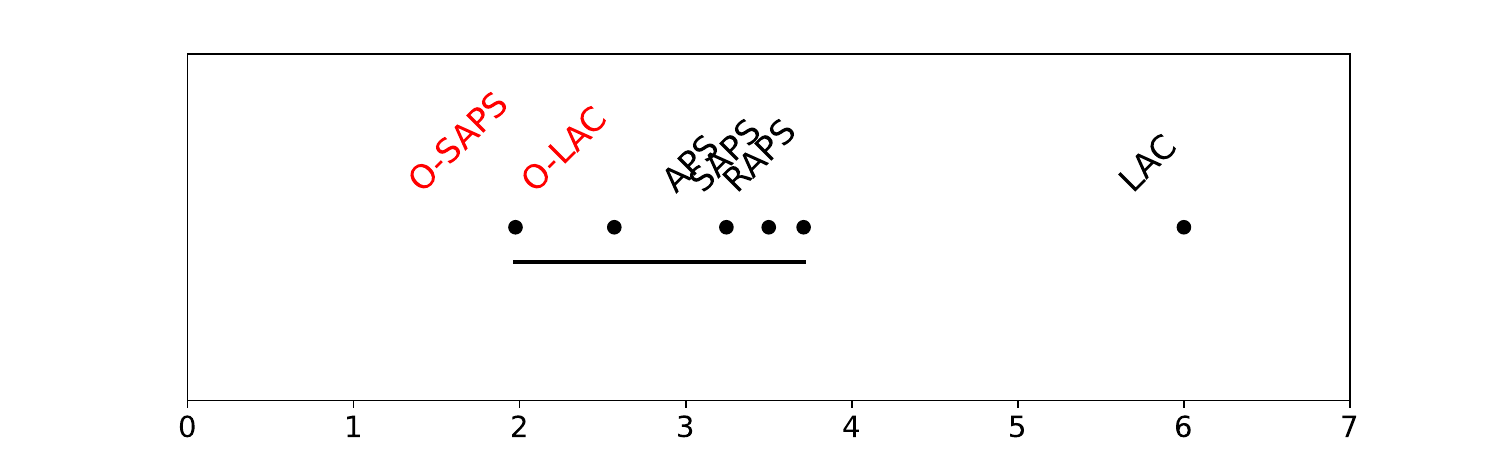}
        \caption{Swin-V2-B}
    \end{subfigure}
    \caption{Critical Difference Diagrams. T-CV, $\alpha=0.20, B=50$. The rank analysis based on these figures is summarized as `Avg. Rank from CD' in Table~\ref{tab:apdx_alg_results_our_metrics} in the main text.}
    \label{fig:apdx_cd_tcv_alpha0.20}
\end{figure}

\begin{figure}[!bt]
    \centering
    \begin{subfigure}[b]{0.45\textwidth}
        \includegraphics[width=\textwidth]{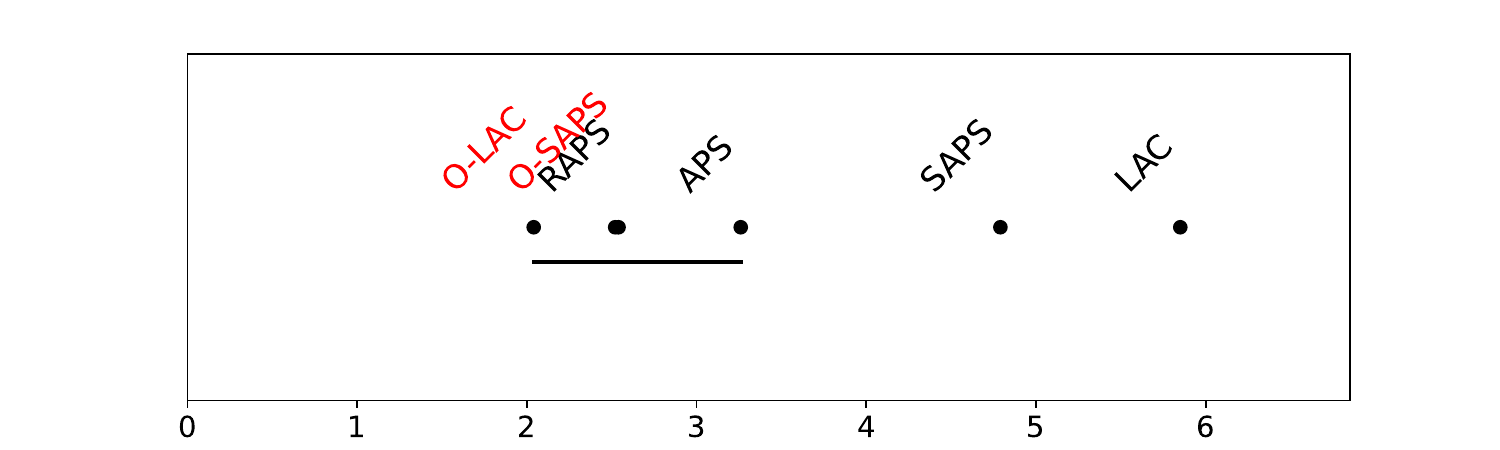}
        \caption{ResNet18}
    \end{subfigure}
    \begin{subfigure}[b]{0.45\textwidth}
        \includegraphics[width=\textwidth]{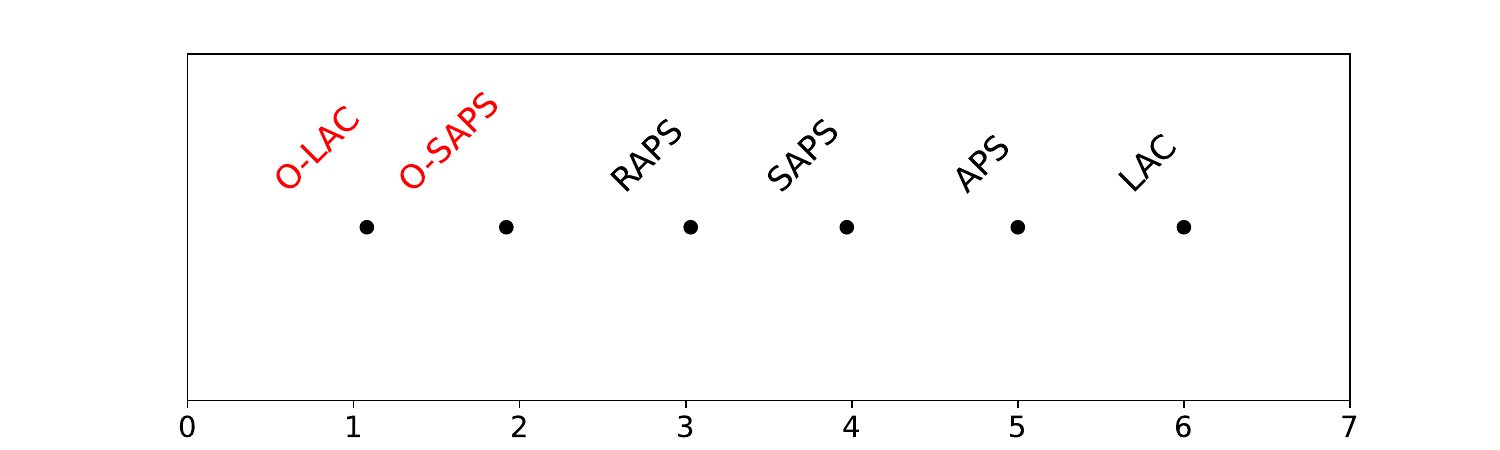}
        \caption{ResNet50}
    \end{subfigure}
    \begin{subfigure}[b]{0.45\textwidth}
        \includegraphics[width=\textwidth]{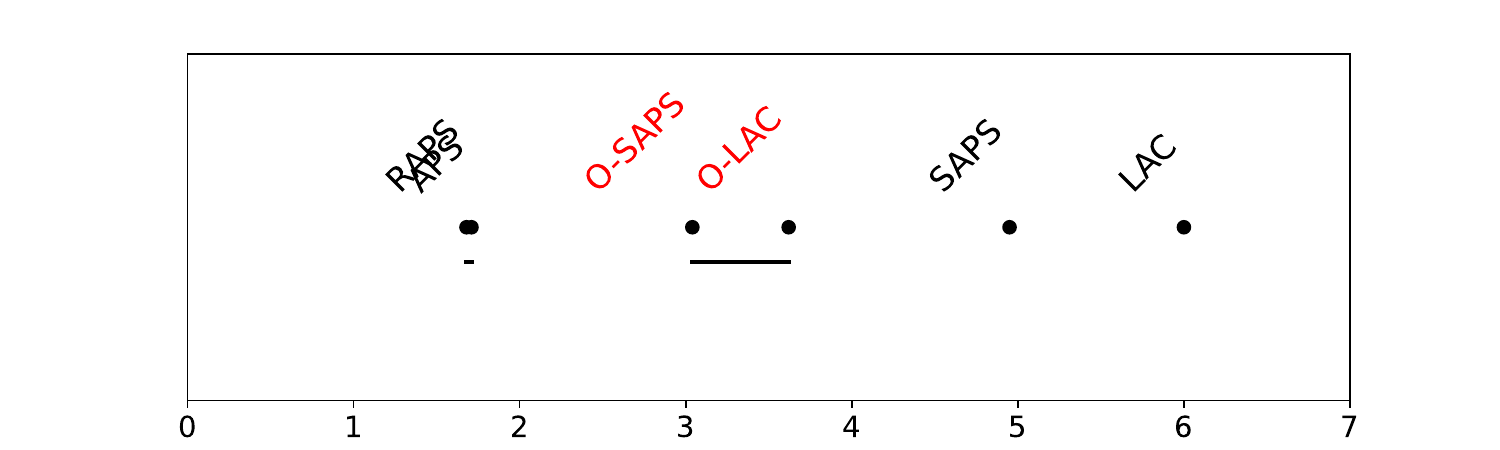}
        \caption{ResNet152}
    \end{subfigure}
    \begin{subfigure}[b]{0.45\textwidth}
        \includegraphics[width=\textwidth]{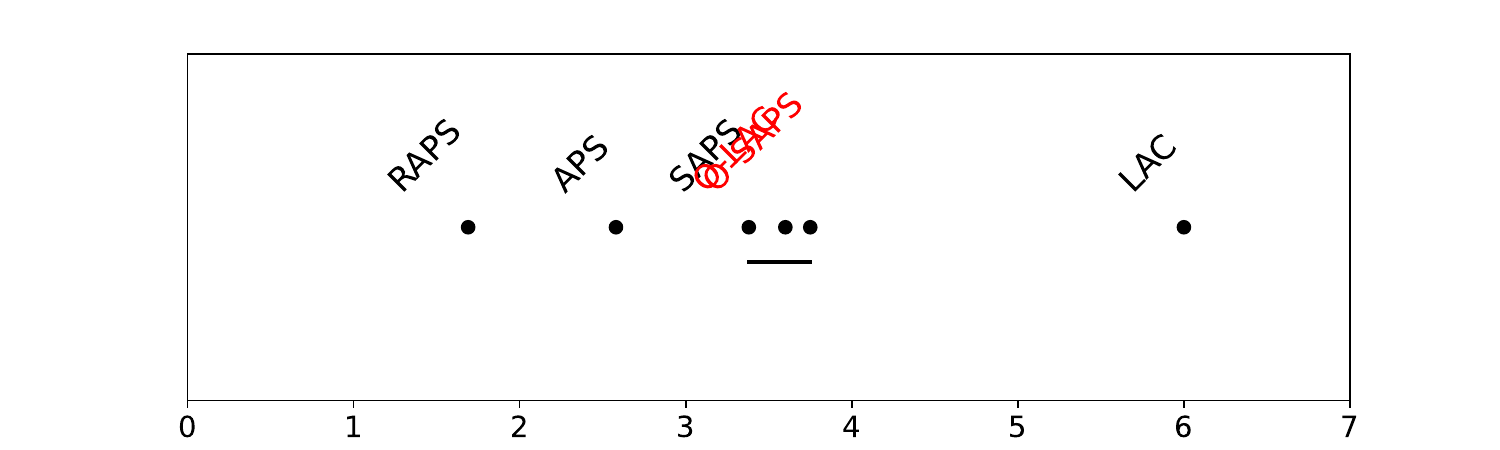}
        \caption{ViT-B-16}
    \end{subfigure}
    \begin{subfigure}[b]{0.45\textwidth}
        \includegraphics[width=\textwidth]{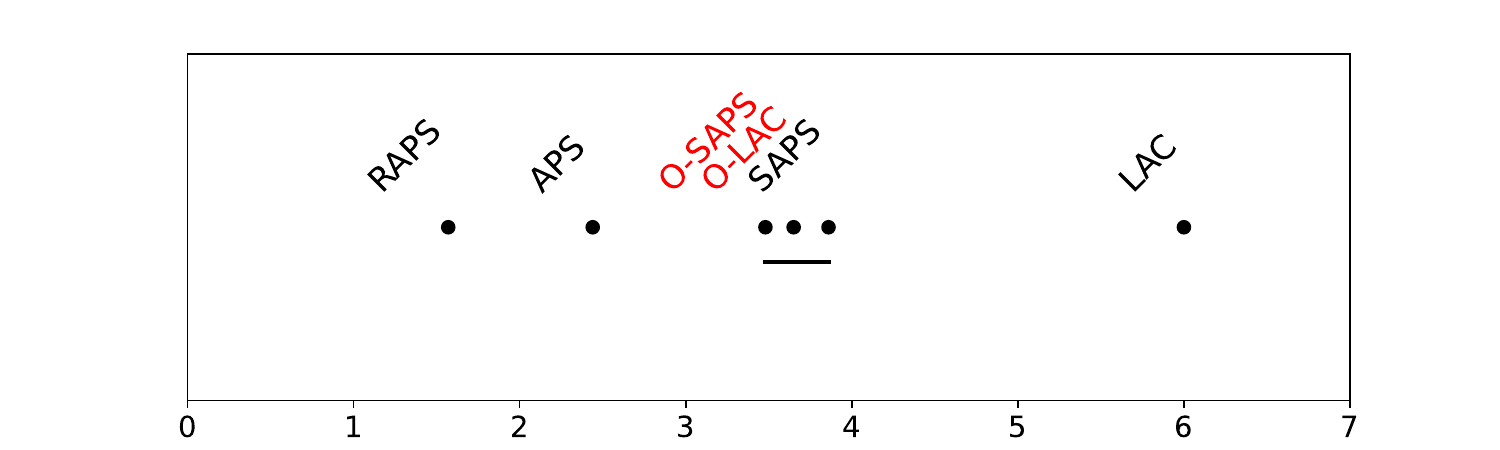}
        \caption{ViT-L-16}
    \end{subfigure}
    \begin{subfigure}[b]{0.45\textwidth}
        \includegraphics[width=\textwidth]{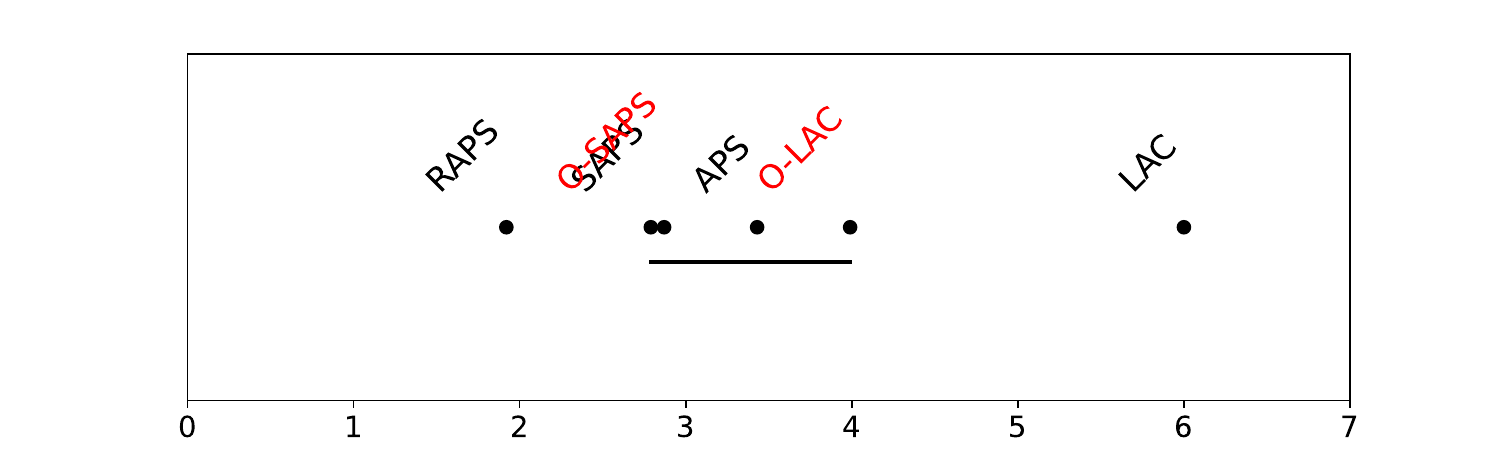}
        \caption{ViT-H-14}
    \end{subfigure}
    \begin{subfigure}[b]{0.45\textwidth}
        \includegraphics[width=\textwidth]{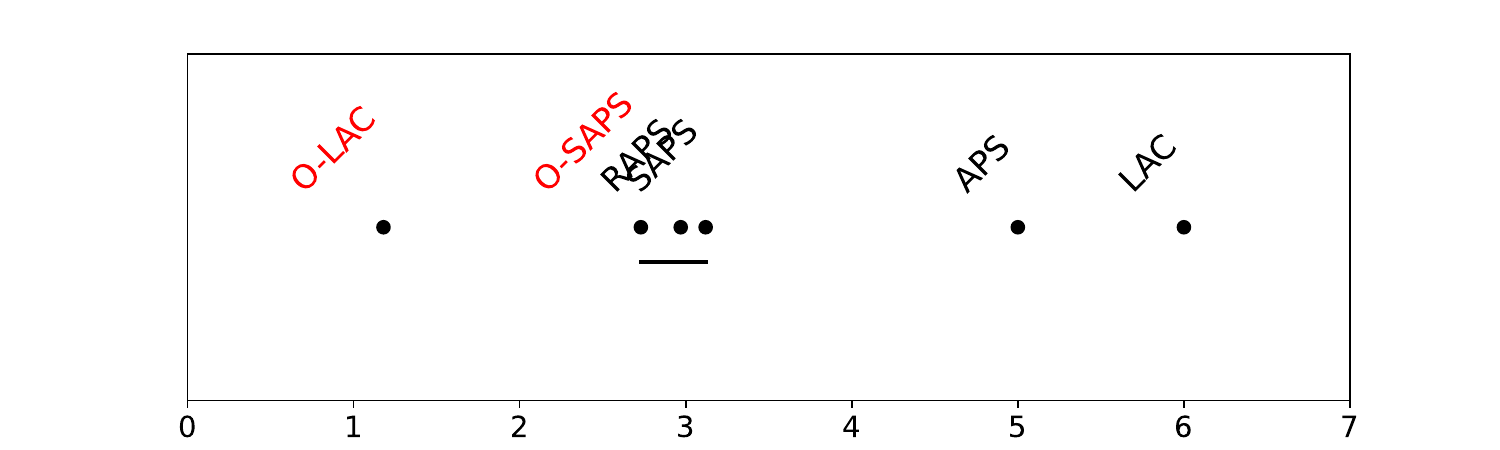}
        \caption{EfficientNet-V2-M}
    \end{subfigure}
    \begin{subfigure}[b]{0.45\textwidth}
        \includegraphics[width=\textwidth]{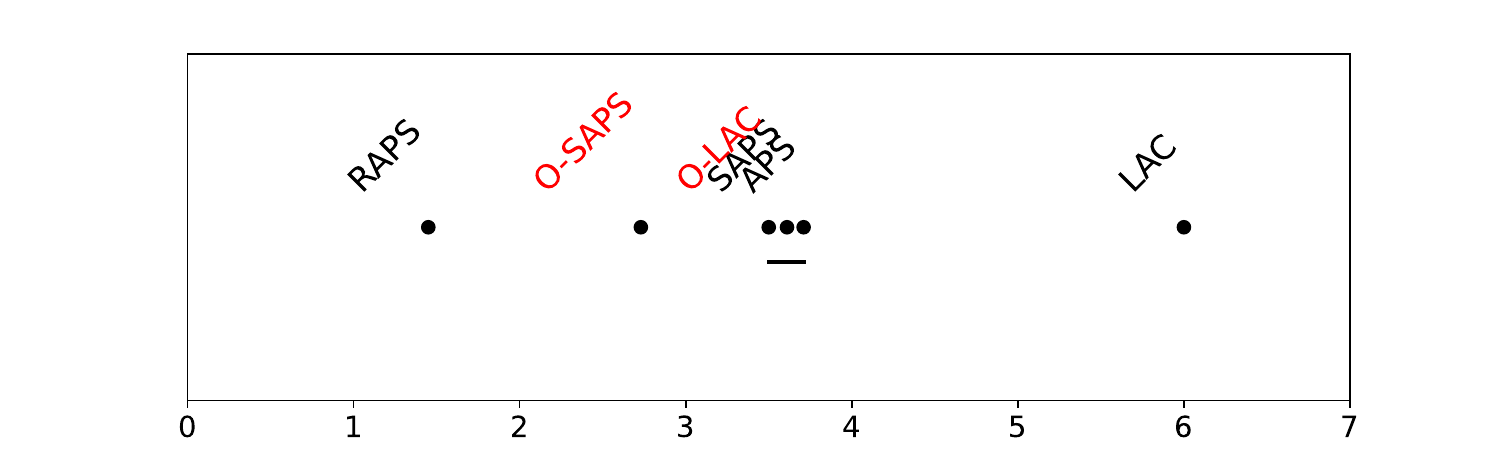}
        \caption{EfficientNet-V2-L}
    \end{subfigure}
    \begin{subfigure}[b]{0.45\textwidth}
        \includegraphics[width=\textwidth]{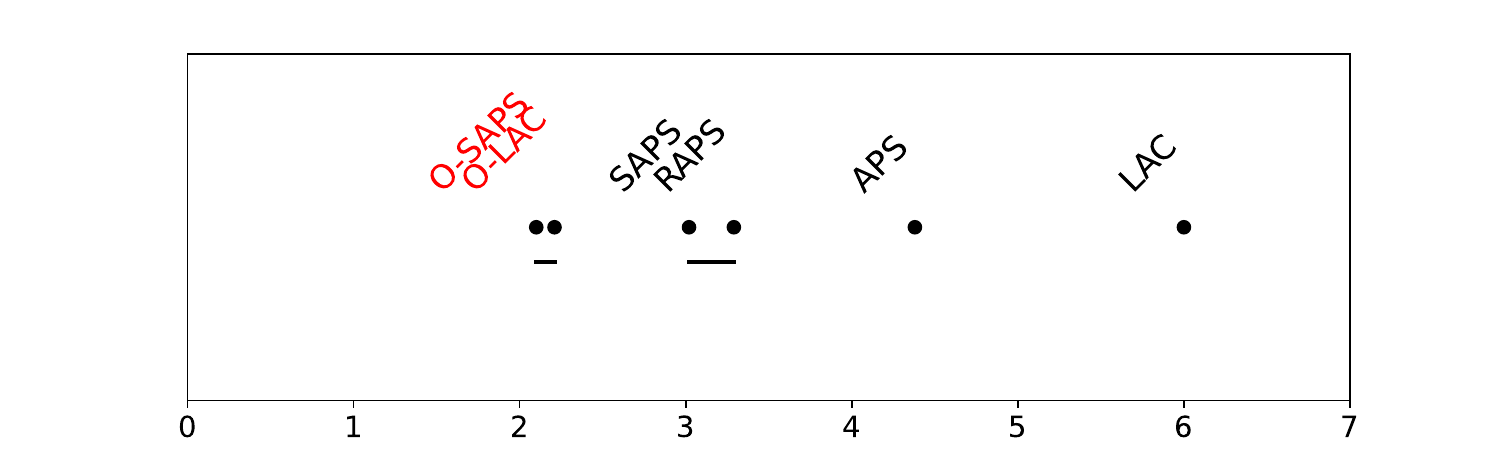}
        \caption{Swin-V2-B}
    \end{subfigure}
    \caption{Critical Difference Diagrams. T-SS, $\alpha=0.20, B=50$. The rank analysis based on these figures is summarized as `Avg. Rank from CD' in Table~\ref{tab:apdx_alg_results_our_metrics} in the main text.}
    \label{fig:apdx_cd_tss_alpha0.20}
\end{figure}

\begin{figure}[!bt]
    \centering
    \begin{subfigure}[b]{0.45\textwidth}
        \includegraphics[width=\textwidth]{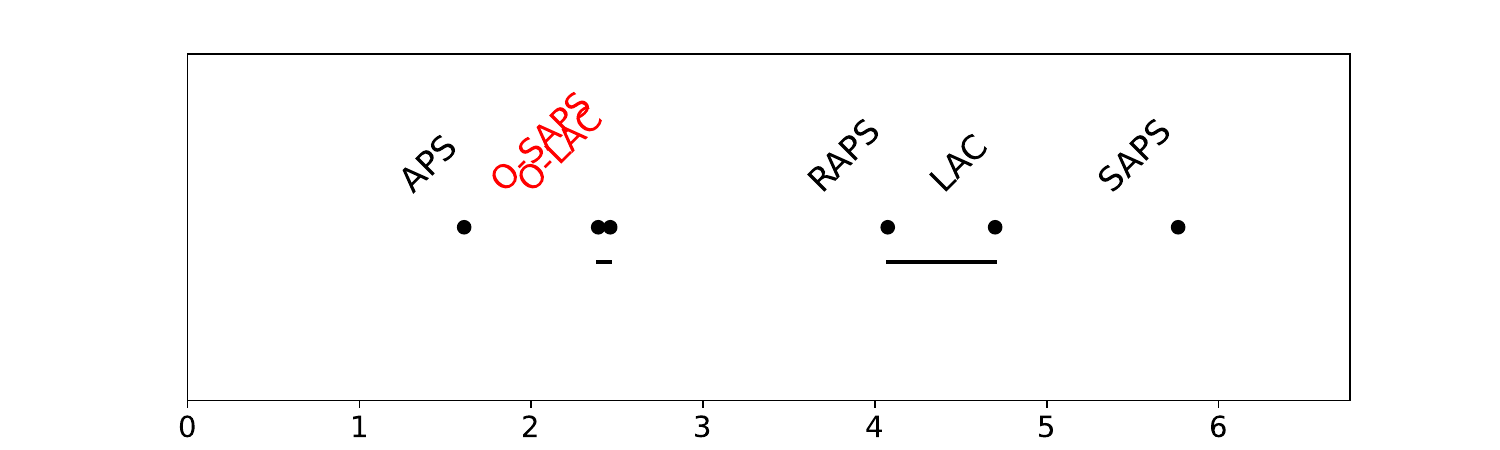}
        \caption{ResNet18}
    \end{subfigure}
    \begin{subfigure}[b]{0.45\textwidth}
        \includegraphics[width=\textwidth]{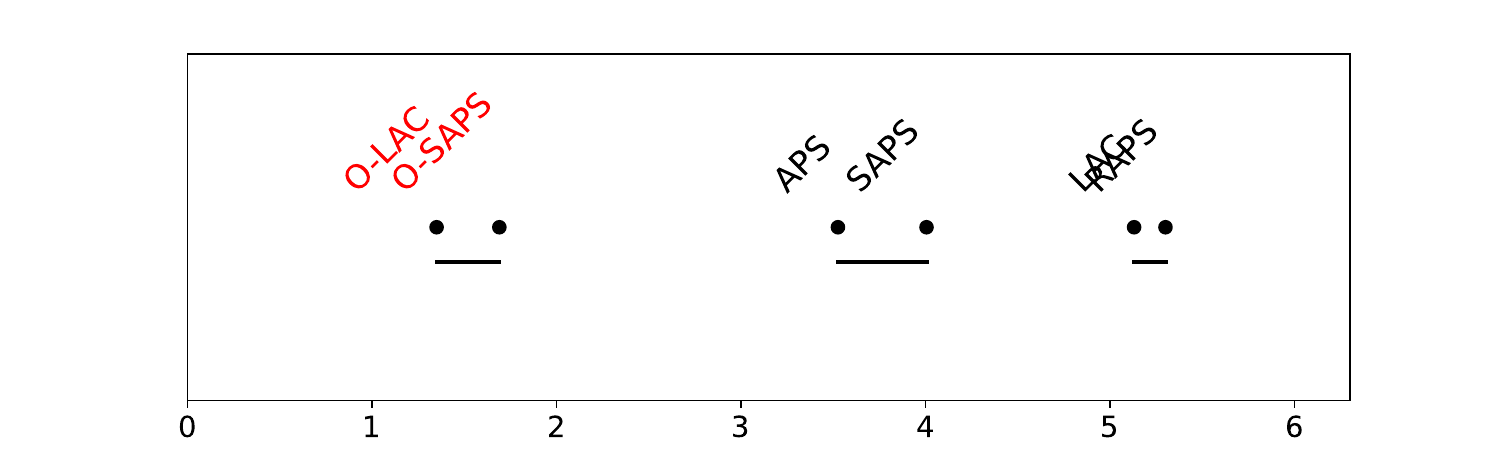}
        \caption{ResNet50}
    \end{subfigure}
    \begin{subfigure}[b]{0.45\textwidth}
        \includegraphics[width=\textwidth]{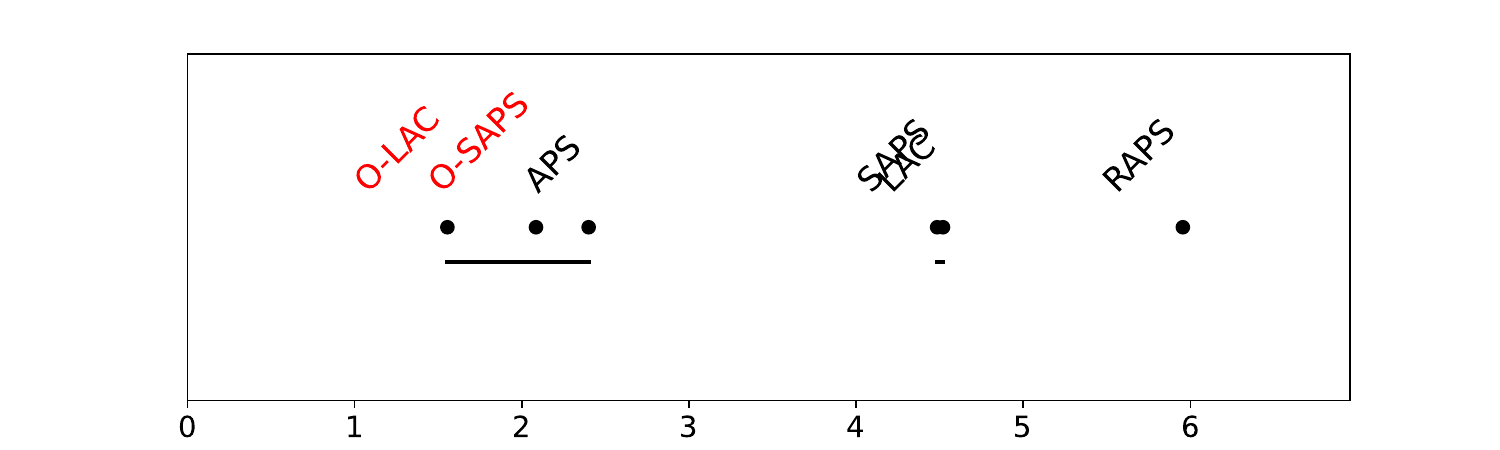}
        \caption{ResNet152}
    \end{subfigure}
    \begin{subfigure}[b]{0.45\textwidth}
        \includegraphics[width=\textwidth]{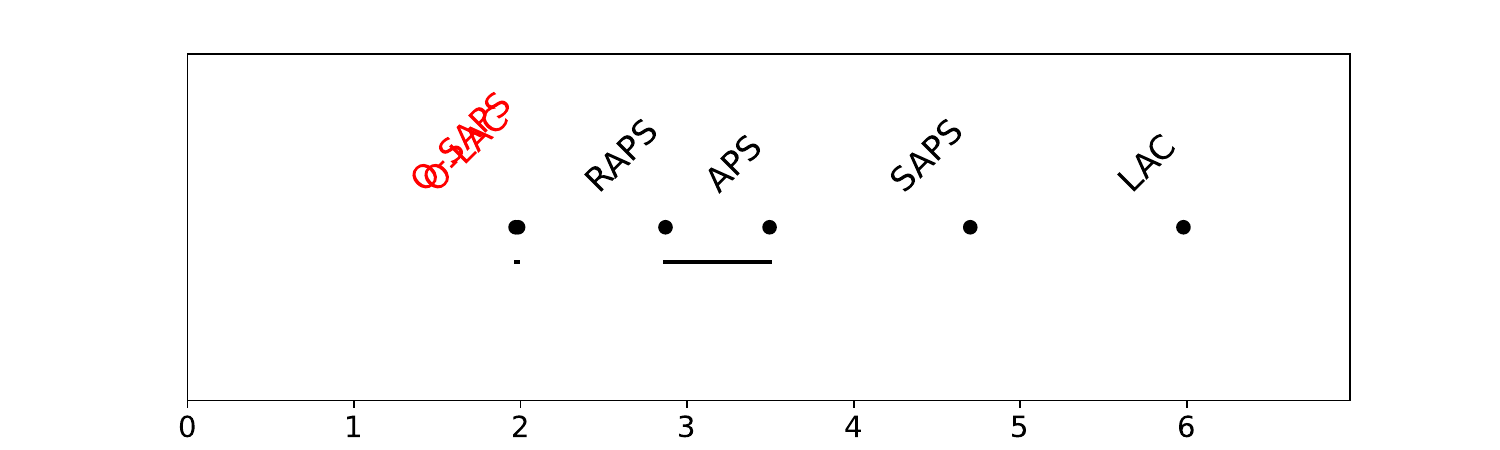}
        \caption{ViT-B-16}
    \end{subfigure}
    \begin{subfigure}[b]{0.45\textwidth}
        \includegraphics[width=\textwidth]{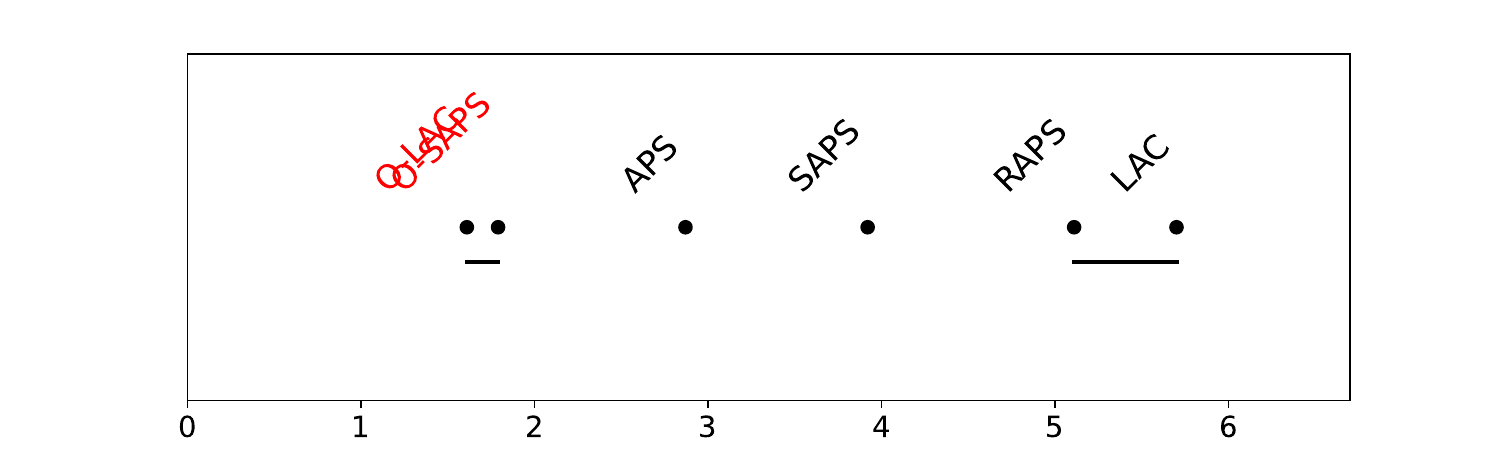}
        \caption{ViT-L-16}
    \end{subfigure}
    \begin{subfigure}[b]{0.45\textwidth}
        \includegraphics[width=\textwidth]{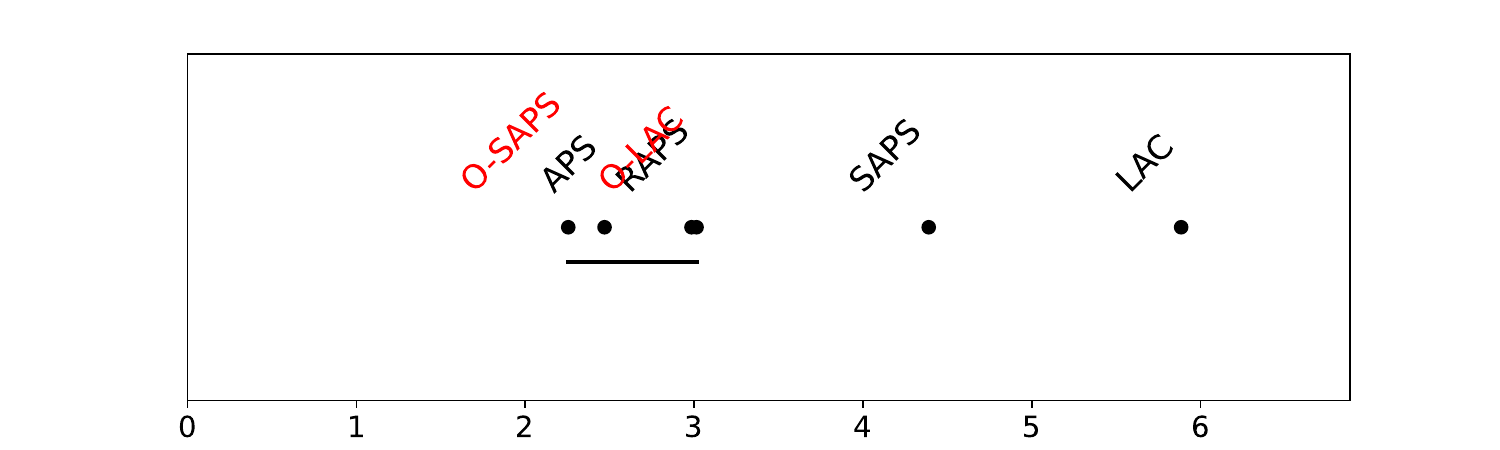}
        \caption{ViT-H-14}
    \end{subfigure}
    \begin{subfigure}[b]{0.45\textwidth}
        \includegraphics[width=\textwidth]{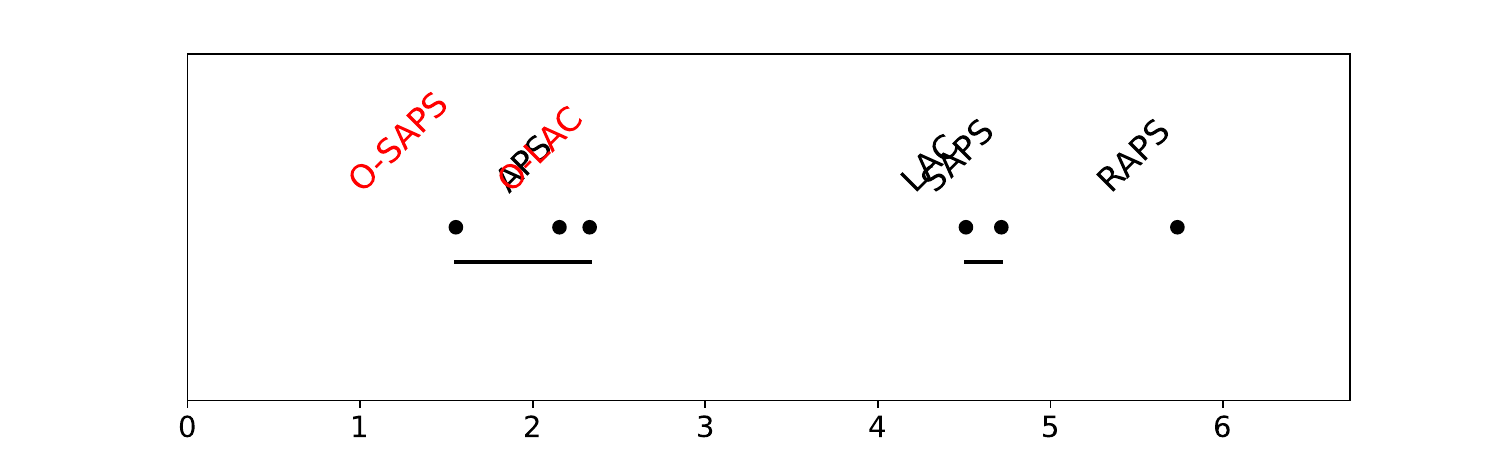}
        \caption{EfficientNet-V2-M}
    \end{subfigure}
    \begin{subfigure}[b]{0.45\textwidth}
        \includegraphics[width=\textwidth]{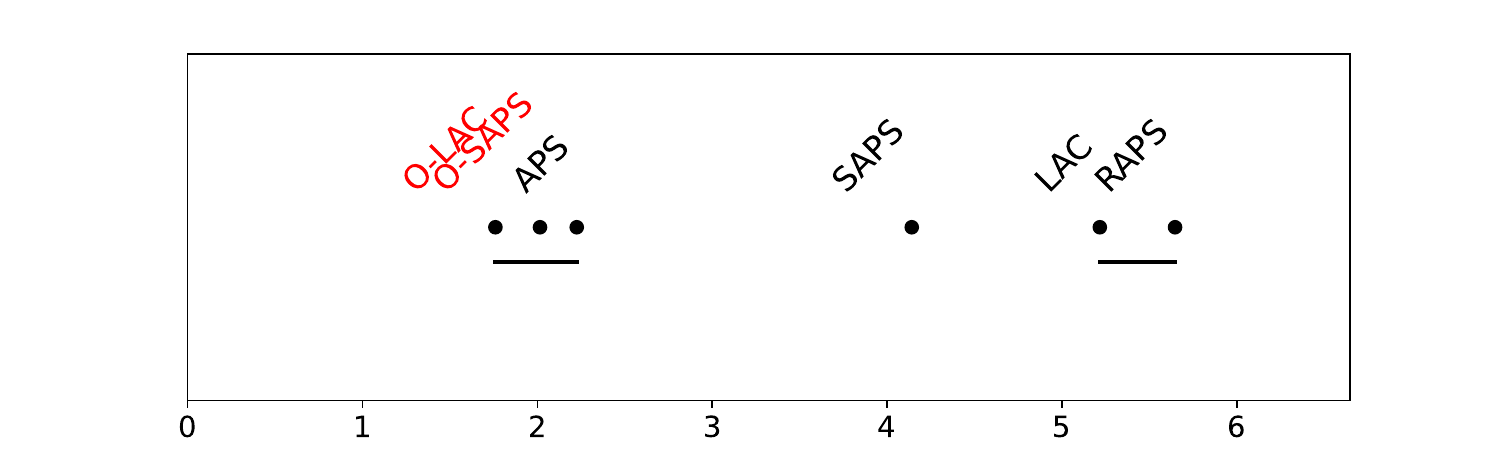}
        \caption{EfficientNet-V2-L}
    \end{subfigure}
    \begin{subfigure}[b]{0.45\textwidth}
        \includegraphics[width=\textwidth]{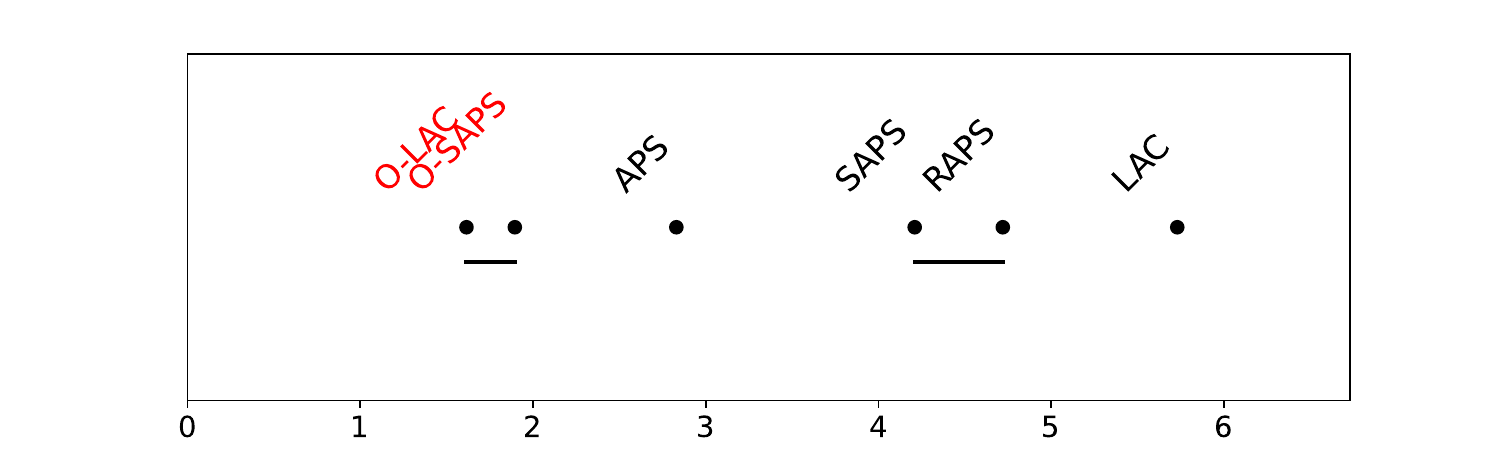}
        \caption{Swin-V2-B}
    \end{subfigure}
    \caption{Critical Difference Diagrams. T-CV, $\alpha=0.05, B=100$. The rank analysis based on these figures is summarized as `Avg. Rank from CD' in Table~\ref{tab:apdx_alg_results_our_metrics_B100} in Appendix.}
    \label{fig:apdx_cd_tcv_B100_alpha0.05}
\end{figure}

\begin{figure}[!bt]
    \centering
    \begin{subfigure}[b]{0.45\textwidth}
        \includegraphics[width=\textwidth]{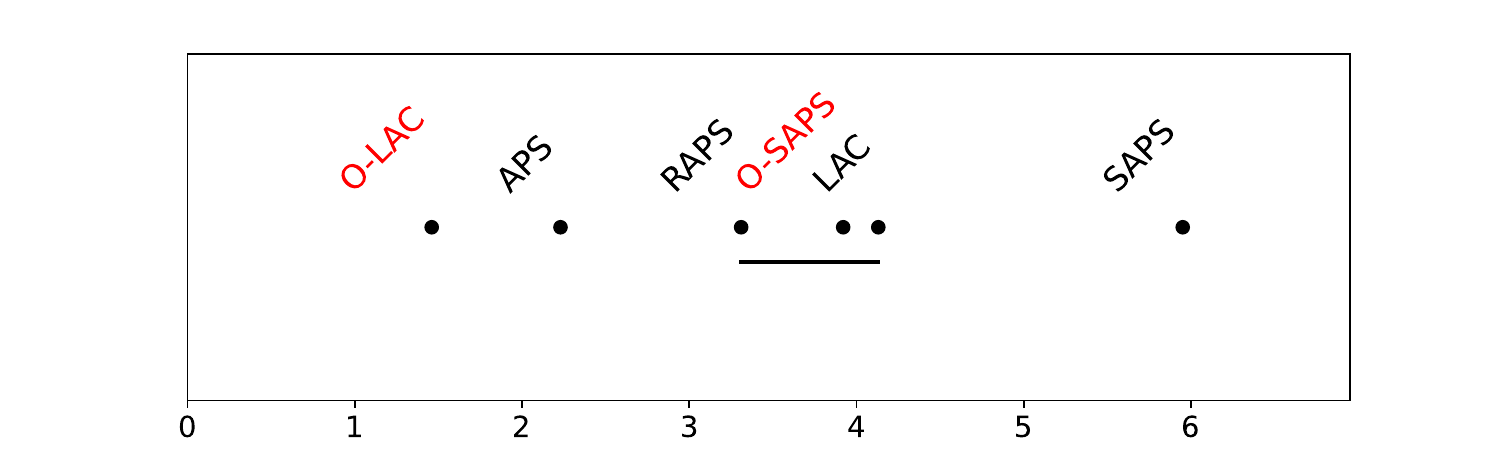}
        \caption{ResNet18}
    \end{subfigure}
    \begin{subfigure}[b]{0.45\textwidth}
        \includegraphics[width=\textwidth]{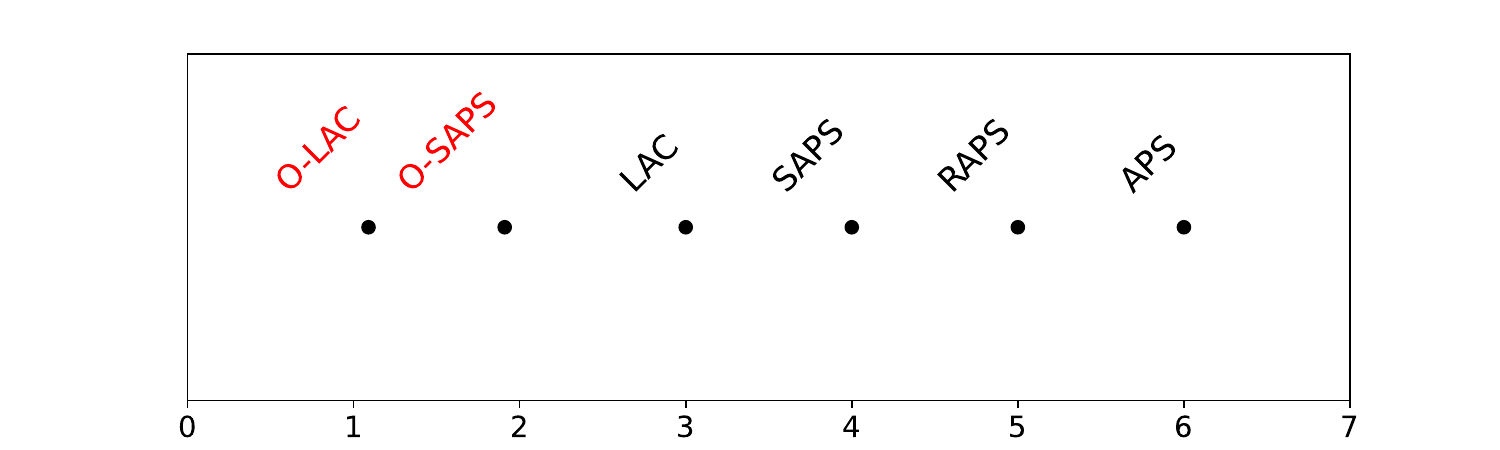}
        \caption{ResNet50}
    \end{subfigure}
    \begin{subfigure}[b]{0.45\textwidth}
        \includegraphics[width=\textwidth]{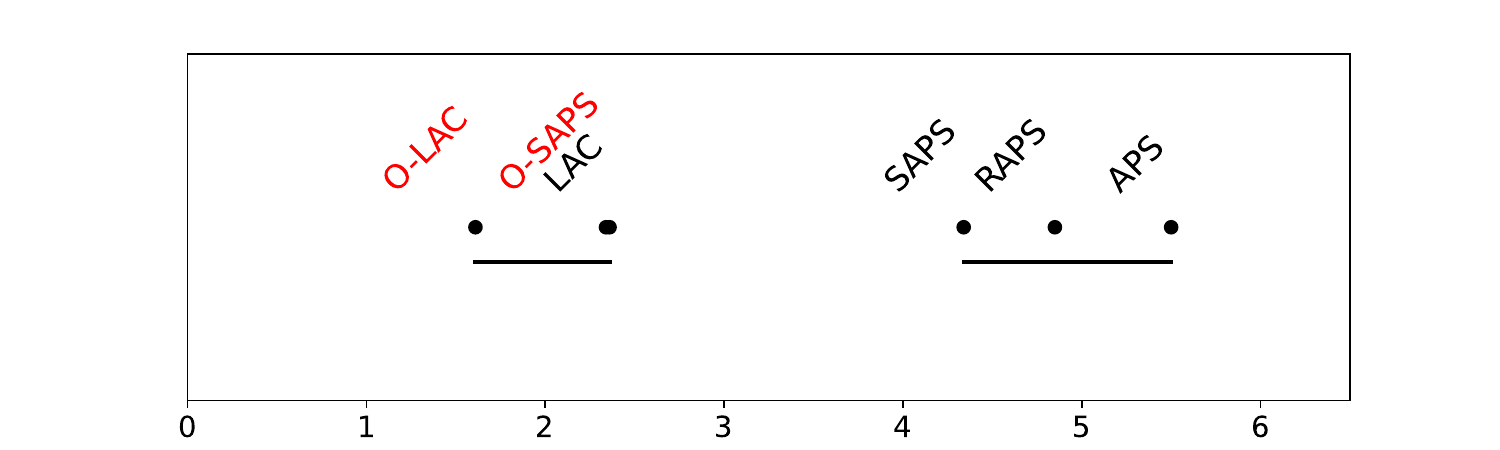}
        \caption{ResNet152}
    \end{subfigure}
    \begin{subfigure}[b]{0.45\textwidth}
        \includegraphics[width=\textwidth]{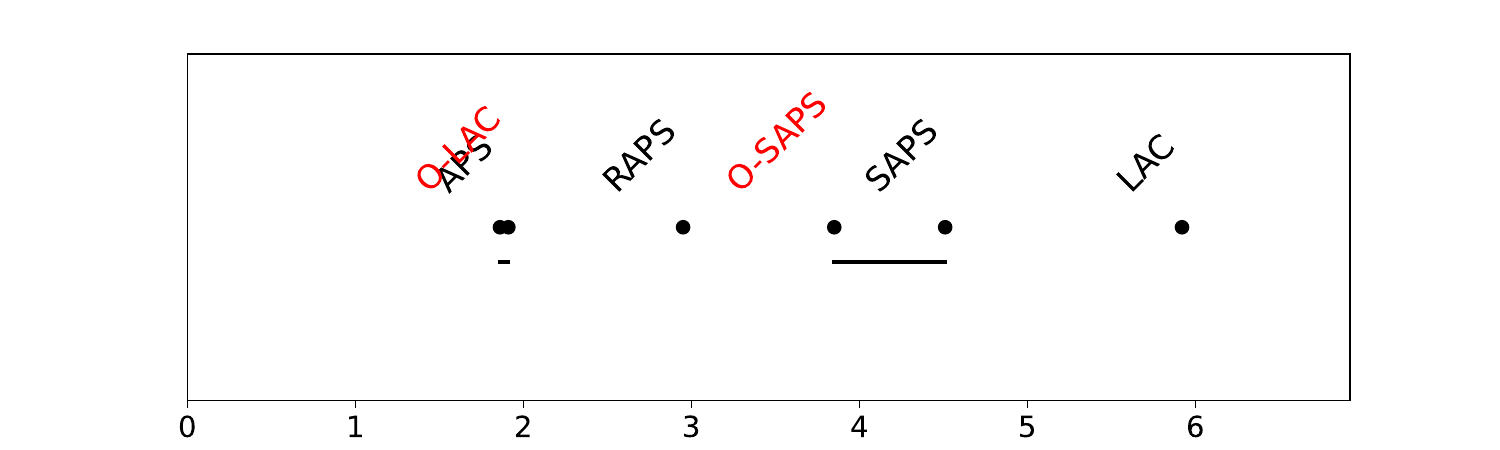}
        \caption{ViT-B-16}
    \end{subfigure}
    \begin{subfigure}[b]{0.45\textwidth}
        \includegraphics[width=\textwidth]{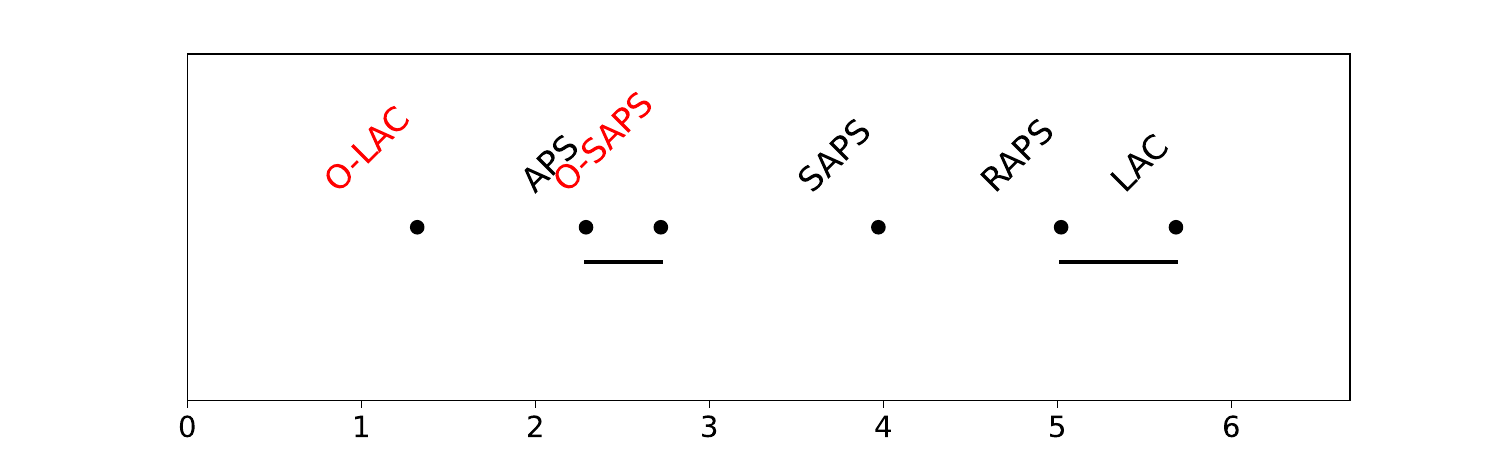}
        \caption{ViT-L-16}
    \end{subfigure}
    \begin{subfigure}[b]{0.45\textwidth}
        \includegraphics[width=\textwidth]{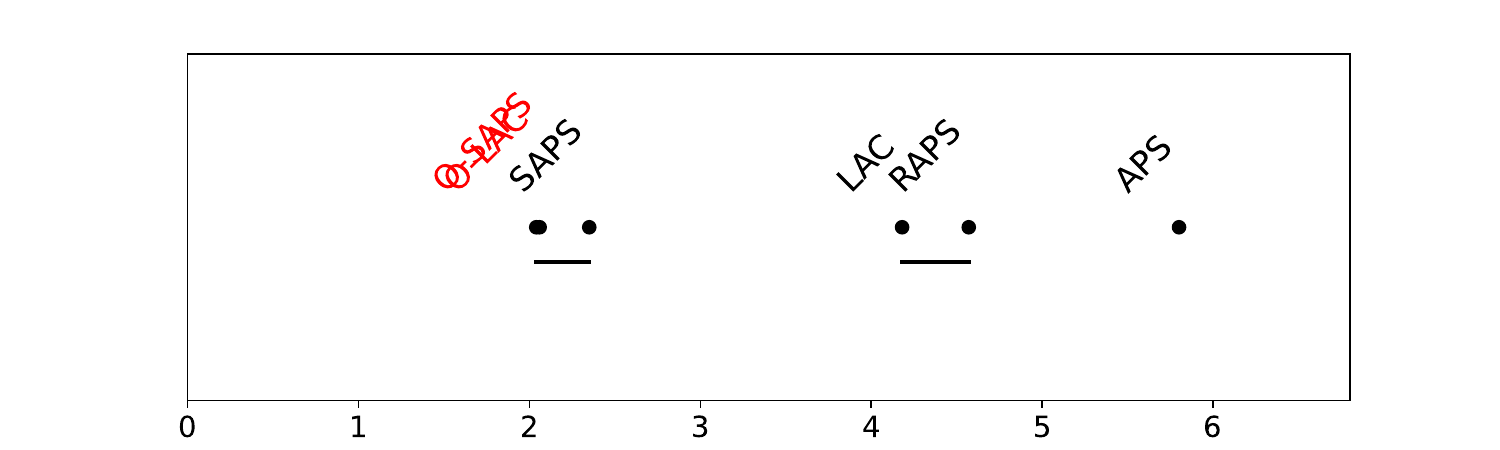}
        \caption{ViT-H-14}
    \end{subfigure}
    \begin{subfigure}[b]{0.45\textwidth}
        \includegraphics[width=\textwidth]{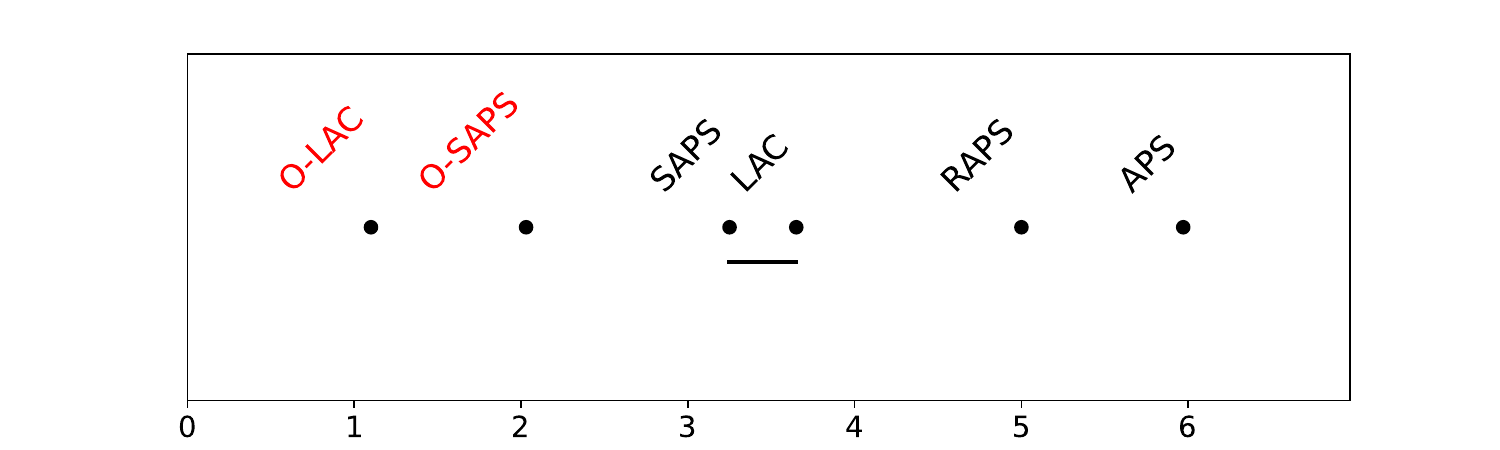}
        \caption{EfficientNet-V2-M}
    \end{subfigure}
    \begin{subfigure}[b]{0.45\textwidth}
        \includegraphics[width=\textwidth]{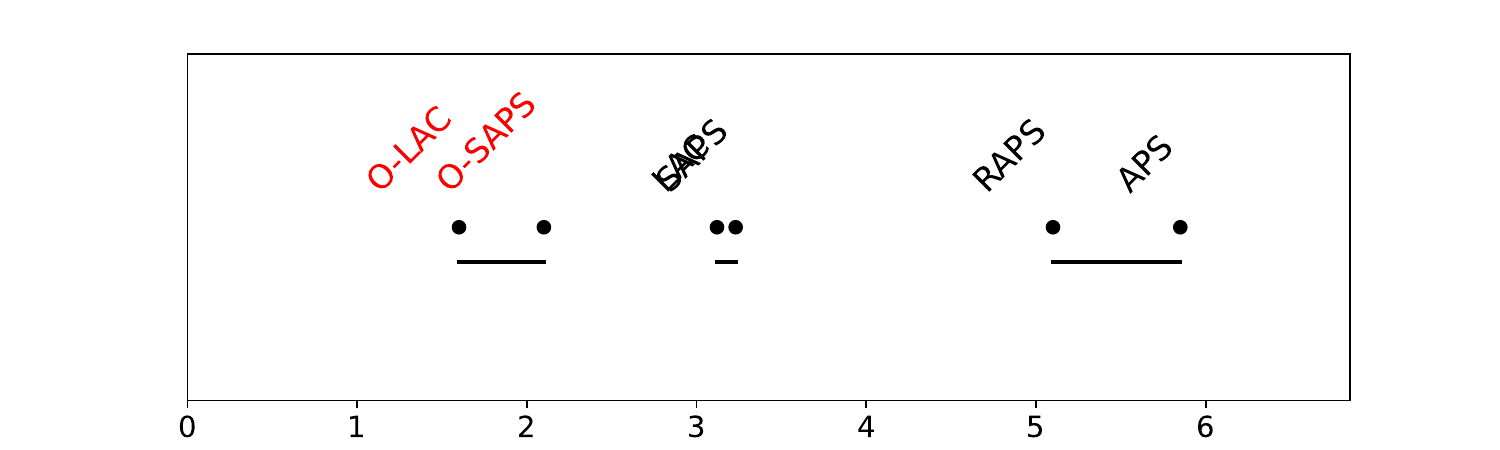}
        \caption{EfficientNet-V2-L}
    \end{subfigure}
    \begin{subfigure}[b]{0.45\textwidth}
        \includegraphics[width=\textwidth]{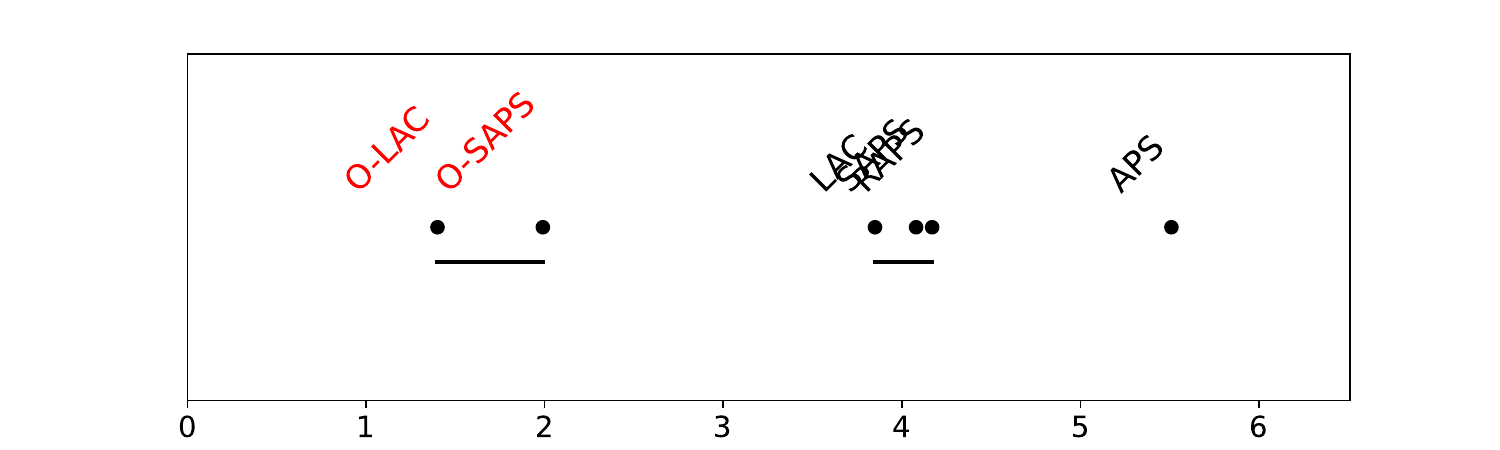}
        \caption{Swin-V2-B}
    \end{subfigure}
    \caption{Critical Difference Diagrams. T-SS, $\alpha=0.05, B=100$. The rank analysis based on these figures is summarized as `Avg. Rank from CD' in Table~\ref{tab:apdx_alg_results_our_metrics_B100} in Appendix.}
    \label{fig:apdx_cd_tss_B100_alpha0.05}
\end{figure}

\begin{figure}[!bt]
    \centering
    \begin{subfigure}[b]{0.45\textwidth}
        \includegraphics[width=\textwidth]{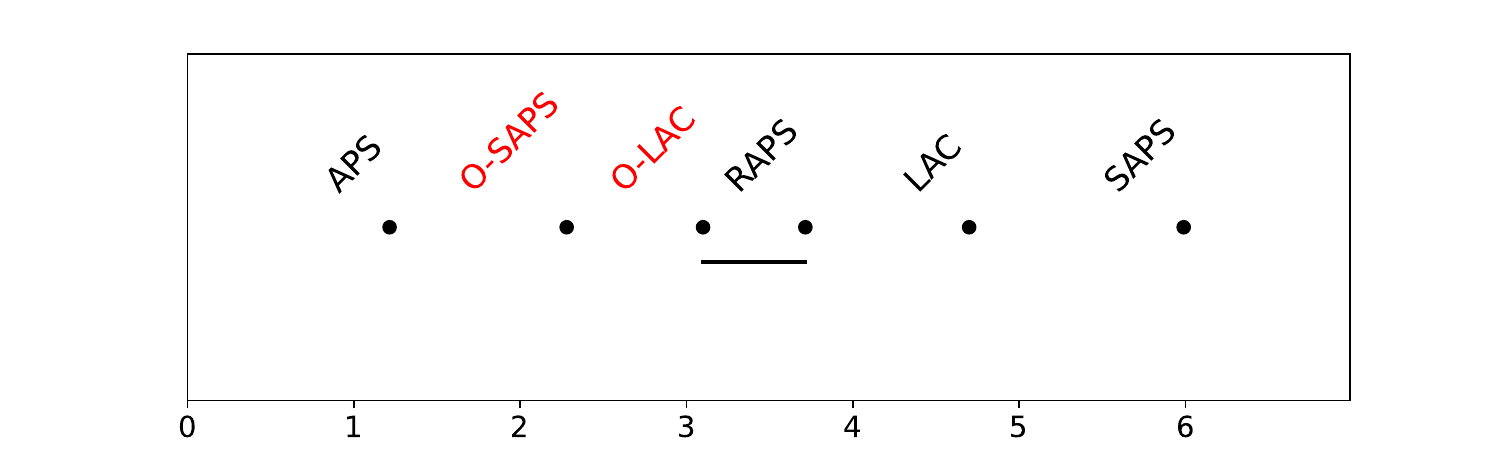}
        \caption{ResNet18}
    \end{subfigure}
    \begin{subfigure}[b]{0.45\textwidth}
        \includegraphics[width=\textwidth]{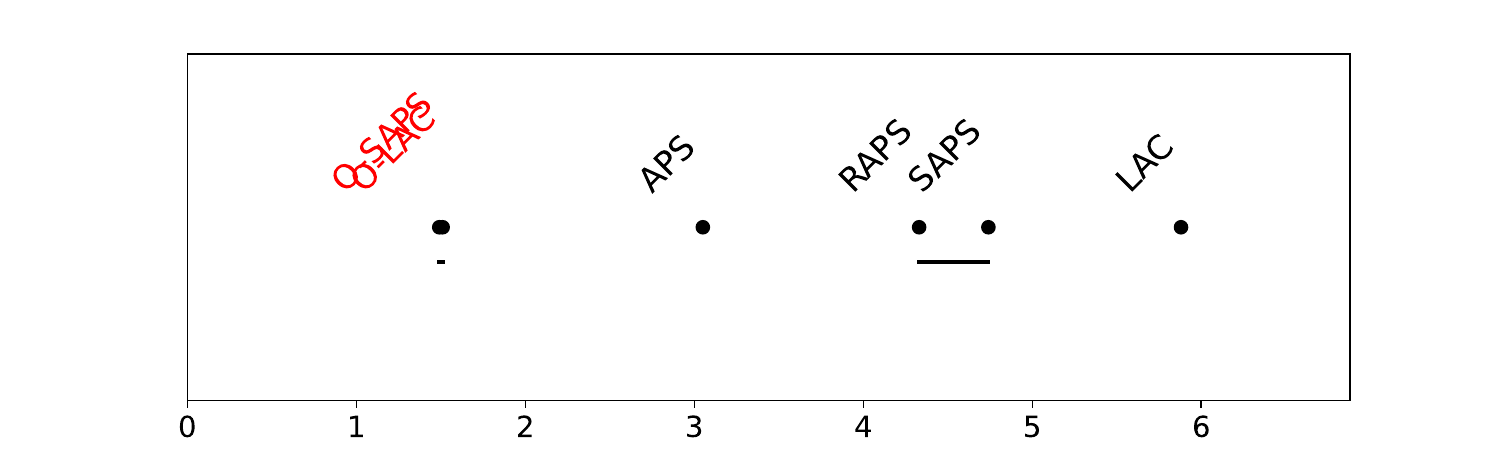}
        \caption{ResNet50}
    \end{subfigure}
    \begin{subfigure}[b]{0.45\textwidth}
        \includegraphics[width=\textwidth]{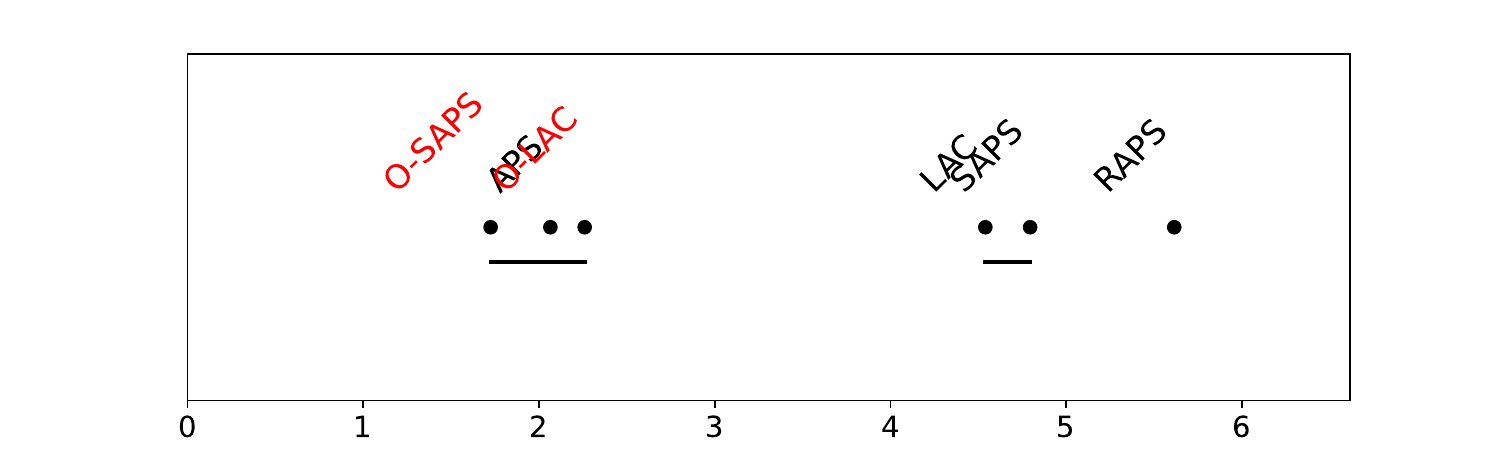}
        \caption{ResNet152}
    \end{subfigure}
    \begin{subfigure}[b]{0.45\textwidth}
        \includegraphics[width=\textwidth]{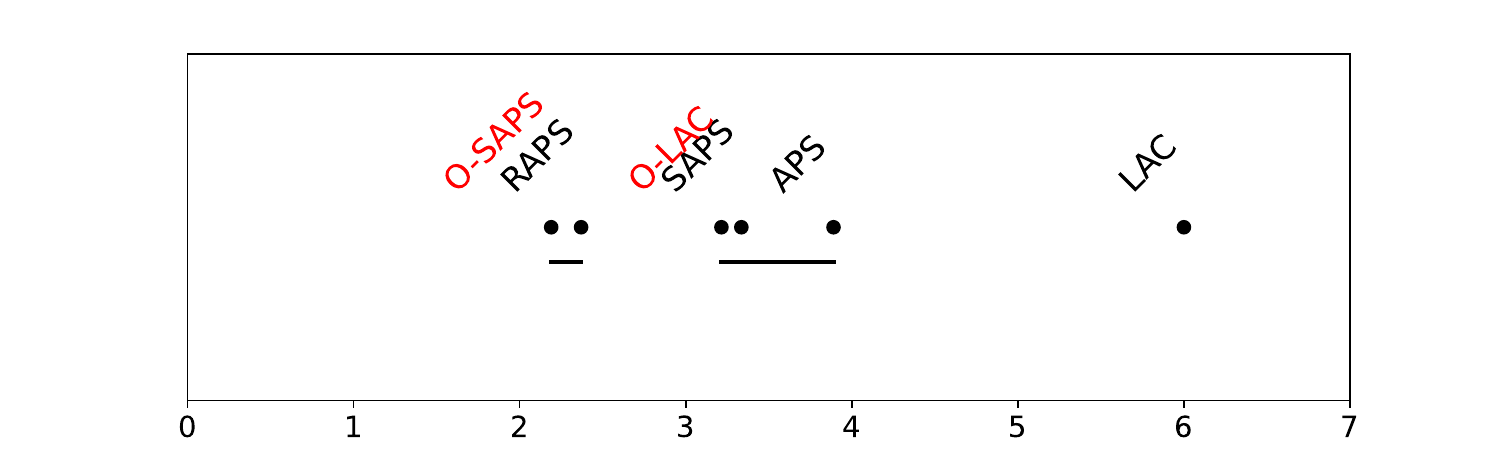}
        \caption{ViT-B-16}
    \end{subfigure}
    \begin{subfigure}[b]{0.45\textwidth}
        \includegraphics[width=\textwidth]{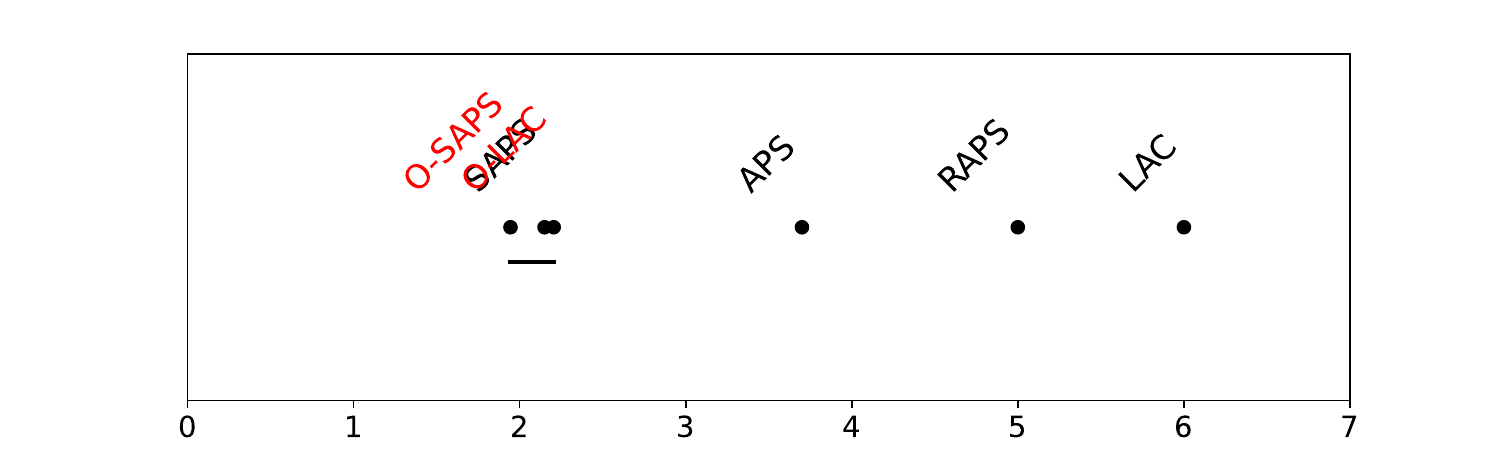}
        \caption{ViT-L-16}
    \end{subfigure}
    \begin{subfigure}[b]{0.45\textwidth}
        \includegraphics[width=\textwidth]{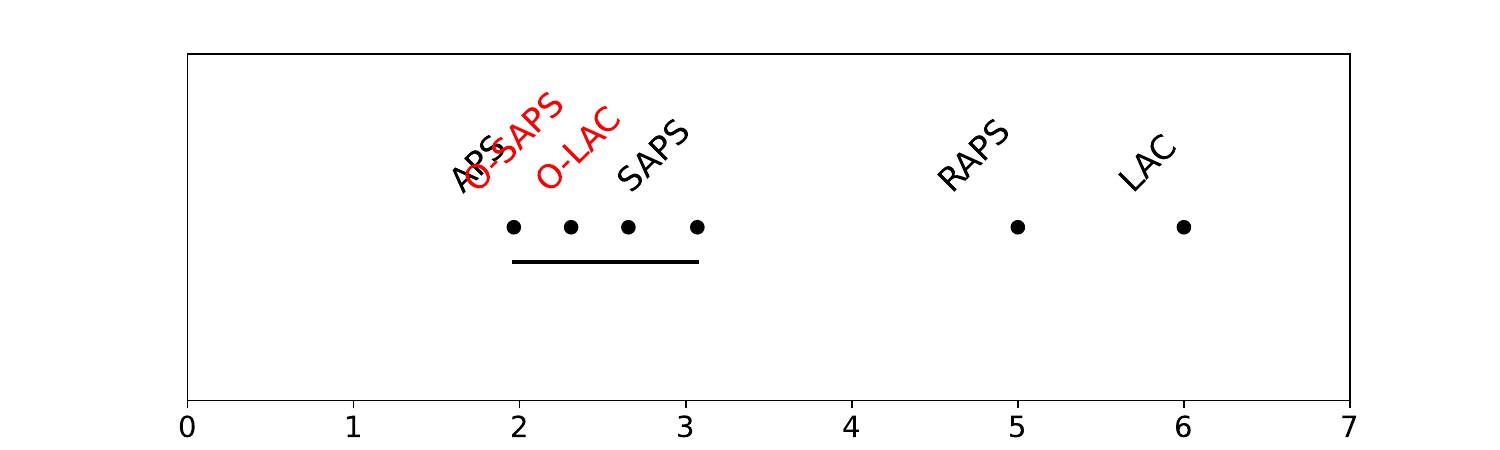}
        \caption{ViT-H-14}
    \end{subfigure}
    \begin{subfigure}[b]{0.45\textwidth}
        \includegraphics[width=\textwidth]{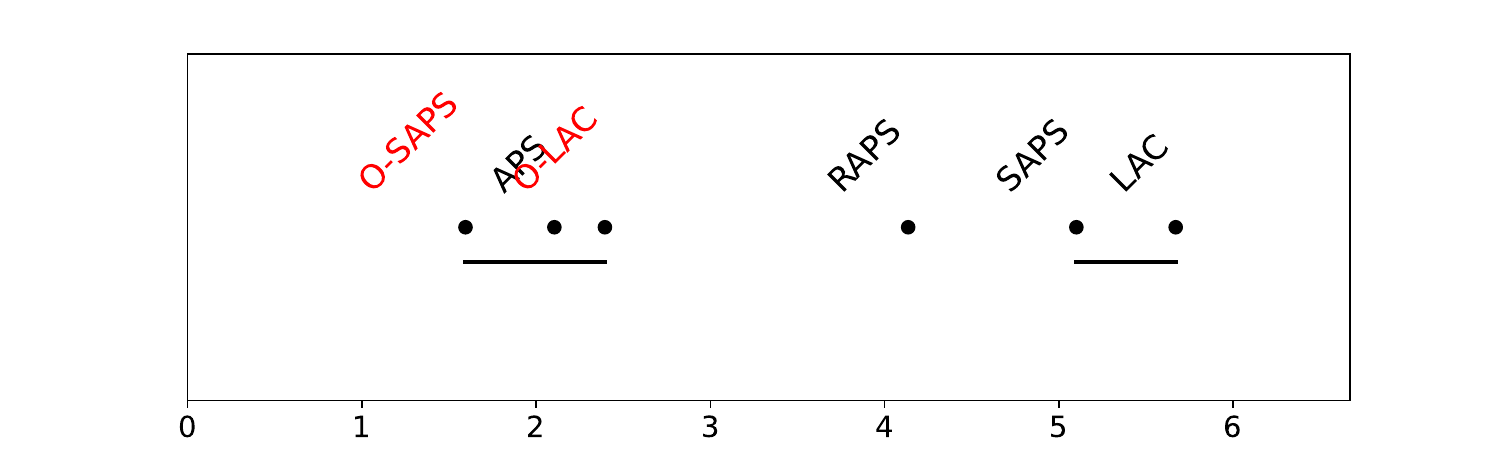}
        \caption{EfficientNet-V2-M}
    \end{subfigure}
    \begin{subfigure}[b]{0.45\textwidth}
        \includegraphics[width=\textwidth]{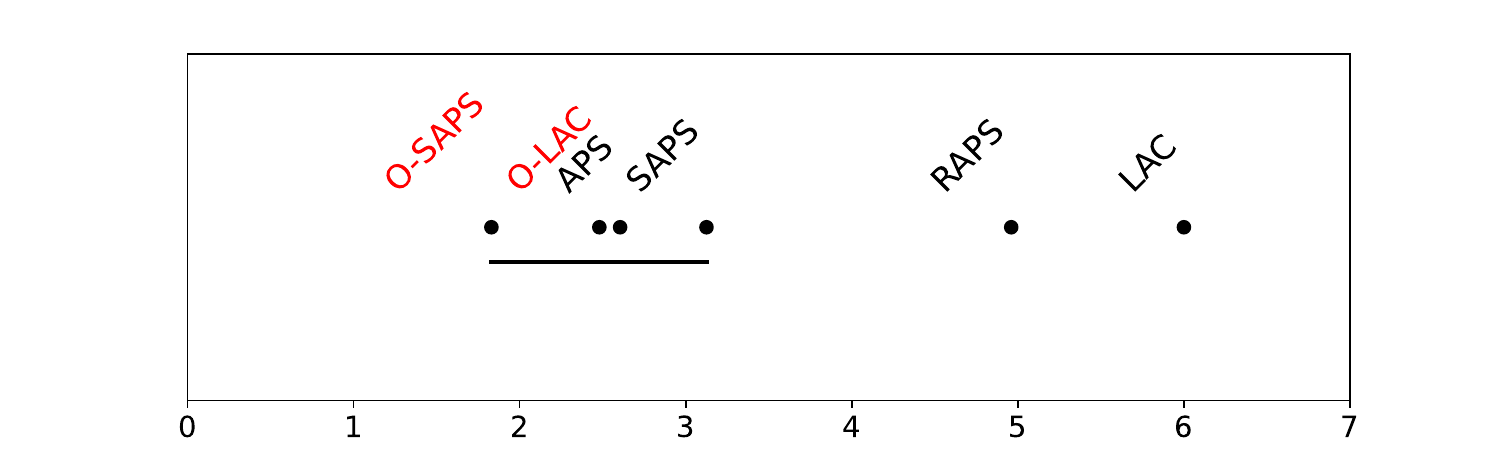}
        \caption{EfficientNet-V2-L}
    \end{subfigure}
    \begin{subfigure}[b]{0.45\textwidth}
        \includegraphics[width=\textwidth]{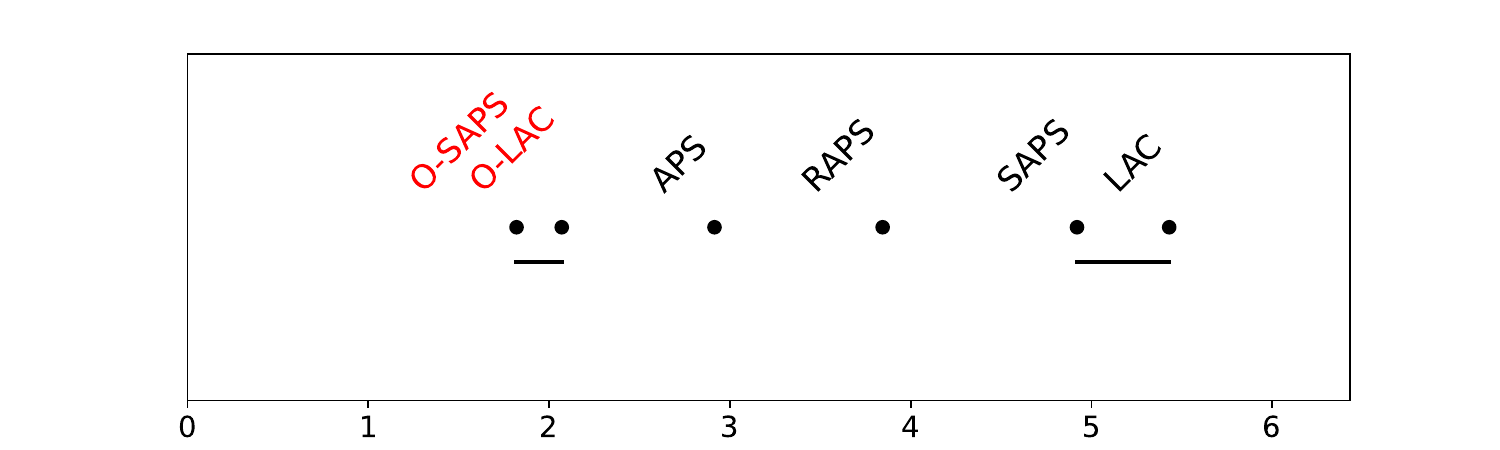}
        \caption{Swin-V2-B}
    \end{subfigure}
    \caption{Critical Difference Diagrams. T-CV, $\alpha=0.10, B=100$. The rank analysis based on these figures is summarized as `Avg. Rank from CD' in Table~\ref{tab:apdx_alg_results_our_metrics_B100} in Appendix.}
    \label{fig:apdx_cd_tcv_B100_alpha0.10}
\end{figure}

\begin{figure}[!bt]
    \centering
    \begin{subfigure}[b]{0.45\textwidth}
        \includegraphics[width=\textwidth]{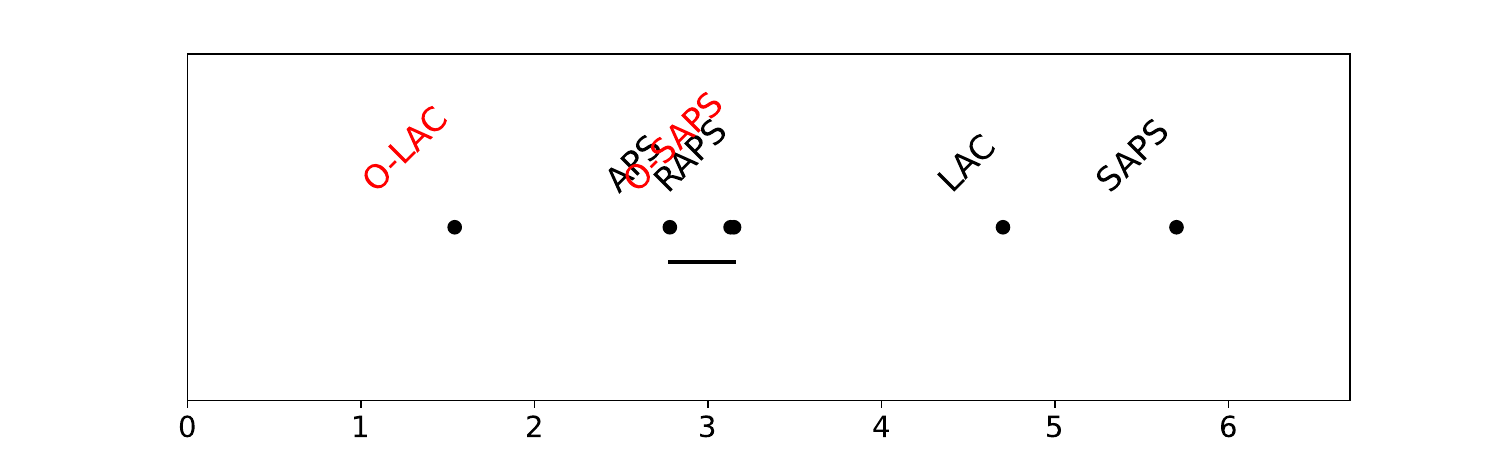}
        \caption{ResNet18}
    \end{subfigure}
    \begin{subfigure}[b]{0.45\textwidth}
        \includegraphics[width=\textwidth]{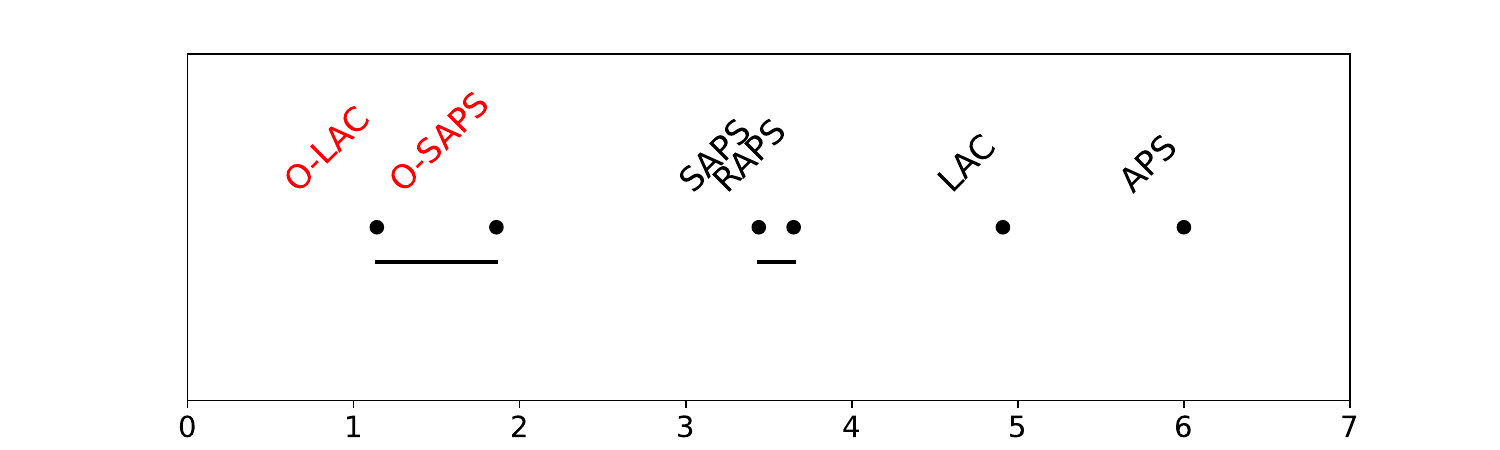}
        \caption{ResNet50}
    \end{subfigure}
    \begin{subfigure}[b]{0.45\textwidth}
        \includegraphics[width=\textwidth]{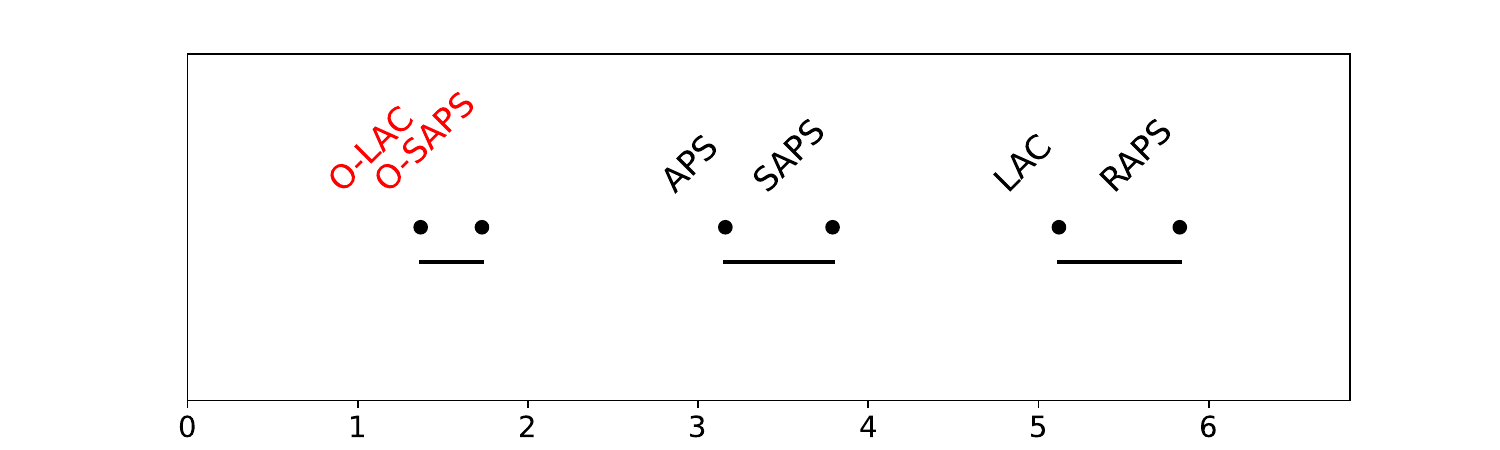}
        \caption{ResNet152}
    \end{subfigure}
    \begin{subfigure}[b]{0.45\textwidth}
        \includegraphics[width=\textwidth]{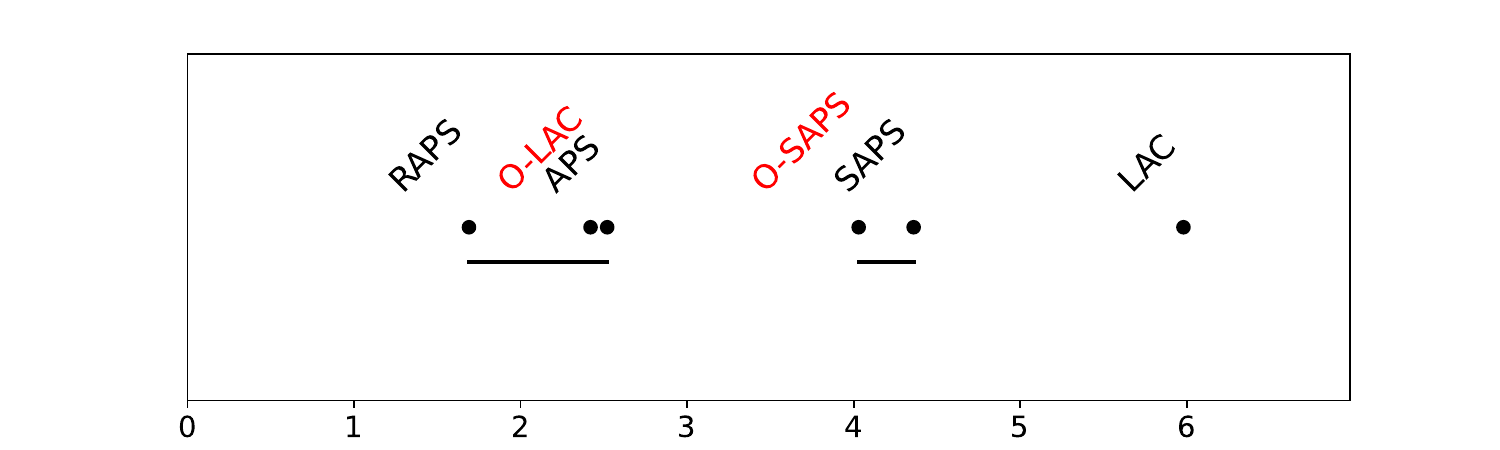}
        \caption{ViT-B-16}
    \end{subfigure}
    \begin{subfigure}[b]{0.45\textwidth}
        \includegraphics[width=\textwidth]{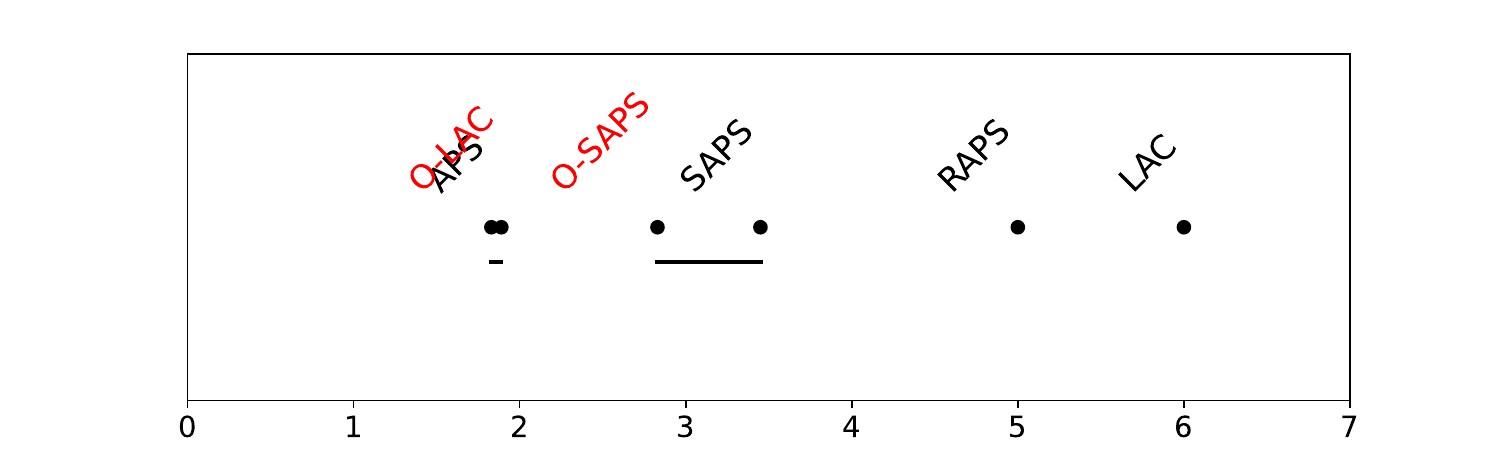}
        \caption{ViT-L-16}
    \end{subfigure}
    \begin{subfigure}[b]{0.45\textwidth}
        \includegraphics[width=\textwidth]{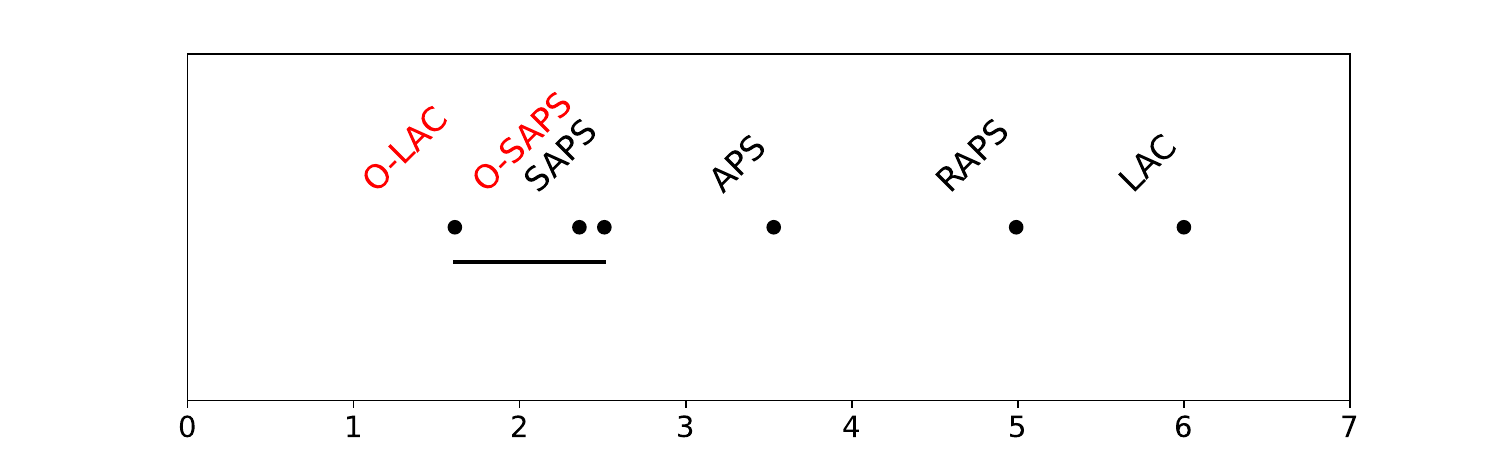}
        \caption{ViT-H-14}
    \end{subfigure}
    \begin{subfigure}[b]{0.45\textwidth}
        \includegraphics[width=\textwidth]{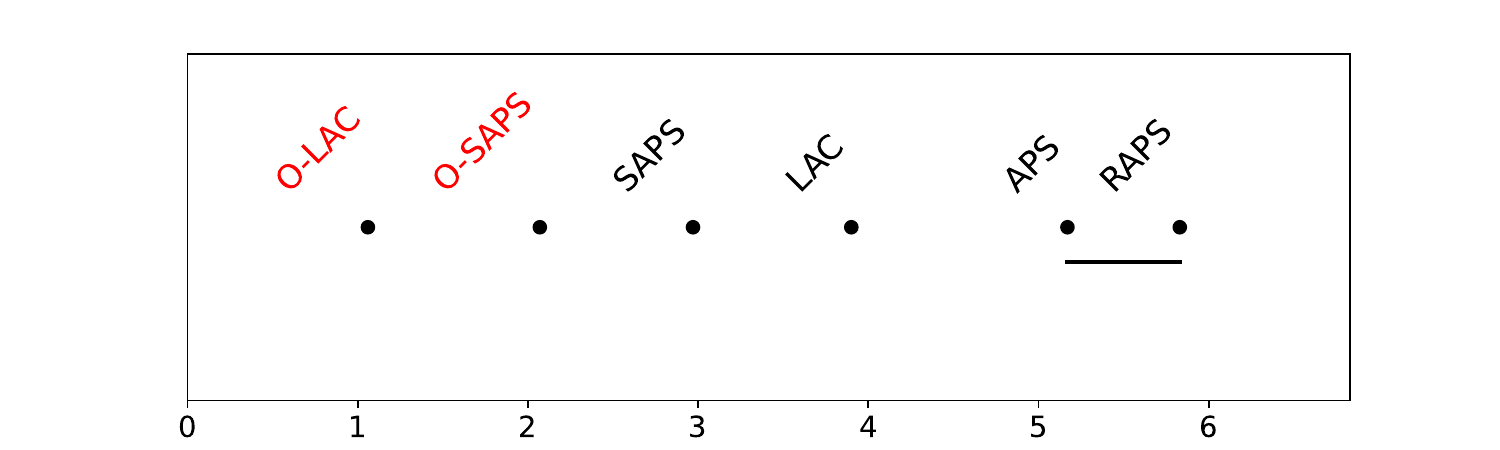}
        \caption{EfficientNet-V2-M}
    \end{subfigure}
    \begin{subfigure}[b]{0.45\textwidth}
        \includegraphics[width=\textwidth]{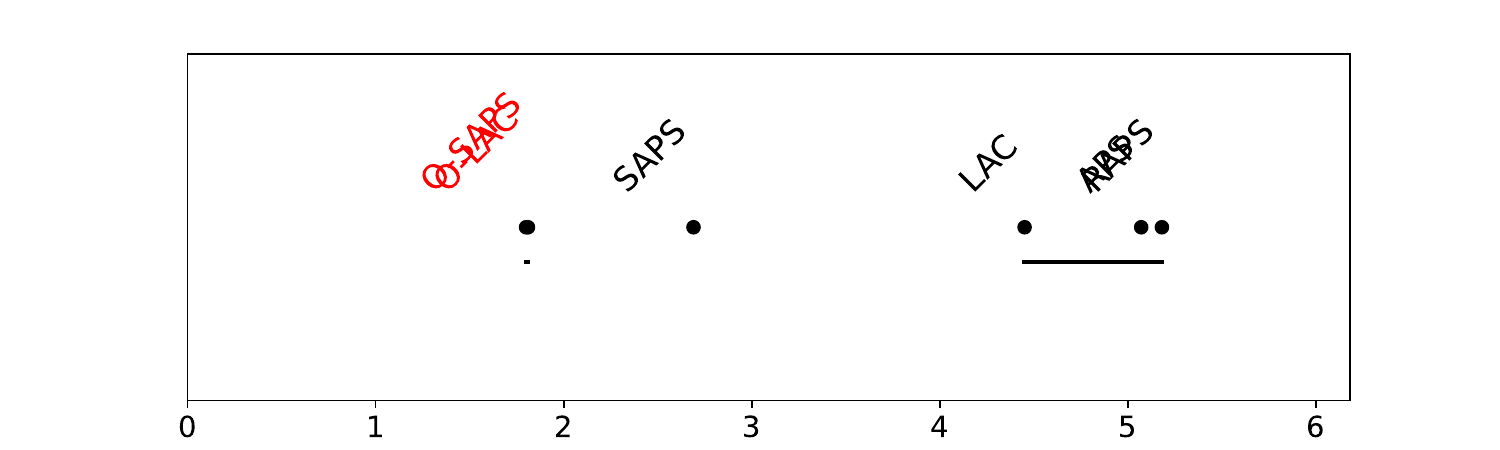}
        \caption{EfficientNet-V2-L}
    \end{subfigure}
    \begin{subfigure}[b]{0.45\textwidth}
        \includegraphics[width=\textwidth]{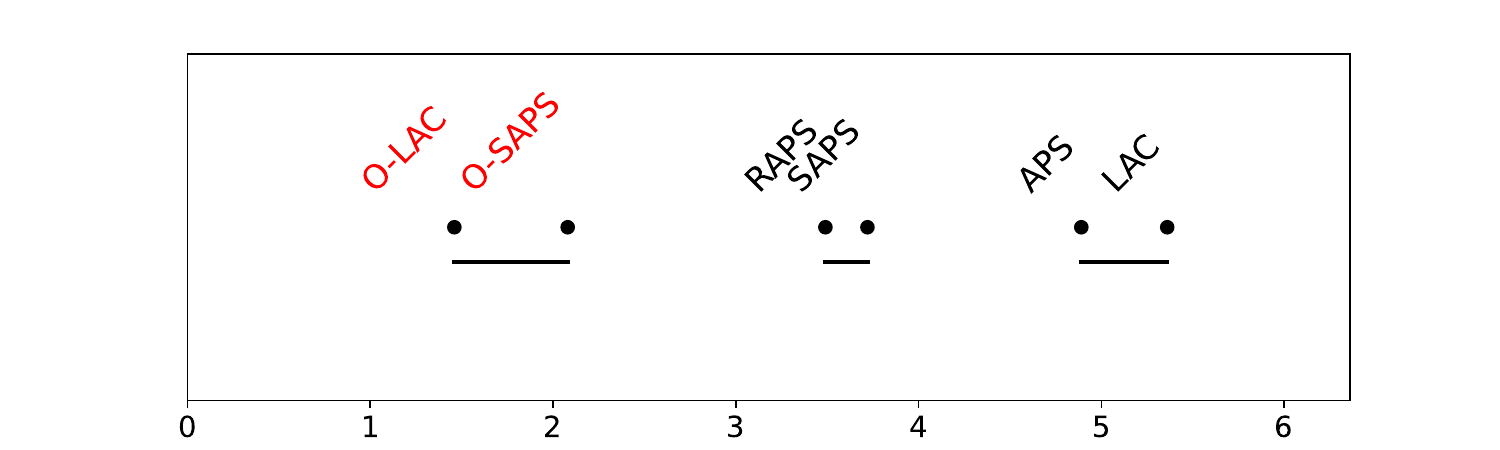}
        \caption{Swin-V2-B}
    \end{subfigure}
    \caption{Critical Difference Diagrams. T-SS, $\alpha=0.10, B=100$. The rank analysis based on these figures is summarized as `Avg. Rank from CD' in Table~\ref{tab:apdx_alg_results_our_metrics_B100} in Appendix.}
    \label{fig:apdx_cd_tss_B100_alpha0.10}
\end{figure}

\begin{figure}[!bt]
    \centering
    \begin{subfigure}[b]{0.45\textwidth}
        \includegraphics[width=\textwidth]{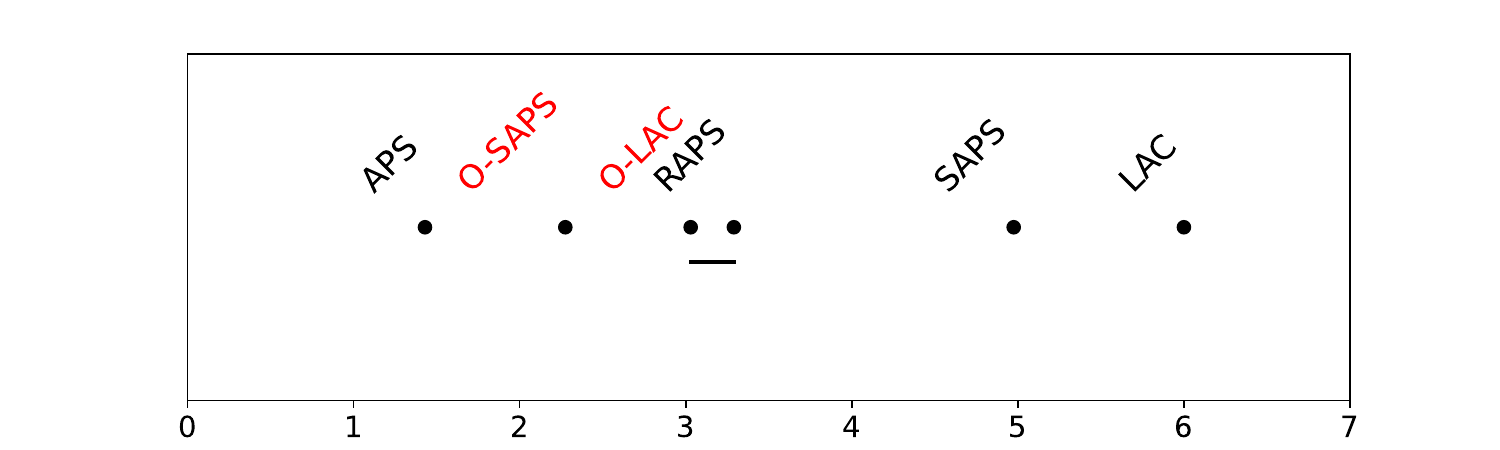}
        \caption{ResNet18}
    \end{subfigure}
    \begin{subfigure}[b]{0.45\textwidth}
        \includegraphics[width=\textwidth]{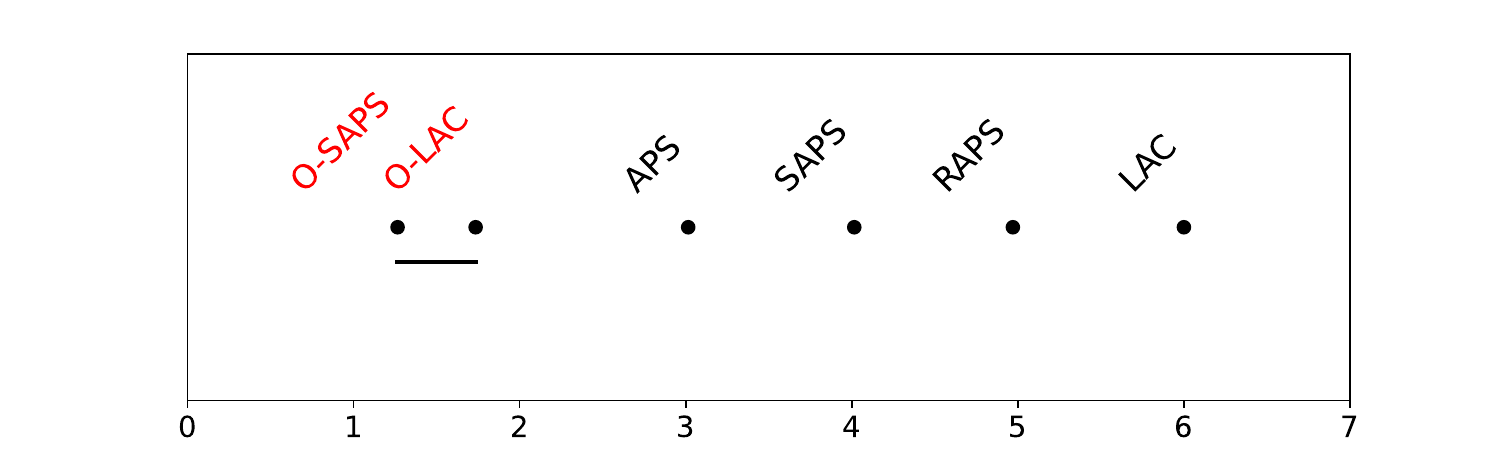}
        \caption{ResNet50}
    \end{subfigure}
    \begin{subfigure}[b]{0.45\textwidth}
        \includegraphics[width=\textwidth]{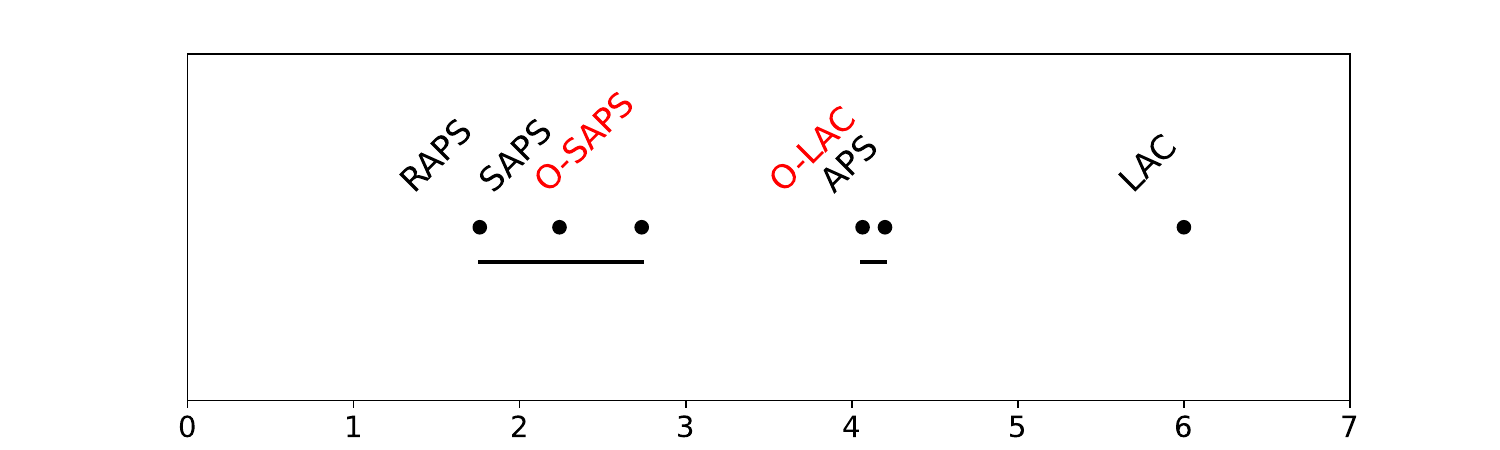}
        \caption{ResNet152}
    \end{subfigure}
    \begin{subfigure}[b]{0.45\textwidth}
        \includegraphics[width=\textwidth]{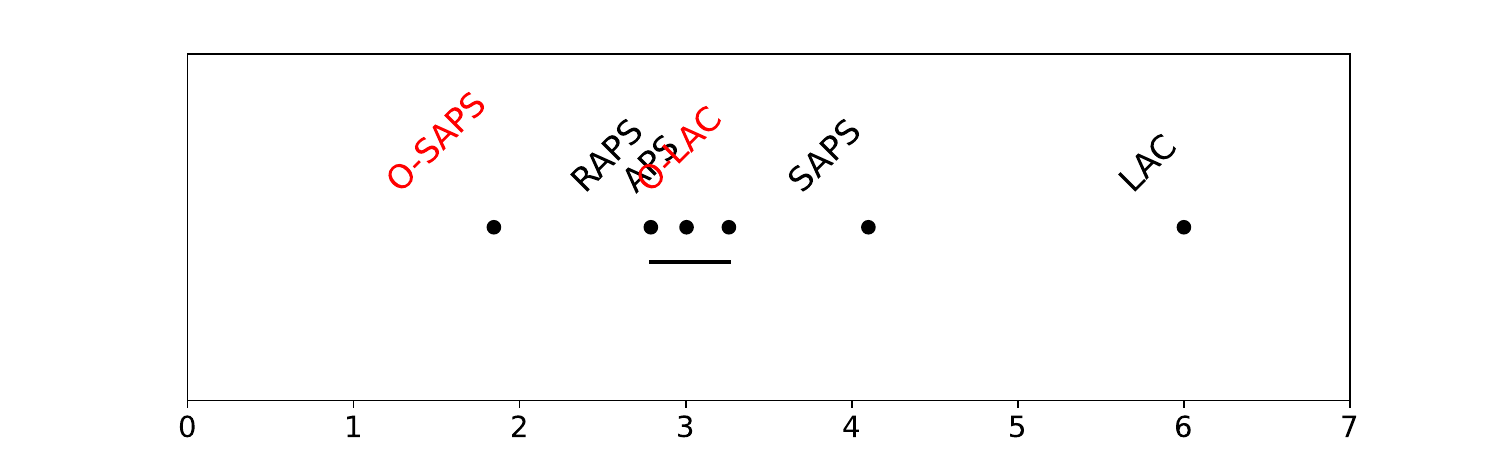}
        \caption{ViT-B-16}
    \end{subfigure}
    \begin{subfigure}[b]{0.45\textwidth}
        \includegraphics[width=\textwidth]{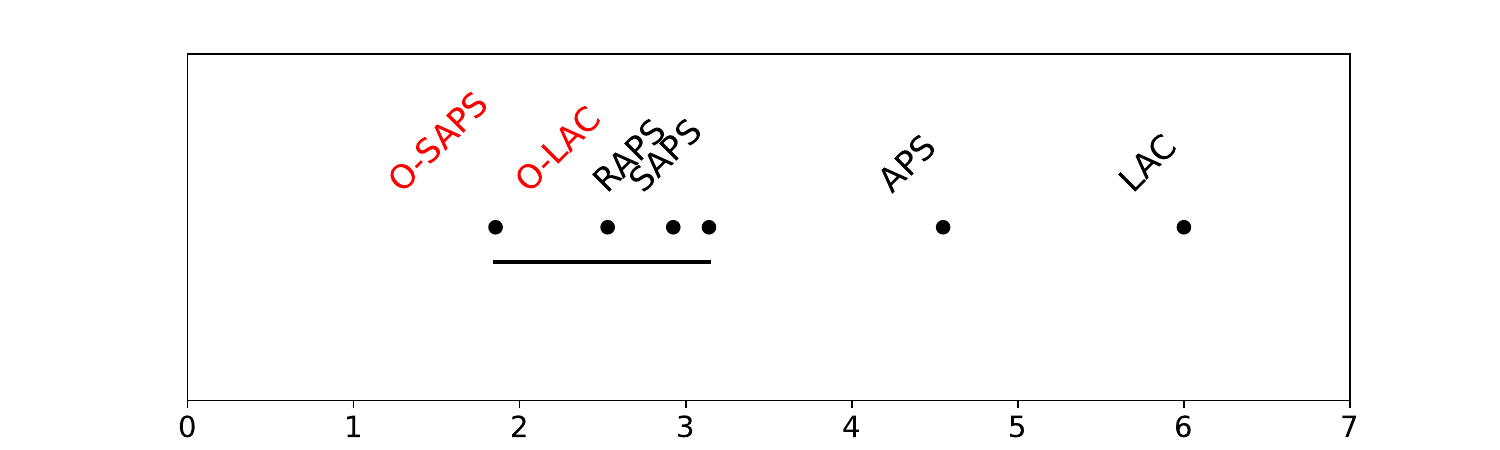}
        \caption{ViT-L-16}
    \end{subfigure}
    \begin{subfigure}[b]{0.45\textwidth}
        \includegraphics[width=\textwidth]{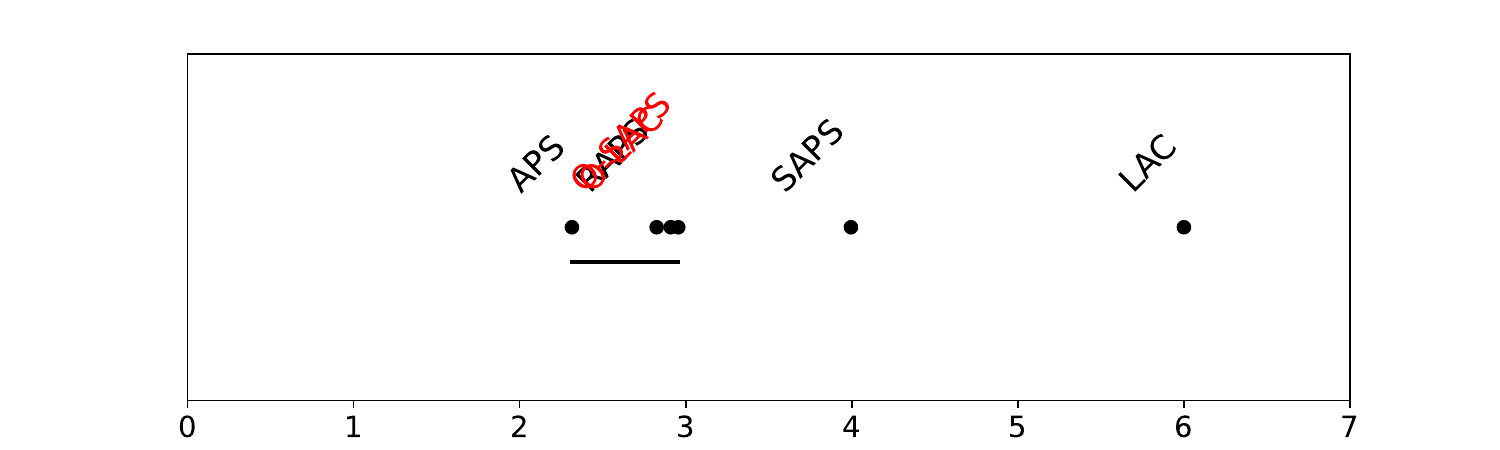}
        \caption{ViT-H-14}
    \end{subfigure}
    \begin{subfigure}[b]{0.45\textwidth}
        \includegraphics[width=\textwidth]{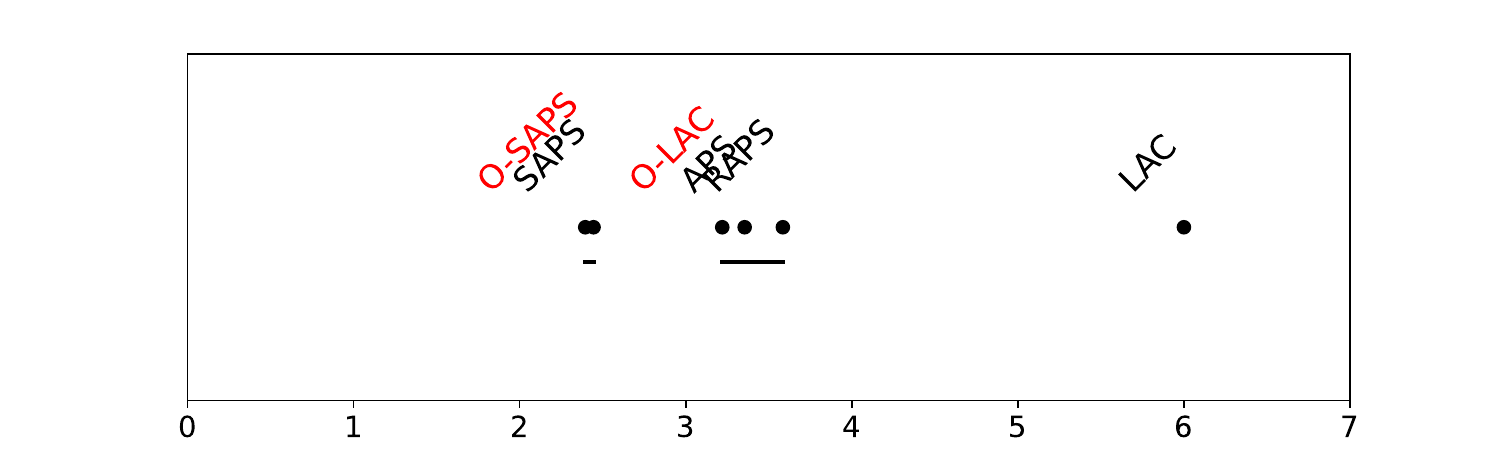}
        \caption{EfficientNet-V2-M}
    \end{subfigure}
    \begin{subfigure}[b]{0.45\textwidth}
        \includegraphics[width=\textwidth]{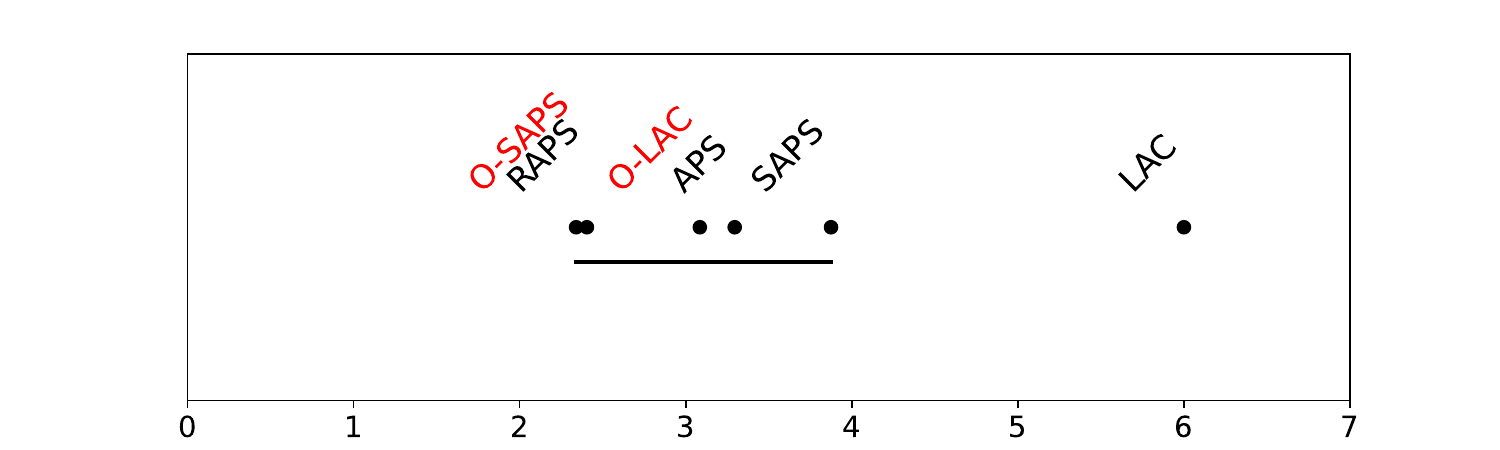}
        \caption{EfficientNet-V2-L}
    \end{subfigure}
    \begin{subfigure}[b]{0.45\textwidth}
        \includegraphics[width=\textwidth]{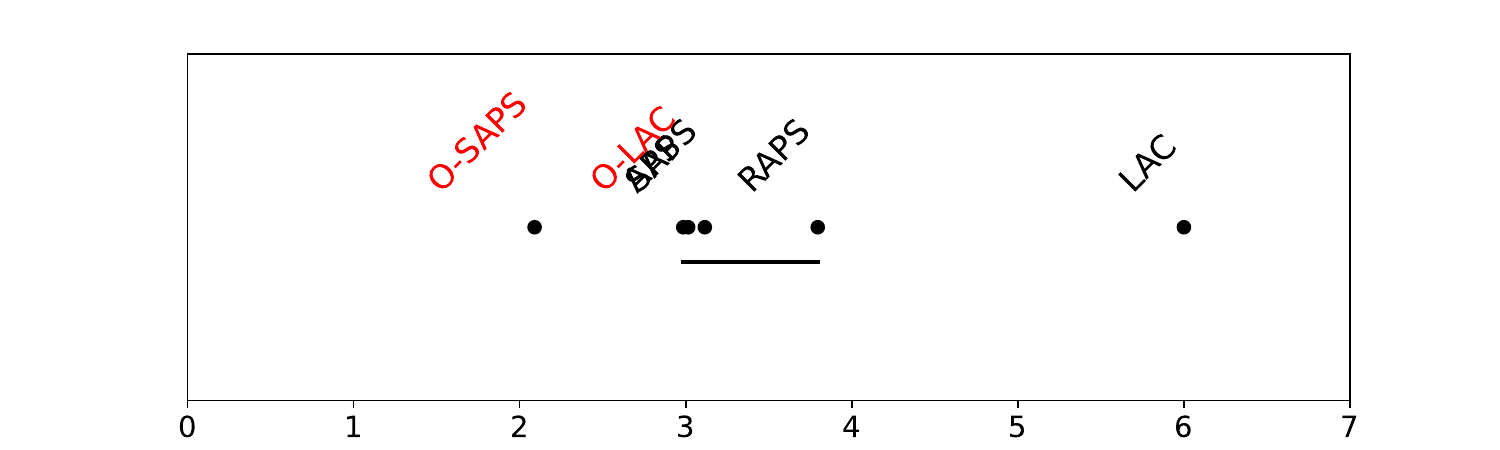}
        \caption{Swin-V2-B}
    \end{subfigure}
    \caption{Critical Difference Diagrams. T-CV, $\alpha=0.15, B=100$. The rank analysis based on these figures is summarized as `Avg. Rank from CD' in Table~\ref{tab:apdx_alg_results_our_metrics_B100} in Appendix.}
    \label{fig:apdx_cd_tcv_B100_alpha0.15}
\end{figure}

\begin{figure}[!bt]
    \centering
    \begin{subfigure}[b]{0.45\textwidth}
        \includegraphics[width=\textwidth]{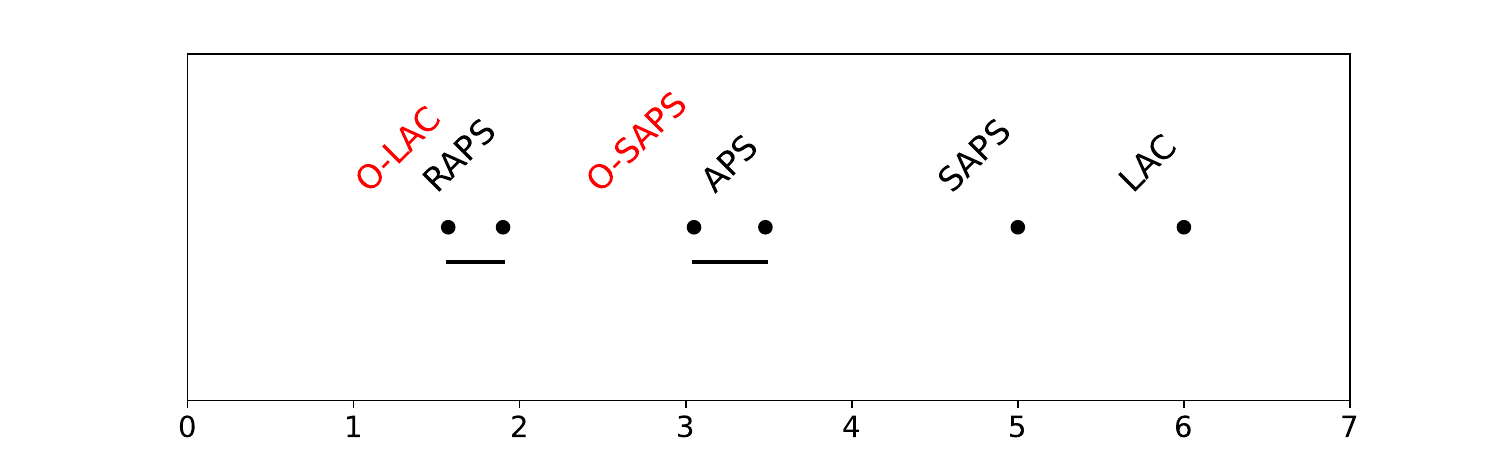}
        \caption{ResNet18}
    \end{subfigure}
    \begin{subfigure}[b]{0.45\textwidth}
        \includegraphics[width=\textwidth]{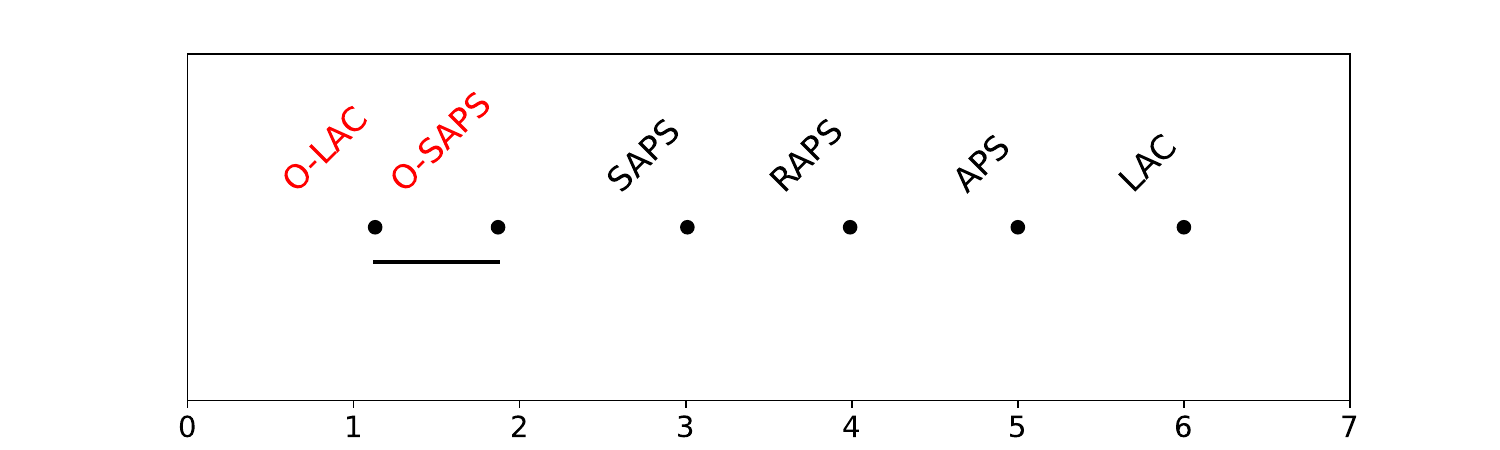}
        \caption{ResNet50}
    \end{subfigure}
    \begin{subfigure}[b]{0.45\textwidth}
        \includegraphics[width=\textwidth]{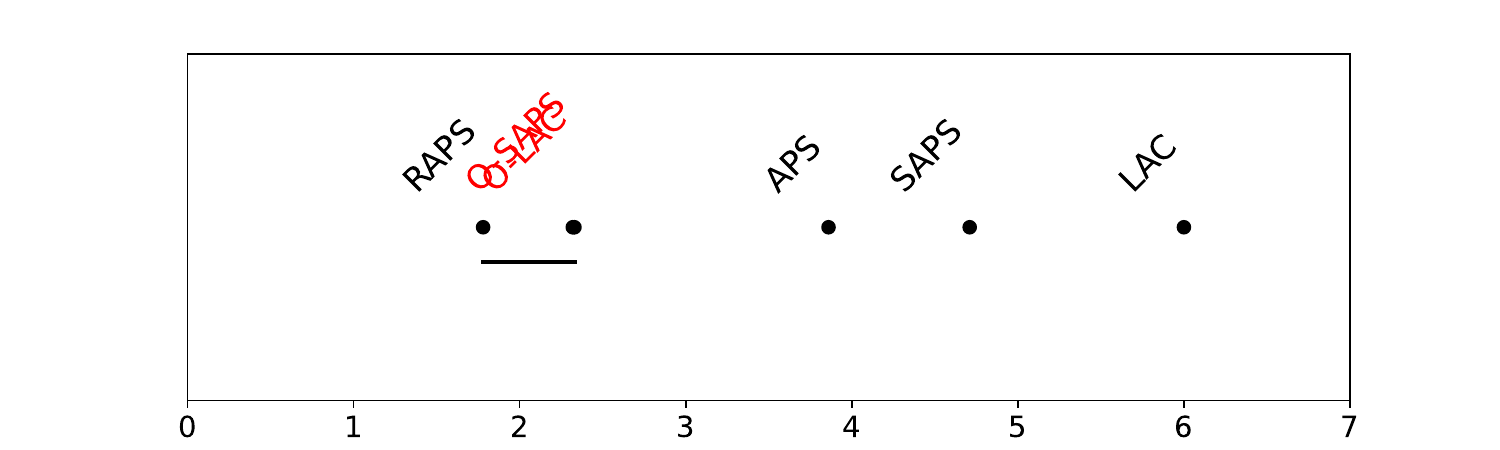}
        \caption{ResNet152}
    \end{subfigure}
    \begin{subfigure}[b]{0.45\textwidth}
        \includegraphics[width=\textwidth]{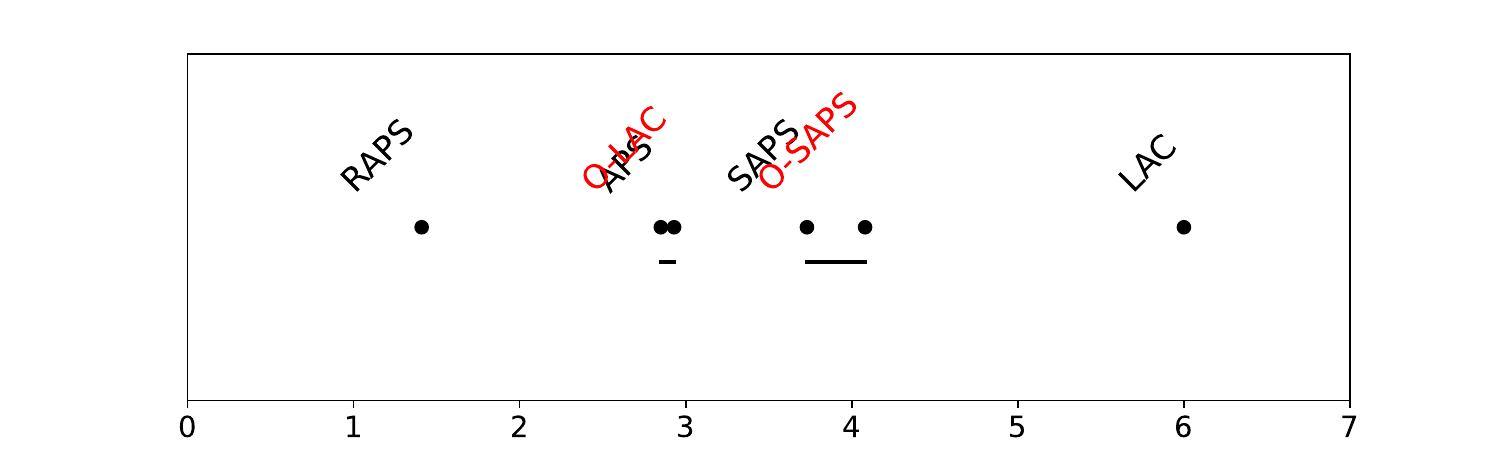}
        \caption{ViT-B-16}
    \end{subfigure}
    \begin{subfigure}[b]{0.45\textwidth}
        \includegraphics[width=\textwidth]{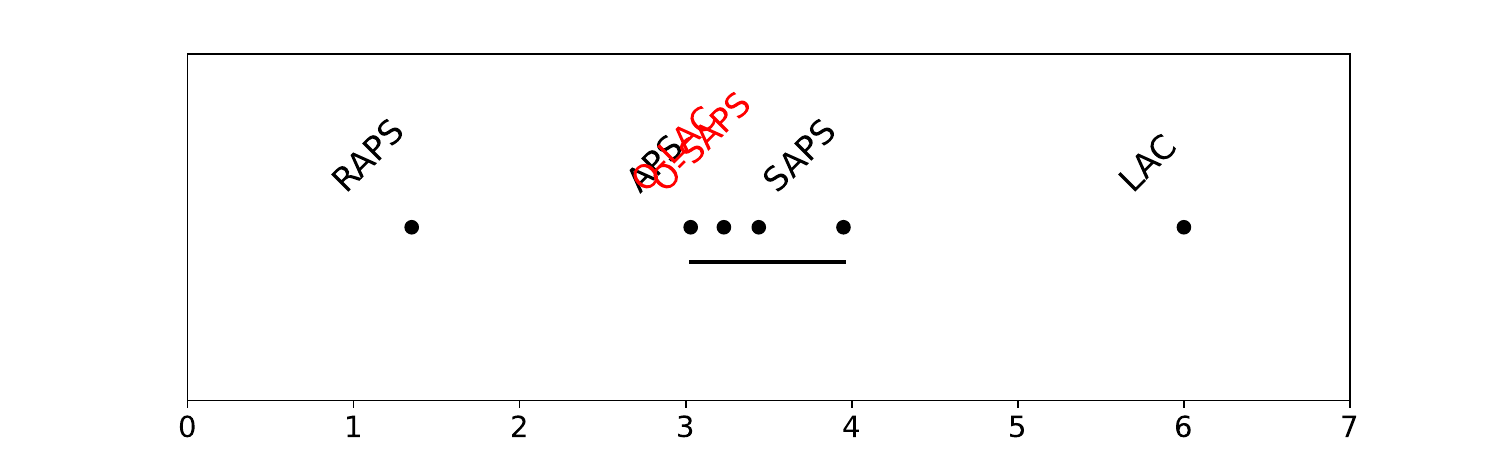}
        \caption{ViT-L-16}
    \end{subfigure}
    \begin{subfigure}[b]{0.45\textwidth}
        \includegraphics[width=\textwidth]{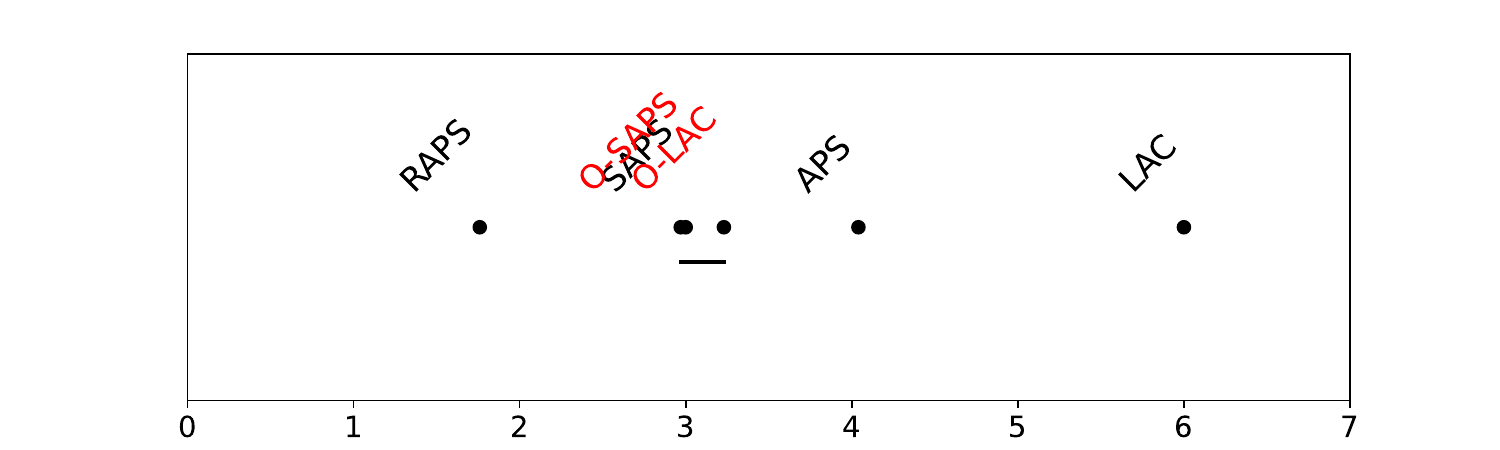}
        \caption{ViT-H-14}
    \end{subfigure}
    \begin{subfigure}[b]{0.45\textwidth}
        \includegraphics[width=\textwidth]{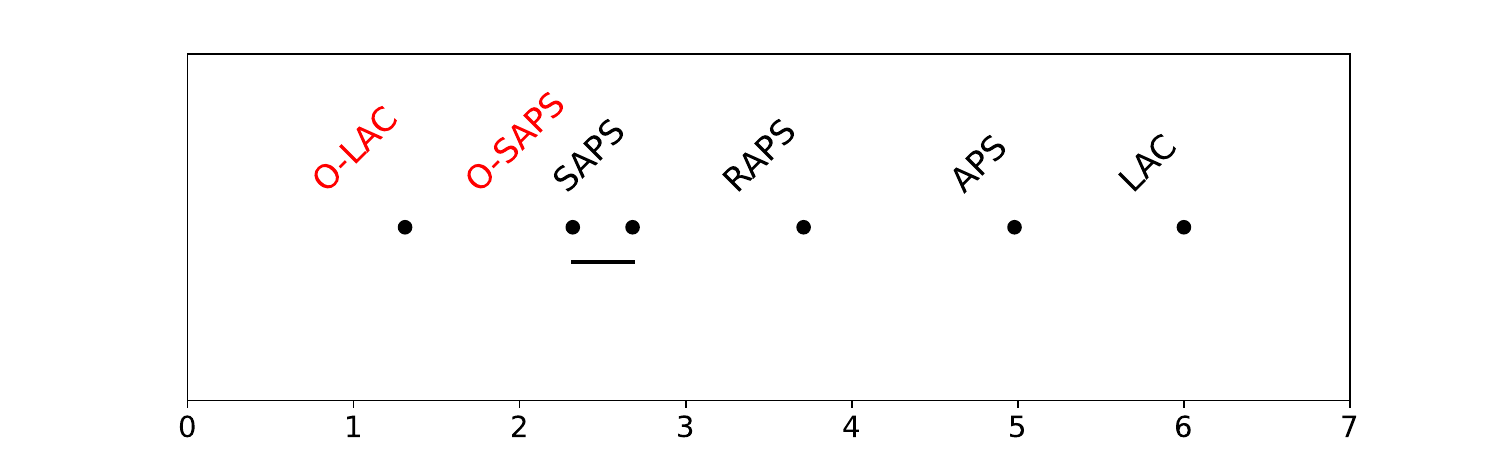}
        \caption{EfficientNet-V2-M}
    \end{subfigure}
    \begin{subfigure}[b]{0.45\textwidth}
        \includegraphics[width=\textwidth]{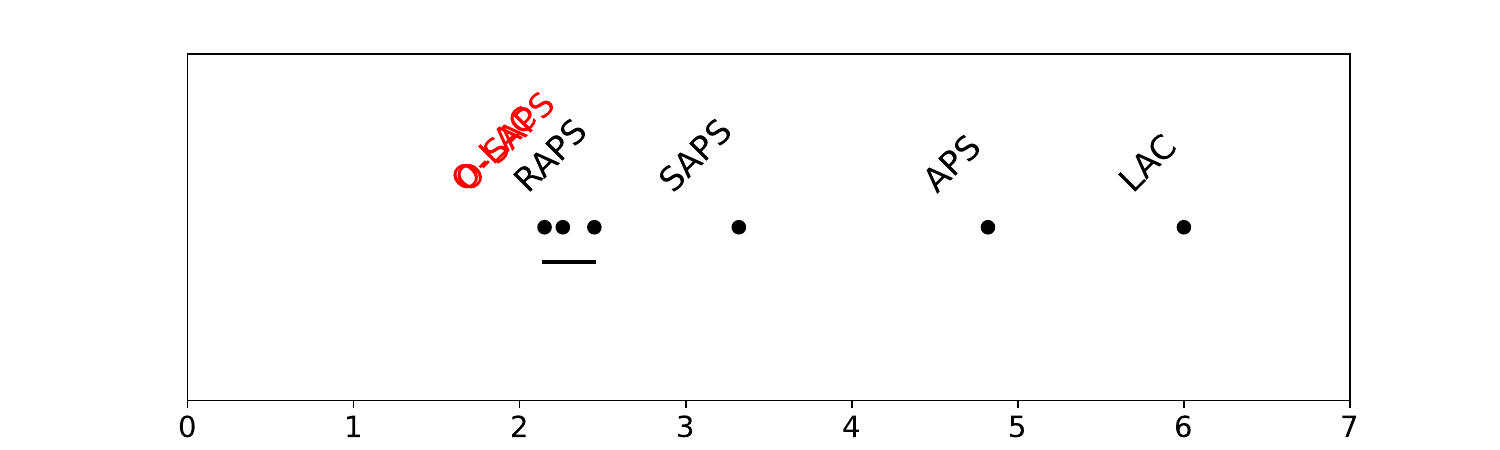}
        \caption{EfficientNet-V2-L}
    \end{subfigure}
    \begin{subfigure}[b]{0.45\textwidth}
        \includegraphics[width=\textwidth]{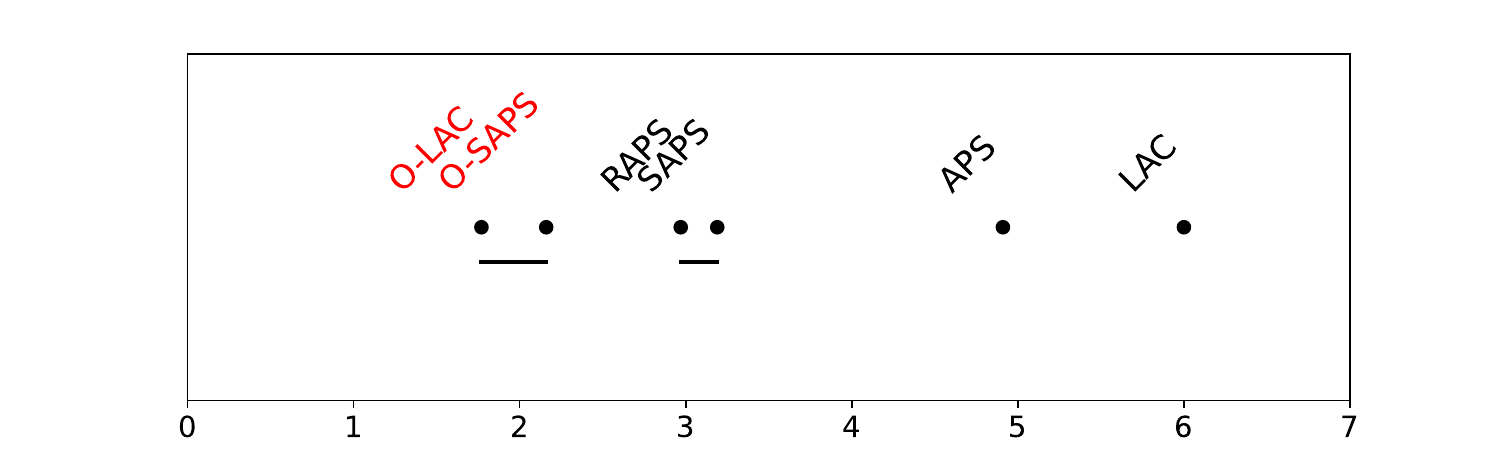}
        \caption{Swin-V2-B}
    \end{subfigure}
    \caption{Critical Difference Diagrams. T-SS, $\alpha=0.15, B=100$. The rank analysis based on these figures is summarized as `Avg. Rank from CD' in Table~\ref{tab:apdx_alg_results_our_metrics_B100} in Appendix.}
    \label{fig:apdx_cd_tss_B100_alpha0.15}
\end{figure}

\begin{figure}[!bt]
    \centering
    \begin{subfigure}[b]{0.45\textwidth}
        \includegraphics[width=\textwidth]{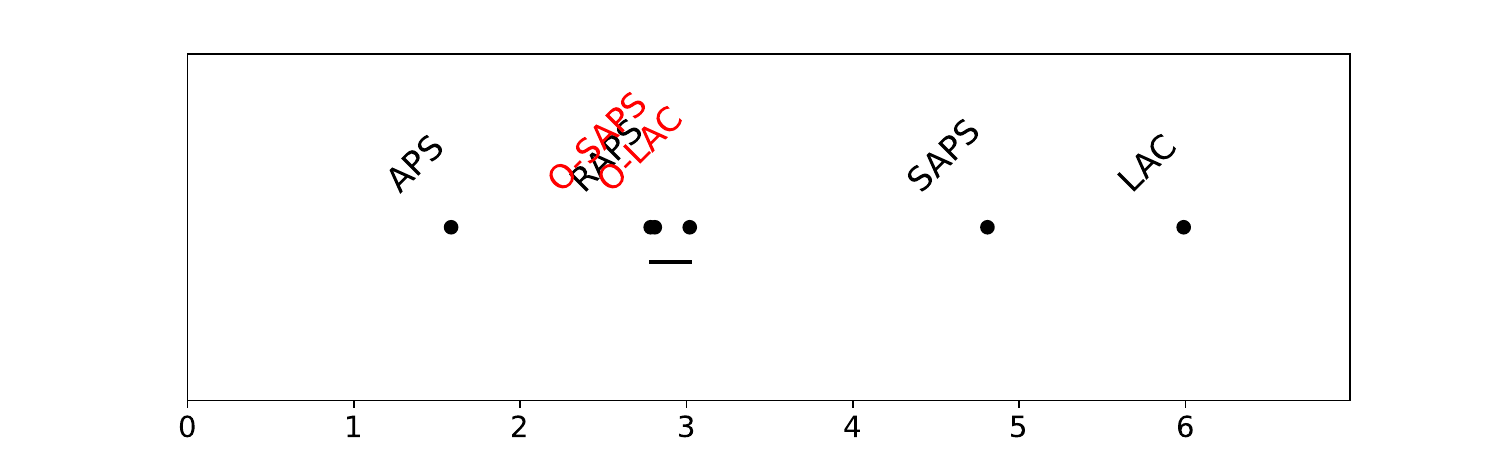}
        \caption{ResNet18}
    \end{subfigure}
    \begin{subfigure}[b]{0.45\textwidth}
        \includegraphics[width=\textwidth]{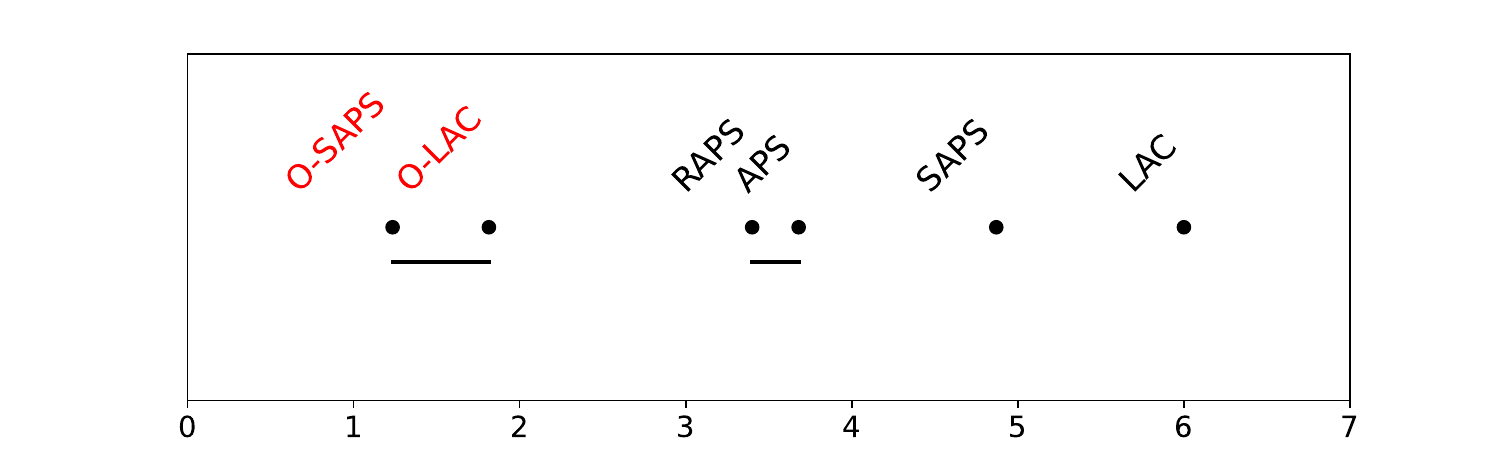}
        \caption{ResNet50}
    \end{subfigure}
    \begin{subfigure}[b]{0.45\textwidth}
        \includegraphics[width=\textwidth]{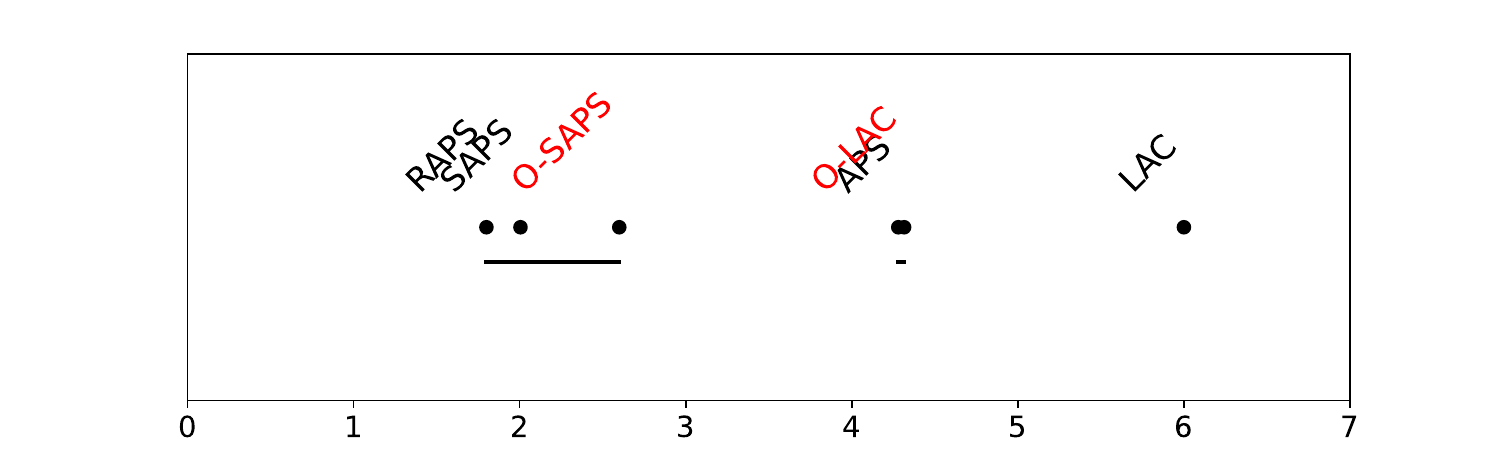}
        \caption{ResNet152}
    \end{subfigure}
    \begin{subfigure}[b]{0.45\textwidth}
        \includegraphics[width=\textwidth]{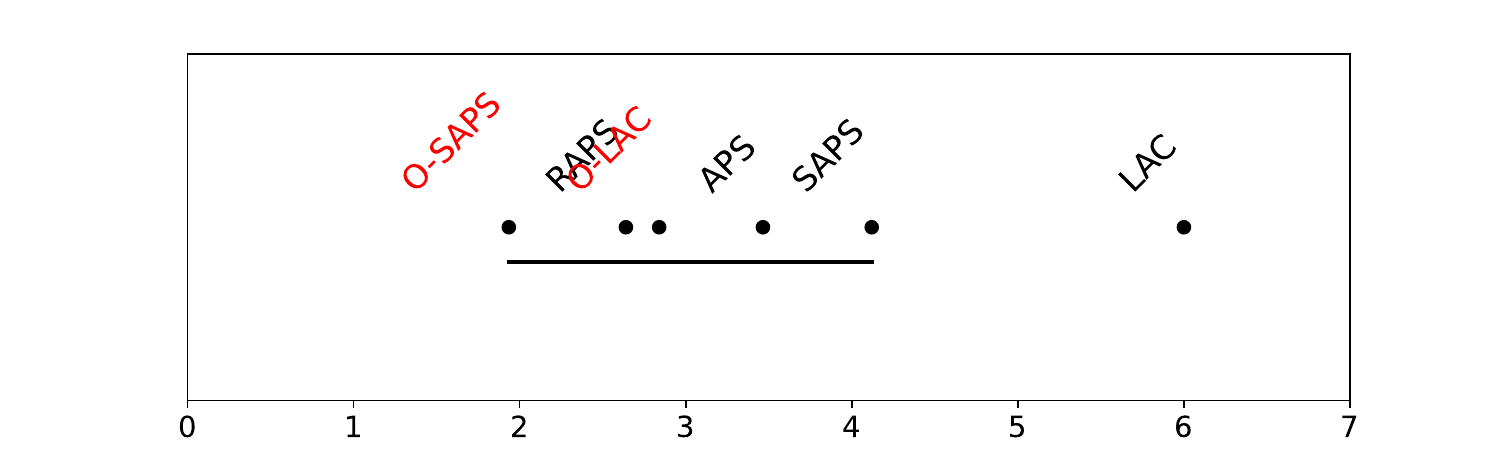}
        \caption{ViT-B-16}
    \end{subfigure}
    \begin{subfigure}[b]{0.45\textwidth}
        \includegraphics[width=\textwidth]{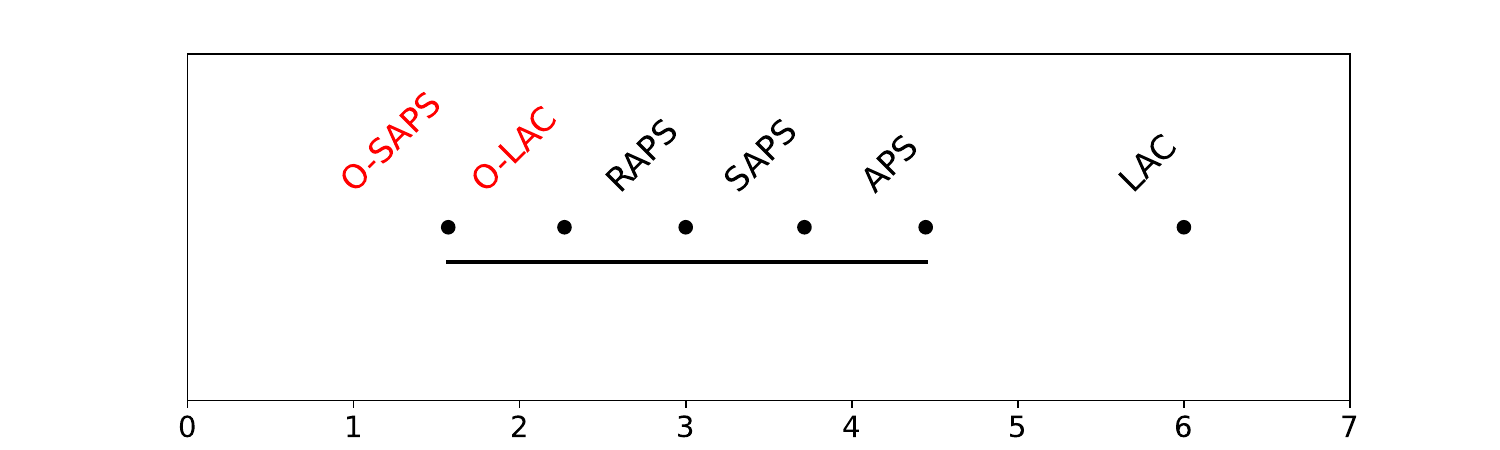}
        \caption{ViT-L-16}
    \end{subfigure}
    \begin{subfigure}[b]{0.45\textwidth}
        \includegraphics[width=\textwidth]{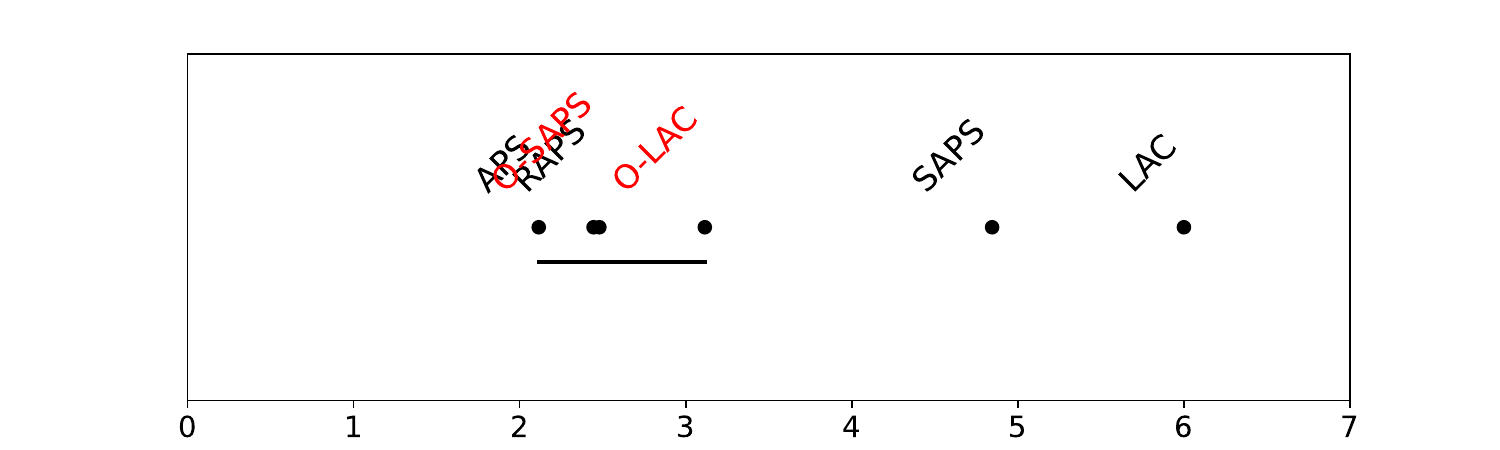}
        \caption{ViT-H-14}
    \end{subfigure}
    \begin{subfigure}[b]{0.45\textwidth}
        \includegraphics[width=\textwidth]{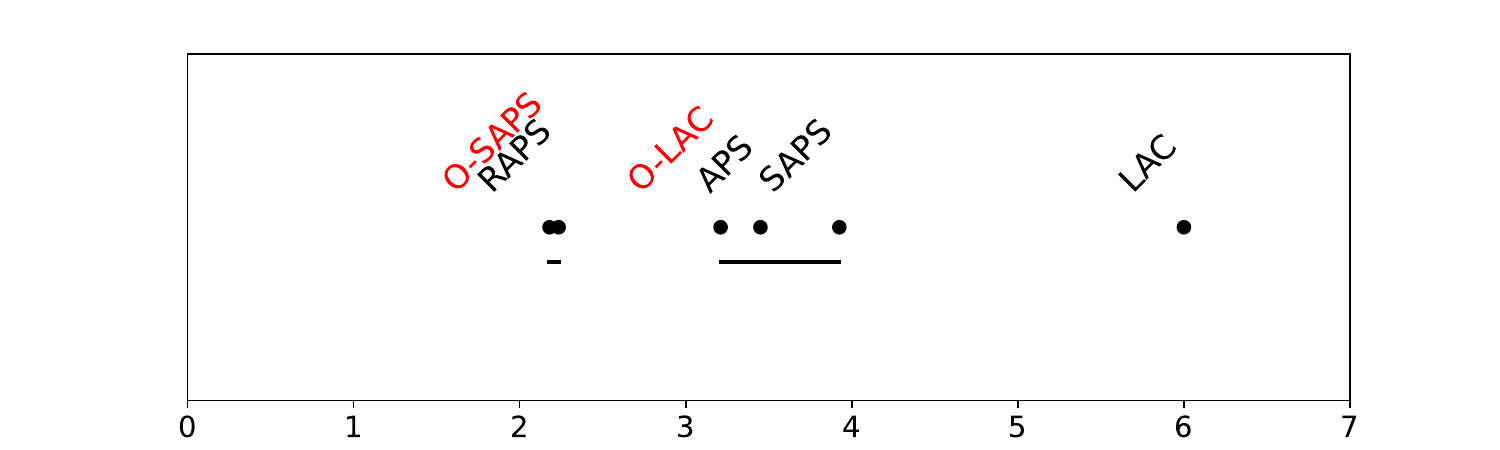}
        \caption{EfficientNet-V2-M}
    \end{subfigure}
    \begin{subfigure}[b]{0.45\textwidth}
        \includegraphics[width=\textwidth]{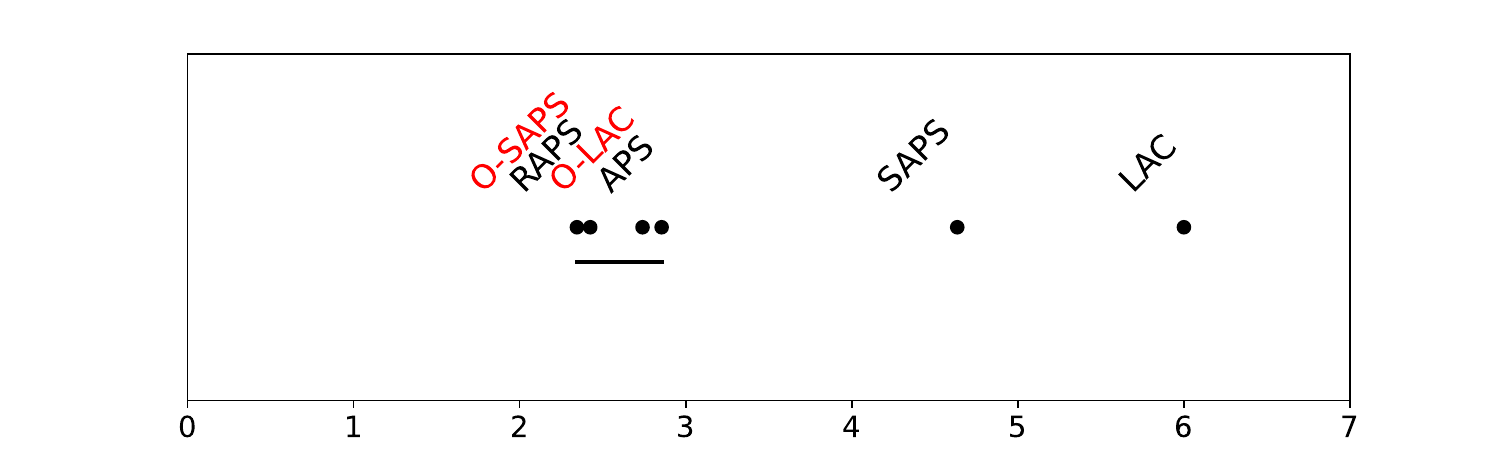}
        \caption{EfficientNet-V2-L}
    \end{subfigure}
    \begin{subfigure}[b]{0.45\textwidth}
        \includegraphics[width=\textwidth]{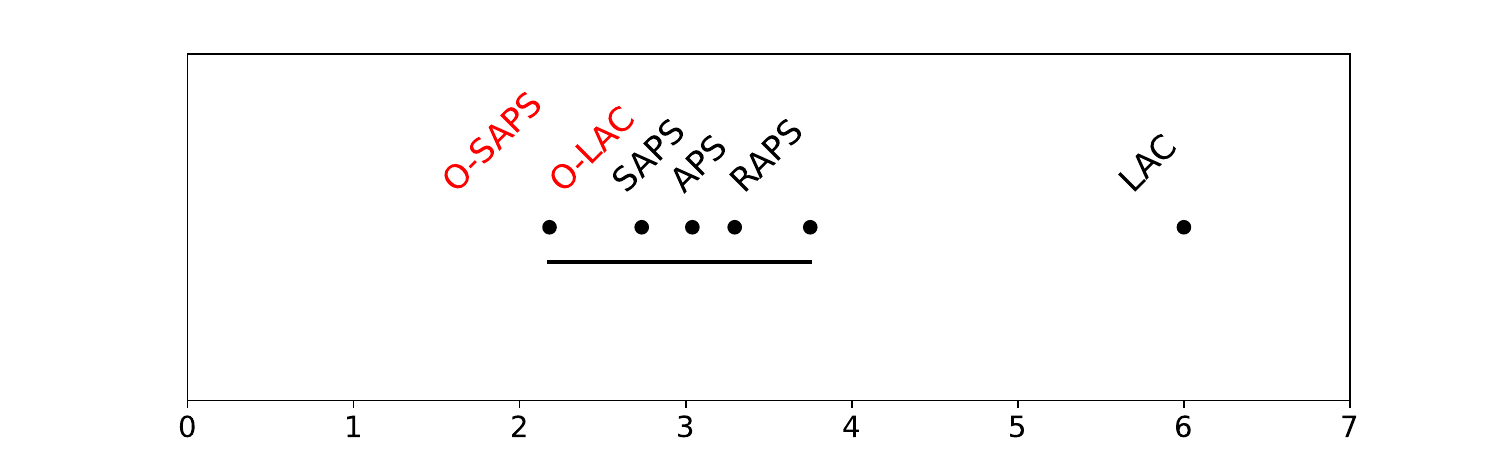}
        \caption{Swin-V2-B}
    \end{subfigure}
    \caption{Critical Difference Diagrams. T-CV, $\alpha=0.20, B=100$. The rank analysis based on these figures is summarized as `Avg. Rank from CD' in Table~\ref{tab:apdx_alg_results_our_metrics_B100} in Appendix.}
    \label{fig:apdx_cd_tcv_B100_alpha0.20}
\end{figure}

\begin{figure}[!bt]
    \centering
    \begin{subfigure}[b]{0.45\textwidth}
        \includegraphics[width=\textwidth]{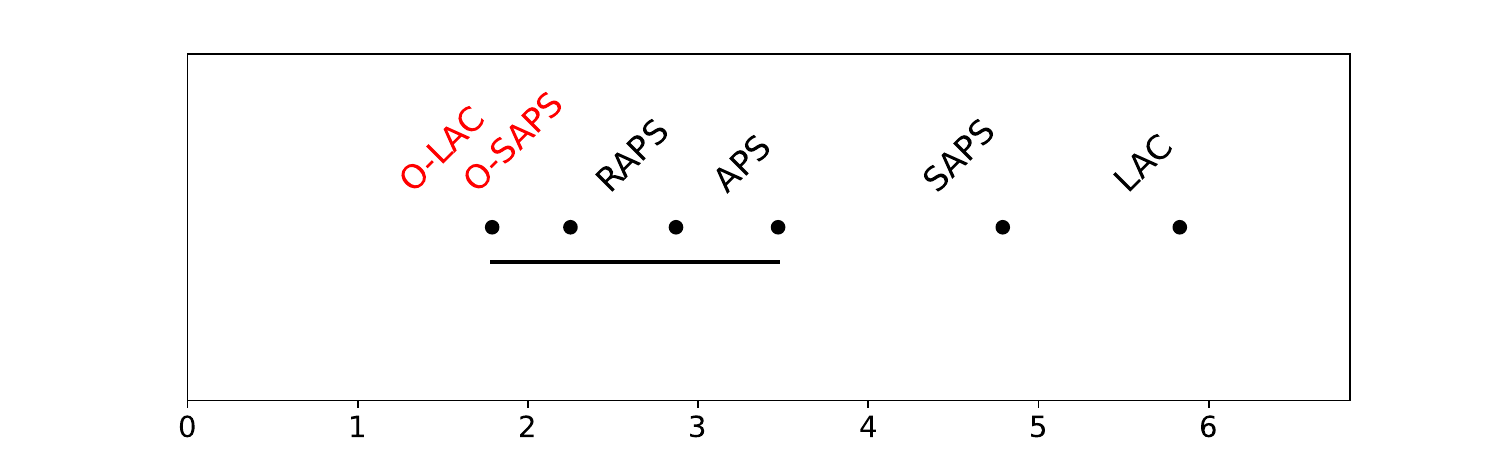}
        \caption{ResNet18}
    \end{subfigure}
    \begin{subfigure}[b]{0.45\textwidth}
        \includegraphics[width=\textwidth]{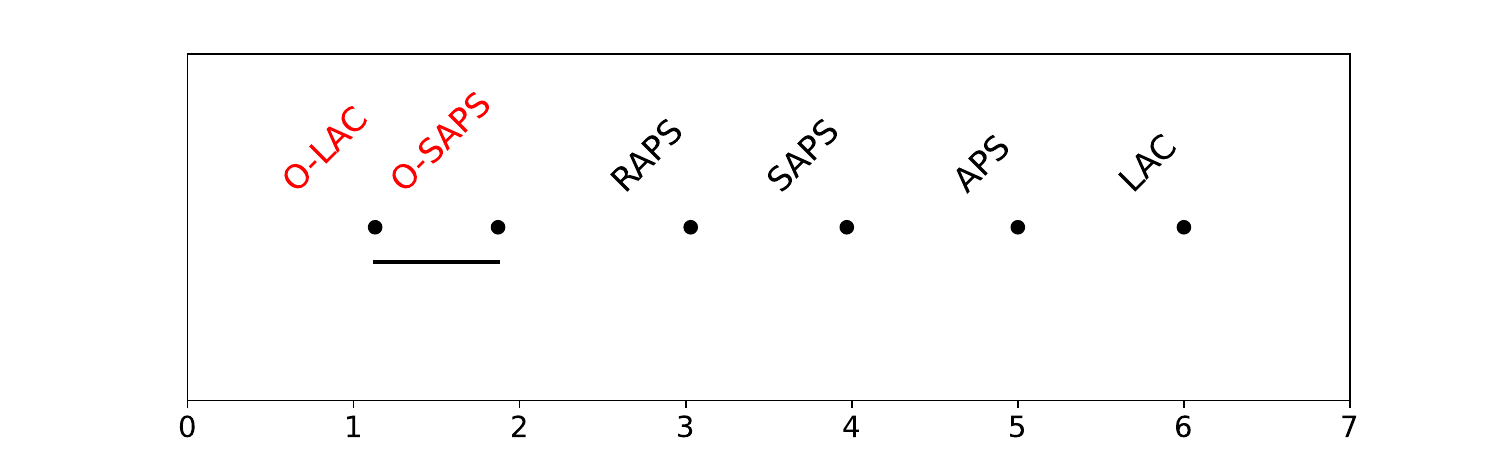}
        \caption{ResNet50}
    \end{subfigure}
    \begin{subfigure}[b]{0.45\textwidth}
        \includegraphics[width=\textwidth]{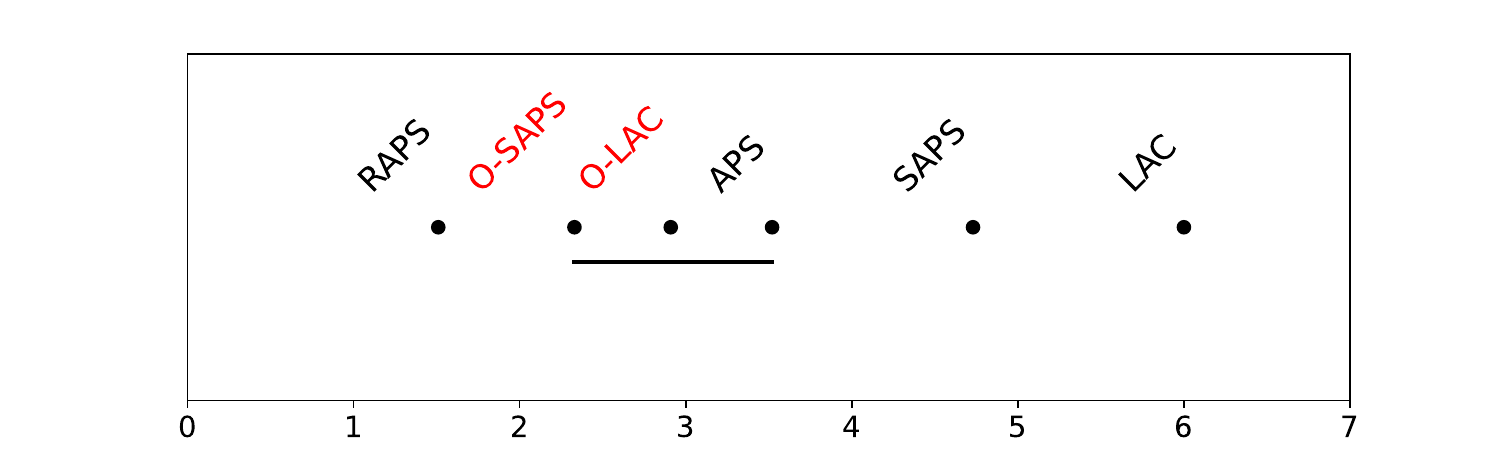}
        \caption{ResNet152}
    \end{subfigure}
    \begin{subfigure}[b]{0.45\textwidth}
        \includegraphics[width=\textwidth]{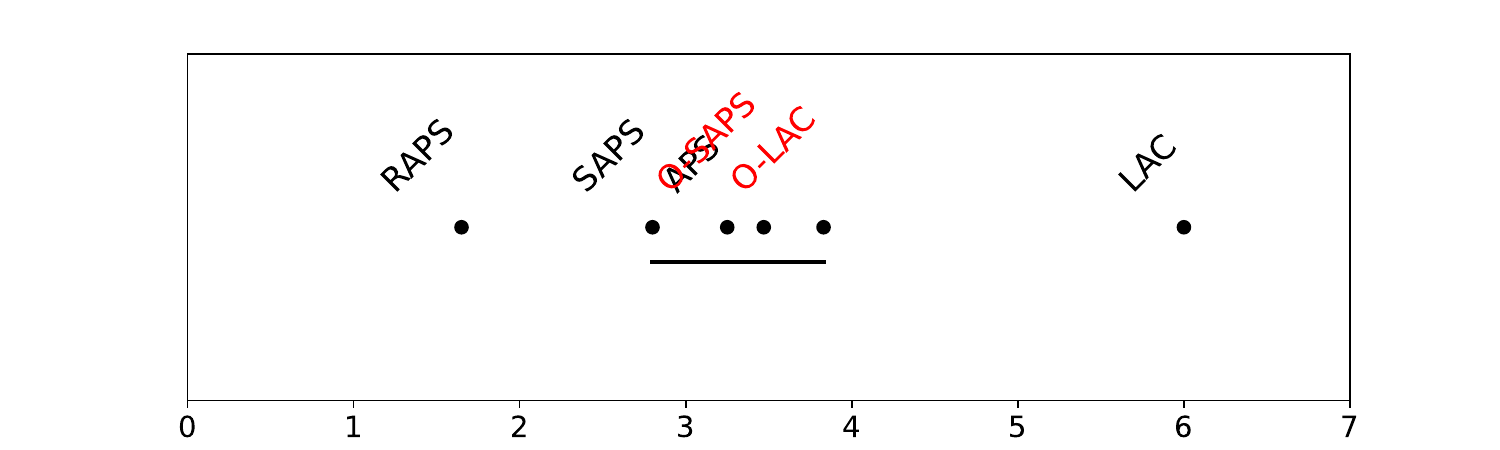}
        \caption{ViT-B-16}
    \end{subfigure}
    \begin{subfigure}[b]{0.45\textwidth}
        \includegraphics[width=\textwidth]{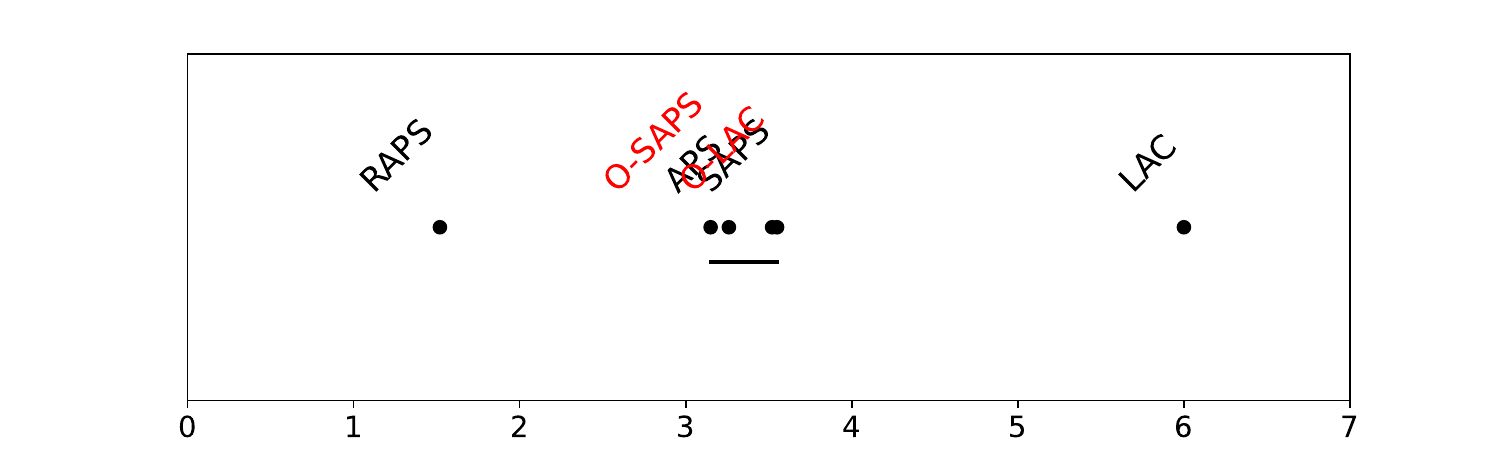}
        \caption{ViT-L-16}
    \end{subfigure}
    \begin{subfigure}[b]{0.45\textwidth}
        \includegraphics[width=\textwidth]{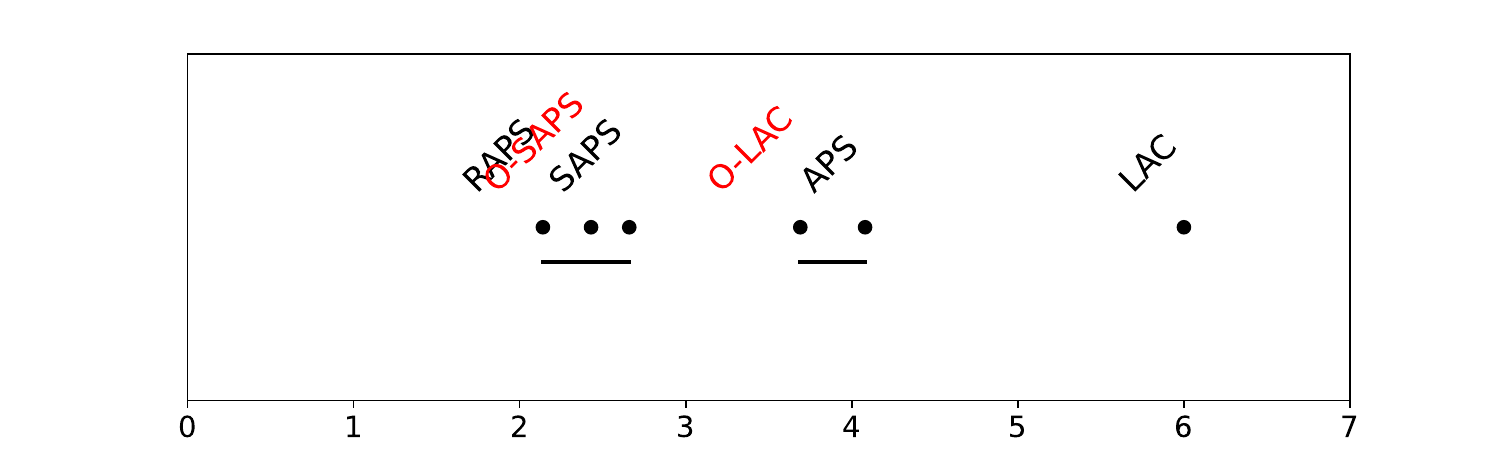}
        \caption{ViT-H-14}
    \end{subfigure}
    \begin{subfigure}[b]{0.45\textwidth}
        \includegraphics[width=\textwidth]{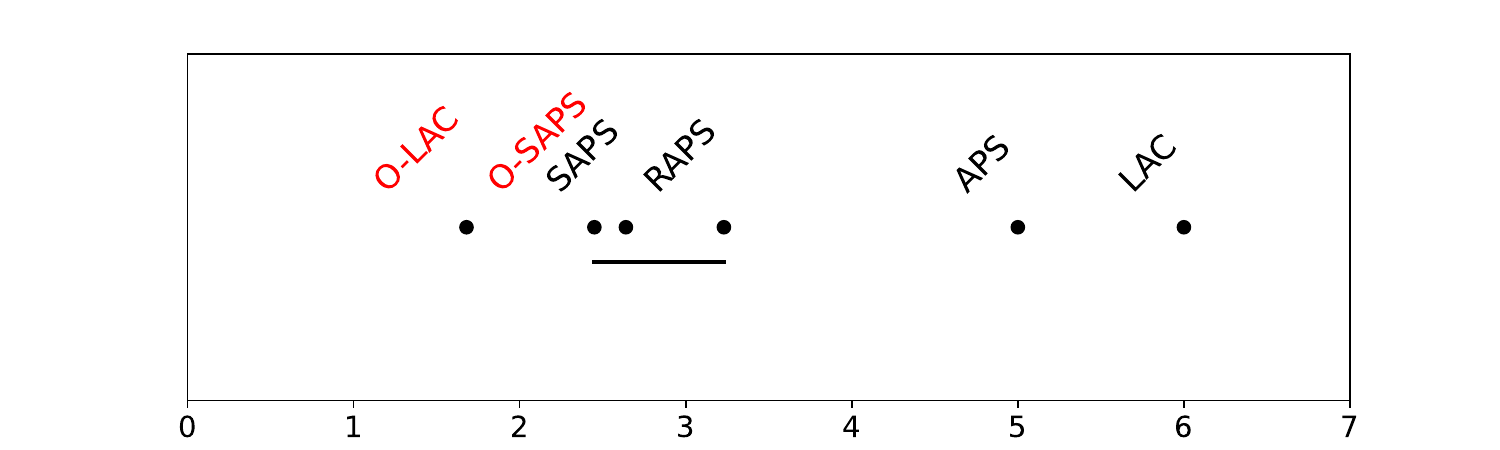}
        \caption{EfficientNet-V2-M}
    \end{subfigure}
    \begin{subfigure}[b]{0.45\textwidth}
        \includegraphics[width=\textwidth]{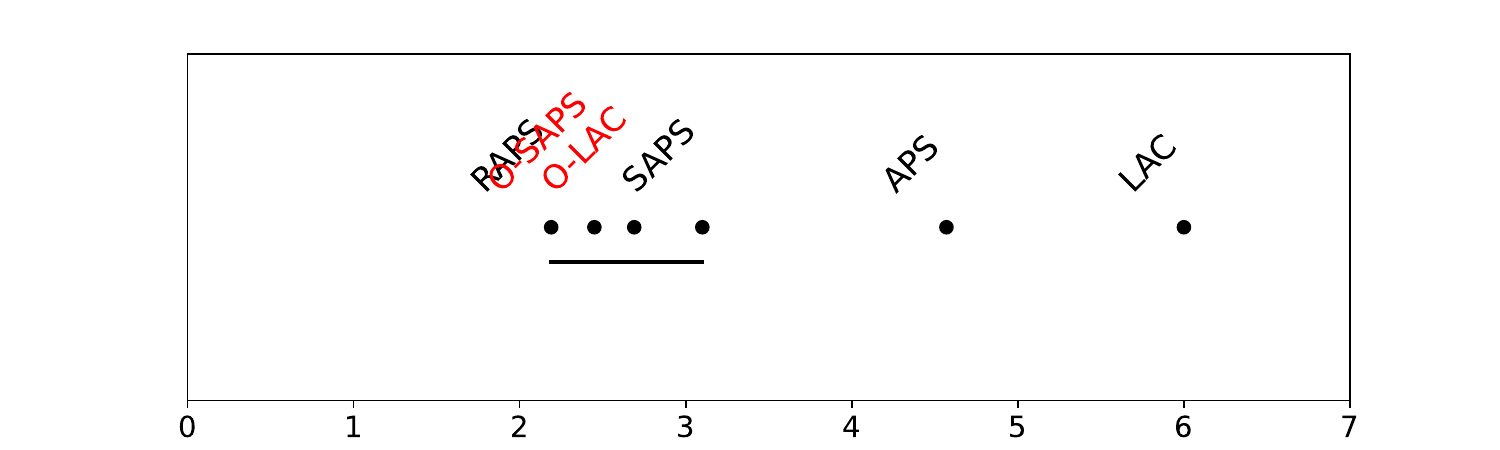}
        \caption{EfficientNet-V2-L}
    \end{subfigure}
    \begin{subfigure}[b]{0.45\textwidth}
        \includegraphics[width=\textwidth]{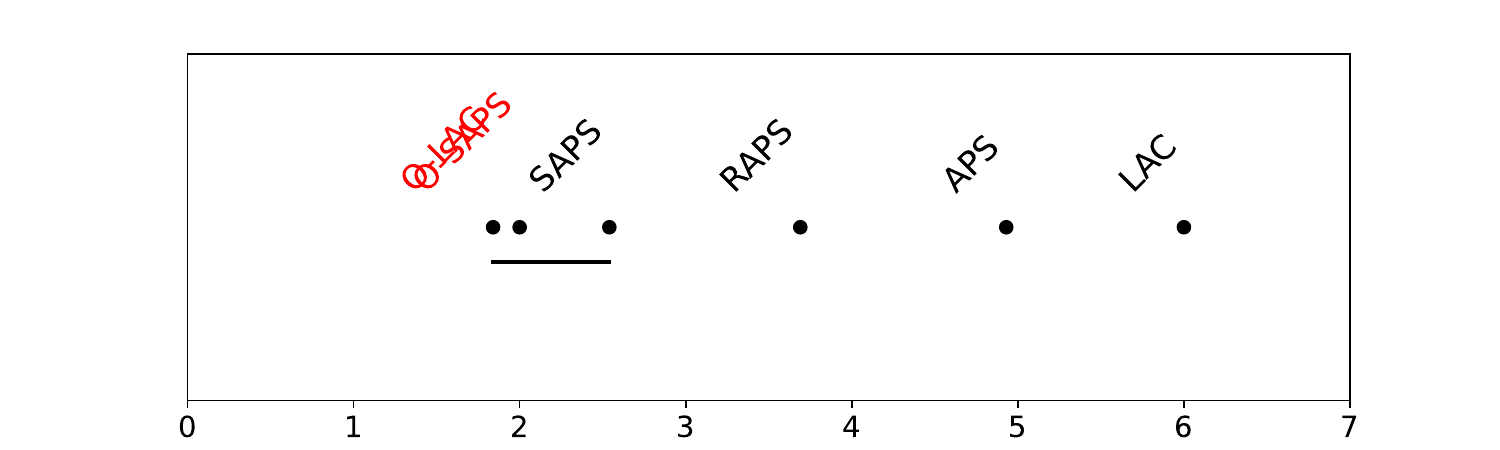}
        \caption{Swin-V2-B}
    \end{subfigure}
    \caption{Critical Difference Diagrams. T-SS, $\alpha=0.20, B=100$. The rank analysis based on these figures is summarized as `Avg. Rank from CD' in Table~\ref{tab:apdx_alg_results_our_metrics_B100} in Appendix.}
    \label{fig:apdx_cd_tss_B100_alpha0.20}
\end{figure}
\begin{figure}[!bt]
    \centering
    \begin{subfigure}[b]{0.45\textwidth}
        \includegraphics[width=\textwidth]{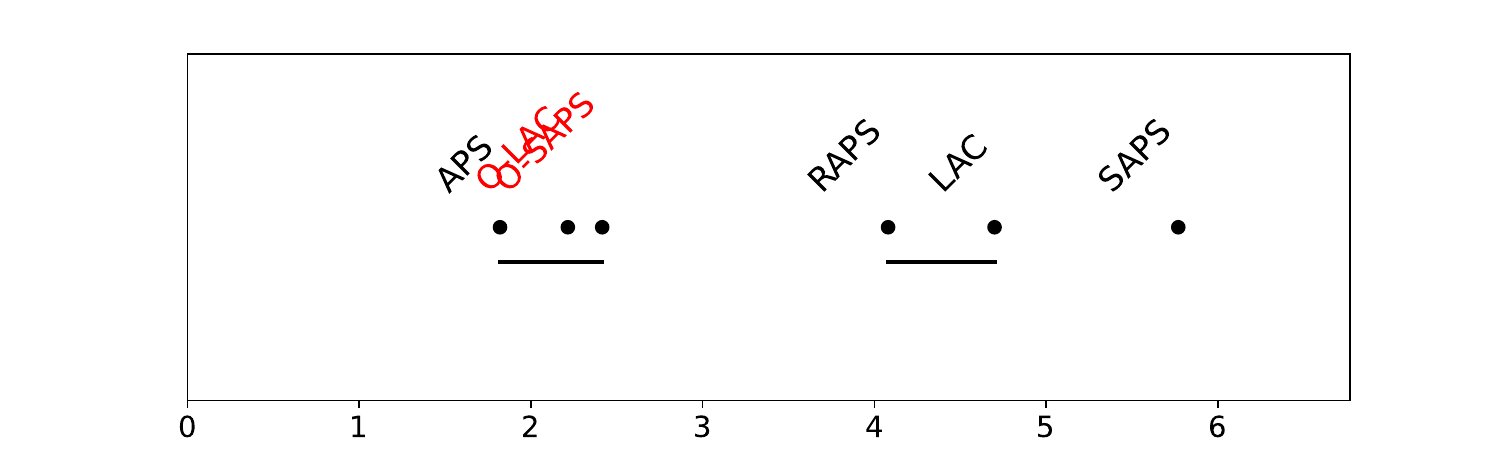}
        \caption{ResNet18}
    \end{subfigure}
    \begin{subfigure}[b]{0.45\textwidth}
        \includegraphics[width=\textwidth]{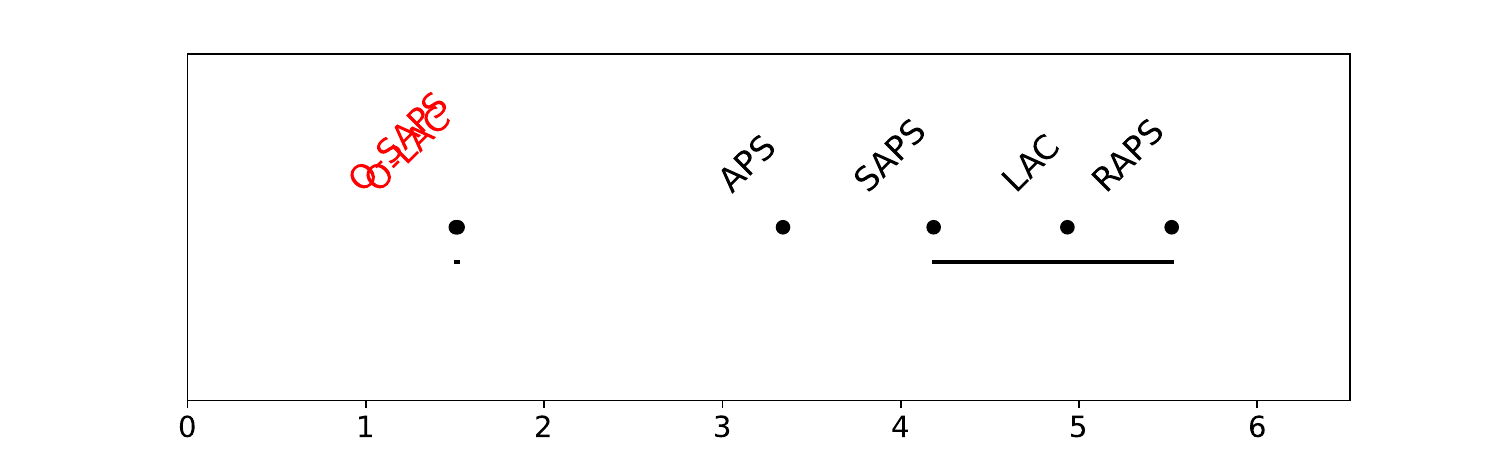}
        \caption{ResNet50}
    \end{subfigure}
    \begin{subfigure}[b]{0.45\textwidth}
        \includegraphics[width=\textwidth]{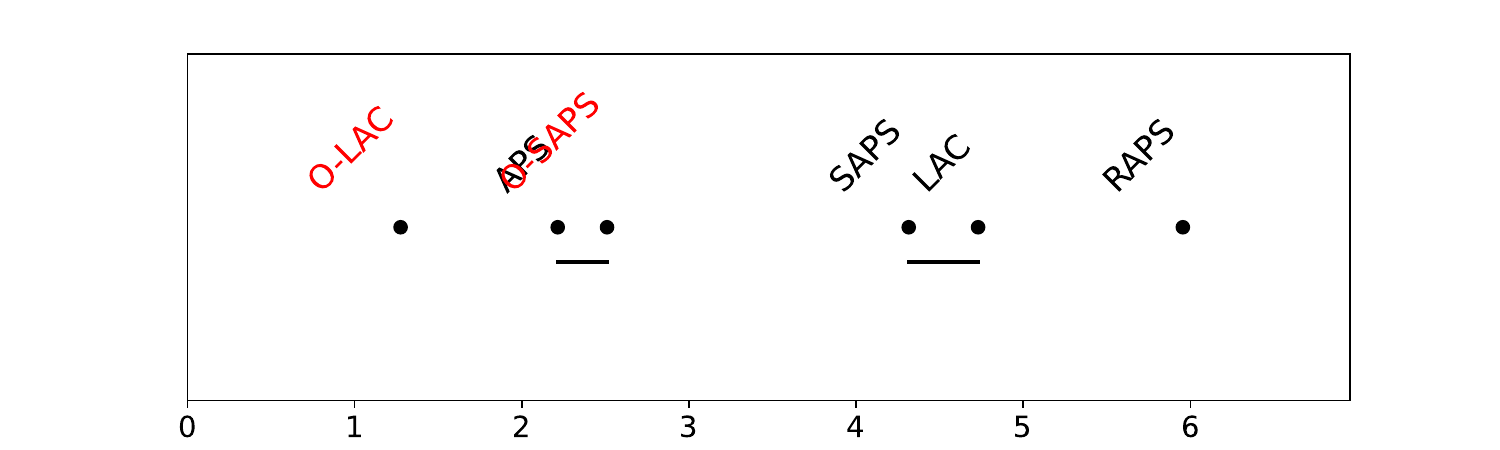}
        \caption{ResNet152}
    \end{subfigure}
    \begin{subfigure}[b]{0.45\textwidth}
        \includegraphics[width=\textwidth]{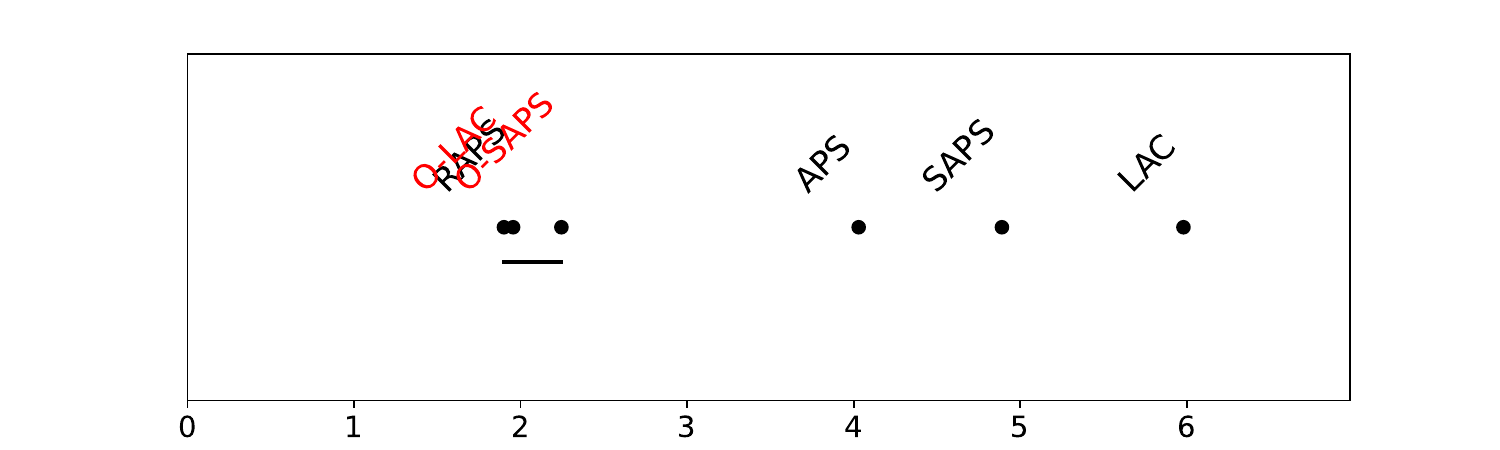}
        \caption{ViT-B-16}
    \end{subfigure}
    \begin{subfigure}[b]{0.45\textwidth}
        \includegraphics[width=\textwidth]{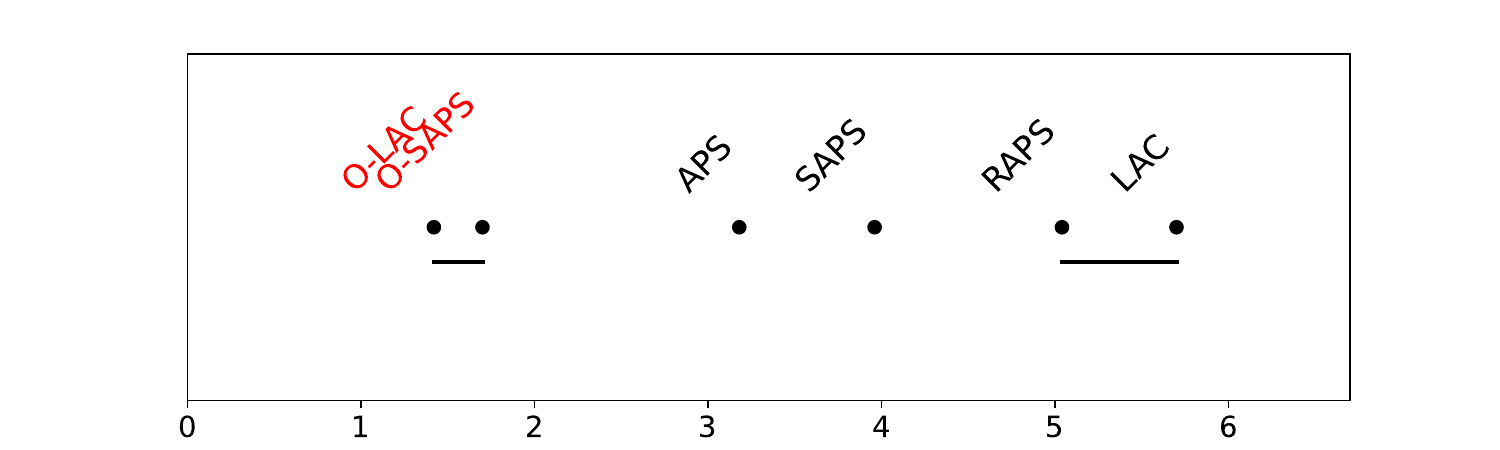}
        \caption{ViT-L-16}
    \end{subfigure}
    \begin{subfigure}[b]{0.45\textwidth}
        \includegraphics[width=\textwidth]{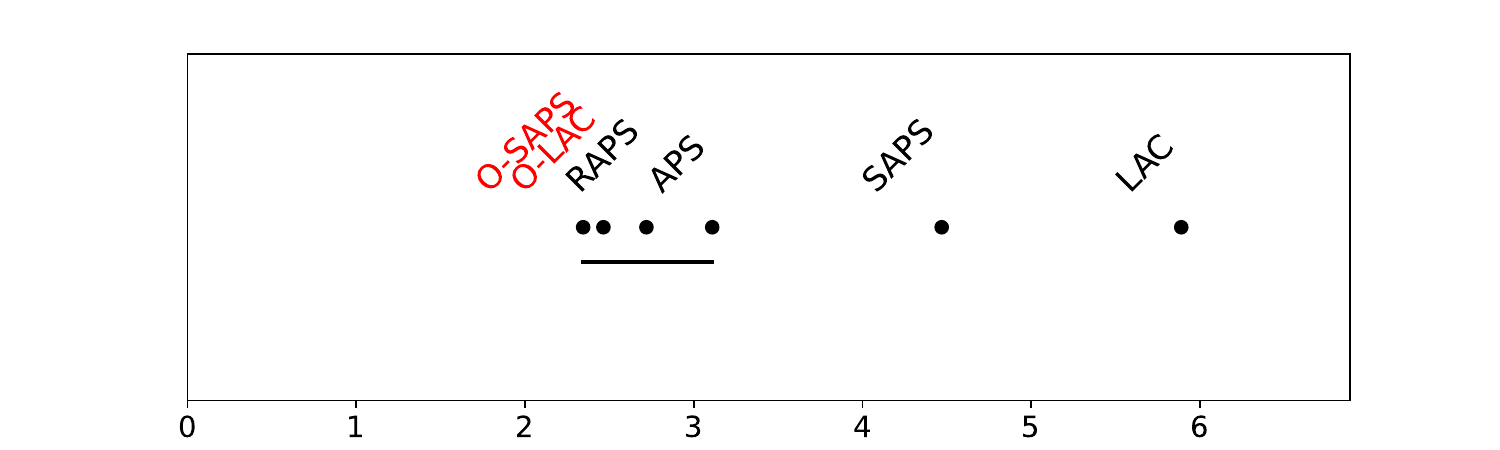}
        \caption{ViT-H-14}
    \end{subfigure}
    \begin{subfigure}[b]{0.45\textwidth}
        \includegraphics[width=\textwidth]{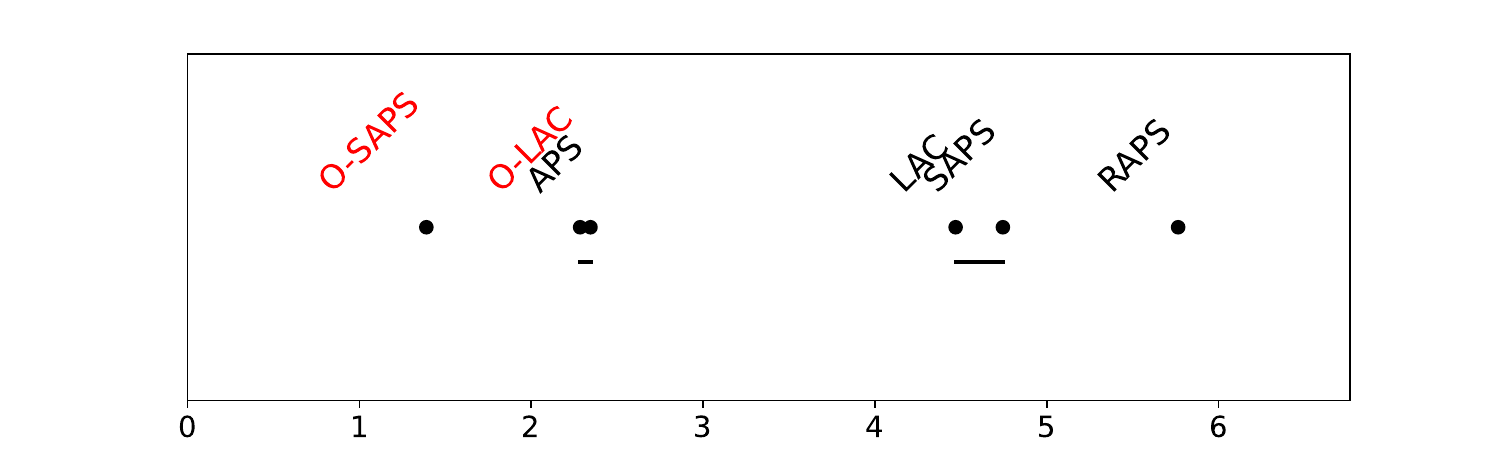}
        \caption{EfficientNet-V2-M}
    \end{subfigure}
    \begin{subfigure}[b]{0.45\textwidth}
        \includegraphics[width=\textwidth]{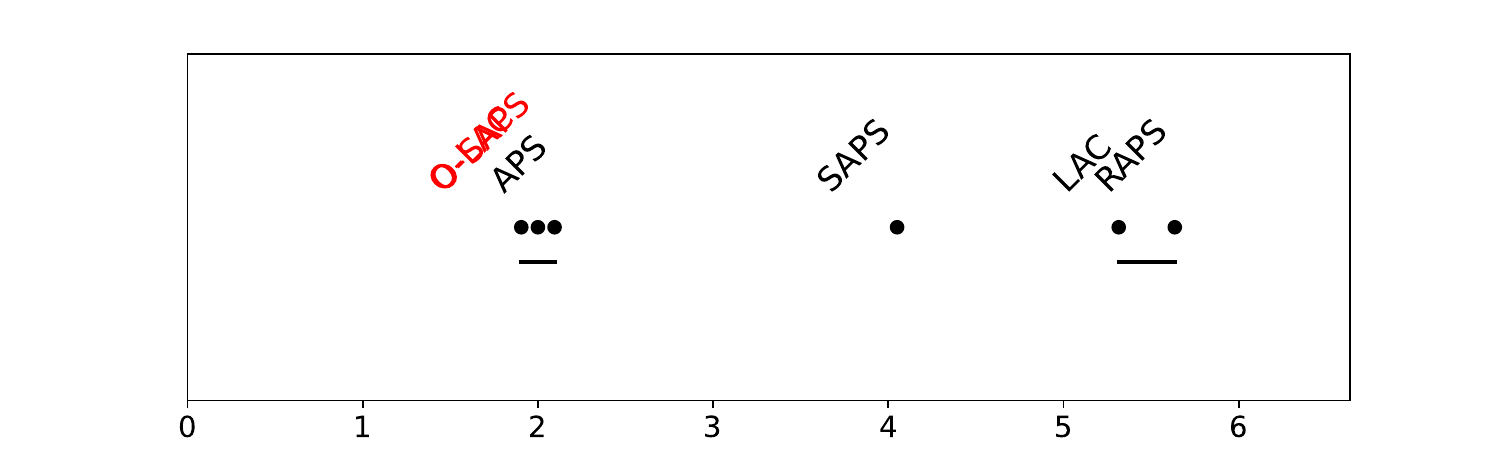}
        \caption{EfficientNet-V2-L}
    \end{subfigure}
    \begin{subfigure}[b]{0.45\textwidth}
        \includegraphics[width=\textwidth]{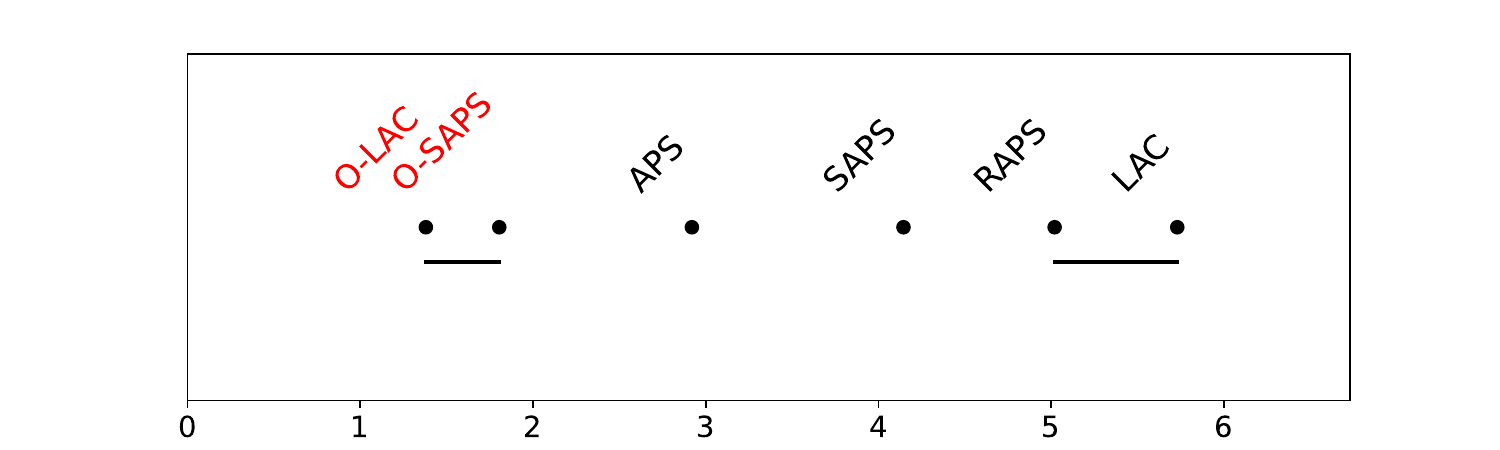}
        \caption{Swin-V2-B}
    \end{subfigure}
    \caption{Critical Difference Diagrams. T-CV, $\alpha=0.05, B=30$. The rank analysis based on these figures is summarized as `Avg. Rank from CD' in Table~\ref{tab:apdx_alg_results_our_metrics_B30} in Appendix.}
    \label{fig:apdx_cd_tcv_B30_alpha0.05}
\end{figure}

\begin{figure}[!bt]
    \centering
    \begin{subfigure}[b]{0.45\textwidth}
        \includegraphics[width=\textwidth]{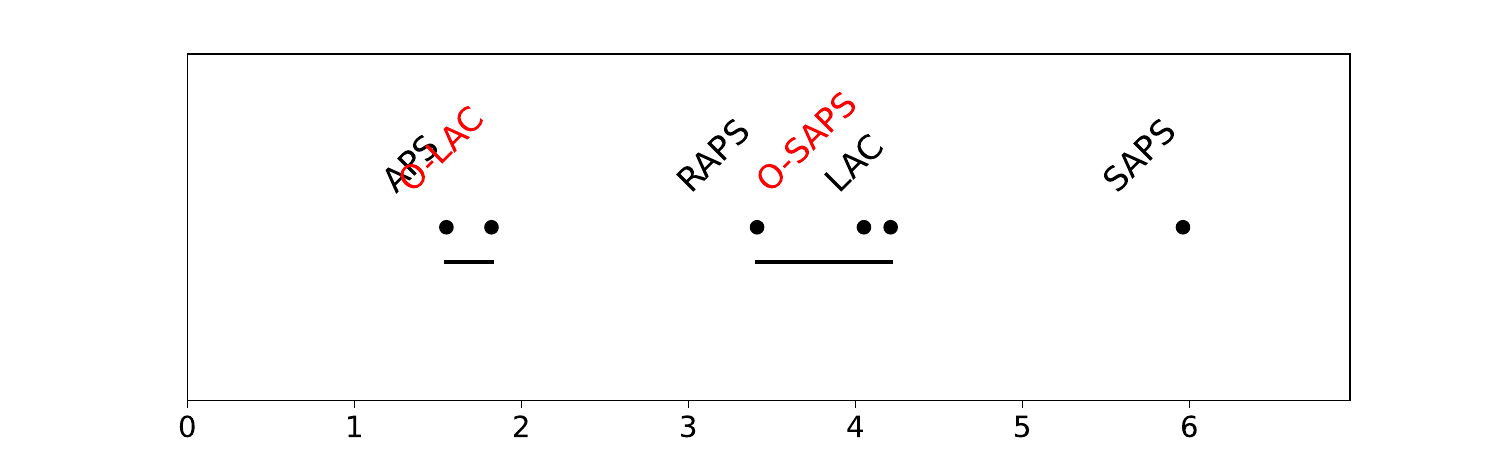}
        \caption{ResNet18}
    \end{subfigure}
    \begin{subfigure}[b]{0.45\textwidth}
        \includegraphics[width=\textwidth]{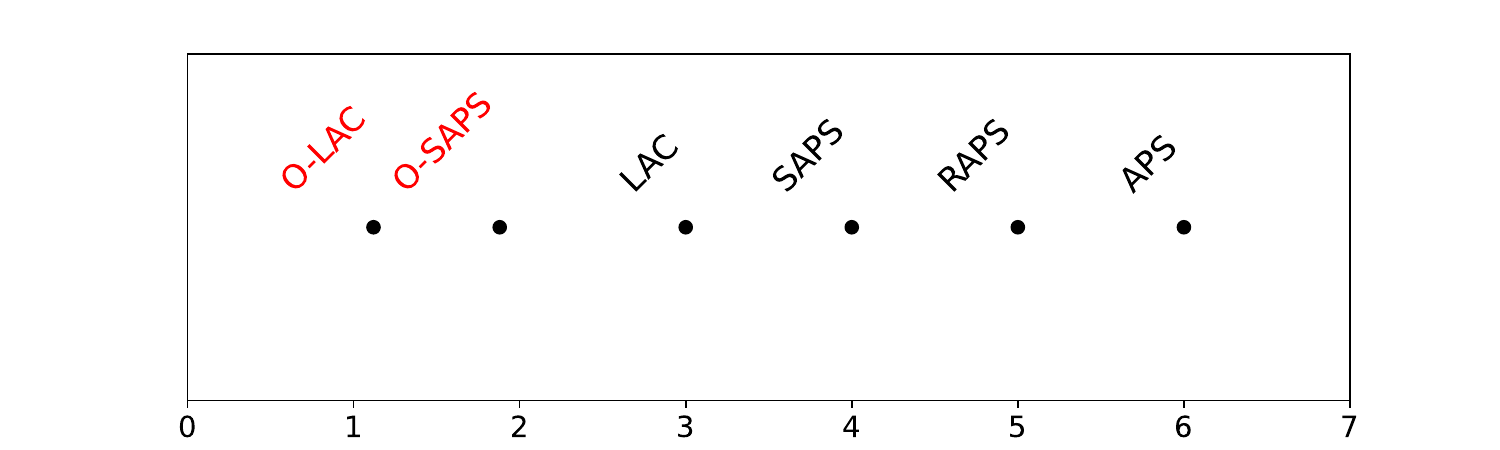}
        \caption{ResNet50}
    \end{subfigure}
    \begin{subfigure}[b]{0.45\textwidth}
        \includegraphics[width=\textwidth]{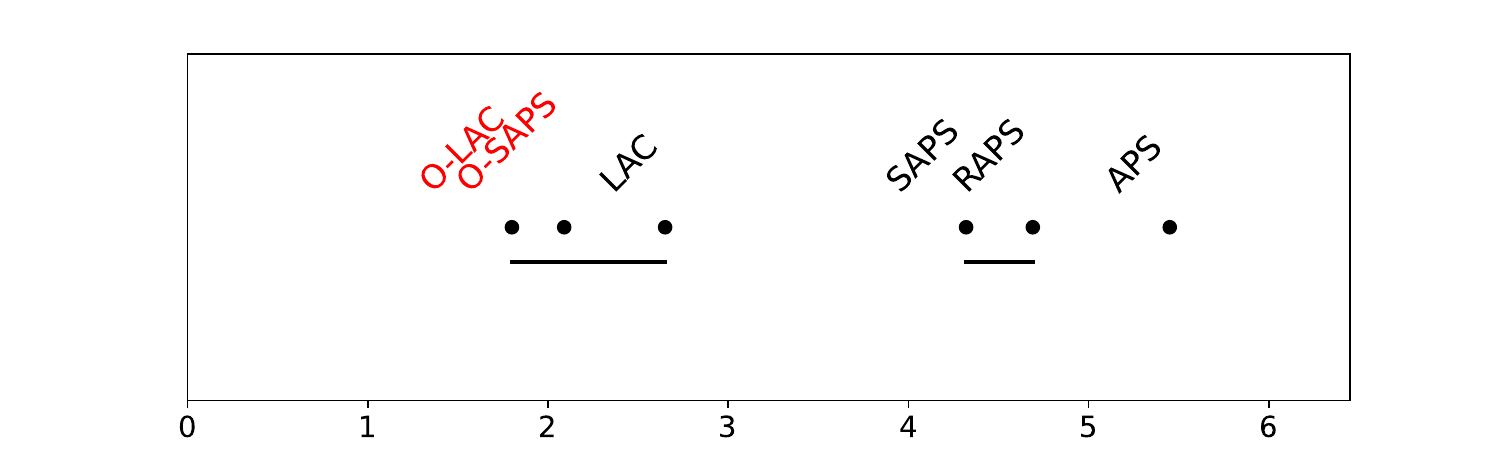}
        \caption{ResNet152}
    \end{subfigure}
    \begin{subfigure}[b]{0.45\textwidth}
        \includegraphics[width=\textwidth]{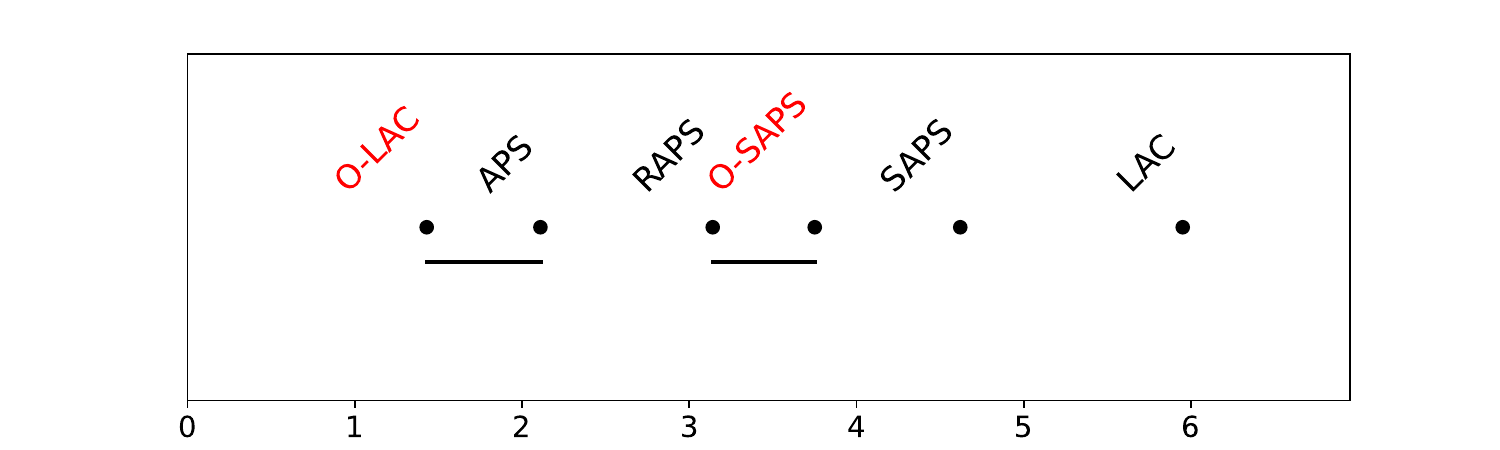}
        \caption{ViT-B-16}
    \end{subfigure}
    \begin{subfigure}[b]{0.45\textwidth}
        \includegraphics[width=\textwidth]{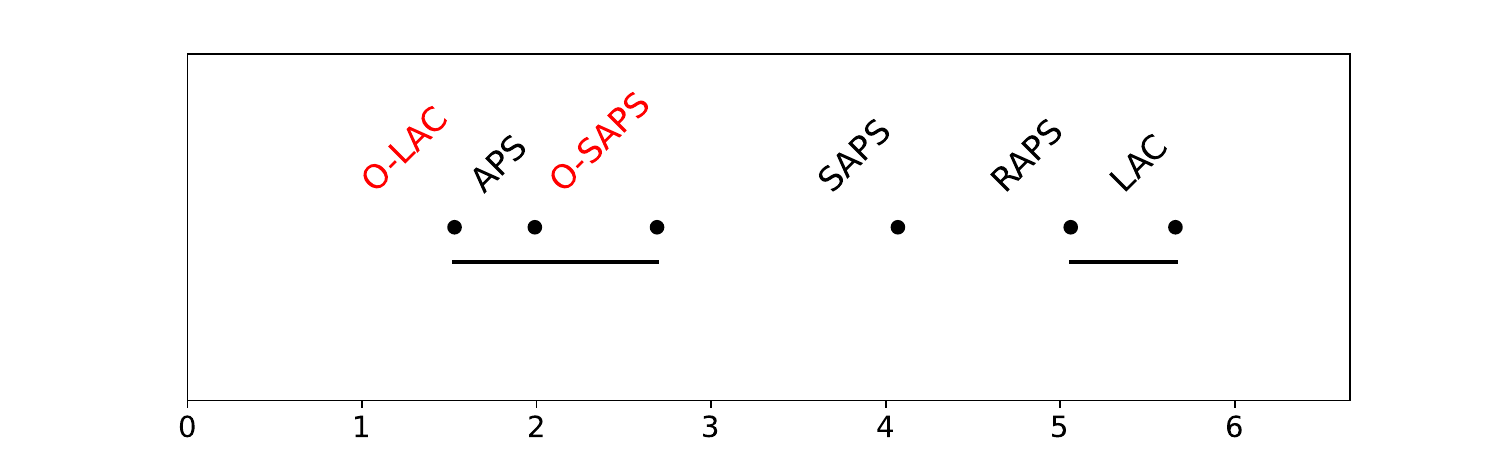}
        \caption{ViT-L-16}
    \end{subfigure}
    \begin{subfigure}[b]{0.45\textwidth}
        \includegraphics[width=\textwidth]{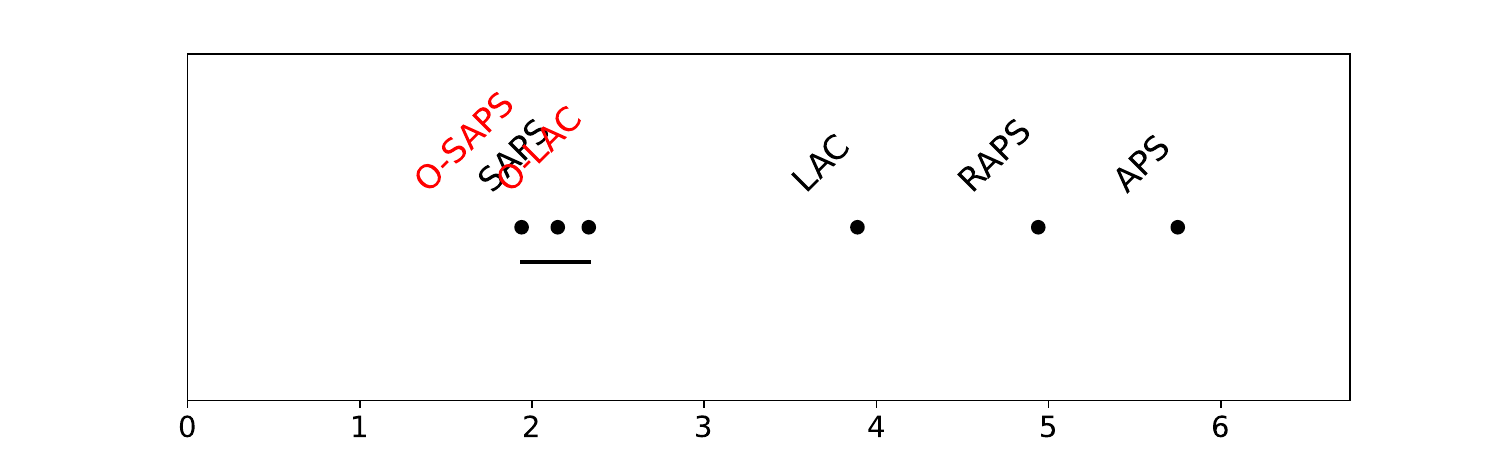}
        \caption{ViT-H-14}
    \end{subfigure}
    \begin{subfigure}[b]{0.45\textwidth}
        \includegraphics[width=\textwidth]{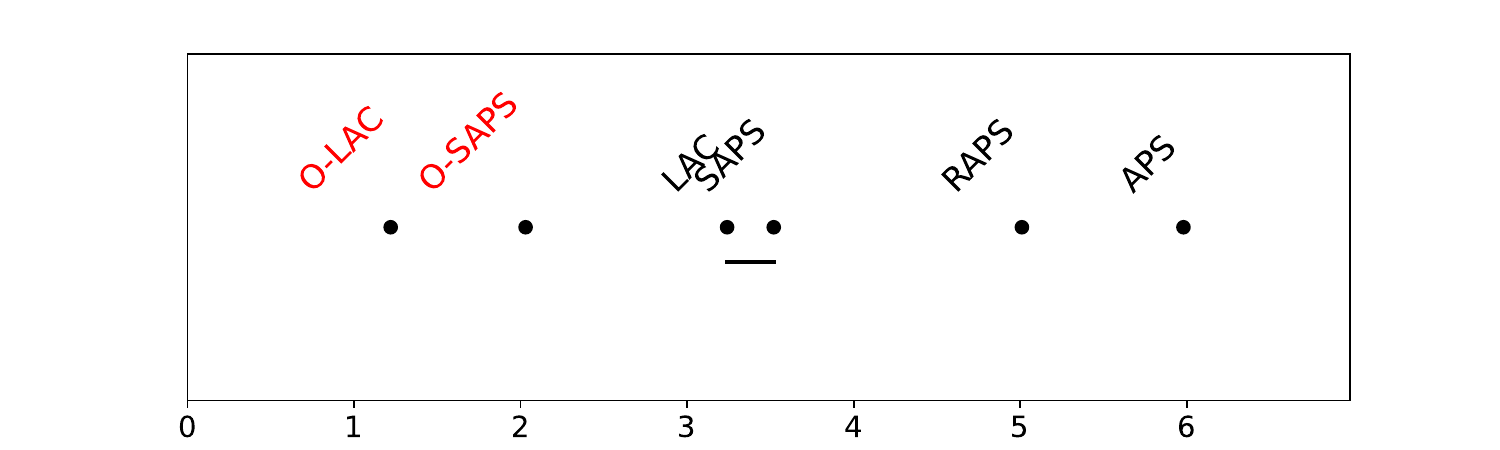}
        \caption{EfficientNet-V2-M}
    \end{subfigure}
    \begin{subfigure}[b]{0.45\textwidth}
        \includegraphics[width=\textwidth]{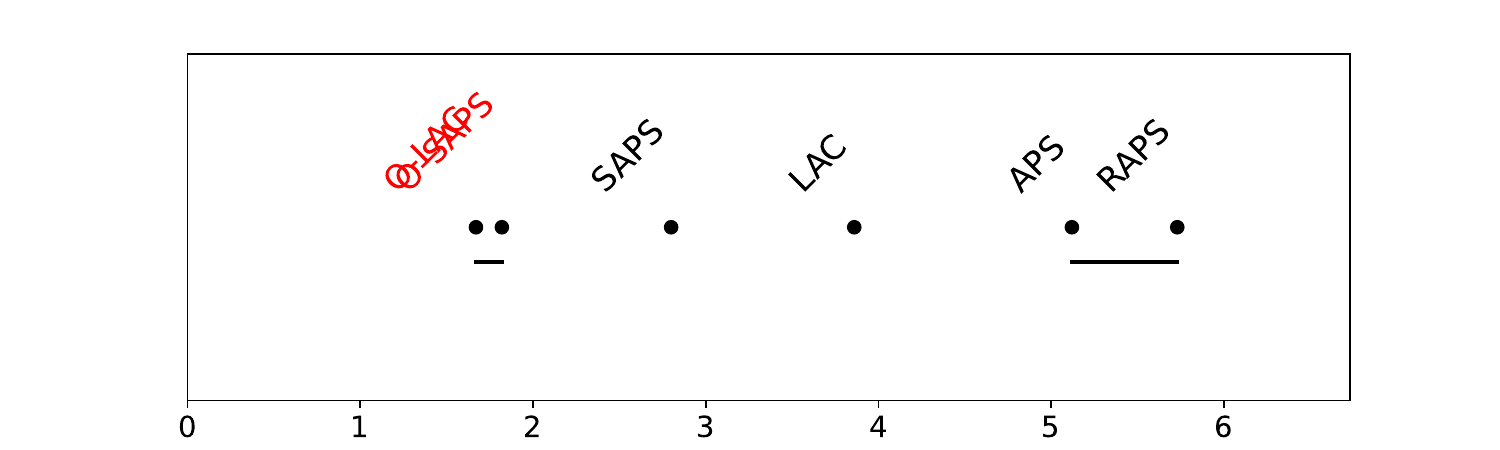}
        \caption{EfficientNet-V2-L}
    \end{subfigure}
    \begin{subfigure}[b]{0.45\textwidth}
        \includegraphics[width=\textwidth]{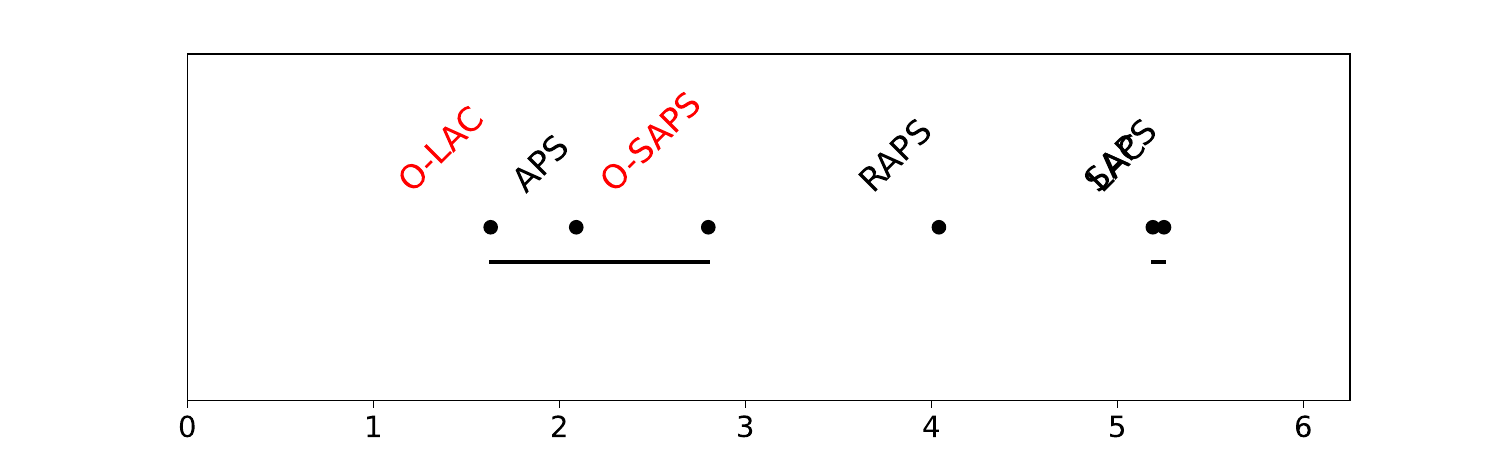}
        \caption{Swin-V2-B}
    \end{subfigure}
    \caption{Critical Difference Diagrams. T-SS, $\alpha=0.05, B=30$. The rank analysis based on these figures is summarized as `Avg. Rank from CD' in Table~\ref{tab:apdx_alg_results_our_metrics_B30} in Appendix.}
    \label{fig:apdx_cd_tss_B30_alpha0.05}
\end{figure}

\begin{figure}[!bt]
    \centering
    \begin{subfigure}[b]{0.45\textwidth}
        \includegraphics[width=\textwidth]{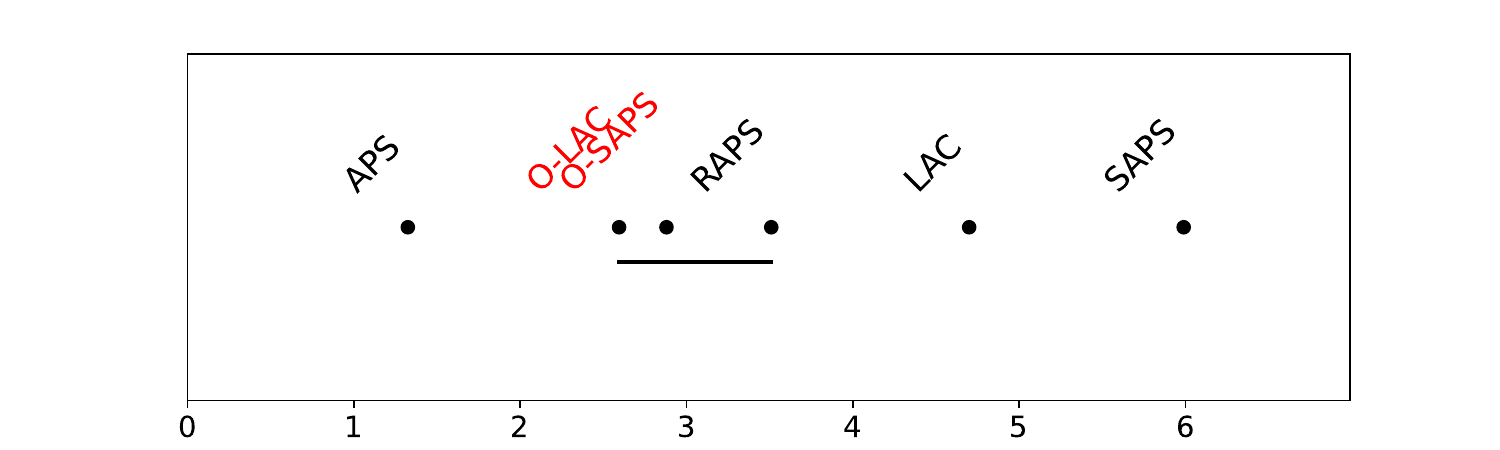}
        \caption{ResNet18}
    \end{subfigure}
    \begin{subfigure}[b]{0.45\textwidth}
        \includegraphics[width=\textwidth]{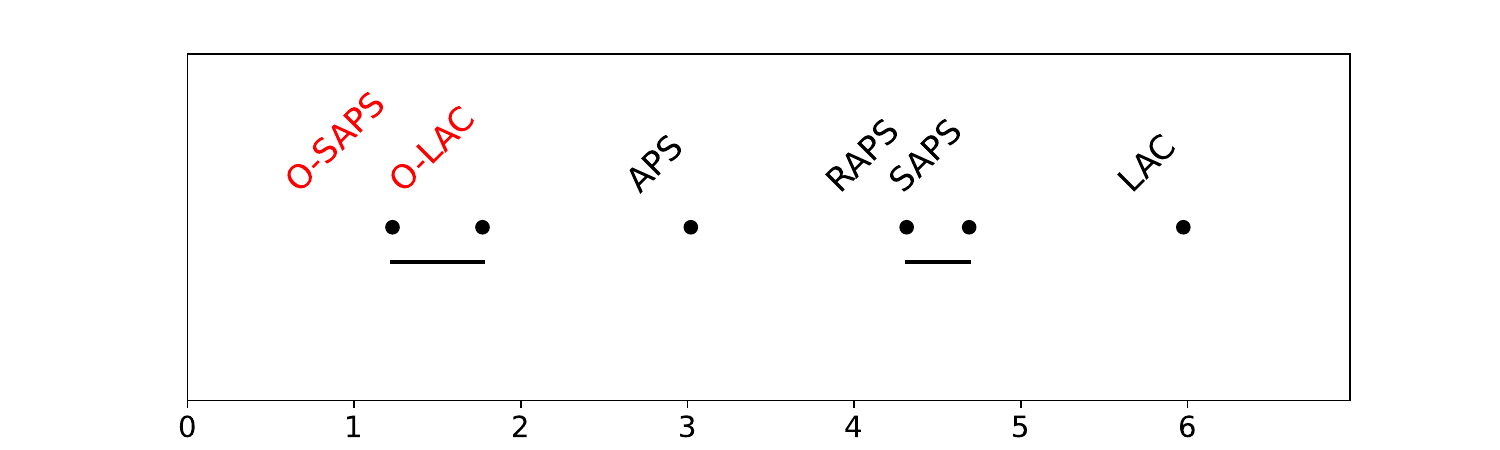}
        \caption{ResNet50}
    \end{subfigure}
    \begin{subfigure}[b]{0.45\textwidth}
        \includegraphics[width=\textwidth]{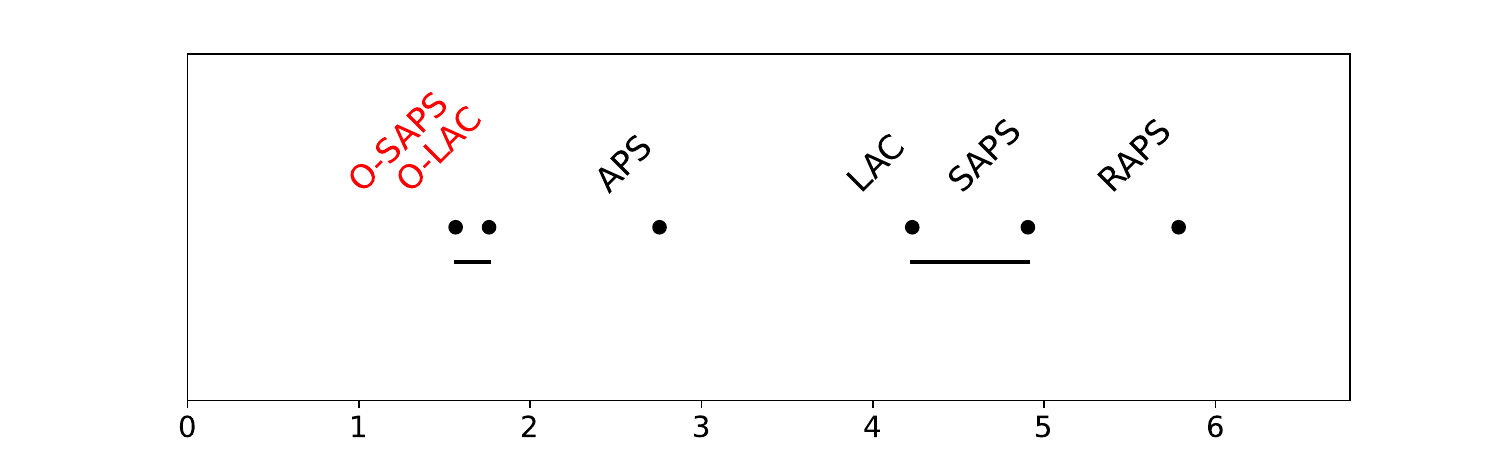}
        \caption{ResNet152}
    \end{subfigure}
    \begin{subfigure}[b]{0.45\textwidth}
        \includegraphics[width=\textwidth]{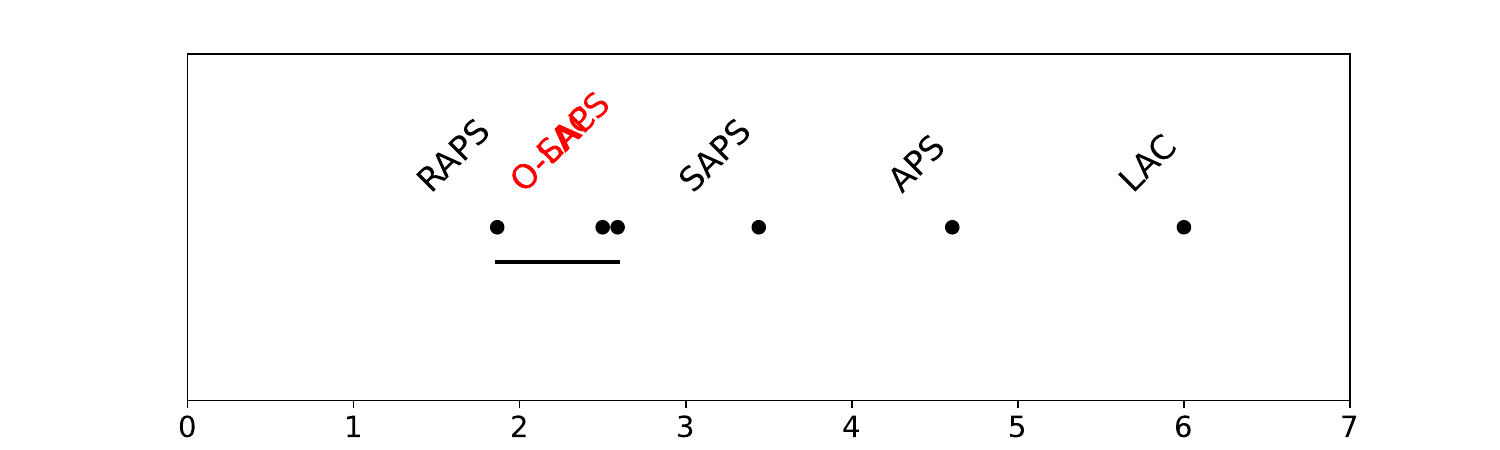}
        \caption{ViT-B-16}
    \end{subfigure}
    \begin{subfigure}[b]{0.45\textwidth}
        \includegraphics[width=\textwidth]{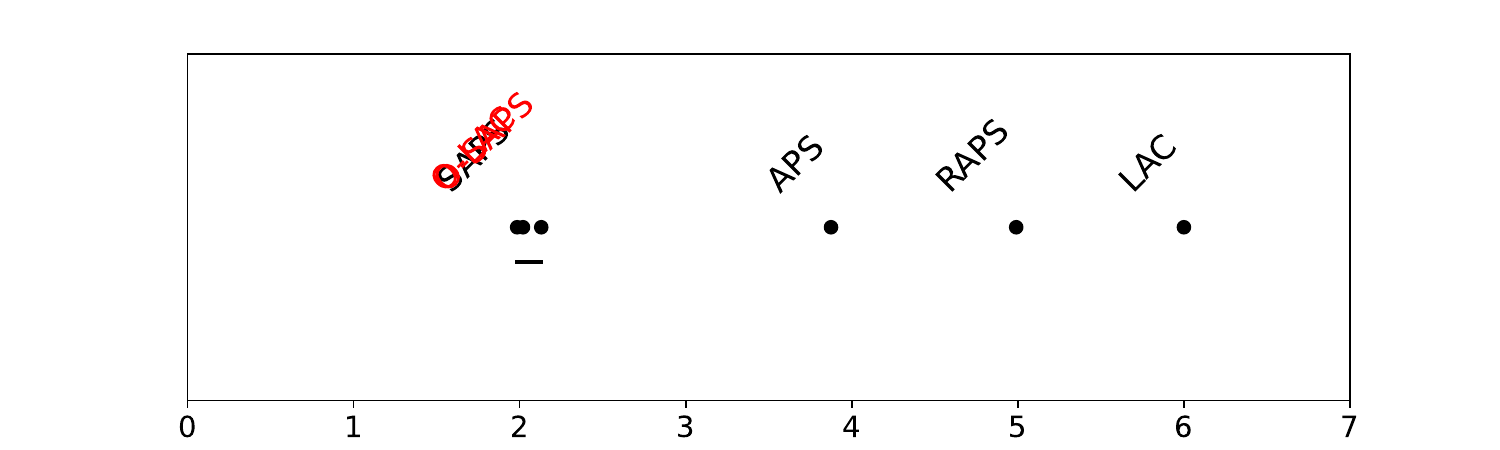}
        \caption{ViT-L-16}
    \end{subfigure}
    \begin{subfigure}[b]{0.45\textwidth}
        \includegraphics[width=\textwidth]{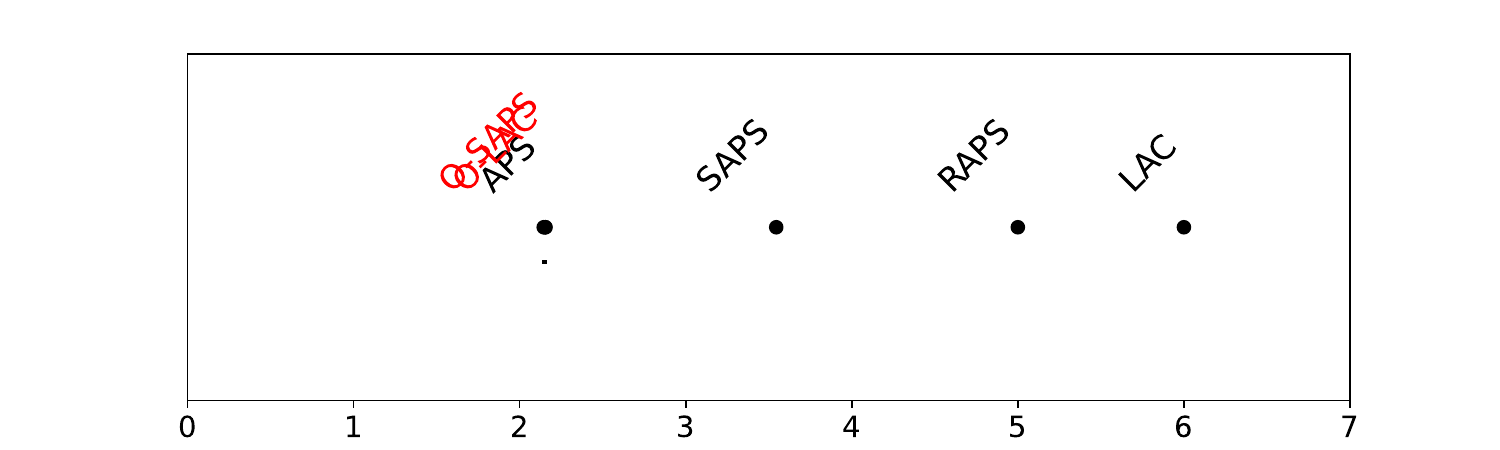}
        \caption{ViT-H-14}
    \end{subfigure}
    \begin{subfigure}[b]{0.45\textwidth}
        \includegraphics[width=\textwidth]{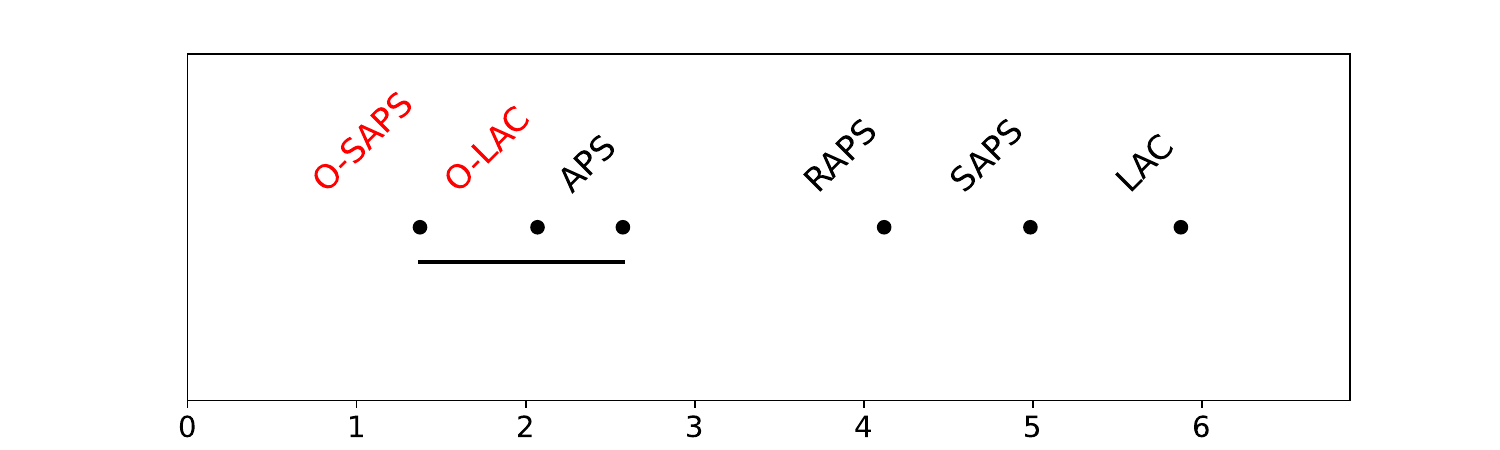}
        \caption{EfficientNet-V2-M}
    \end{subfigure}
    \begin{subfigure}[b]{0.45\textwidth}
        \includegraphics[width=\textwidth]{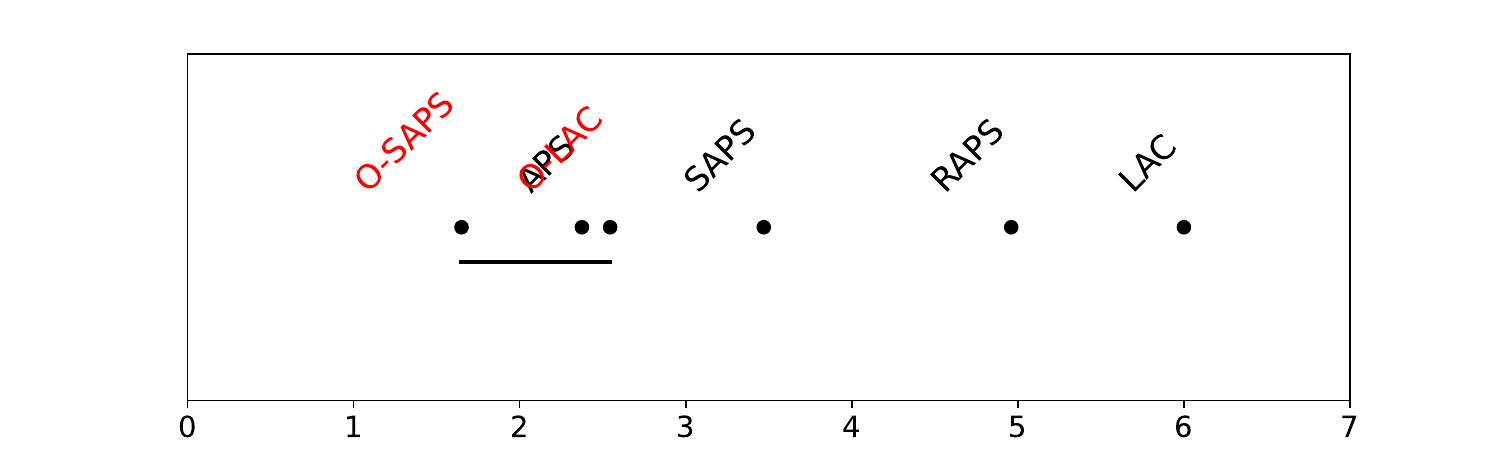}
        \caption{EfficientNet-V2-L}
    \end{subfigure}
    \begin{subfigure}[b]{0.45\textwidth}
        \includegraphics[width=\textwidth]{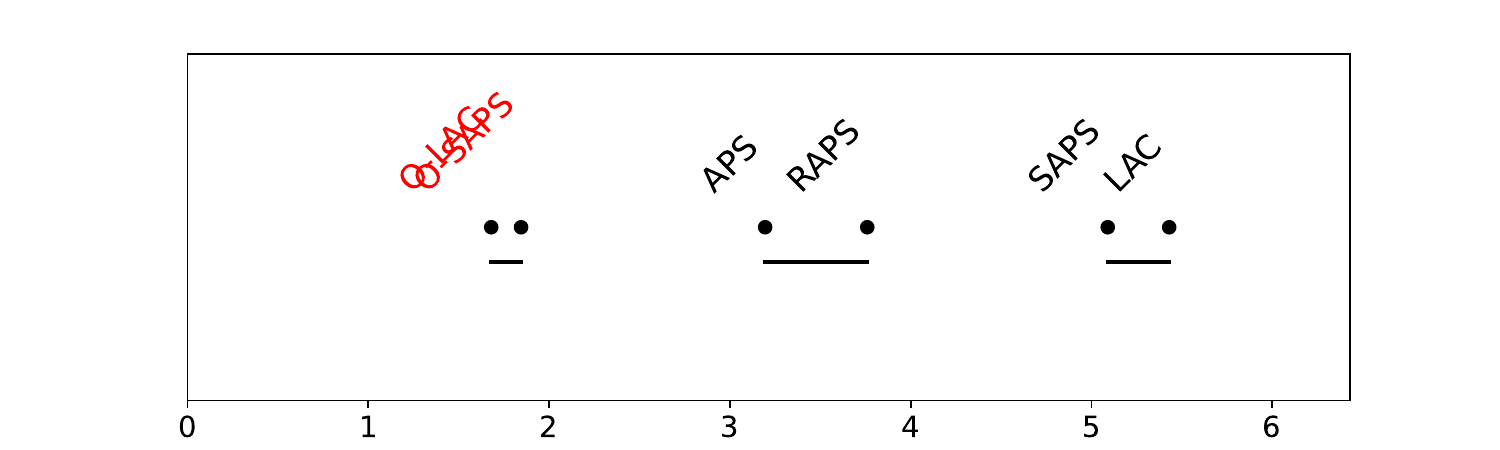}
        \caption{Swin-V2-B}
    \end{subfigure}
    \caption{Critical Difference Diagrams. T-CV, $\alpha=0.10, B=30$. The rank analysis based on these figures is summarized as `Avg. Rank from CD' in Table~\ref{tab:apdx_alg_results_our_metrics_B30} in Appendix.}
    \label{fig:apdx_cd_tcv_B30_alpha0.10}
\end{figure}

\begin{figure}[!bt]
    \centering
    \begin{subfigure}[b]{0.45\textwidth}
        \includegraphics[width=\textwidth]{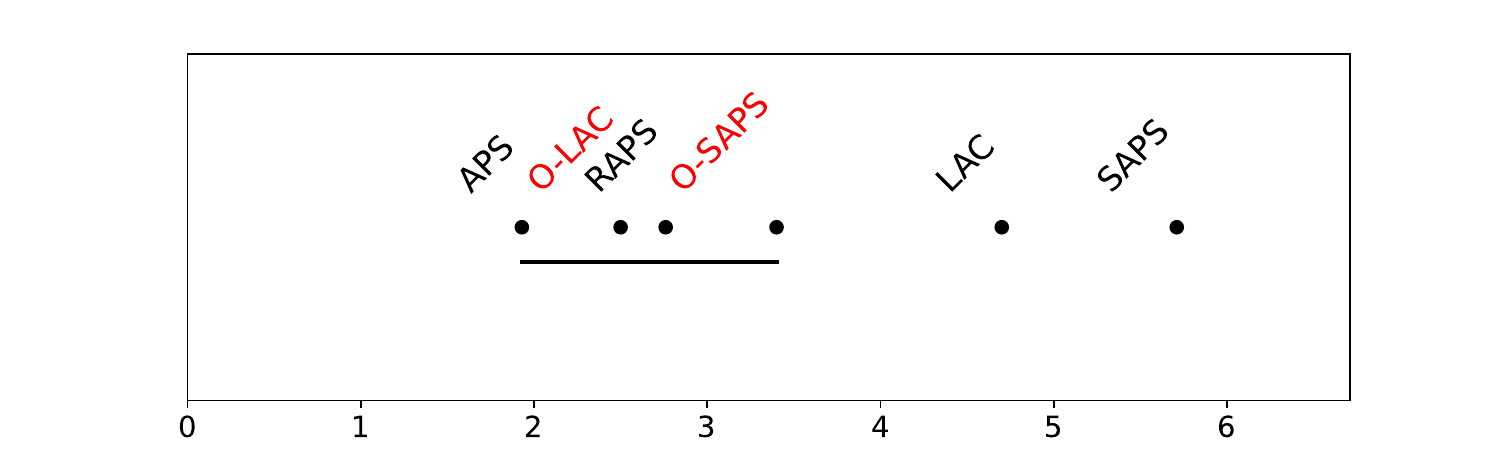}
        \caption{ResNet18}
    \end{subfigure}
    \begin{subfigure}[b]{0.45\textwidth}
        \includegraphics[width=\textwidth]{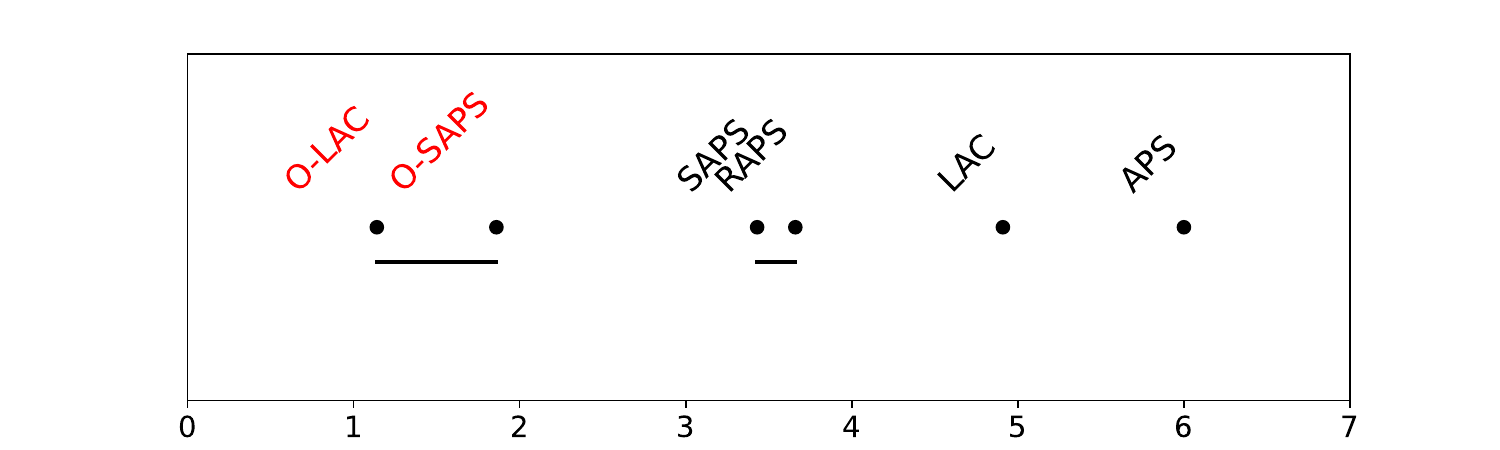}
        \caption{ResNet50}
    \end{subfigure}
    \begin{subfigure}[b]{0.45\textwidth}
        \includegraphics[width=\textwidth]{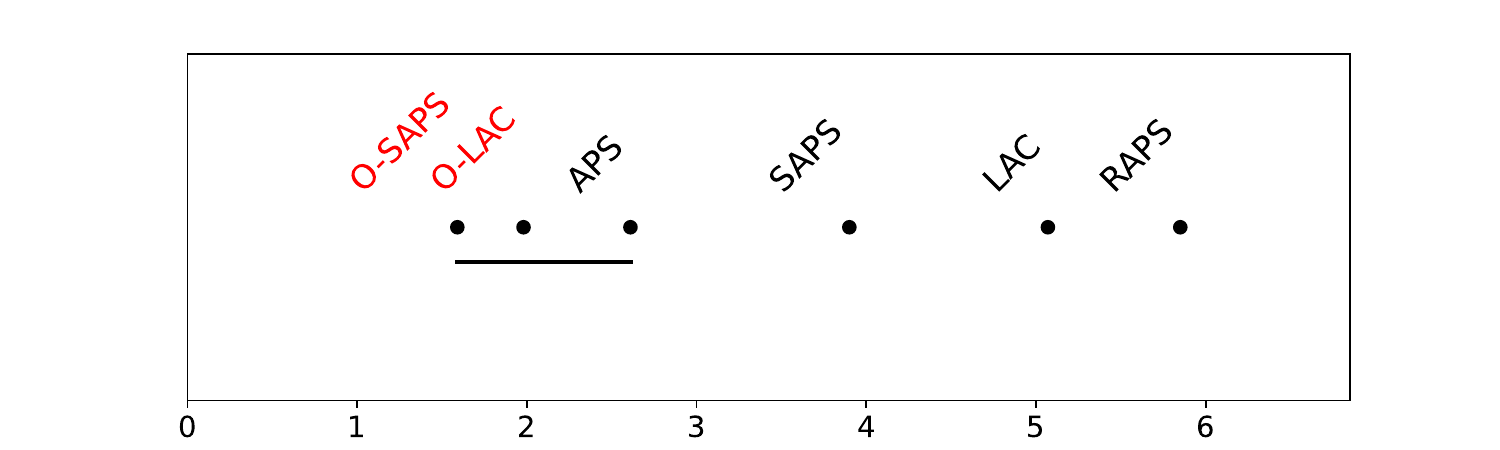}
        \caption{ResNet152}
    \end{subfigure}
    \begin{subfigure}[b]{0.45\textwidth}
        \includegraphics[width=\textwidth]{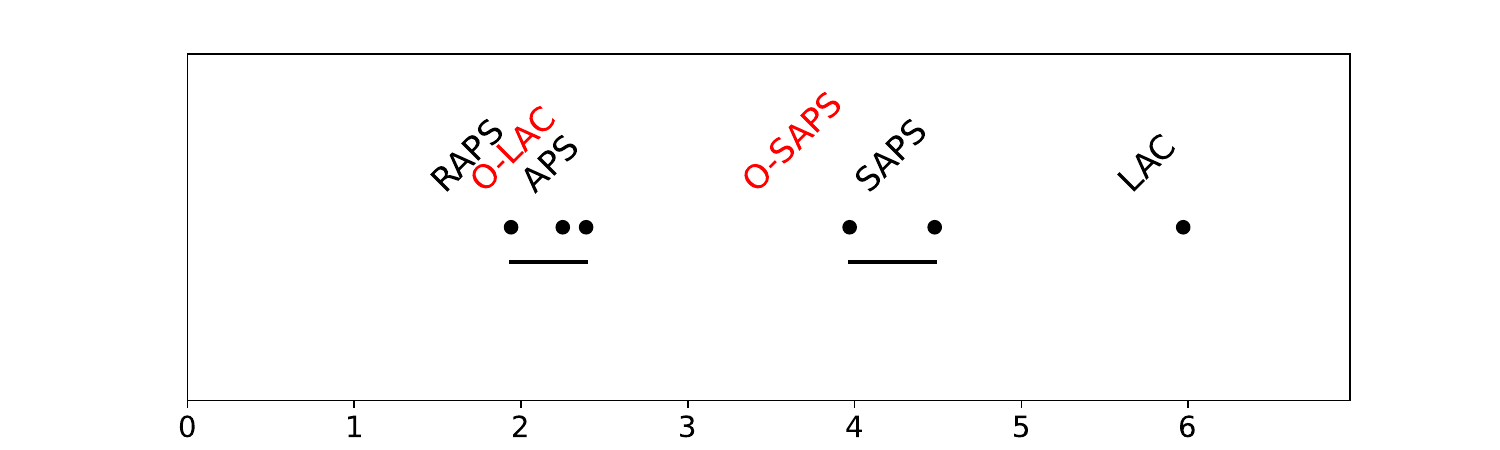}
        \caption{ViT-B-16}
    \end{subfigure}
    \begin{subfigure}[b]{0.45\textwidth}
        \includegraphics[width=\textwidth]{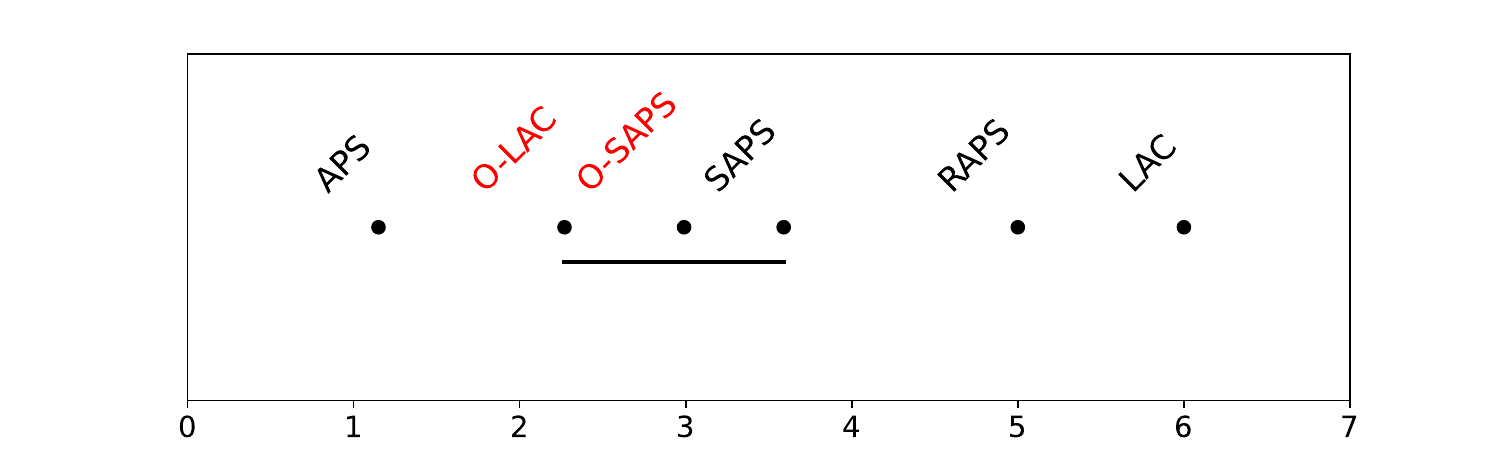}
        \caption{ViT-L-16}
    \end{subfigure}
    \begin{subfigure}[b]{0.45\textwidth}
        \includegraphics[width=\textwidth]{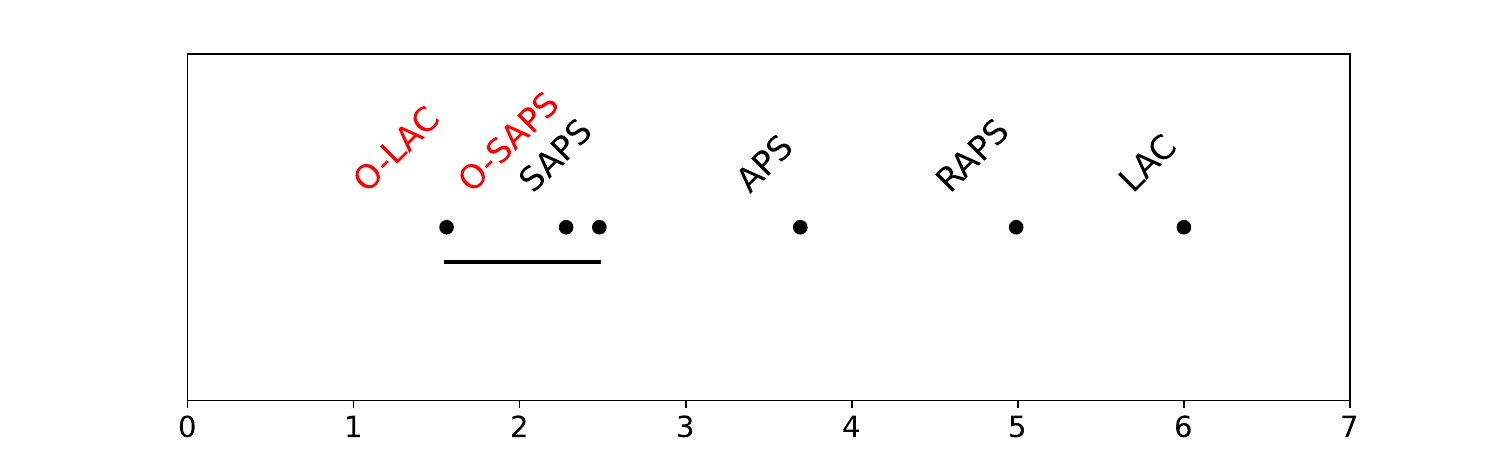}
        \caption{ViT-H-14}
    \end{subfigure}
    \begin{subfigure}[b]{0.45\textwidth}
        \includegraphics[width=\textwidth]{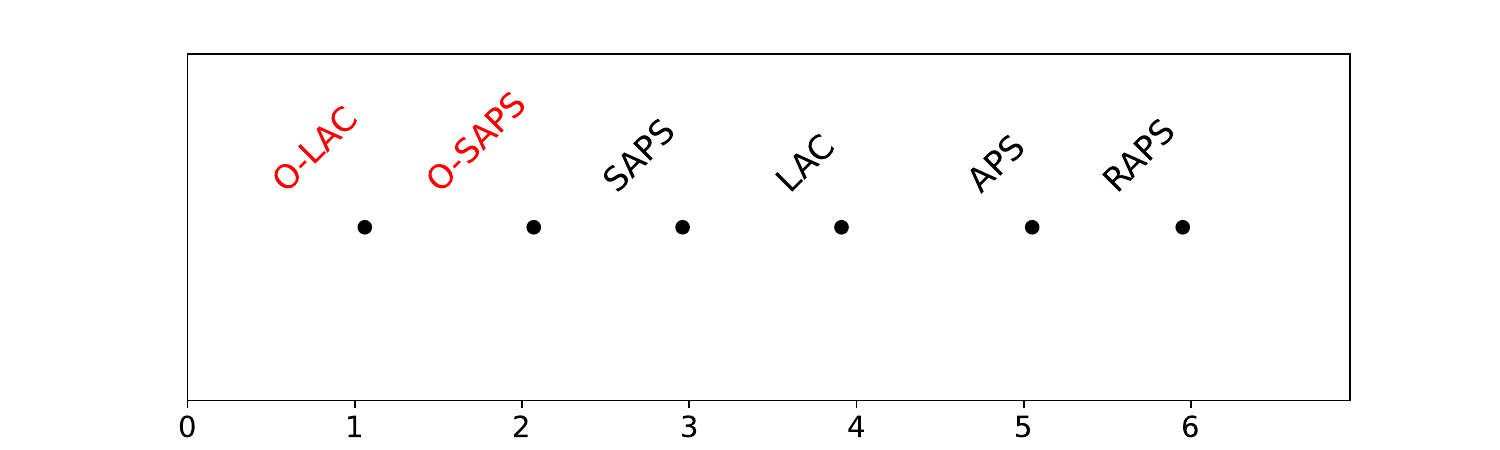}
        \caption{EfficientNet-V2-M}
    \end{subfigure}
    \begin{subfigure}[b]{0.45\textwidth}
        \includegraphics[width=\textwidth]{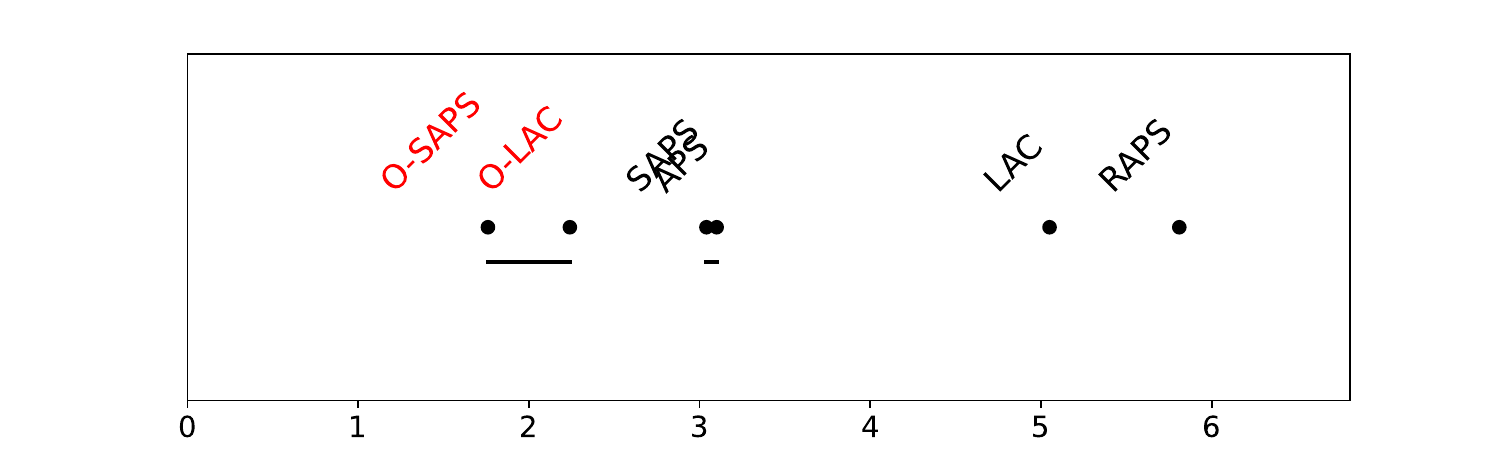}
        \caption{EfficientNet-V2-L}
    \end{subfigure}
    \begin{subfigure}[b]{0.45\textwidth}
        \includegraphics[width=\textwidth]{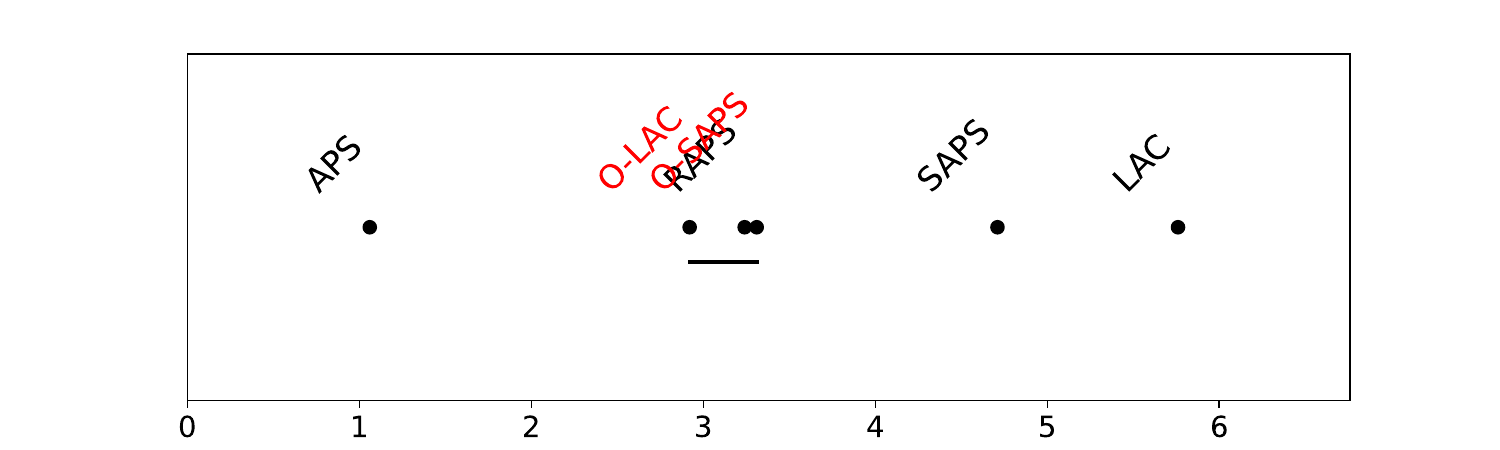}
        \caption{Swin-V2-B}
    \end{subfigure}
    \caption{Critical Difference Diagrams. T-SS, $\alpha=0.10, B=30$. The rank analysis based on these figures is summarized as `Avg. Rank from CD' in Table~\ref{tab:apdx_alg_results_our_metrics_B30} in Appendix.}
    \label{fig:apdx_cd_tss_B30_alpha0.10}
\end{figure}

\begin{figure}[!bt]
    \centering
    \begin{subfigure}[b]{0.45\textwidth}
        \includegraphics[width=\textwidth]{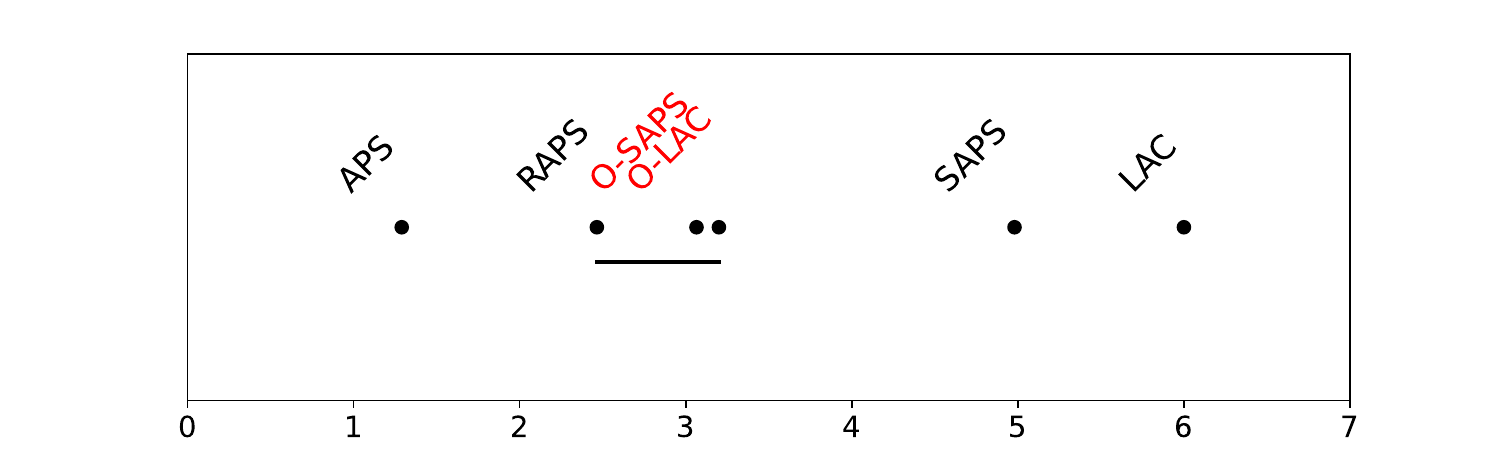}
        \caption{ResNet18}
    \end{subfigure}
    \begin{subfigure}[b]{0.45\textwidth}
        \includegraphics[width=\textwidth]{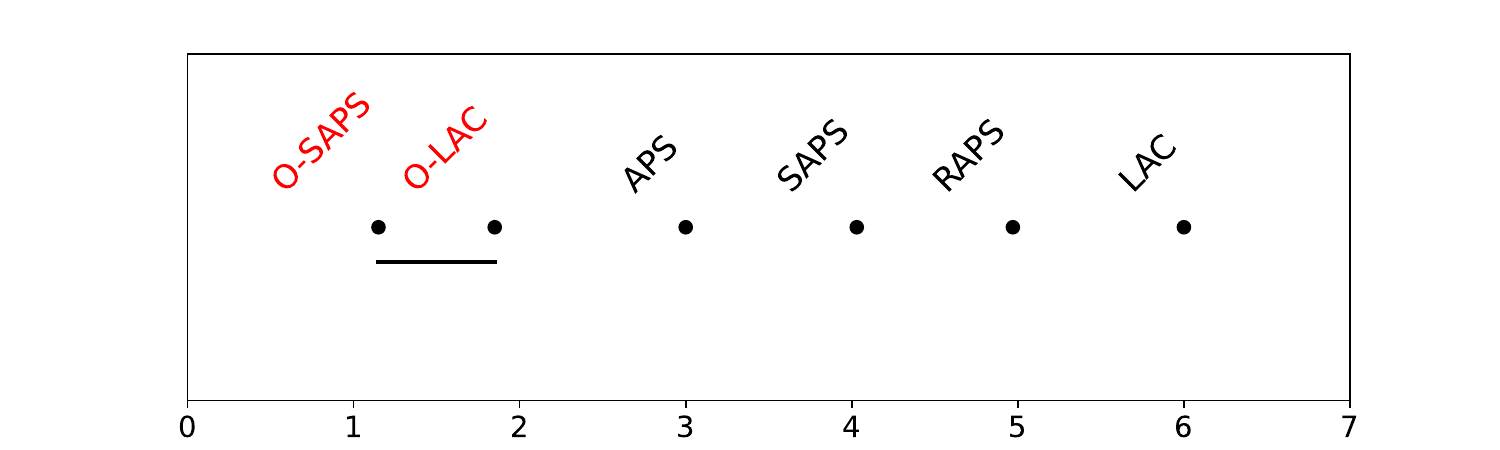}
        \caption{ResNet50}
    \end{subfigure}
    \begin{subfigure}[b]{0.45\textwidth}
        \includegraphics[width=\textwidth]{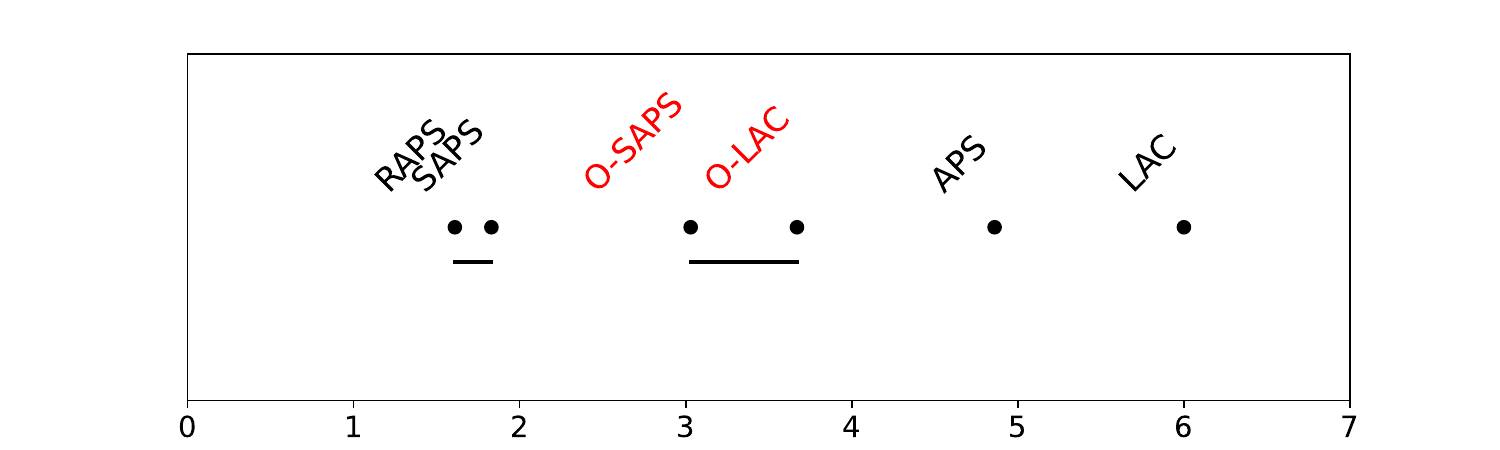}
        \caption{ResNet152}
    \end{subfigure}
    \begin{subfigure}[b]{0.45\textwidth}
        \includegraphics[width=\textwidth]{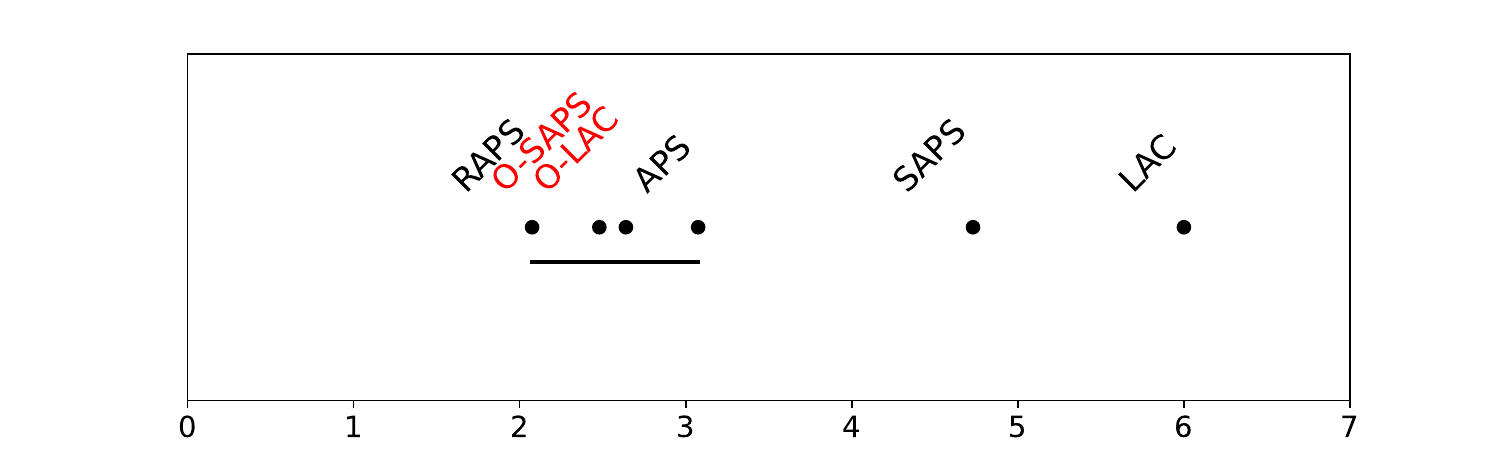}
        \caption{ViT-B-16}
    \end{subfigure}
    \begin{subfigure}[b]{0.45\textwidth}
        \includegraphics[width=\textwidth]{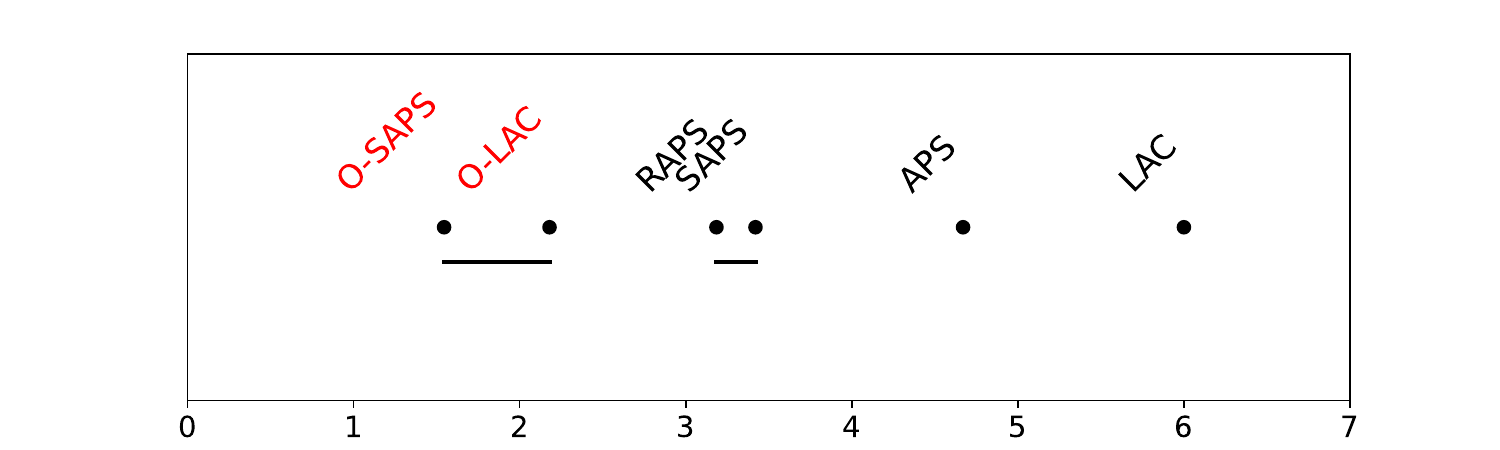}
        \caption{ViT-L-16}
    \end{subfigure}
    \begin{subfigure}[b]{0.45\textwidth}
        \includegraphics[width=\textwidth]{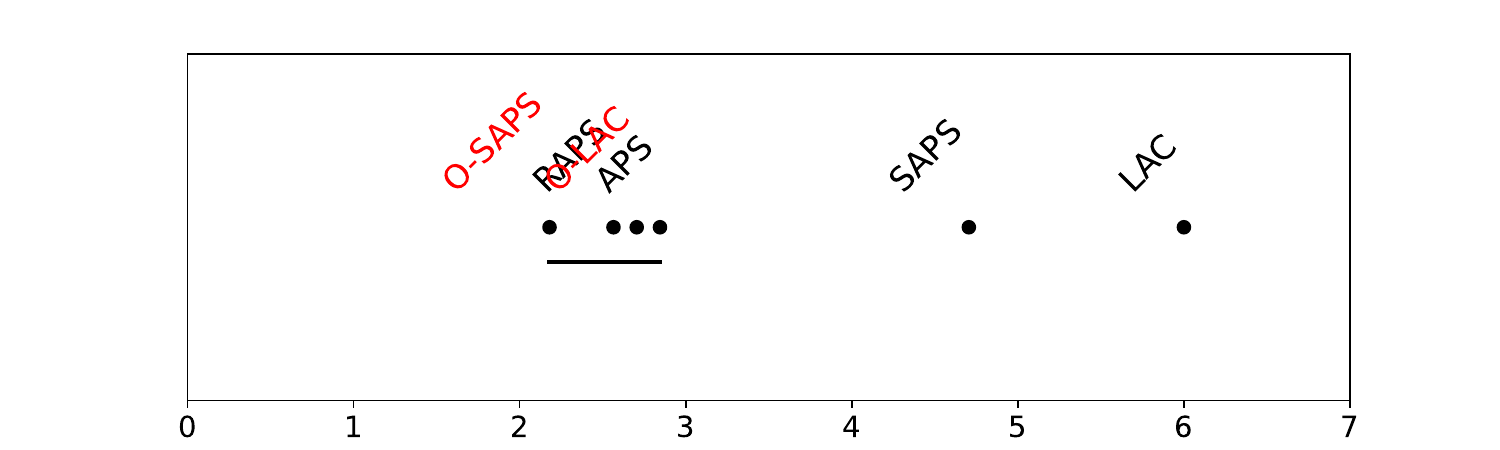}
        \caption{ViT-H-14}
    \end{subfigure}
    \begin{subfigure}[b]{0.45\textwidth}
        \includegraphics[width=\textwidth]{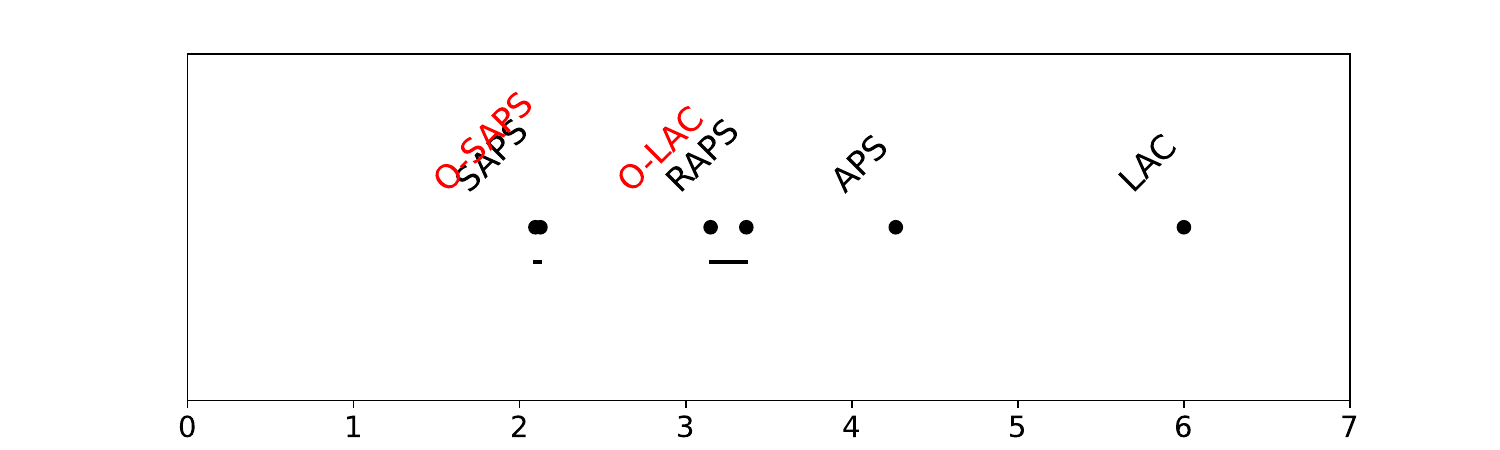}
        \caption{EfficientNet-V2-M}
    \end{subfigure}
    \begin{subfigure}[b]{0.45\textwidth}
        \includegraphics[width=\textwidth]{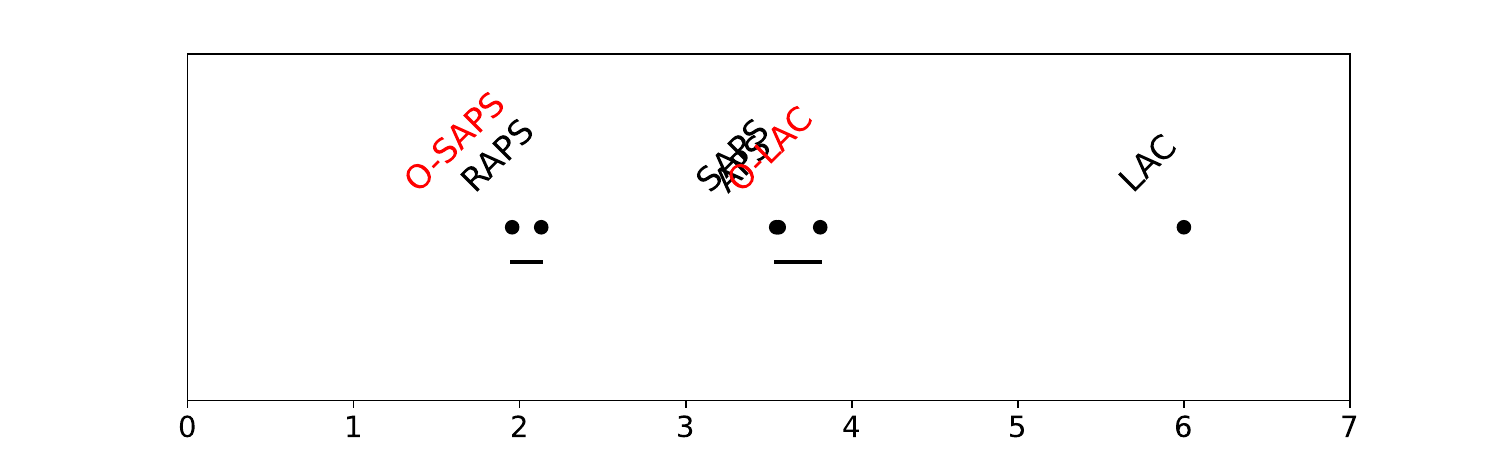}
        \caption{EfficientNet-V2-L}
    \end{subfigure}
    \begin{subfigure}[b]{0.45\textwidth}
        \includegraphics[width=\textwidth]{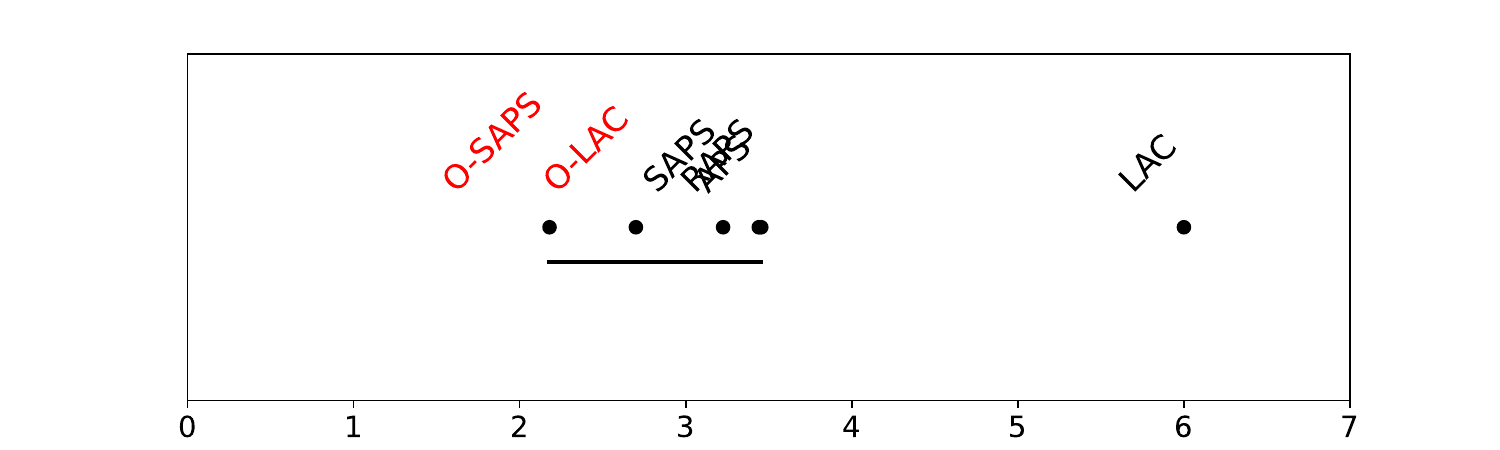}
        \caption{Swin-V2-B}
    \end{subfigure}
    \caption{Critical Difference Diagrams. T-CV, $\alpha=0.15, B=30$. The rank analysis based on these figures is summarized as `Avg. Rank from CD' in Table~\ref{tab:apdx_alg_results_our_metrics_B30} in Appendix.}
    \label{fig:apdx_cd_tcv_B30_alpha0.15}
\end{figure}

\begin{figure}[!bt]
    \centering
    \begin{subfigure}[b]{0.45\textwidth}
        \includegraphics[width=\textwidth]{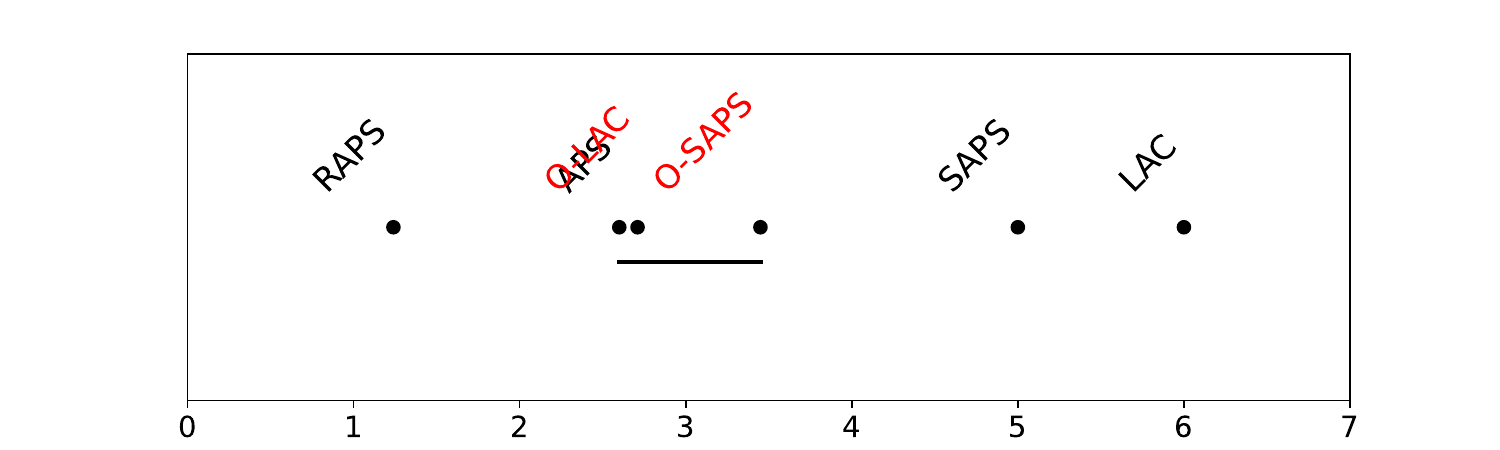}
        \caption{ResNet18}
    \end{subfigure}
    \begin{subfigure}[b]{0.45\textwidth}
        \includegraphics[width=\textwidth]{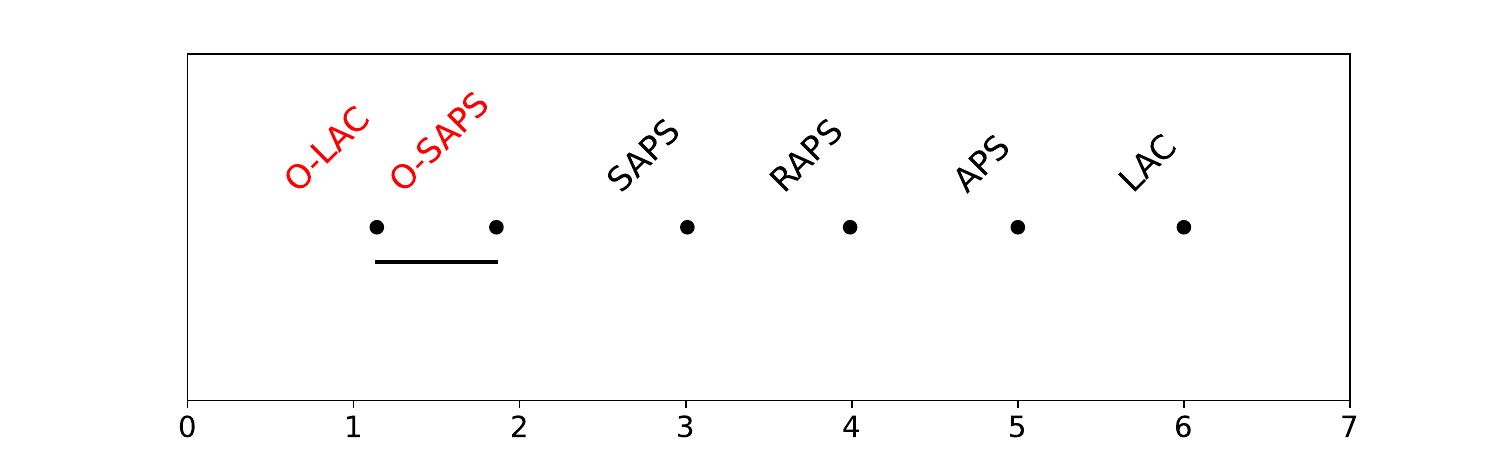}
        \caption{ResNet50}
    \end{subfigure}
    \begin{subfigure}[b]{0.45\textwidth}
        \includegraphics[width=\textwidth]{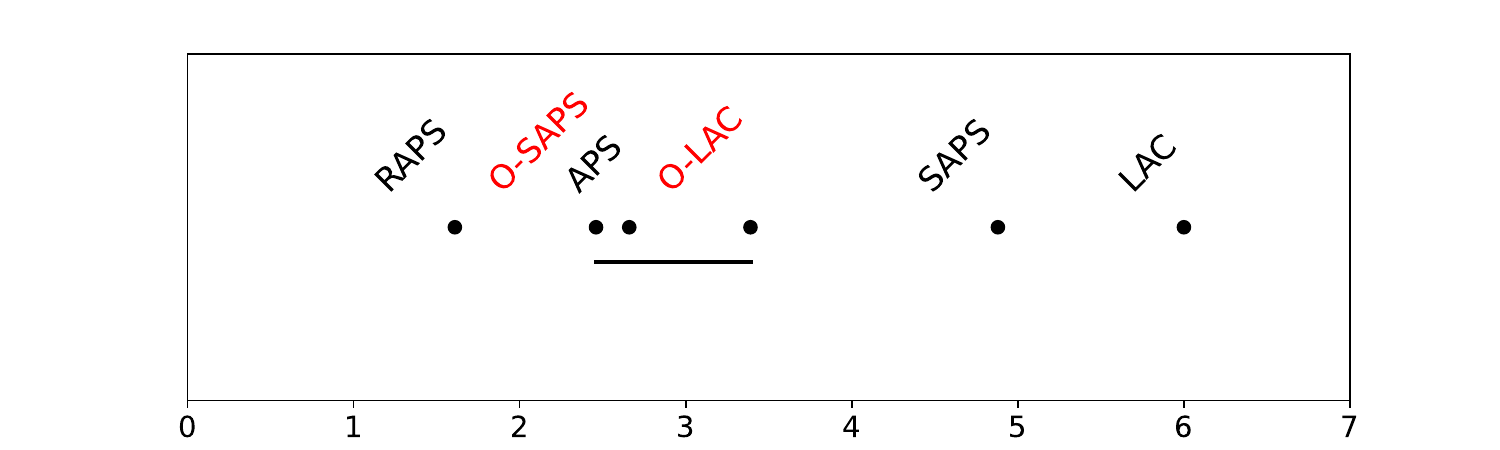}
        \caption{ResNet152}
    \end{subfigure}
    \begin{subfigure}[b]{0.45\textwidth}
        \includegraphics[width=\textwidth]{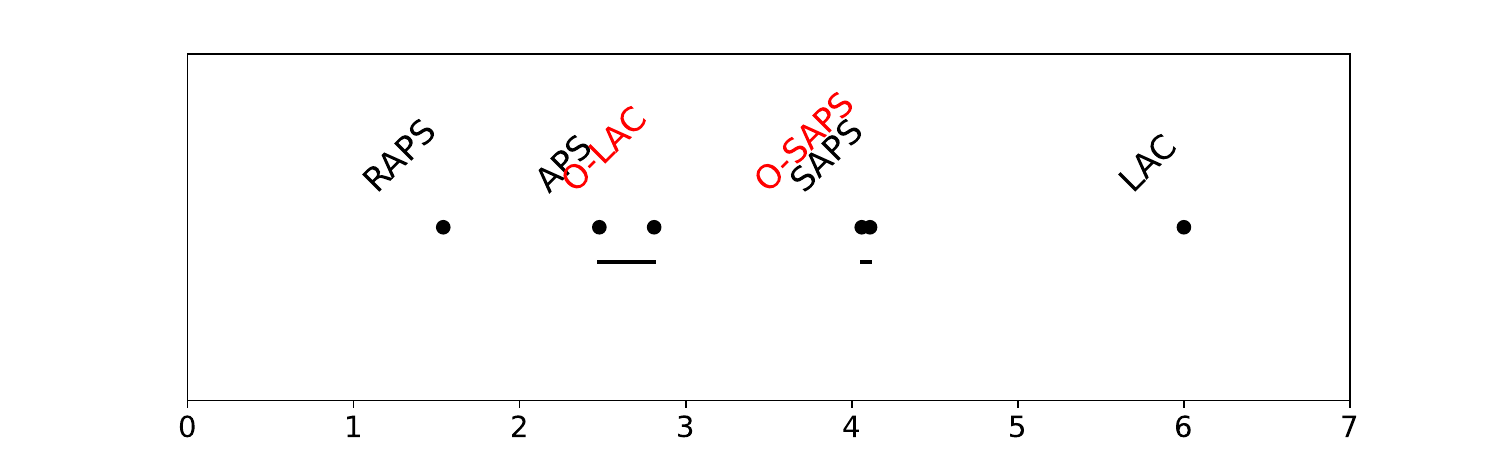}
        \caption{ViT-B-16}
    \end{subfigure}
    \begin{subfigure}[b]{0.45\textwidth}
        \includegraphics[width=\textwidth]{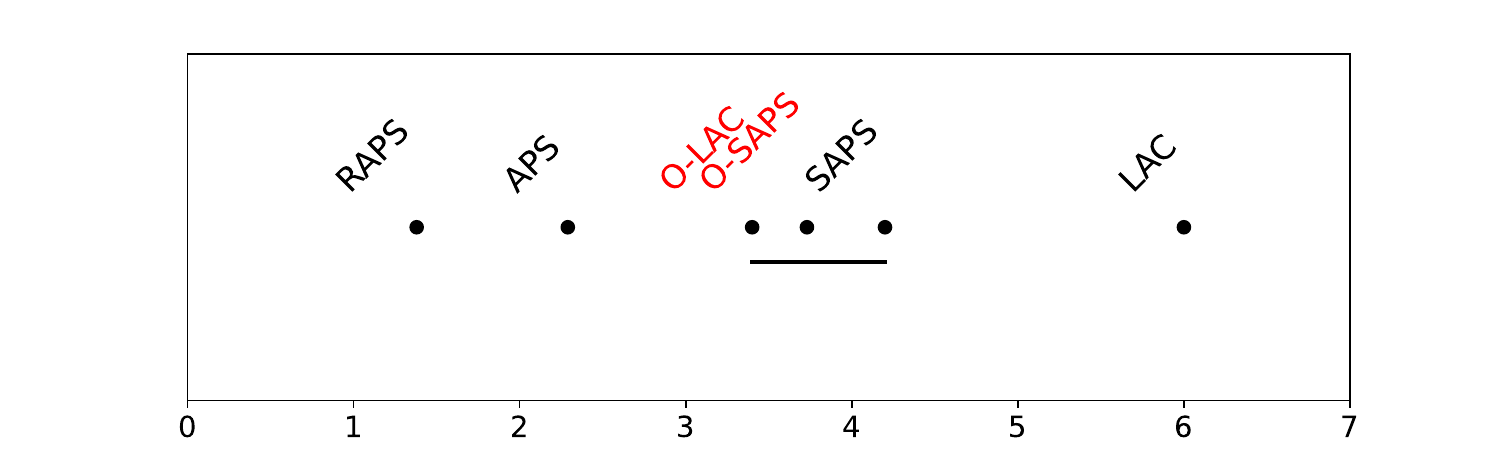}
        \caption{ViT-L-16}
    \end{subfigure}
    \begin{subfigure}[b]{0.45\textwidth}
        \includegraphics[width=\textwidth]{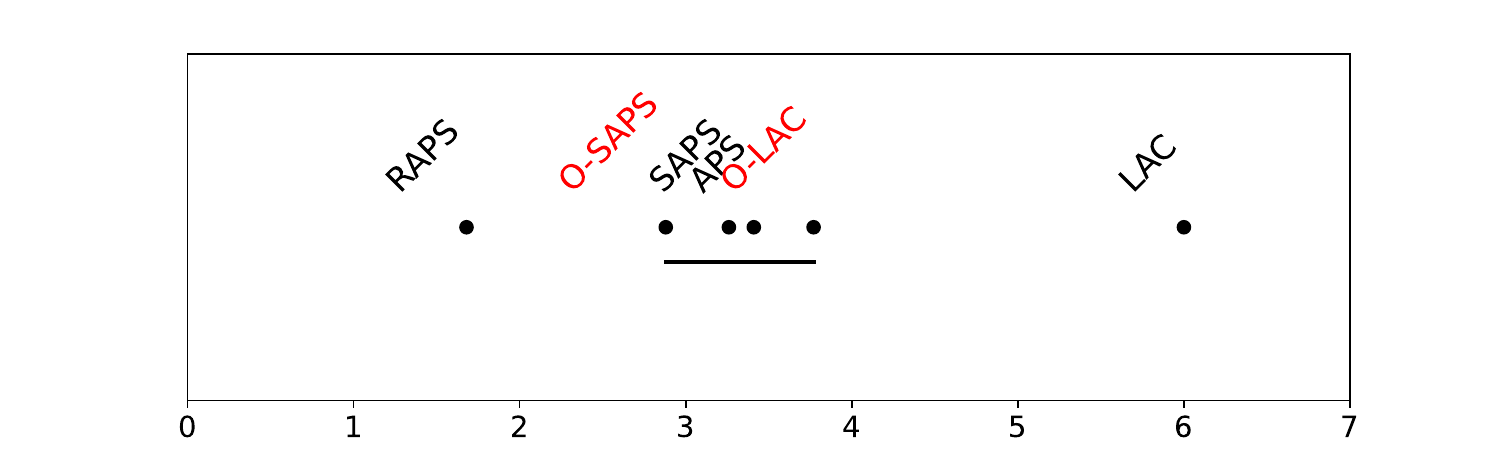}
        \caption{ViT-H-14}
    \end{subfigure}
    \begin{subfigure}[b]{0.45\textwidth}
        \includegraphics[width=\textwidth]{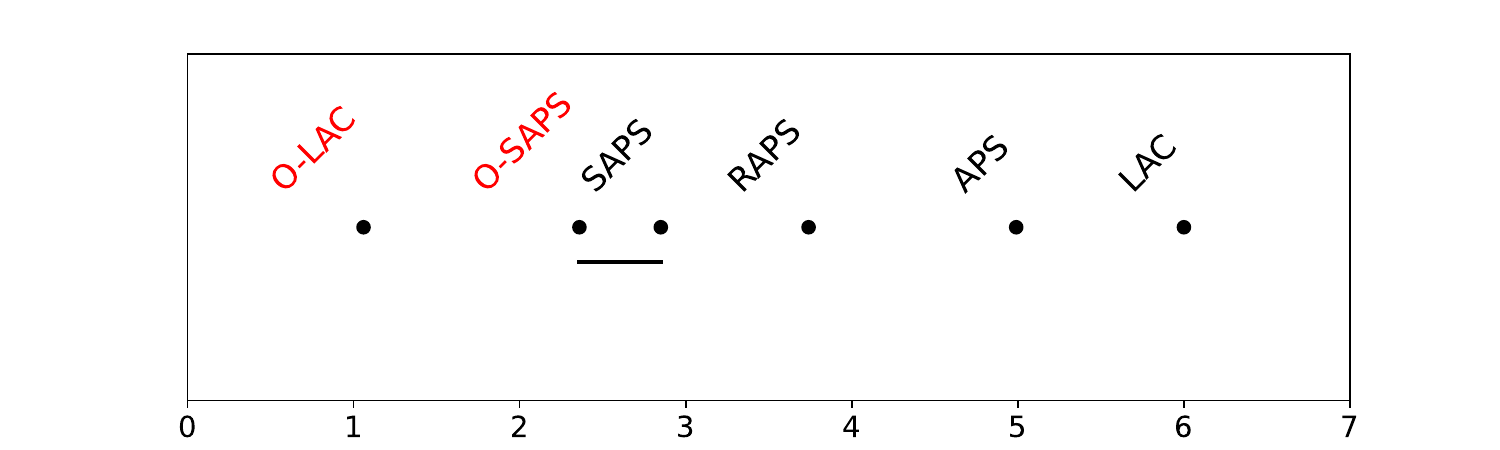}
        \caption{EfficientNet-V2-M}
    \end{subfigure}
    \begin{subfigure}[b]{0.45\textwidth}
        \includegraphics[width=\textwidth]{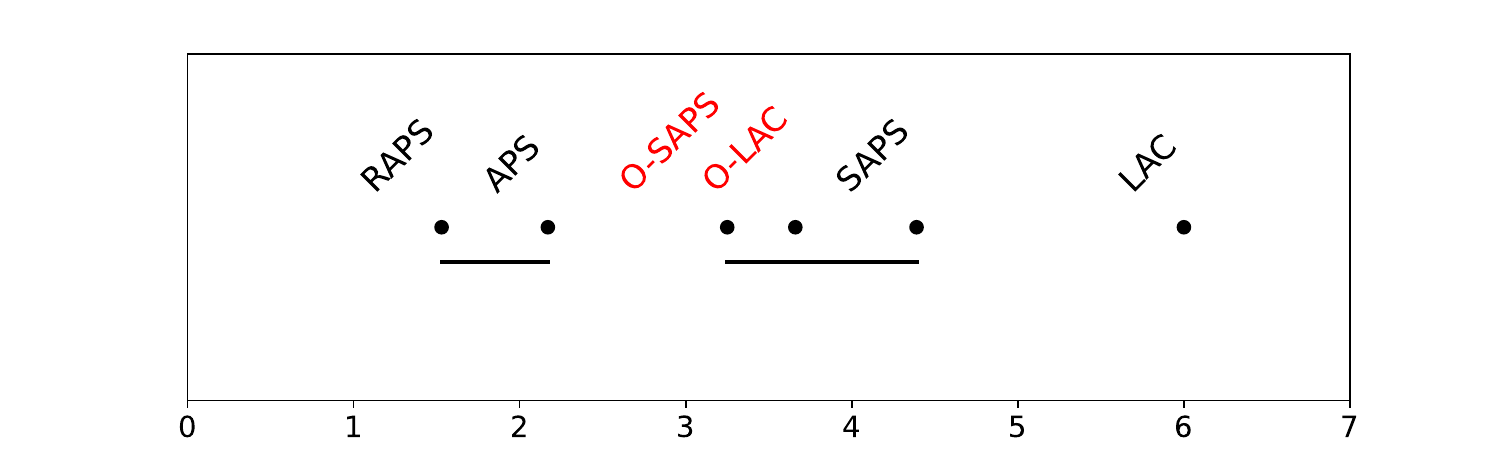}
        \caption{EfficientNet-V2-L}
    \end{subfigure}
    \begin{subfigure}[b]{0.45\textwidth}
        \includegraphics[width=\textwidth]{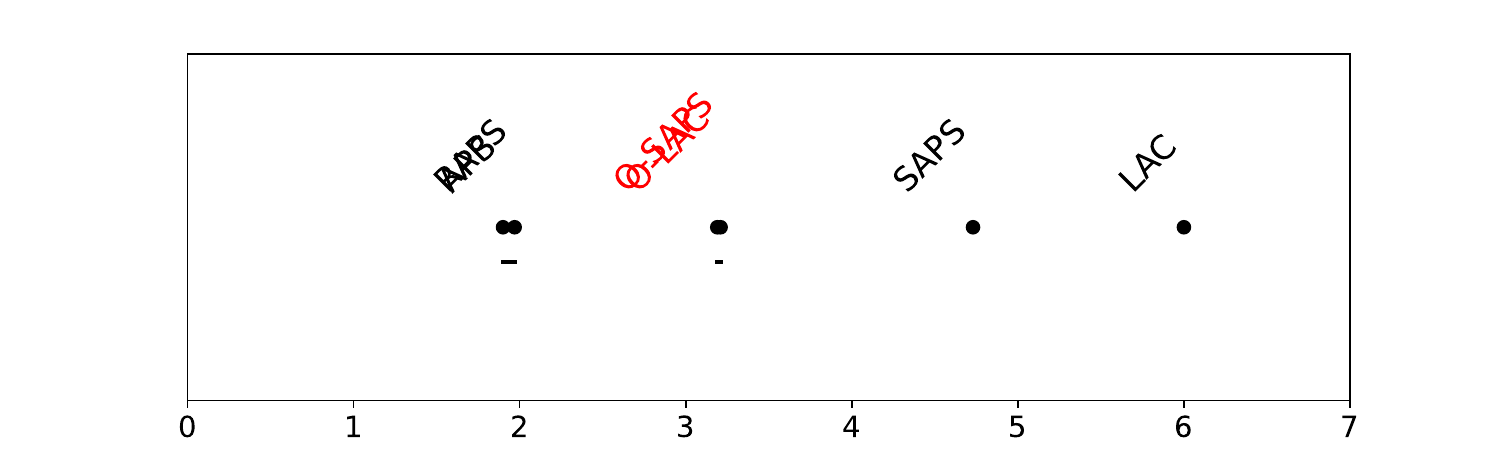}
        \caption{Swin-V2-B}
    \end{subfigure}
    \caption{Critical Difference Diagrams. T-SS, $\alpha=0.15, B=30$. The rank analysis based on these figures is summarized as `Avg. Rank from CD' in Table~\ref{tab:apdx_alg_results_our_metrics_B30} in Appendix.}
    \label{fig:apdx_cd_tss_B30_alpha0.15}
\end{figure}

\begin{figure}[!bt]
    \centering
    \begin{subfigure}[b]{0.45\textwidth}
        \includegraphics[width=\textwidth]{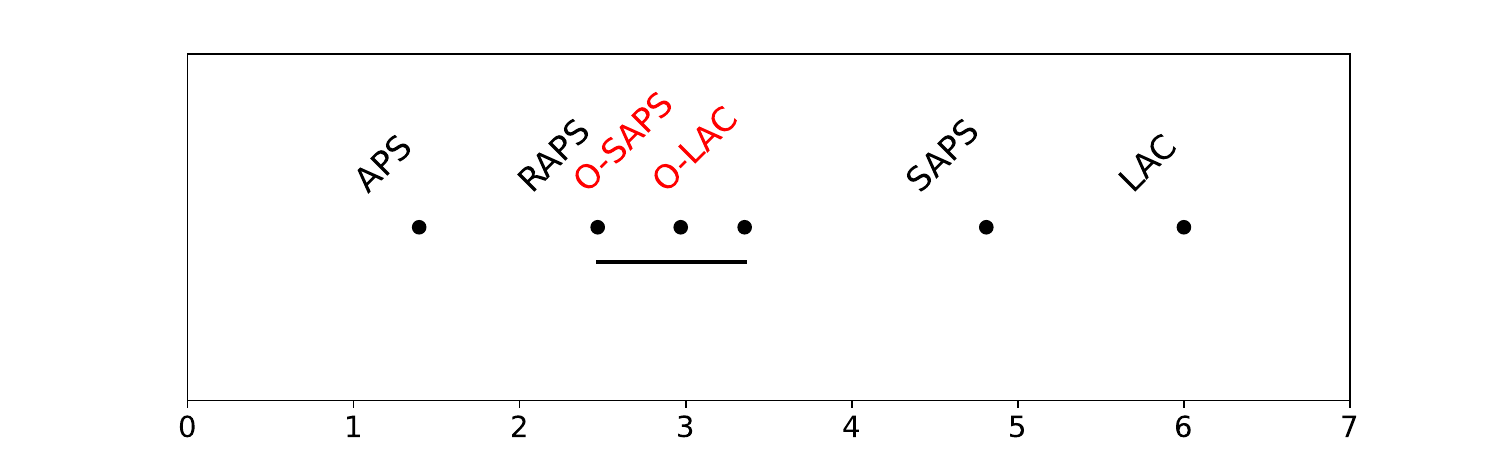}
        \caption{ResNet18}
    \end{subfigure}
    \begin{subfigure}[b]{0.45\textwidth}
        \includegraphics[width=\textwidth]{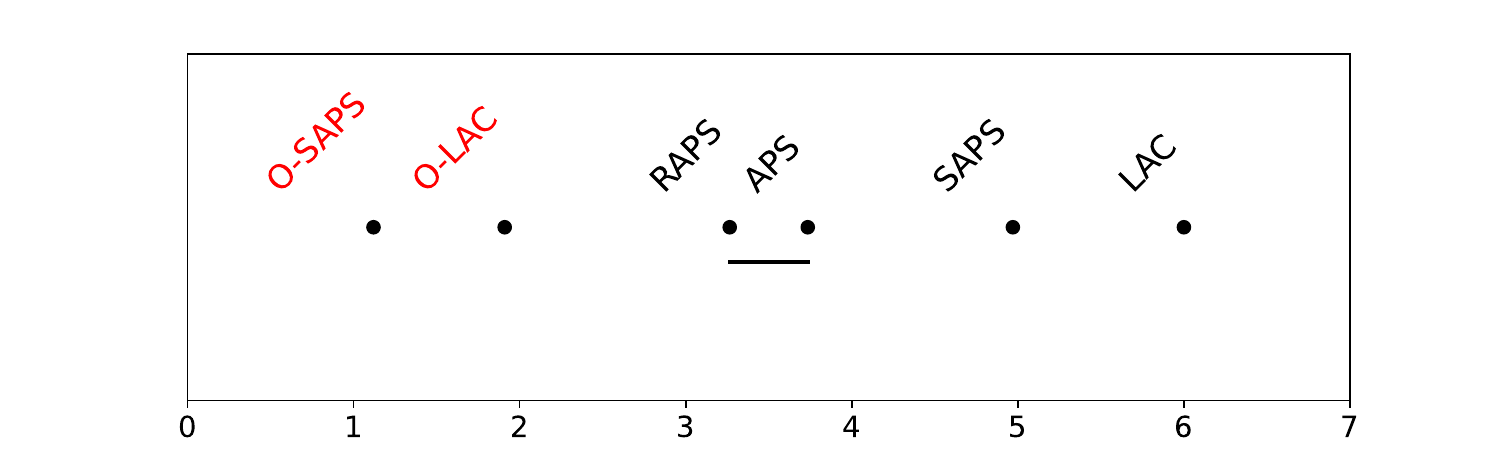}
        \caption{ResNet50}
    \end{subfigure}
    \begin{subfigure}[b]{0.45\textwidth}
        \includegraphics[width=\textwidth]{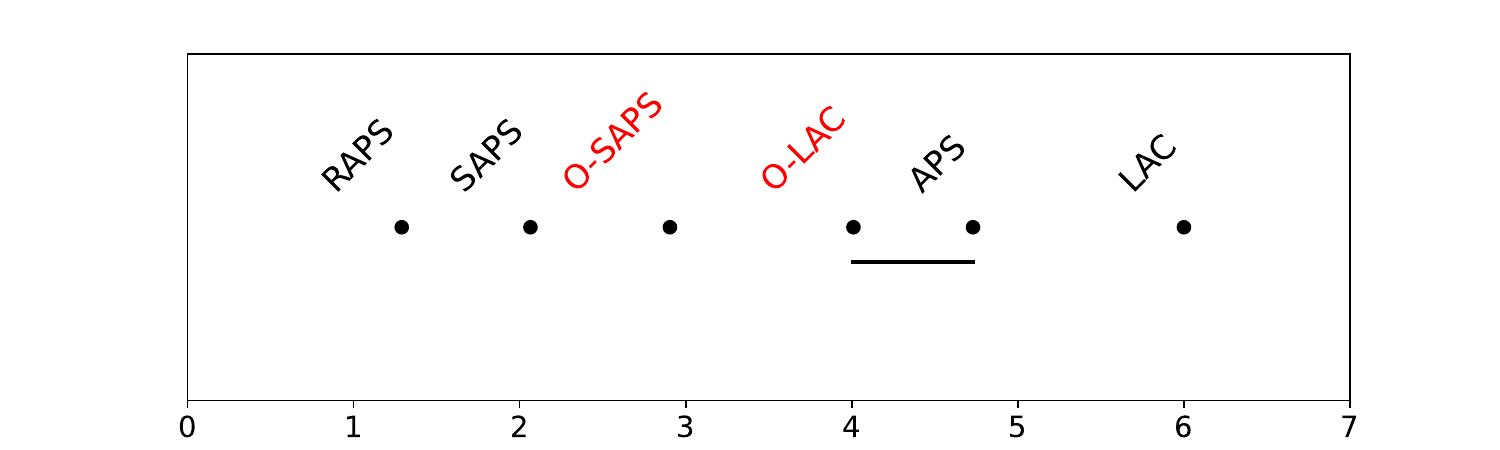}
        \caption{ResNet152}
    \end{subfigure}
    \begin{subfigure}[b]{0.45\textwidth}
        \includegraphics[width=\textwidth]{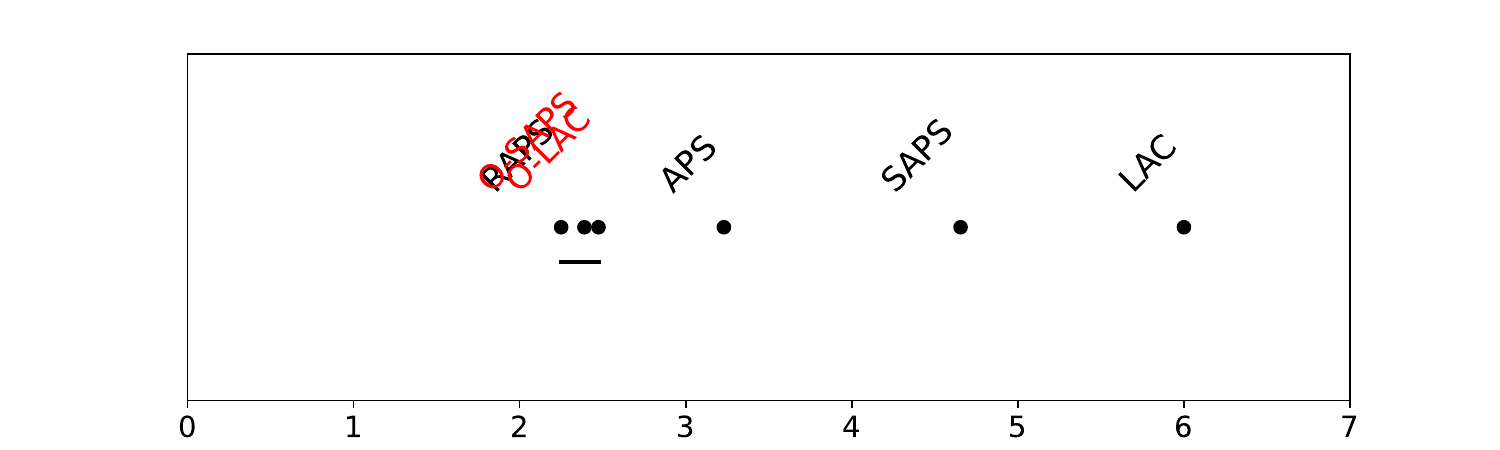}
        \caption{ViT-B-16}
    \end{subfigure}
    \begin{subfigure}[b]{0.45\textwidth}
        \includegraphics[width=\textwidth]{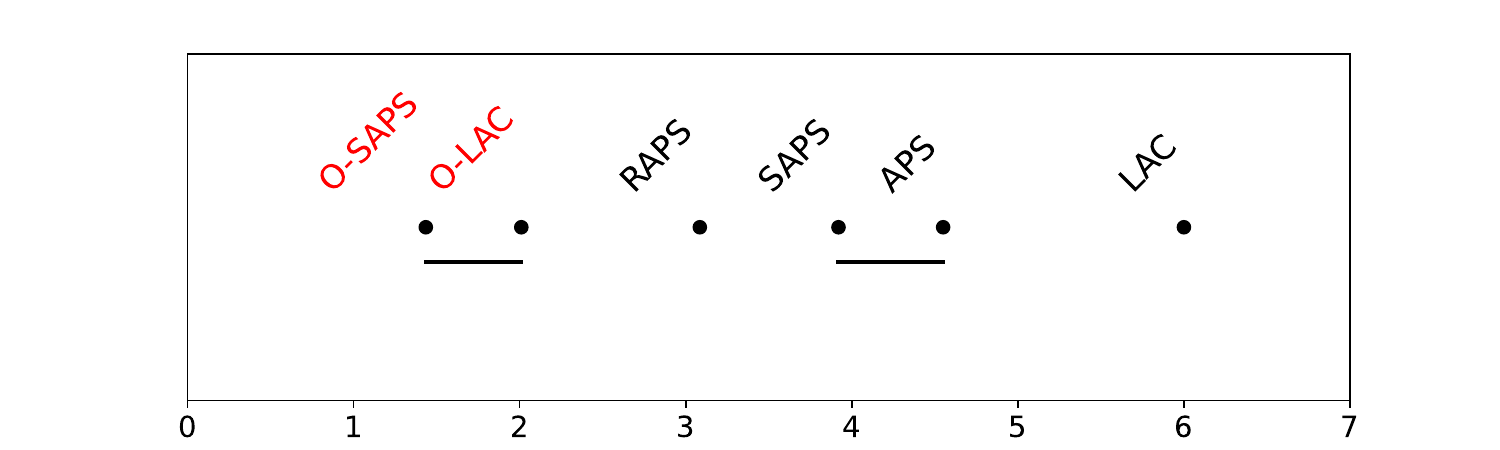}
        \caption{ViT-L-16}
    \end{subfigure}
    \begin{subfigure}[b]{0.45\textwidth}
        \includegraphics[width=\textwidth]{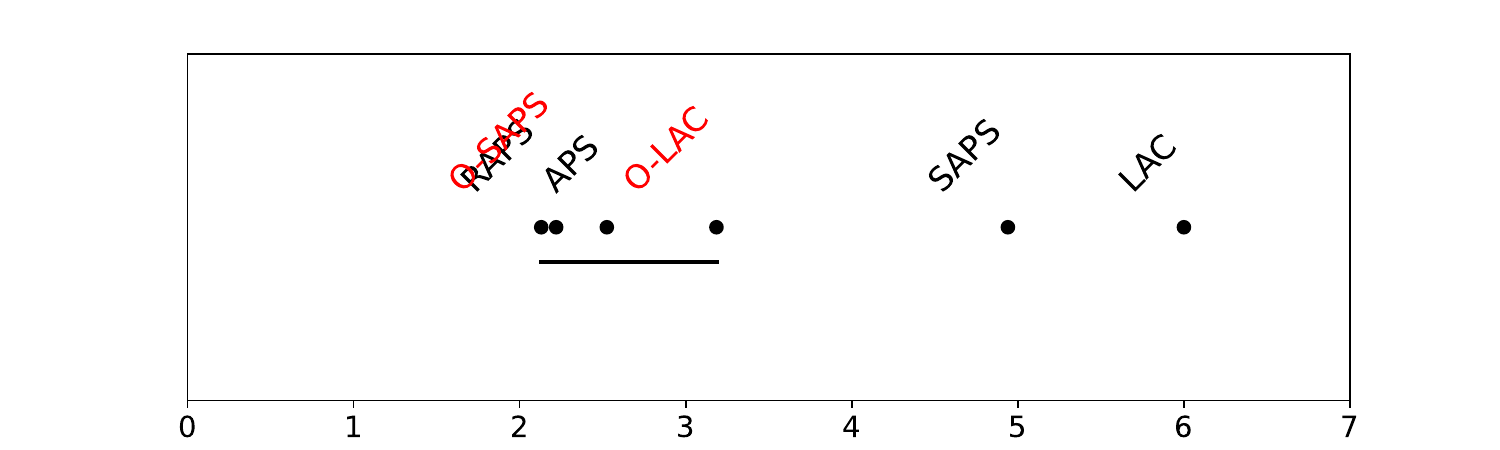}
        \caption{ViT-H-14}
    \end{subfigure}
    \begin{subfigure}[b]{0.45\textwidth}
        \includegraphics[width=\textwidth]{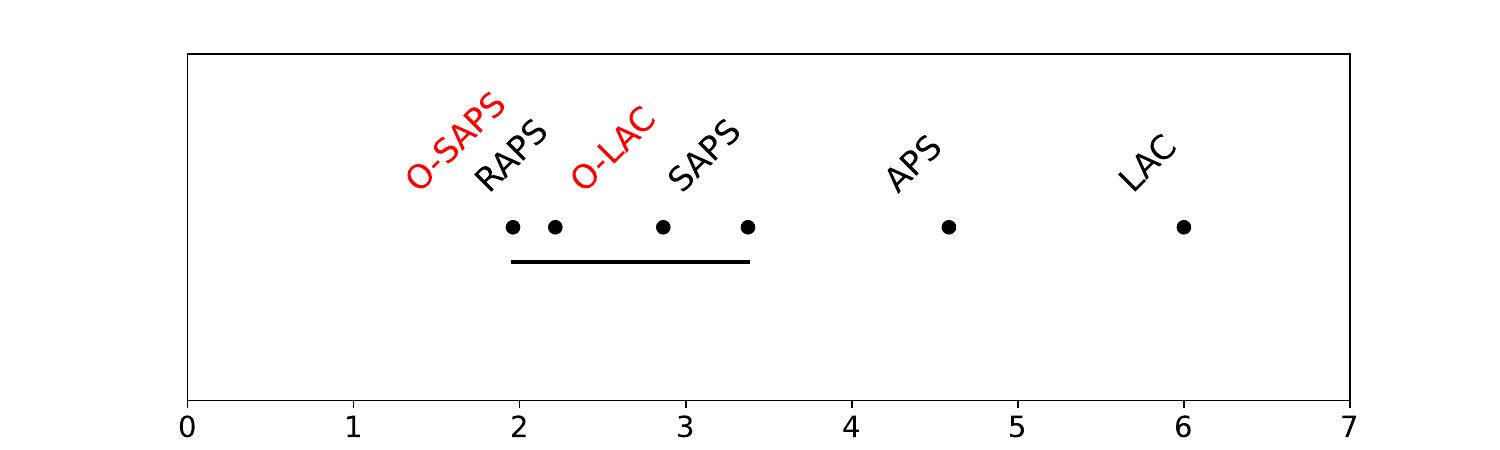}
        \caption{EfficientNet-V2-M}
    \end{subfigure}
    \begin{subfigure}[b]{0.45\textwidth}
        \includegraphics[width=\textwidth]{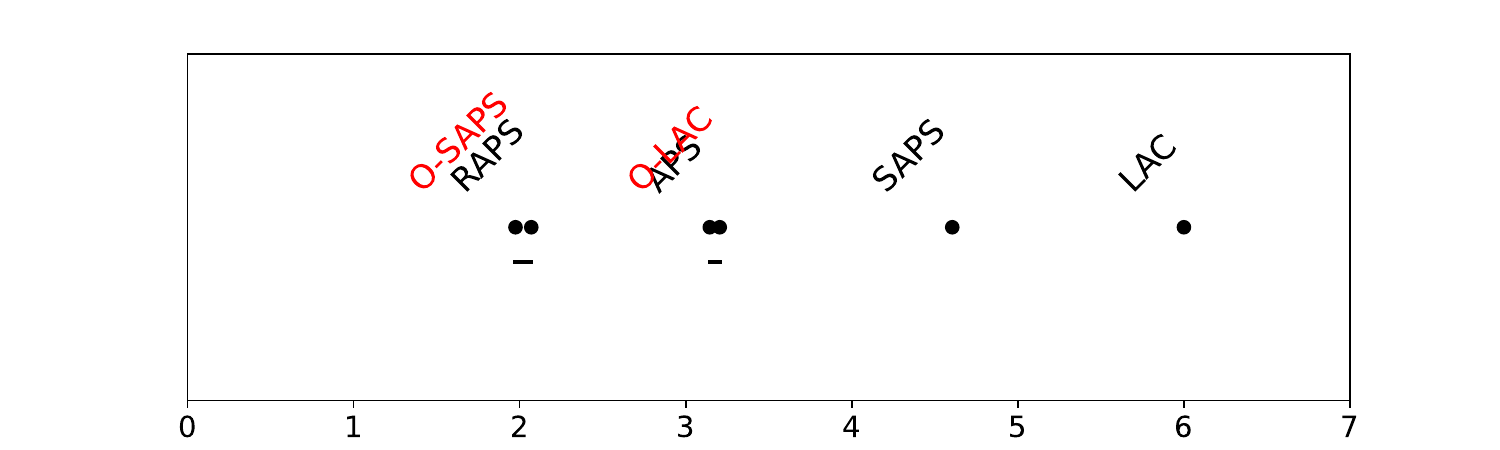}
        \caption{EfficientNet-V2-L}
    \end{subfigure}
    \begin{subfigure}[b]{0.45\textwidth}
        \includegraphics[width=\textwidth]{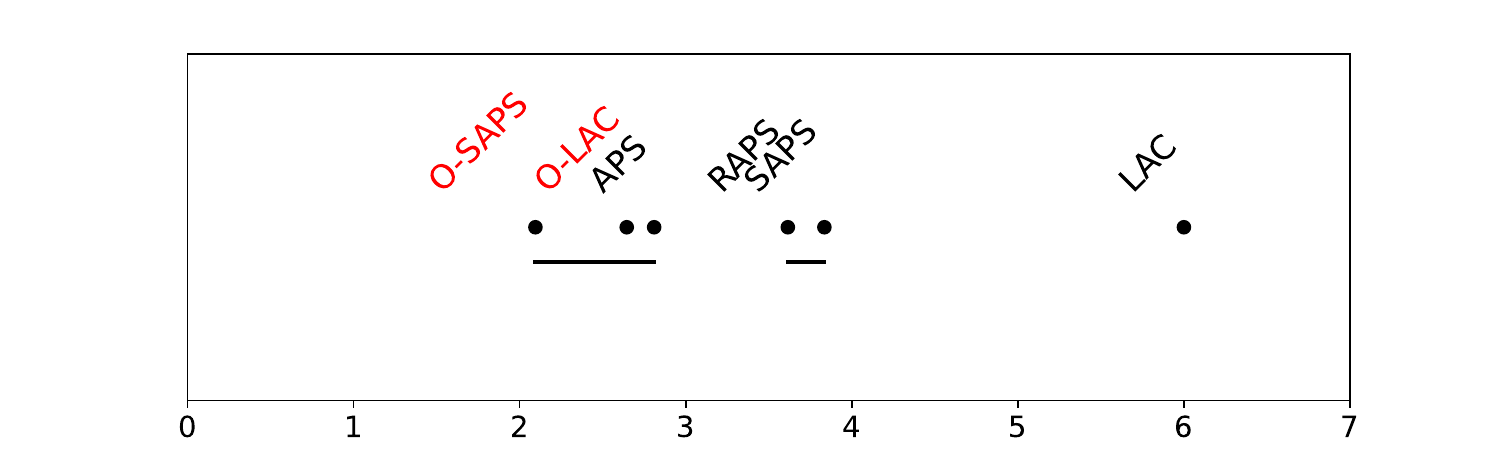}
        \caption{Swin-V2-B}
    \end{subfigure}
    \caption{Critical Difference Diagrams. T-CV, $\alpha=0.20, B=30$. The rank analysis based on these figures is summarized as `Avg. Rank from CD' in Table~\ref{tab:apdx_alg_results_our_metrics_B30} in Appendix.}
    \label{fig:apdx_cd_tcv_B30_alpha0.20}
\end{figure}

\begin{figure}[!bt]
    \centering
    \begin{subfigure}[b]{0.45\textwidth}
        \includegraphics[width=\textwidth]{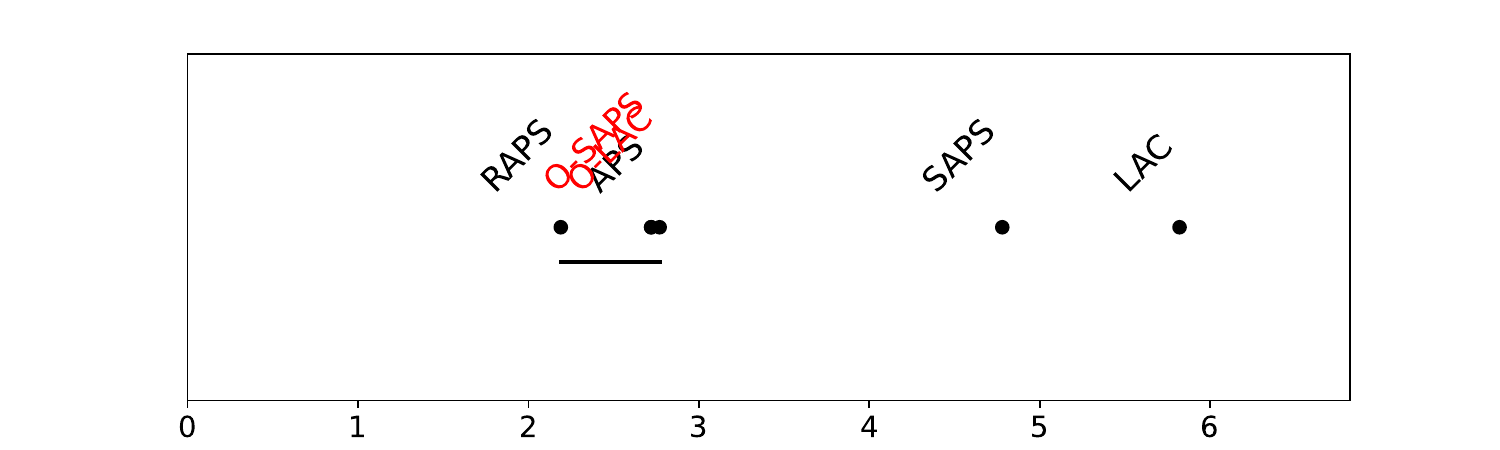}
        \caption{ResNet18}
    \end{subfigure}
    \begin{subfigure}[b]{0.45\textwidth}
        \includegraphics[width=\textwidth]{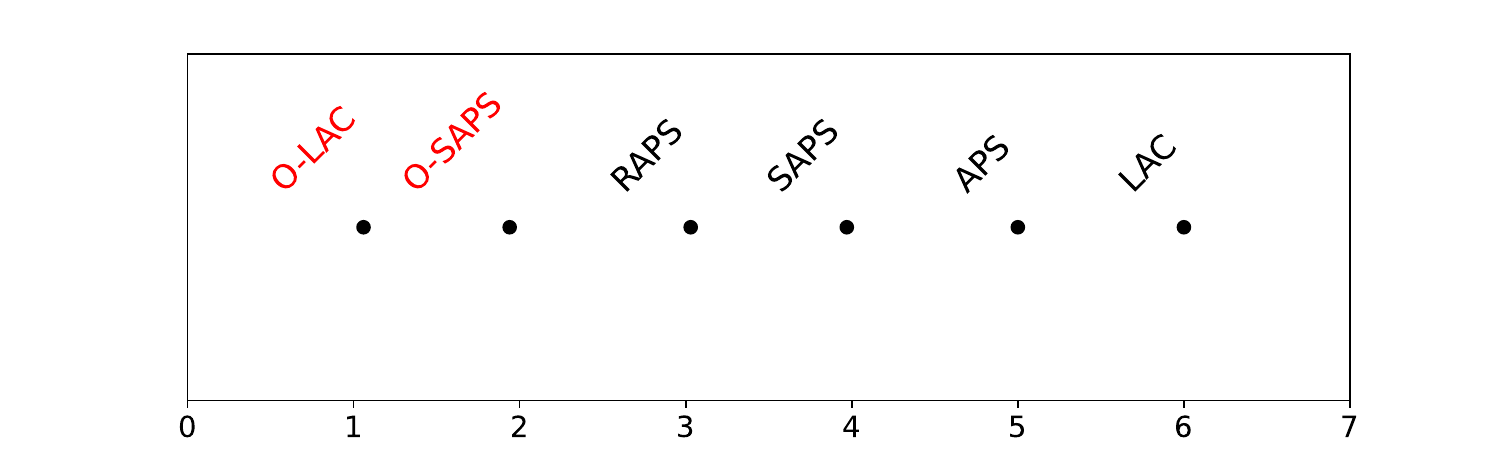}
        \caption{ResNet50}
    \end{subfigure}
    \begin{subfigure}[b]{0.45\textwidth}
        \includegraphics[width=\textwidth]{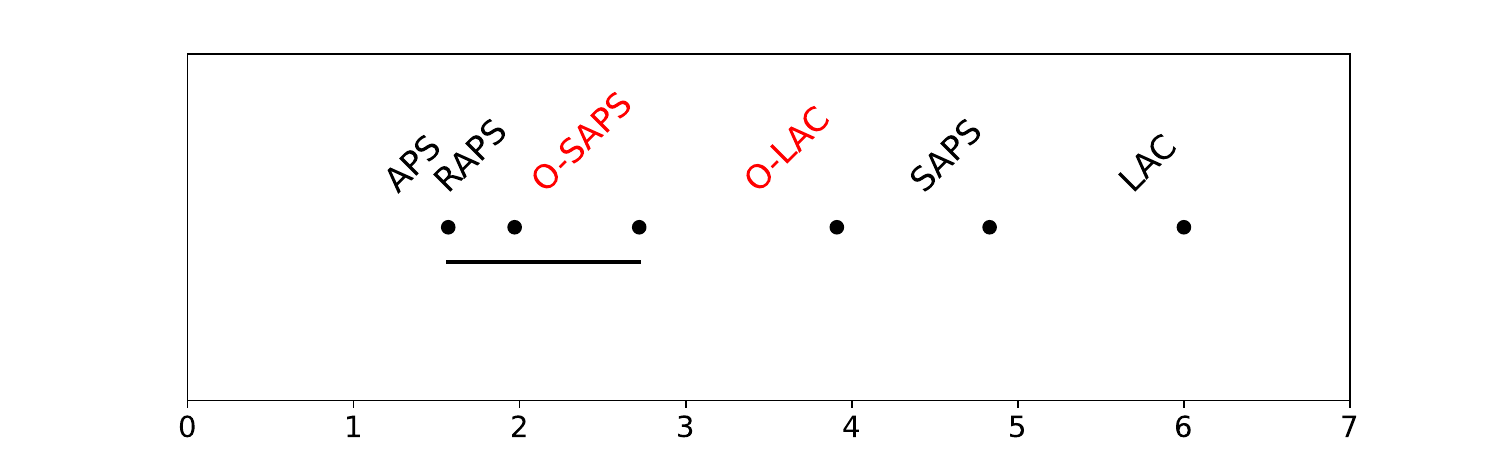}
        \caption{ResNet152}
    \end{subfigure}
    \begin{subfigure}[b]{0.45\textwidth}
        \includegraphics[width=\textwidth]{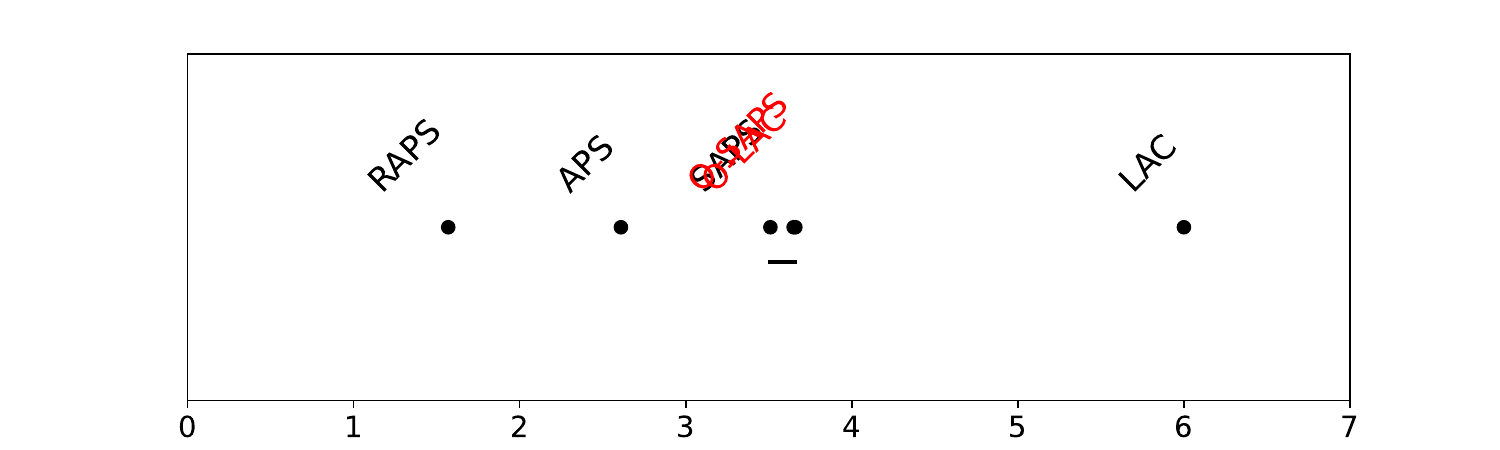}
        \caption{ViT-B-16}
    \end{subfigure}
    \begin{subfigure}[b]{0.45\textwidth}
        \includegraphics[width=\textwidth]{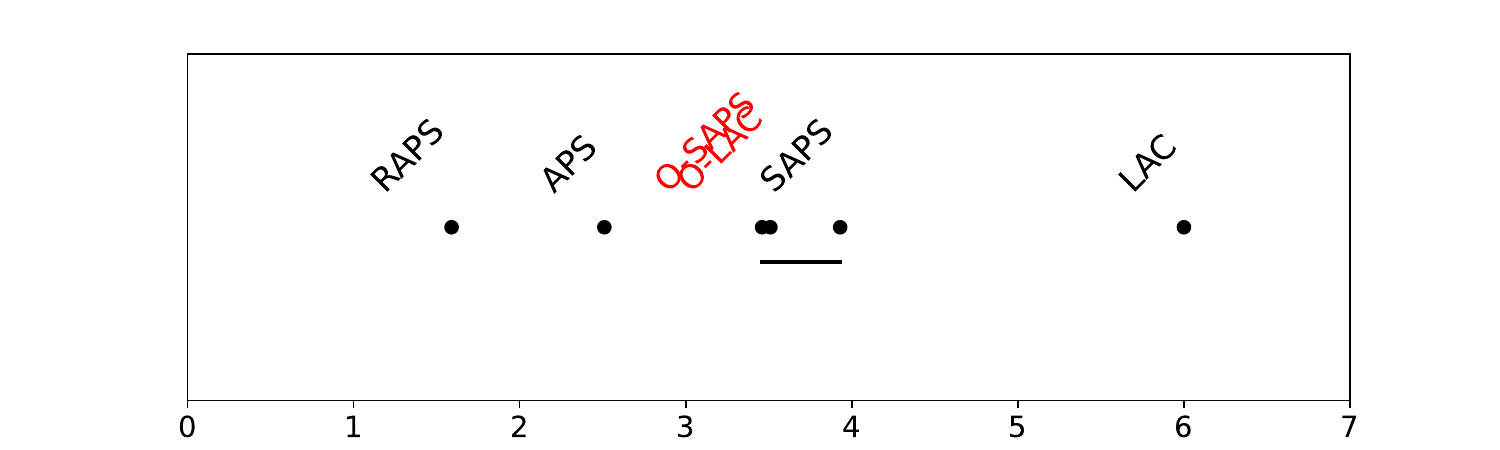}
        \caption{ViT-L-16}
    \end{subfigure}
    \begin{subfigure}[b]{0.45\textwidth}
        \includegraphics[width=\textwidth]{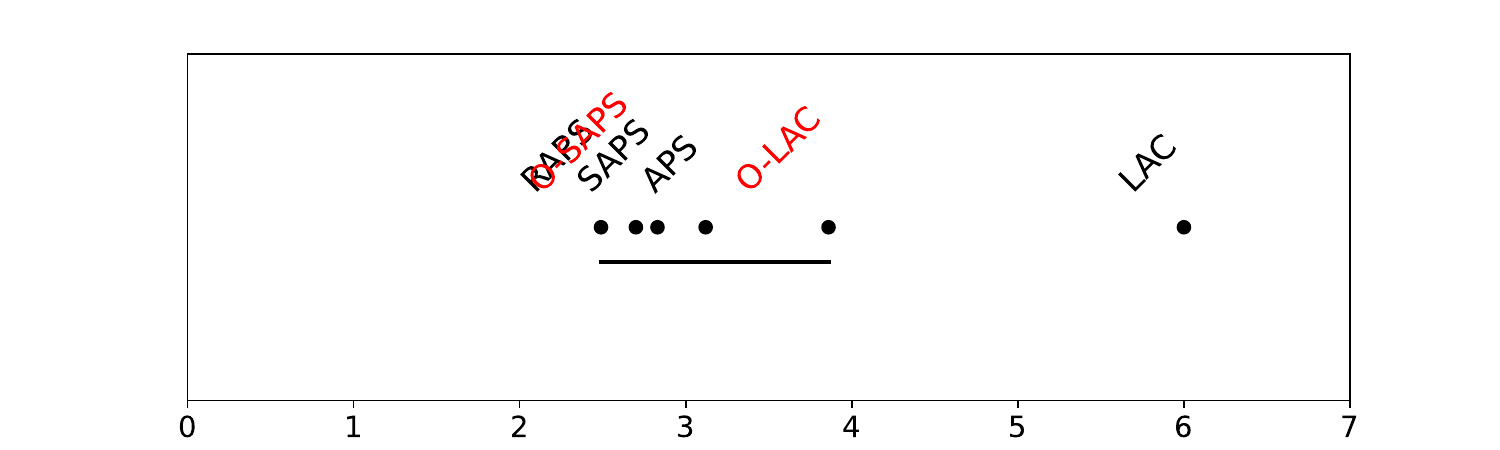}
        \caption{ViT-H-14}
    \end{subfigure}
    \begin{subfigure}[b]{0.45\textwidth}
        \includegraphics[width=\textwidth]{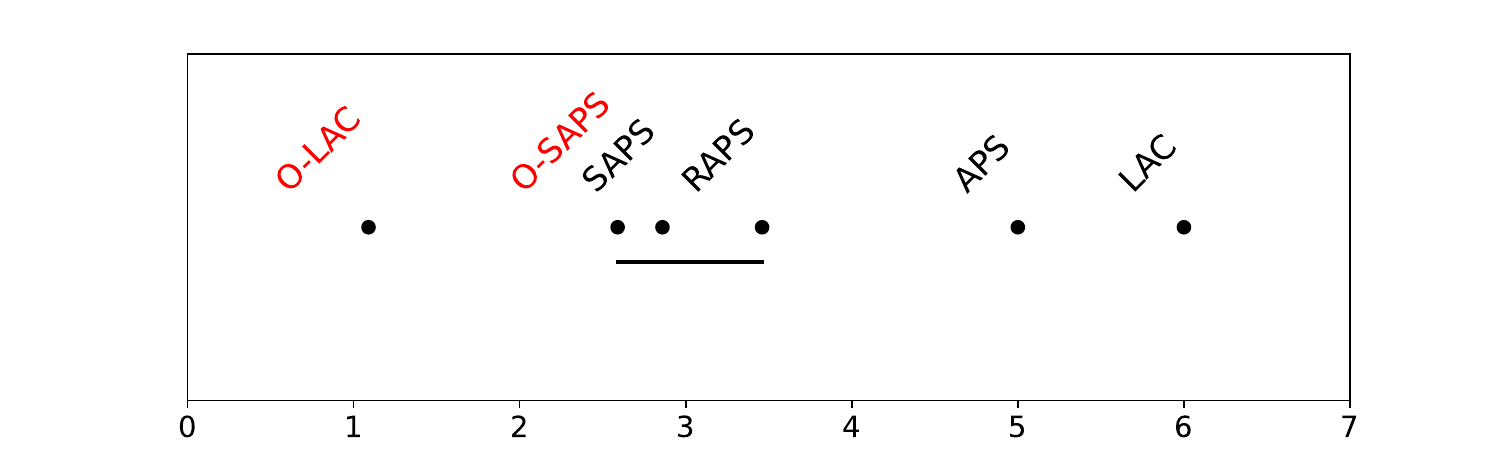}
        \caption{EfficientNet-V2-M}
    \end{subfigure}
    \begin{subfigure}[b]{0.45\textwidth}
        \includegraphics[width=\textwidth]{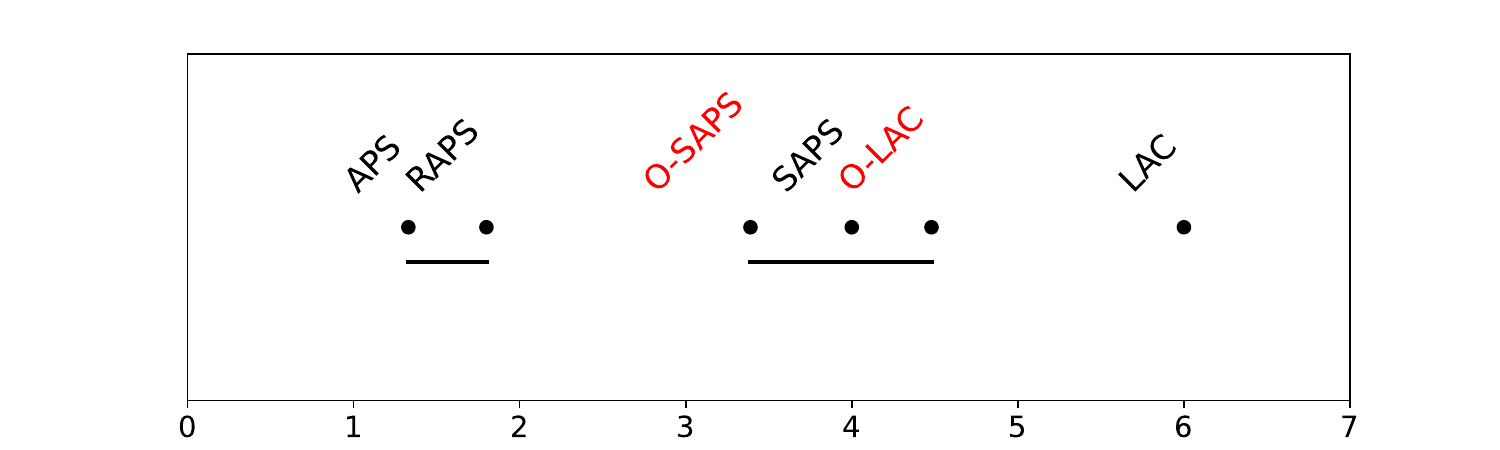}
        \caption{EfficientNet-V2-L}
    \end{subfigure}
    \begin{subfigure}[b]{0.45\textwidth}
        \includegraphics[width=\textwidth]{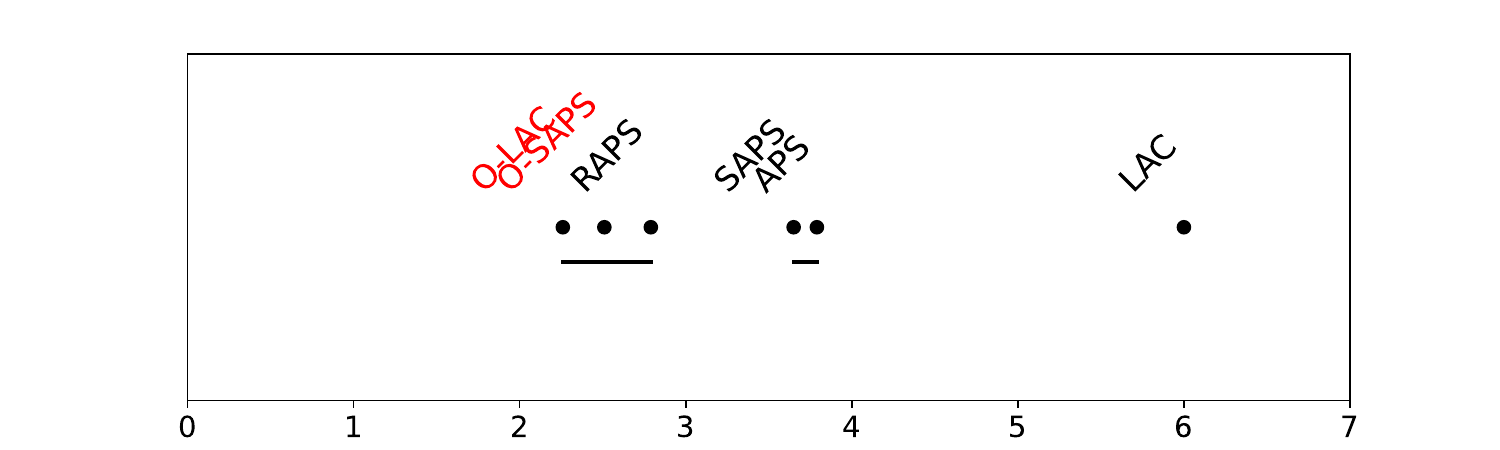}
        \caption{Swin-V2-B}
    \end{subfigure}
    \caption{Critical Difference Diagrams. T-SS, $\alpha=0.20, B=30$. The rank analysis based on these figures is summarized as `Avg. Rank from CD' in Table~\ref{tab:apdx_alg_results_our_metrics_B30} in Appendix.}
    \label{fig:apdx_cd_tss_B30_alpha0.20}
\end{figure}
\FloatBarrier
\subsection{Visual Acuity Prediction Experiments}
\begin{figure*}
    \centering
    \begin{subfigure}{0.49\linewidth}
        \includegraphics[width=0.9\linewidth]{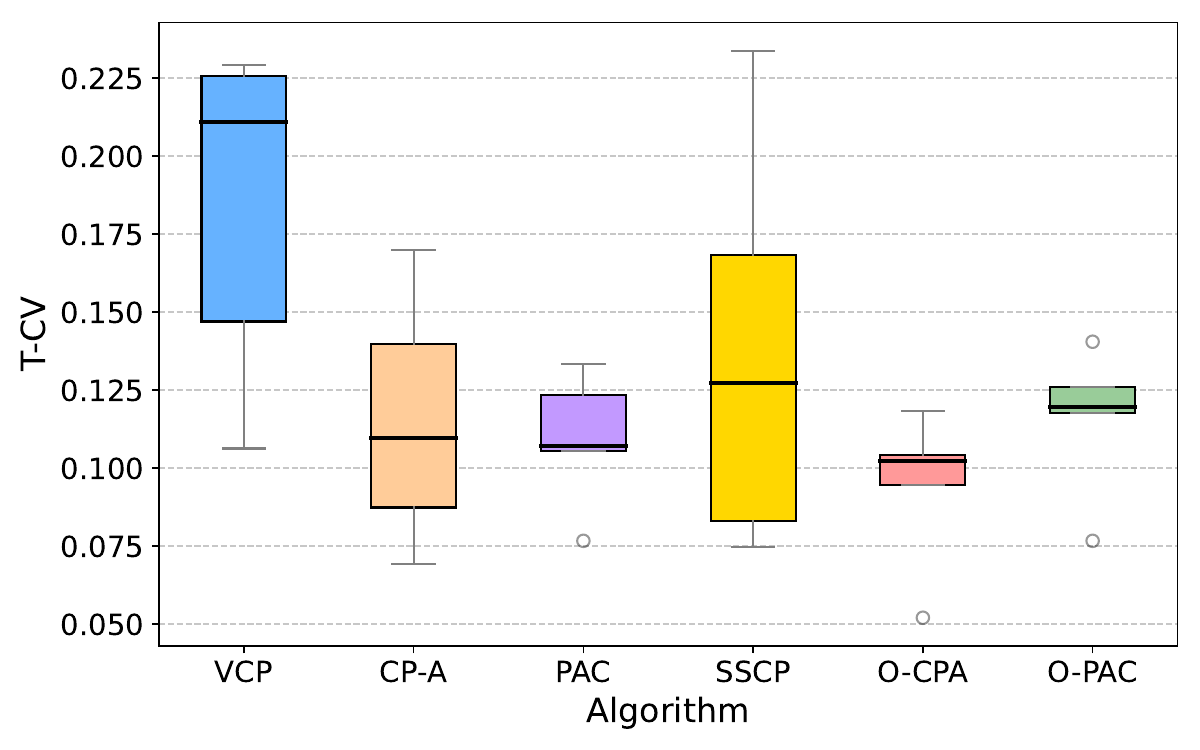}
        \caption{T-CV. Simple-CNN.}
    \end{subfigure}%
    \hfill
    \begin{subfigure}{0.49\linewidth}
        \includegraphics[width=0.9\linewidth]{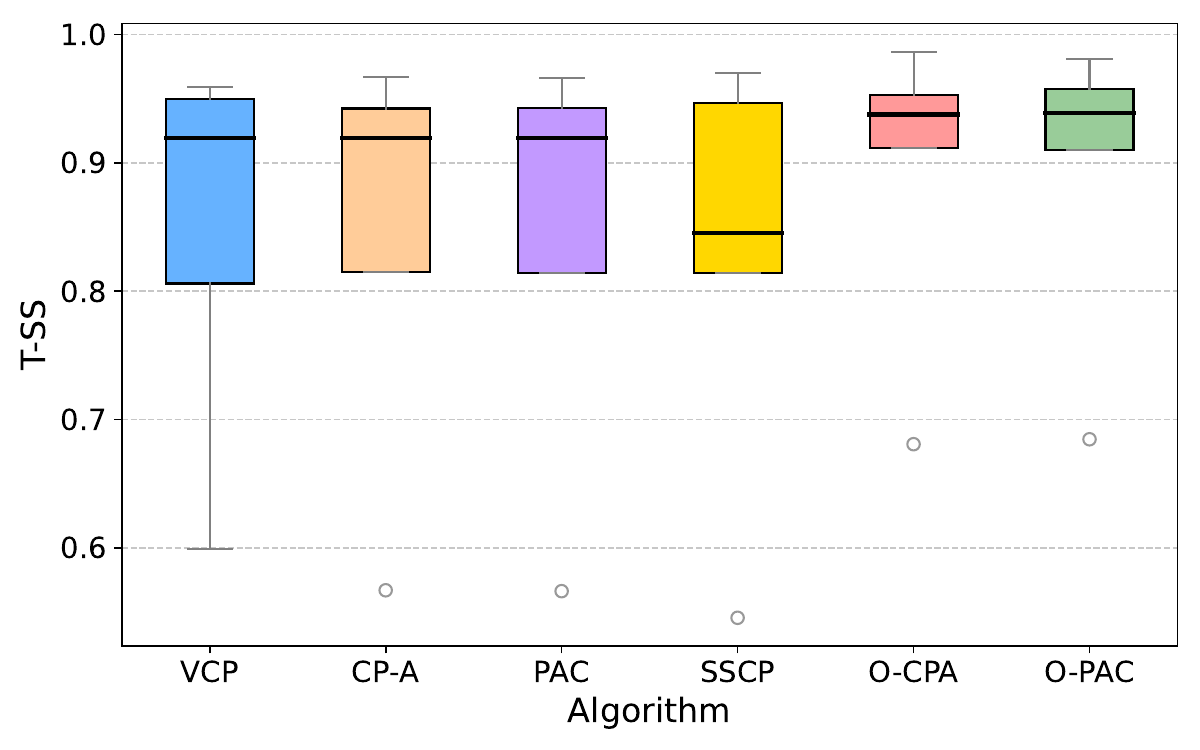}
        \caption{T-SS. Simple-CNN.}
    \end{subfigure}
    \begin{subfigure}{0.49\linewidth}
        \includegraphics[width=0.9\linewidth]{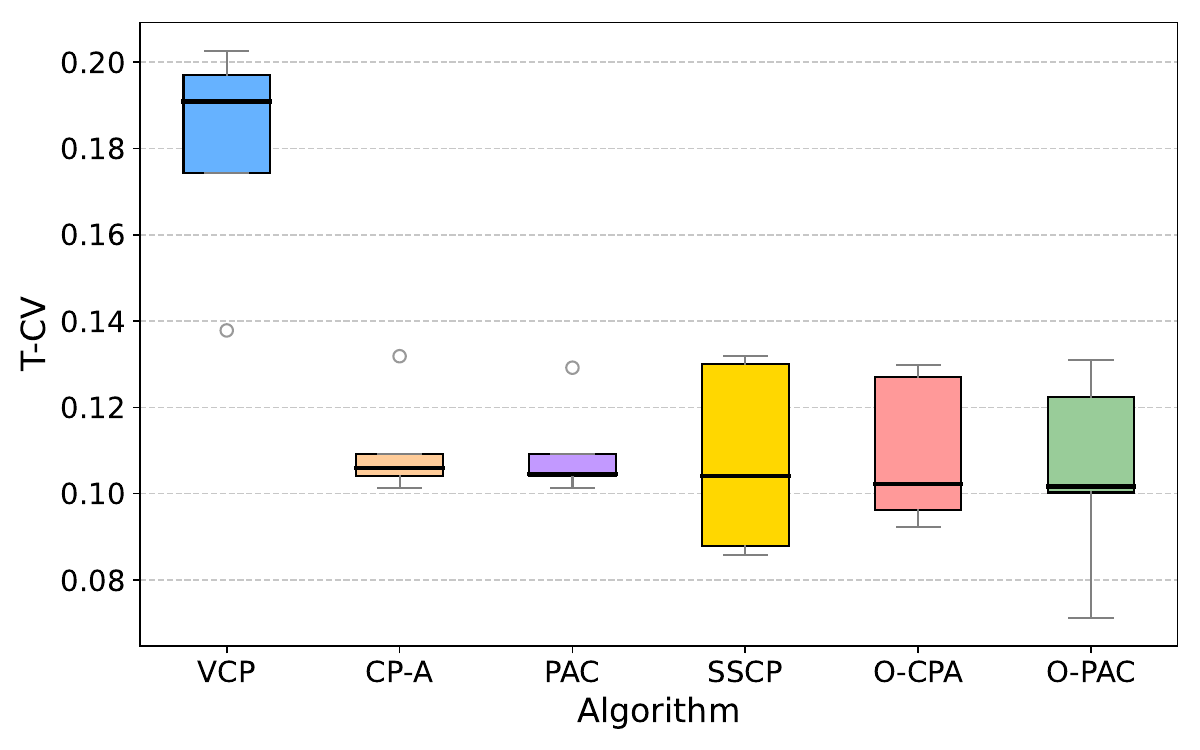}
        \caption{T-CV. ResNet18.}
    \end{subfigure}%
    \hfill
    \begin{subfigure}{0.49\linewidth}
        \includegraphics[width=0.9\linewidth]{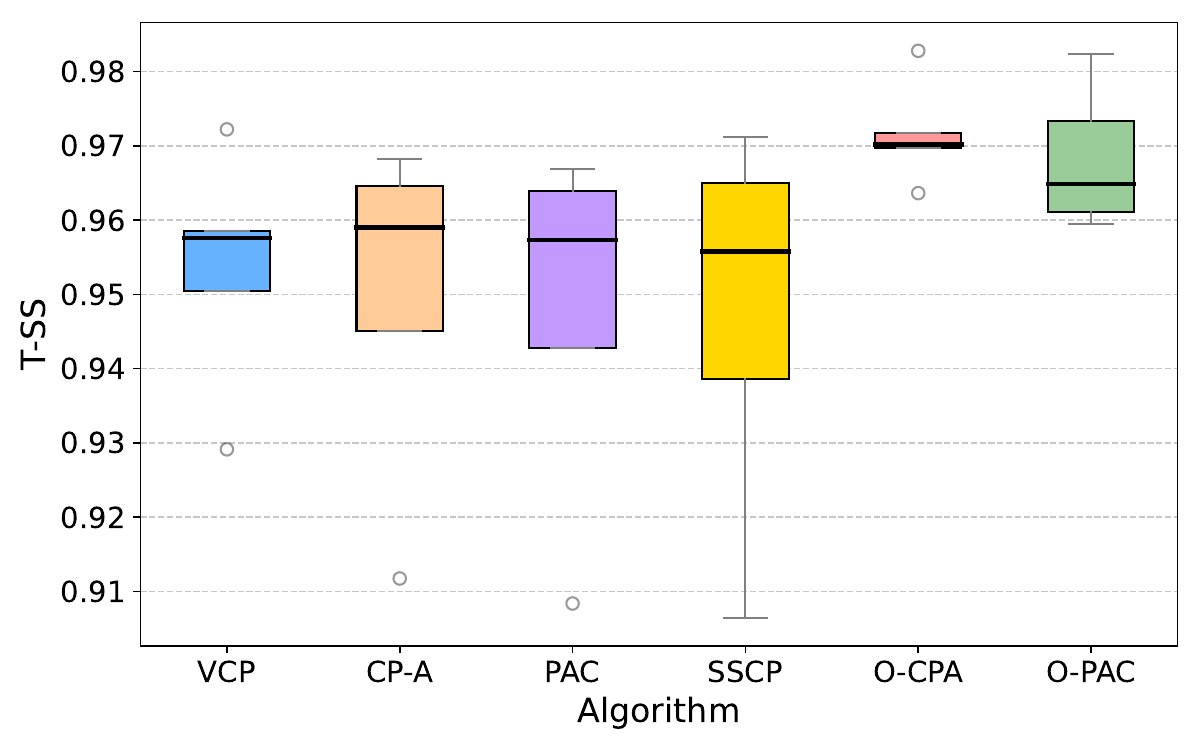}
        \caption{T-SS. ResNet18.}
    \end{subfigure}
    \begin{subfigure}{0.49\linewidth}
        \includegraphics[width=0.9\linewidth]{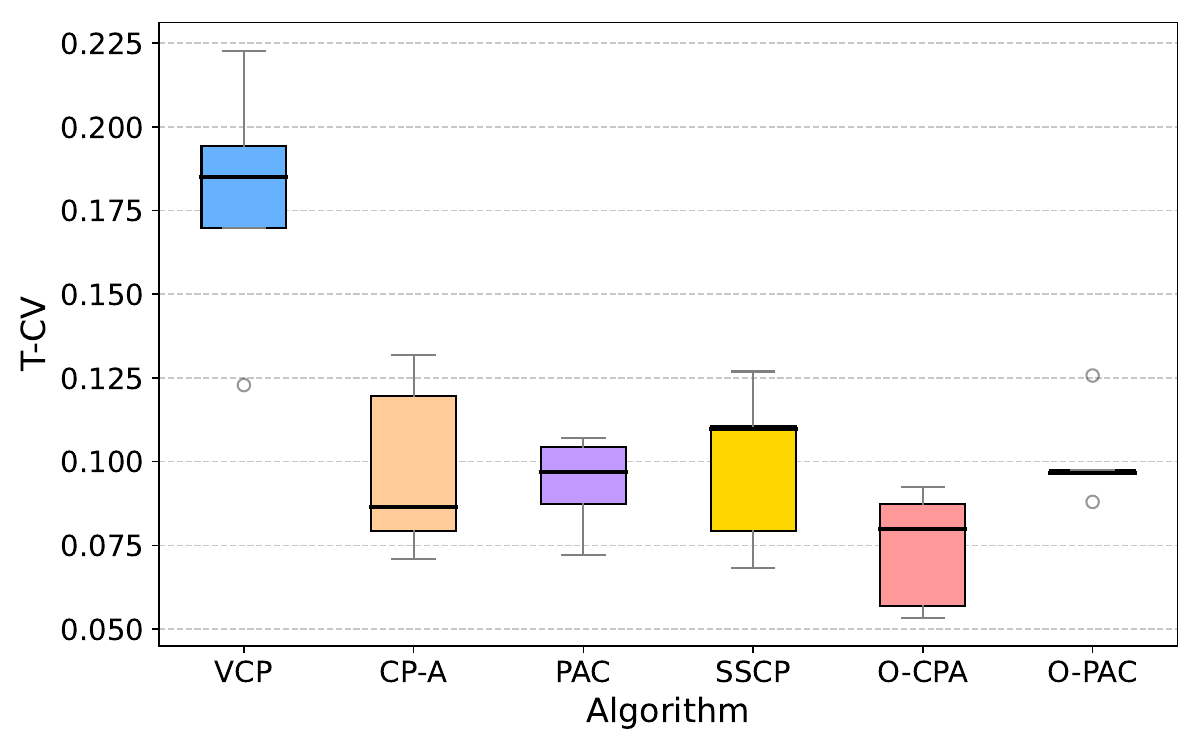}
        \caption{T-CV. ResNet50.}
    \end{subfigure}%
    \hfill
    \begin{subfigure}{0.49\linewidth}
        \includegraphics[width=0.9\linewidth]{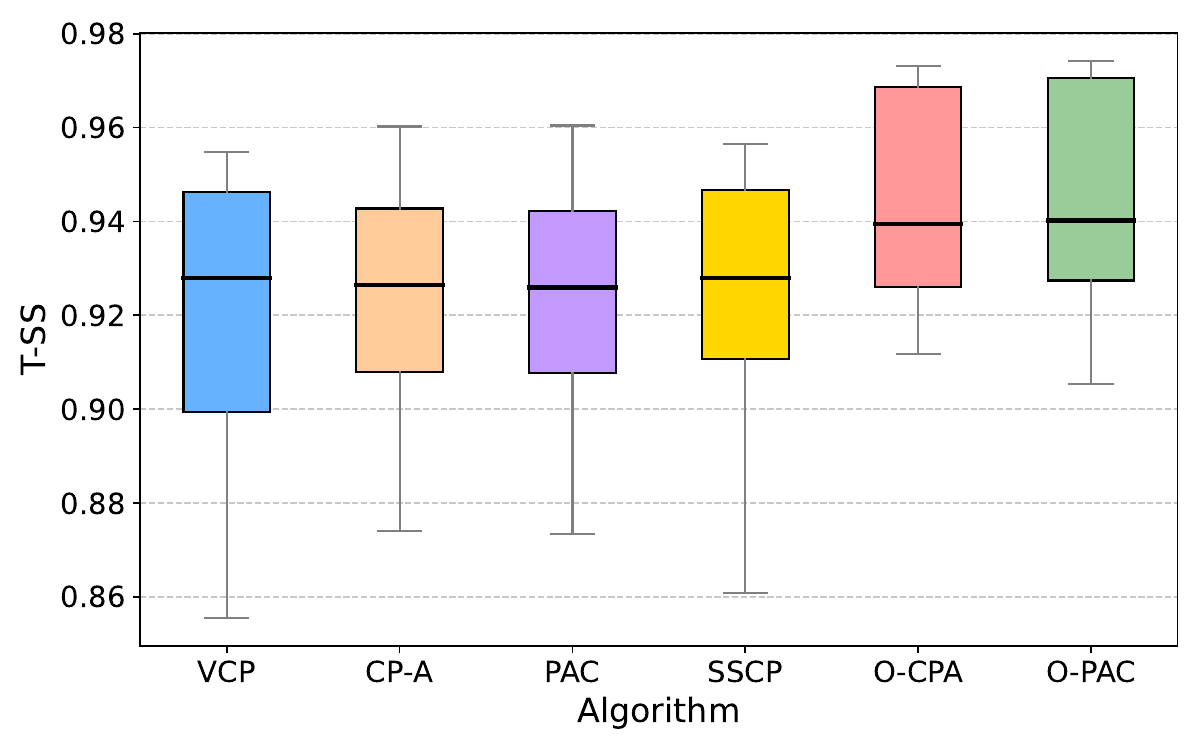}
        \caption{T-SS. ResNet50.}
    \end{subfigure}
    \begin{subfigure}{0.49\linewidth}
        \includegraphics[width=0.9\linewidth]{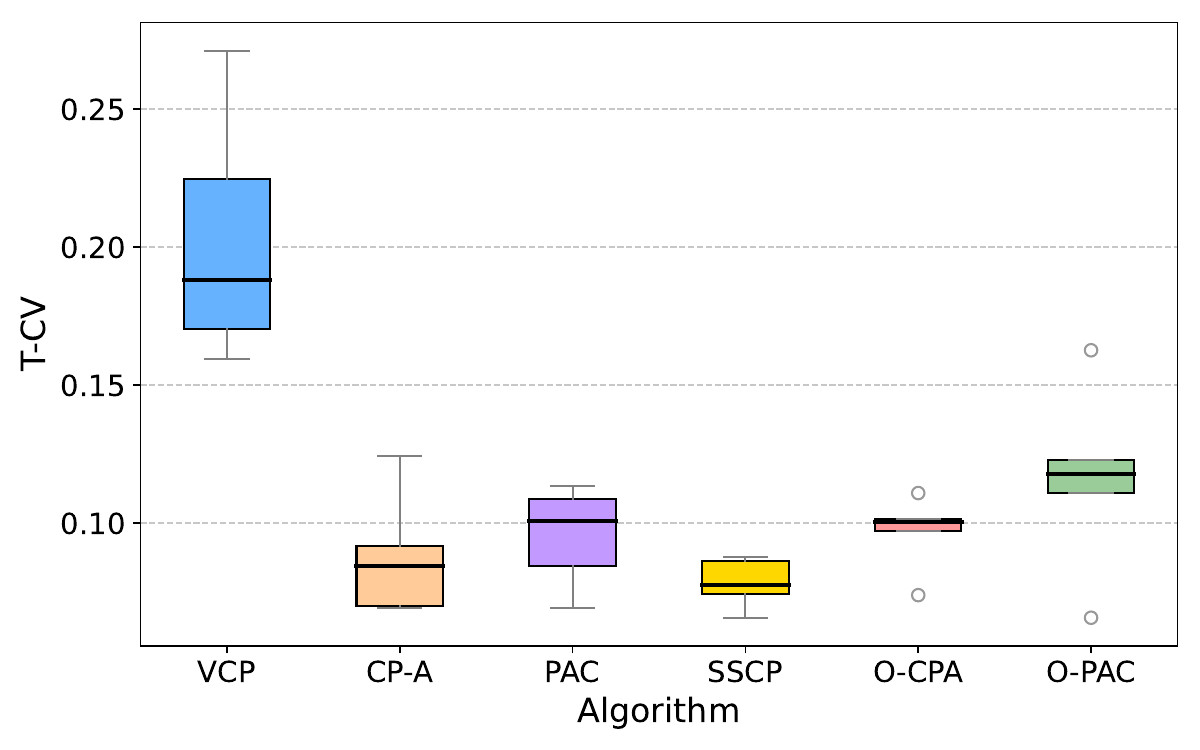}
        \caption{T-CV. EfficientNet-V2-S.}
    \end{subfigure}%
    \hfill
    \begin{subfigure}{0.49\linewidth}
        \includegraphics[width=0.9\linewidth]{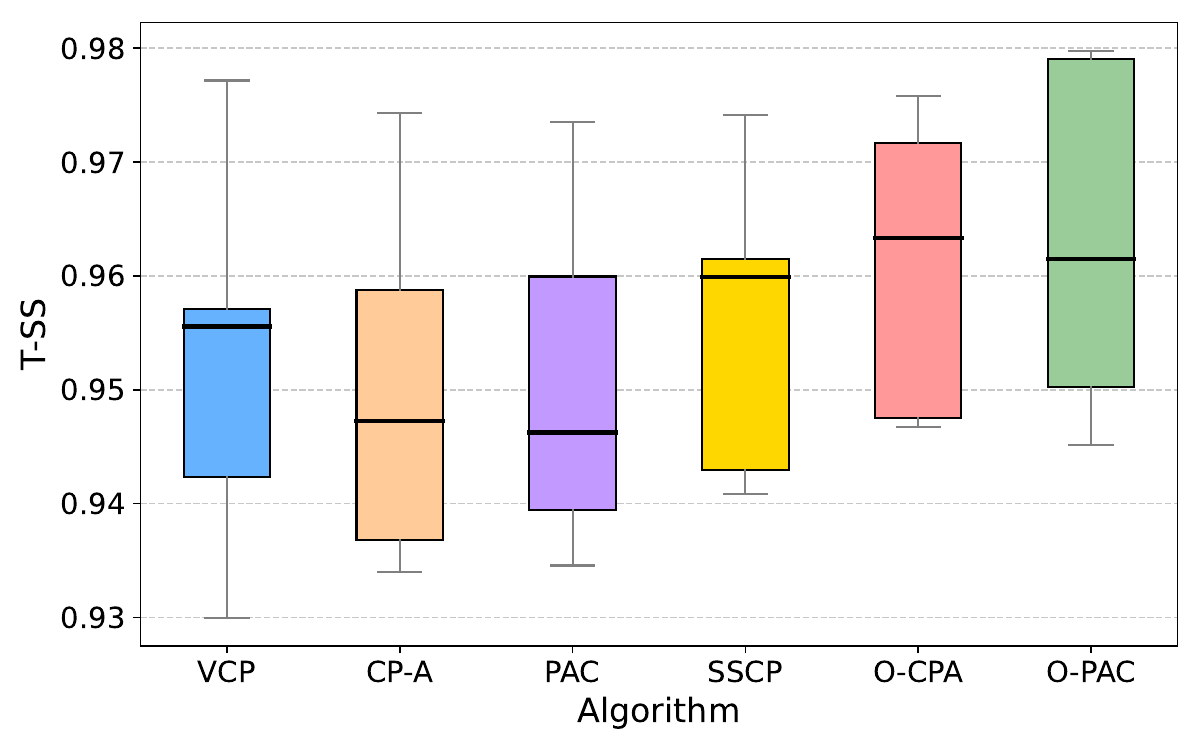}
        \caption{T-SS. EfficientNet-V2-S.}
    \end{subfigure}
    
    \caption{Visual Acuity Prediction Result. $\alpha=0.20$}
    \label{fig:va_adaptivity_alpha0.20}
\end{figure*}

\begin{figure*}
    \centering
    \begin{subfigure}{0.49\linewidth}
        \centering 
        \includegraphics[width=0.9\linewidth]{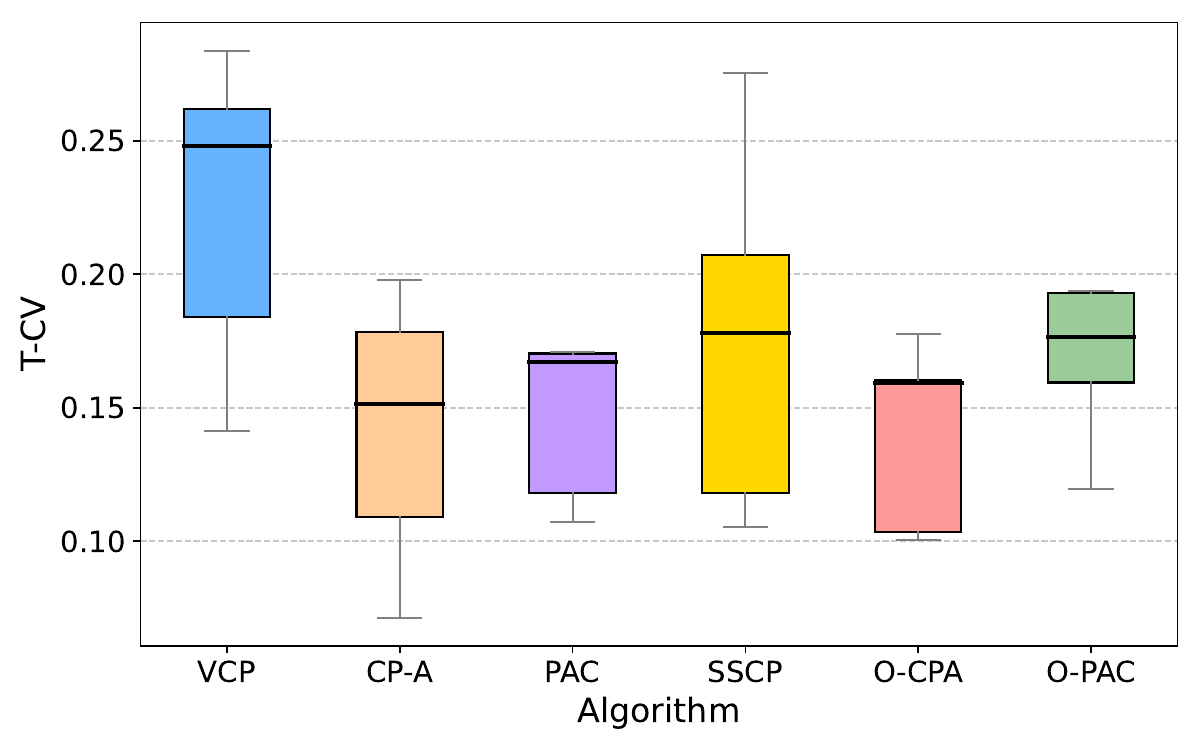}
        \caption{T-CV. Simple-CNN.}
    \end{subfigure}%
    \hfill
    \begin{subfigure}[b]{0.49\linewidth}
        \centering 
        \includegraphics[width=0.9\linewidth]{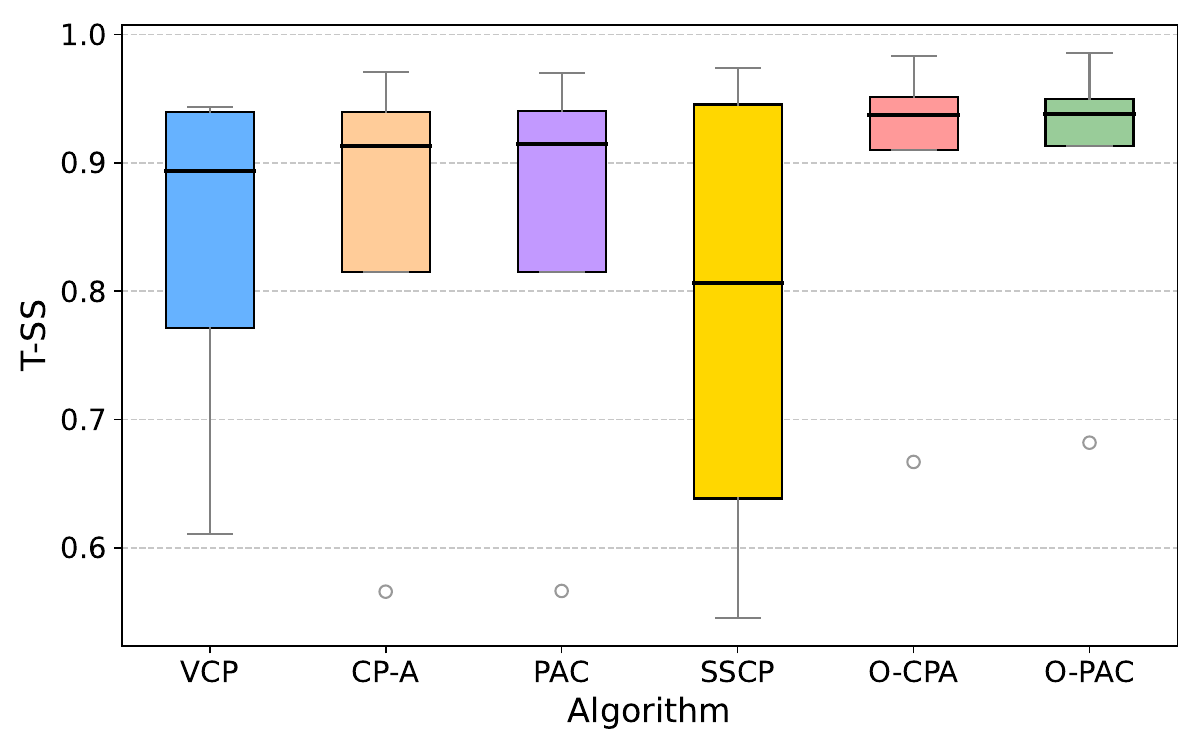}
        \caption{T-SS. Simple-CNN.}
    \end{subfigure}
    \begin{subfigure}[b]{0.49\linewidth}
        \centering 
        \includegraphics[width=0.9\linewidth]{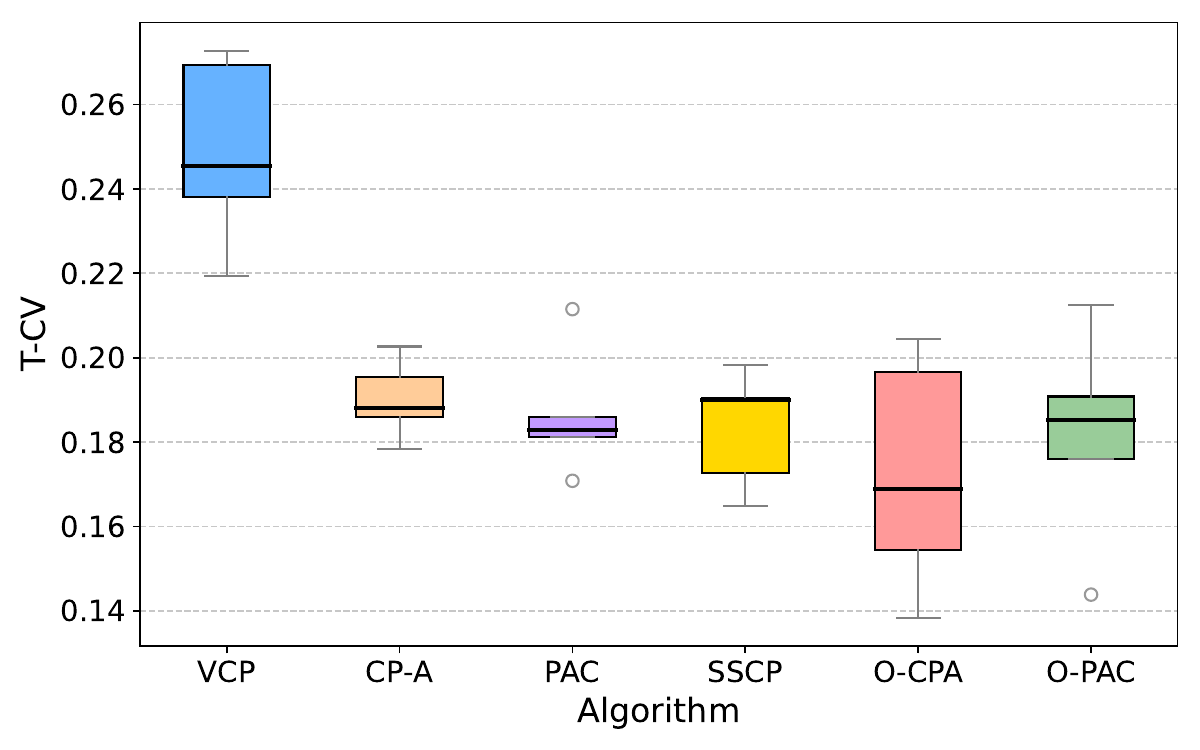}
        \caption{T-CV. ResNet18.}
    \end{subfigure}%
    \hfill
    \begin{subfigure}{0.49\linewidth}
        \centering 
        \includegraphics[width=0.9\linewidth]{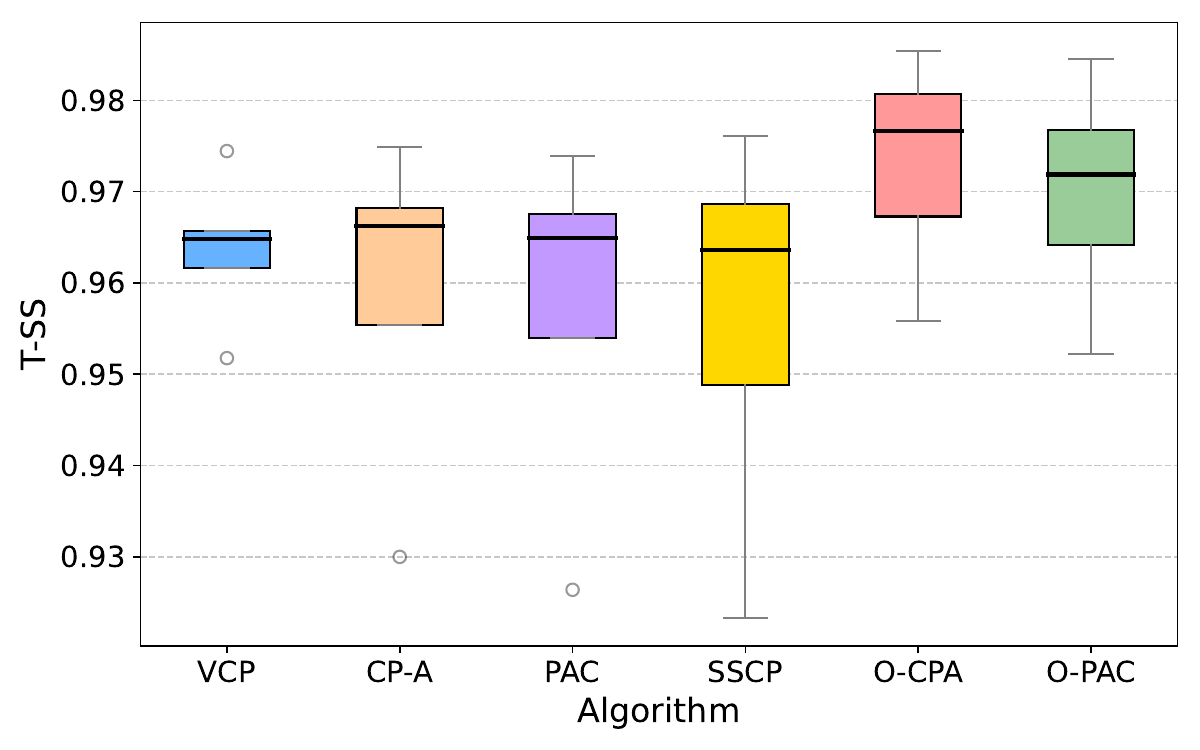}
        \caption{T-SS. ResNet18.}
    \end{subfigure}
    \begin{subfigure}{0.49\linewidth}
        \centering 
        \includegraphics[width=0.9\linewidth]{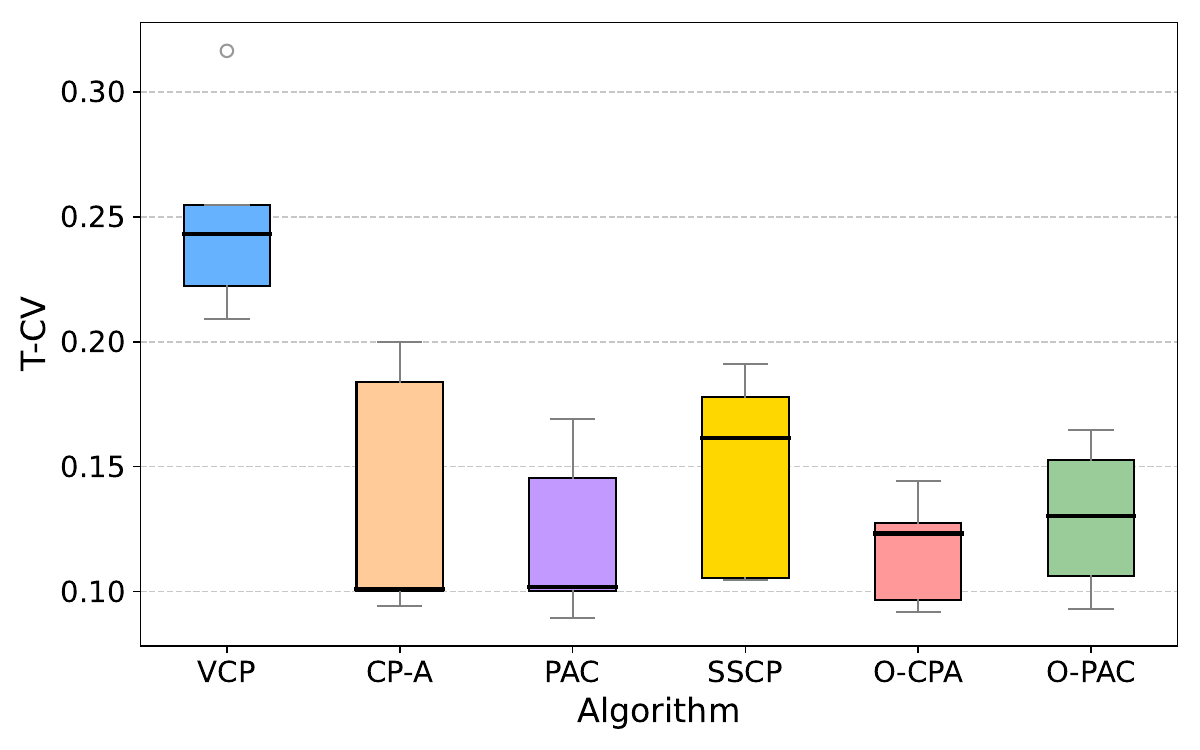}
        \caption{T-CV. ResNet50.}
    \end{subfigure}%
    \hfill
    \begin{subfigure}{0.49\linewidth}
        \centering
        \includegraphics[width=0.9\linewidth]{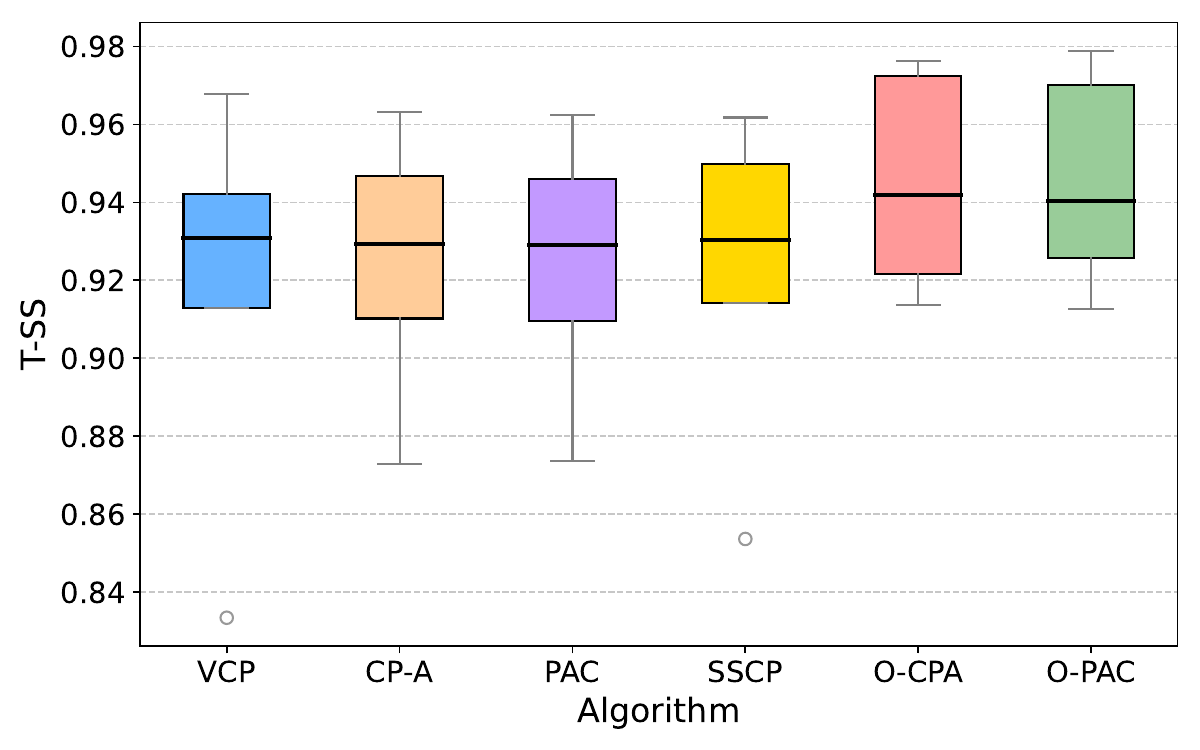}
        \caption{T-SS. ResNet50.}
    \end{subfigure}
    \begin{subfigure}{0.49\linewidth}
        \centering 
        \includegraphics[width=0.9\linewidth]{figures/va/alpha0.30/efficientnetv2s-pd_alpha0.3_tcv.pdf}
        \caption{T-CV. EfficientNet-V2-S.}
    \end{subfigure}%
    \hfill
    \begin{subfigure}{0.49\linewidth}
        \centering 
        \includegraphics[width=0.9\linewidth]{figures/va/alpha0.30/efficientnetv2s-pd_alpha0.3_tss.pdf}
        \caption{T-SS. EfficientNet-V2-S.}
    \end{subfigure}
    \caption{Visual Acuity Prediction Result. $\alpha=0.30$}
    \label{fig:va_adaptivity_alpha0.30}
\end{figure*}

\begin{figure*}
    \centering
    \begin{subfigure}{0.48\linewidth}
        \includegraphics[width=0.9\linewidth]{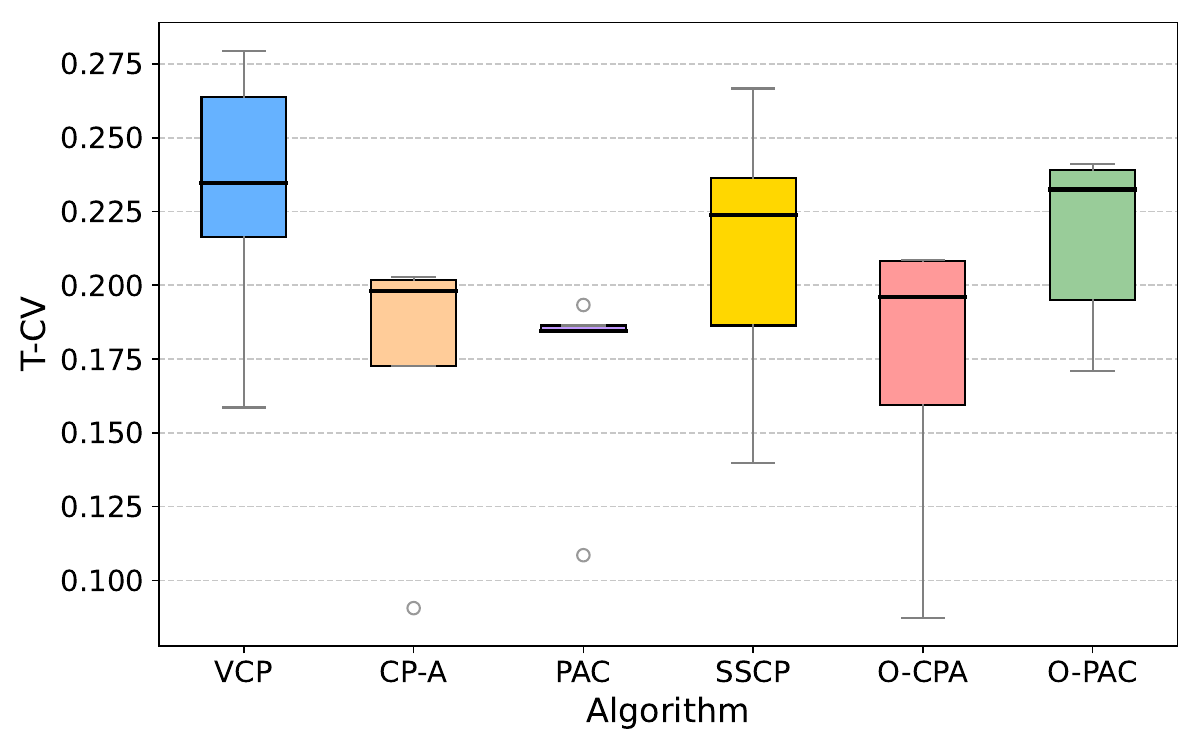}
        \caption{T-CV. Simple-CNN.}
    \end{subfigure}%
    \hfill
    \begin{subfigure}{0.48\linewidth}
        \includegraphics[width=0.9\linewidth]{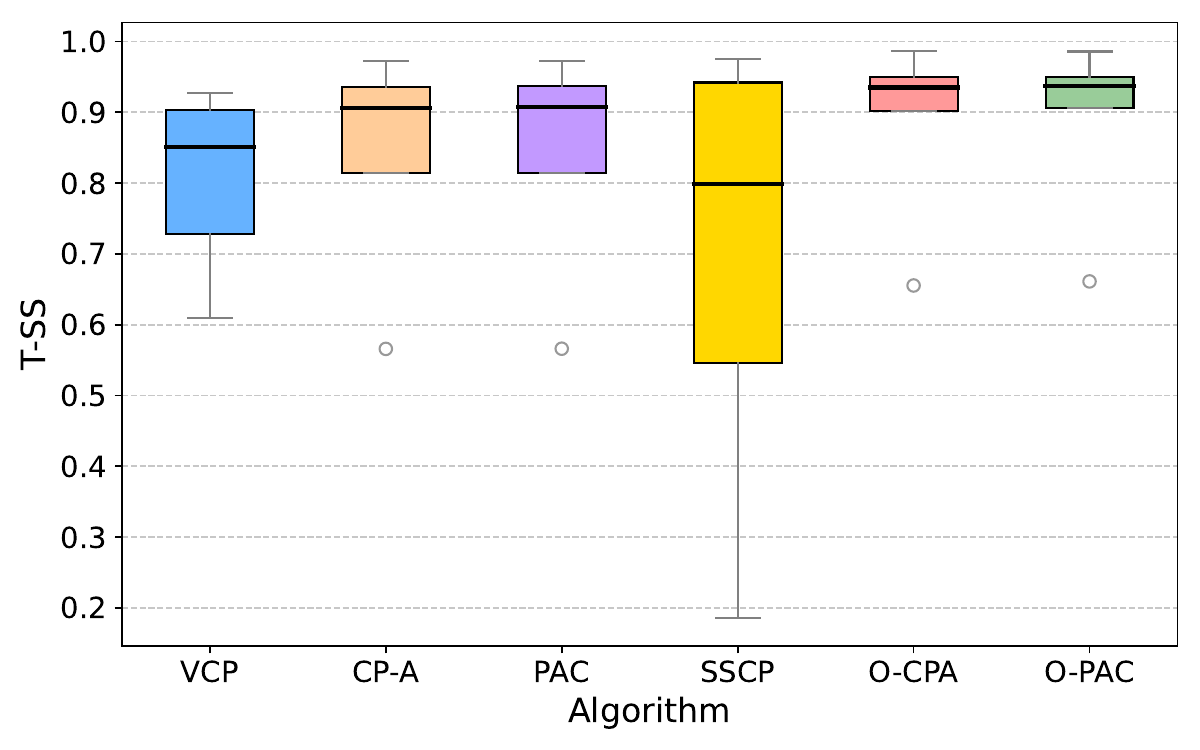}
        \caption{T-SS. Simple-CNN.}
    \end{subfigure}
    \begin{subfigure}{0.48\linewidth}
        \includegraphics[width=0.9\linewidth]{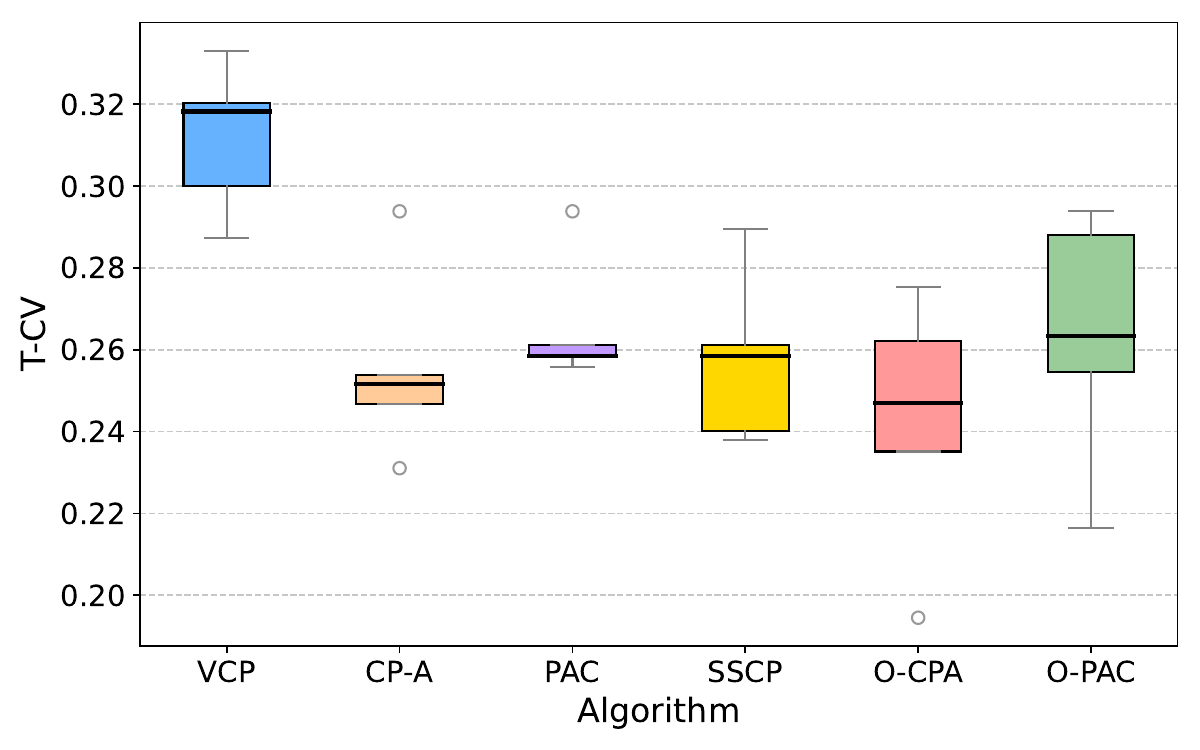}
        \caption{T-CV. ResNet18.}
    \end{subfigure}%
    \hfill
    \begin{subfigure}{0.48\linewidth}
        \includegraphics[width=0.9\linewidth]{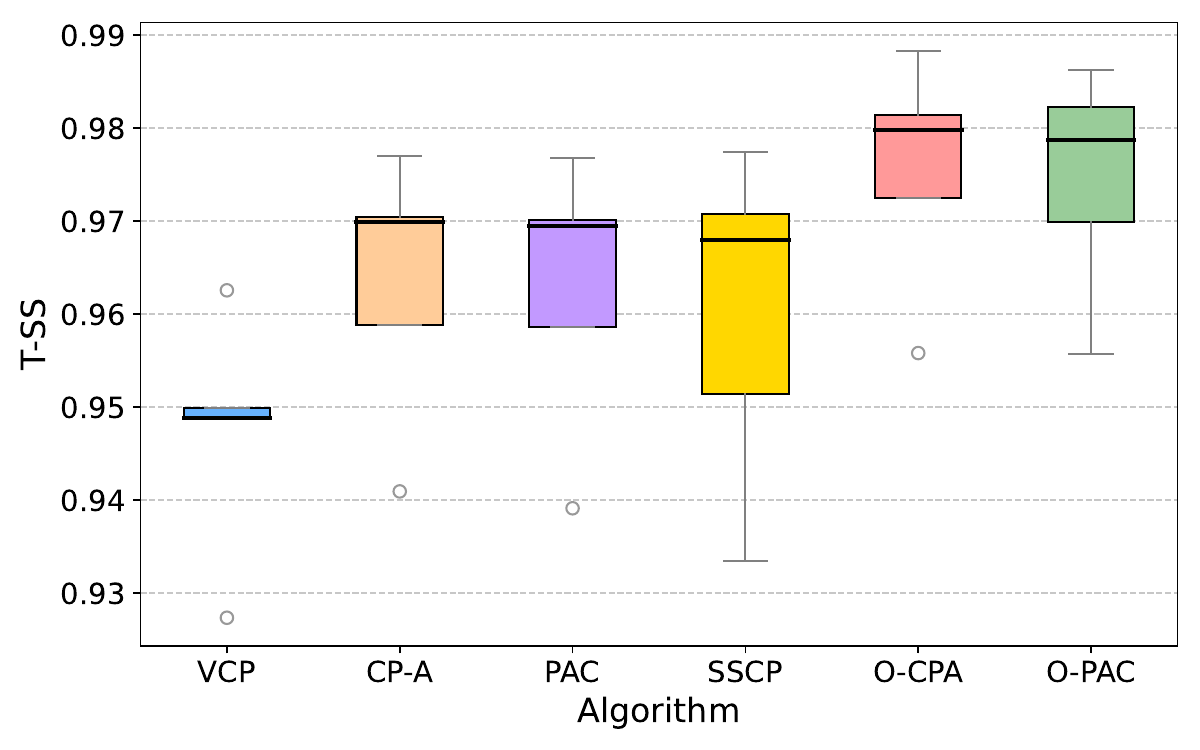}
        \caption{T-SS. ResNet18.}
    \end{subfigure}
    \begin{subfigure}{0.48\linewidth}
        \includegraphics[width=0.9\linewidth]{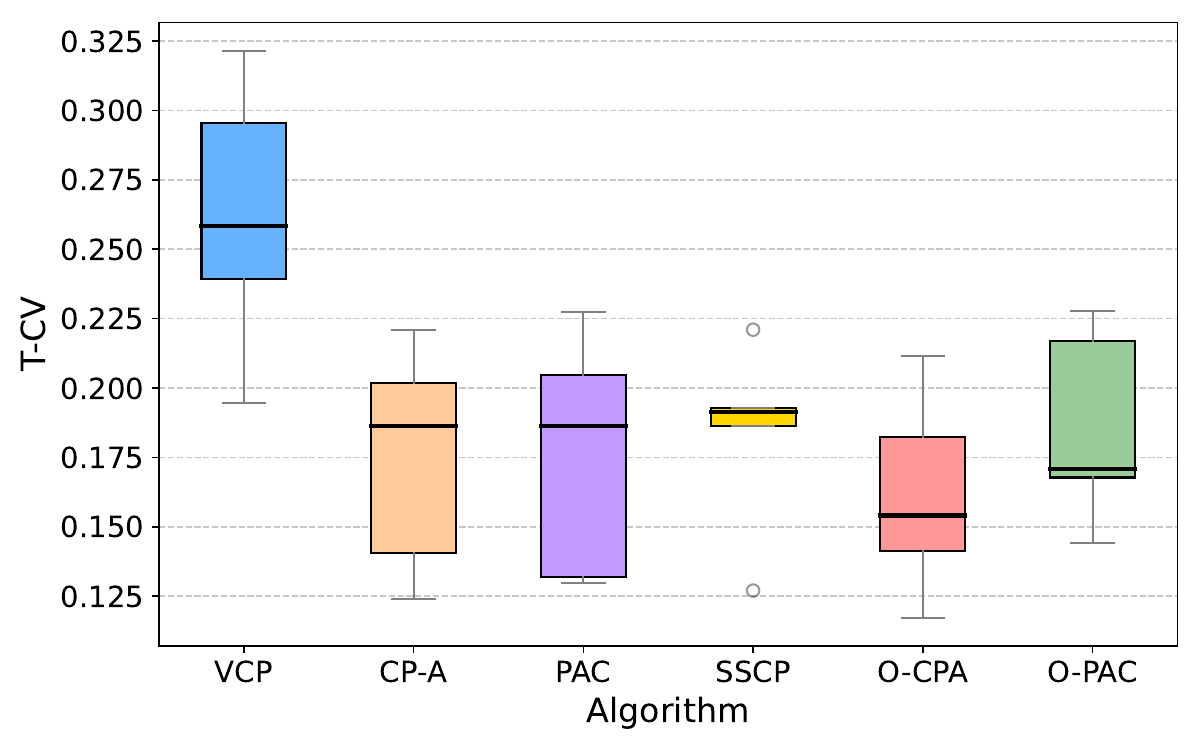}
        \caption{T-CV. ResNet50.}
    \end{subfigure}%
    \hfill
    \begin{subfigure}{0.48\linewidth}
        \includegraphics[width=0.9\linewidth]{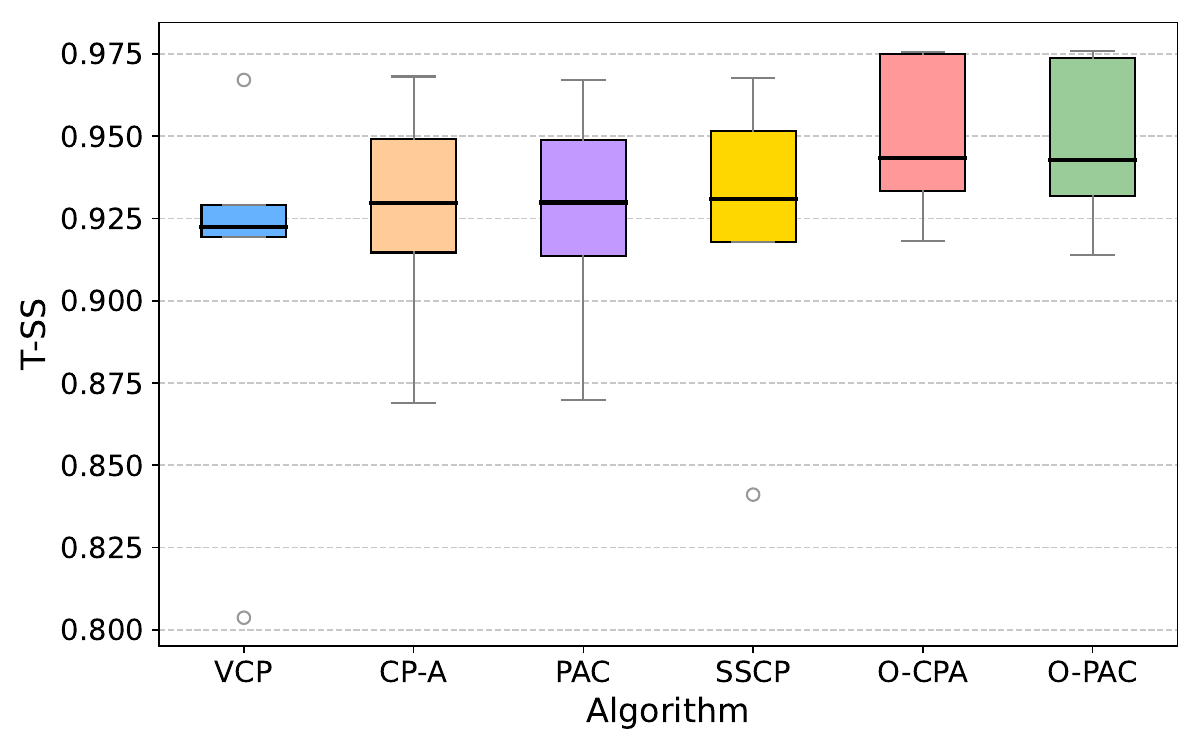}
        \caption{T-SS. ResNet50.}
    \end{subfigure}
    \begin{subfigure}{0.48\linewidth}
        \includegraphics[width=0.9\linewidth]{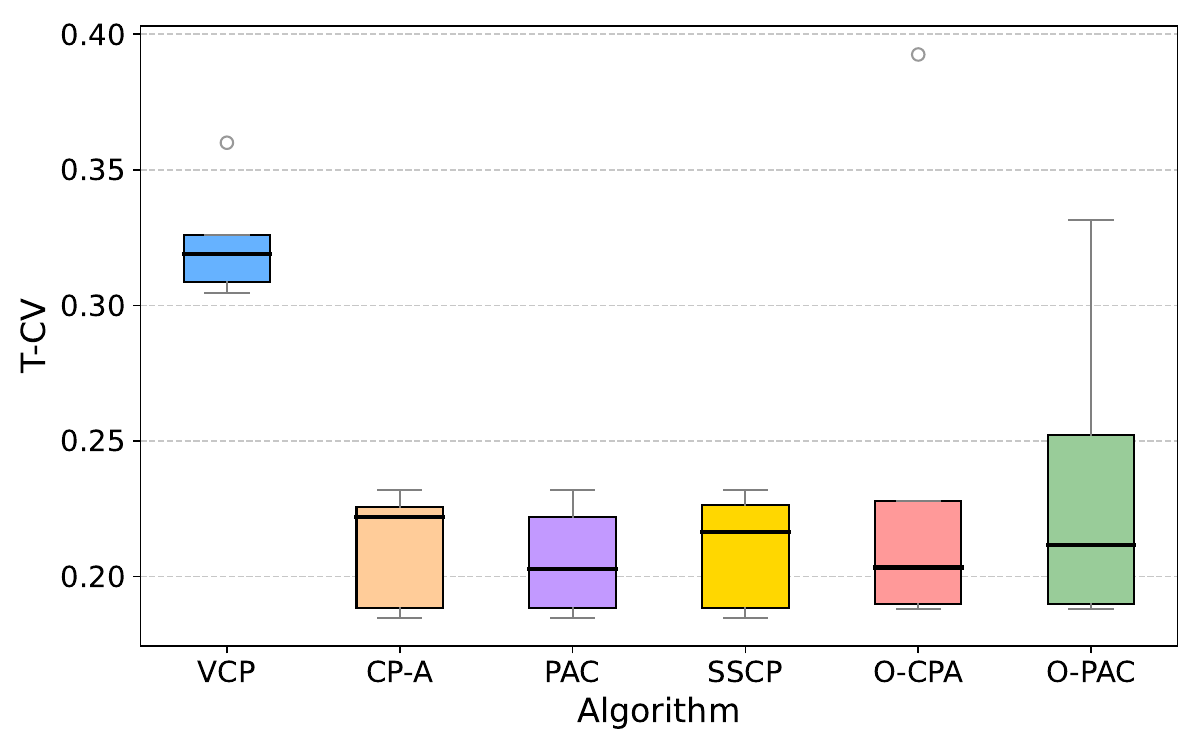}
        \caption{T-CV. EfficientNet-V2-S.}
    \end{subfigure}%
    \hfill
    \begin{subfigure}{0.48\linewidth}
        \includegraphics[width=0.9\linewidth]{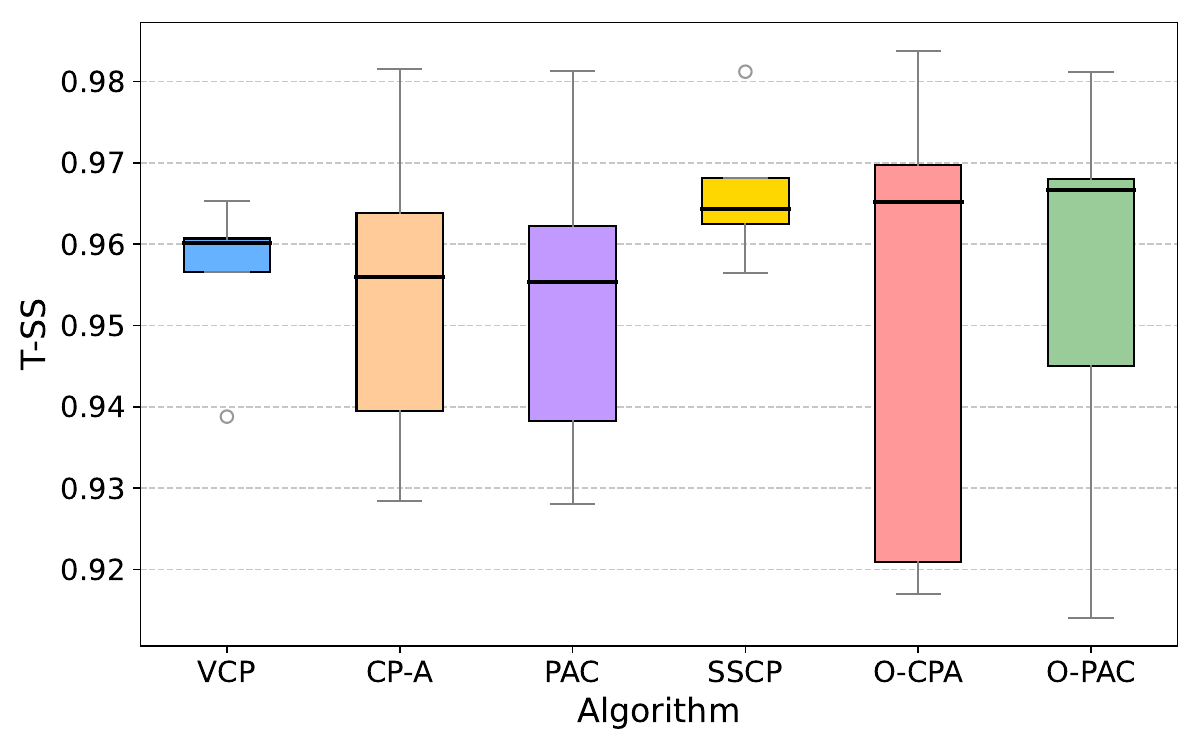}
        \caption{T-SS. EfficientNet-V2-S.}
    \end{subfigure}
    
    \caption{Visual Acuity Prediction Result. $\alpha=0.40$}
    \label{fig:va_adaptivity_alpha0.40}
\end{figure*}

Tables \ref{tab:res_va_adaptivity} and \ref{tab:res_va_general} provide the results of adaptivity, coverage rate, and average interval width analysis.
Figures \ref{fig:va_adaptivity_alpha0.20} - \ref{fig:va_adaptivity_alpha0.40} compare the adaptivity of each algorithm across different base models.

\begin{table}[!hb]
    \centering
    \scriptsize
    \setlength{\tabcolsep}{3pt}
    \begin{tabular}{c cccccc cccccc}
    \toprule
    \multirow{2}{*}{Model} & \multicolumn{6}{c}{T-CV} & \multicolumn{6}{c}{T-SS}\\
    & CP & CP-A & PAC-PI & SSCP & O-CP-A & O-PAC & CP & CP-A & PAC-PI & SSCP&  O-CP-A & O-PAC\\
    \midrule
    \multicolumn{13}{c}{$\alpha=0.40$}\\
    \midrule
    Simple-CNN & 0.235 & 0.198 & \textbf{0.184} & 0.224 & \underline{0.196} & 0.232 & 0.851 & 0.906 & 0.908 & 0.799 & \underline{0.935} & \textbf{0.937}\\
    ResNet18 & 0.318 & \underline{0.252} & 0.258 & 0.258 & \textbf{0.247} & 0.263 & 0.949 & 0.970 & 0.969 & 0.968 & \textbf{0.980} & \underline{0.979}\\
    ResNet50 & 0.258 & 0.186 & 0.186 & 0.191 & \textbf{0.154} & \underline{0.171} & 0.922 & 0.930 & 0.930 & \underline{0.931} & \textbf{0.943} & \textbf{0.943}\\
    EfficientNet-V2-S & 0.319 & 0.222 & \textbf{0.203} & 0.216 & \textbf{0.203} & \underline{0.211} & 0.960 & 0.956 & 0.955 & 0.964 & \underline{0.965} & \textbf{0.967}\\
    \midrule
    \multicolumn{13}{c}{$\alpha=0.30$}\\
    \midrule
    Simple-CNN & 0.248 & \textbf{0.151} & 0.167 & 0.178 & \underline{0.159} & 0.176 & 0.893 & 0.913 & 0.915 & 0.806 & \underline{0.937} & \textbf{0.938}\\
    ResNet18 & 0.245 & 0.188 & \underline{0.183} & 0.190 & \textbf{0.169} & 0.185 & 0.965 & 0.966 & 0.965 & 0.964 & \textbf{0.977} & \underline{0.972}\\
    ResNet50 & 0.243 & \textbf{0.101} & \underline{0.102} & 0.162 & 0.123 & 0.130 & 0.931 & 0.929 & 0.929 & 0.930 & \textbf{0.942} & \underline{0.940}\\
    EfficientNet-V2-S & 0.267 & \underline{0.136} & \textbf{0.134} & 0.141 & 0.173 & 0.159 & 0.960 & 0.955 & 0.954 & 0.961 & \textbf{0.968} & \underline{0.965}\\
    \midrule
    \multicolumn{13}{c}{$\alpha=0.20$}\\
    \midrule
    Simple-CNN & 0.211 & 0.110 & \underline{0.107} & 0.127 & \textbf{0.102} & 0.120 & 0.919 & 0.919 & 0.919 & 0.846 &  \underline{0.938} & \textbf{0.939}\\
    ResNet18 & 0.191 & 0.106 & 0.105 & \underline{0.104} & \textbf{0.102} & \textbf{0.102} & 0.958 & 0.959 & 0.957 & 0.956 & \textbf{0.970} & \underline{0.965}\\
    ResNet50 & 0.185 & \underline{0.086} & 0.097 & 0.110 & \textbf{0.080} & 0.097 & 0.928 & 0.926 & 0.926 & 0.928 & \underline{0.939} & \textbf{0.940}\\
    EfficientNet-V2-S & 0.188 & \underline{0.085} & 0.101 & \textbf{0.078}  &\underline{0.101} & 0.118 & 0.956 & 0.947 & 0.946 & 0.960 & \textbf{0.963} & \underline{0.961}\\
    \bottomrule
    \end{tabular}
    \caption{Visual Acuity Prediction. Adaptivity Evaluation. O-CPA and O-PAC generally outperform other baselines. }
    \label{tab:res_va_adaptivity}
\end{table}

\begin{table}[!bt]
    \centering
    \scriptsize
    \setlength{\tabcolsep}{3pt}
    \begin{tabular}{c cccccc cccccc}
    \toprule
    \multirow{2}{*}{Model} & \multicolumn{6}{c}{Coverage Rate} & \multicolumn{6}{c}{Average Interval Width}\\
    & CP & CP-A & PAC-PI & SSCP & O-CP-A & O-PAC & CP & CP-A & PAC-PI & SSCP & O-CP-A & O-PAC\\
    \midrule
    \multicolumn{13}{c}{$\alpha=0.40$}\\
    \midrule
    Simple-CNN & 0.594 & 0.599 & 0.624 & 0.599 & 0.663 & 0.706 & \textbf{3.075} & 3.282 & 3.441 & \underline{3.162} & 3.495 & 3.912\\
    ResNet18 & 0.599 & 0.595 & 0.620 & 0.600 & 0.682 & 0.723 & \textbf{2.499} & \underline{2.537} & 2.657 & 2.571 & 2.936 & 3.270\\
    ResNet50 & 0.589 & 0.595 & 0.618 & 0.596 & 0.645 & 0.698 & \textbf{2.312} & \underline{2.389} & 2.512 & 2.460 & 2.651 & 2.984\\
    EfficientNet-V2-S & 0.594 & 0.591 & 0.612 & 0.592 & 0.604 & 0.662 & \textbf{2.231} & 2.259 & 2.382 & \underline{2.235} & 2.313 & 2.654\\
    \midrule
    \multicolumn{13}{c}{$\alpha=0.30$}\\
    \midrule
    Simple-CNN & 0.698 & 0.700 & 0.721 & 0.702 & 0.748 & 0.783 & 3.940 & \underline{3.911} & 4.056 & \textbf{3.879} & 4.092 & 4.452\\
    ResNet18 & 0.696 & 0.695 & 0.719 & 0.698 & 0.743 & 0.791 & 3.283 & \textbf{3.131} & 3.291 & \underline{3.143} & 3.452 & 3.806\\
    ResNet50 & 0.691 & 0.701 & 0.721 & 0.698 & 0.704 & 0.753 & 3.038 & \underline{3.019} & 3.174& \textbf{3.013} & 3.047 & 3.473\\
    EfficientNet-V2-S & 0.693 & 0.694 & 0.714 & 0.692 & 0.699 & 0.777 & 2.993 & 2.908 & 3.039 & \underline{2.891} & \textbf{2.834} & 3.539\\
    \midrule
    \multicolumn{13}{c}{$\alpha=0.20$}\\
    \midrule
    Simple-CNN & 0.796 & 0.798 & 0.815 & 0.799 & 0.821 & 0.864 & 4.932 & \textbf{4.666} & 4.827 & \underline{4.756} & 4.767 & 5.188\\
    ResNet18 & 0.800 & 0.798 & 0.817 & 0.799 & 0.801 & 0.851 & 4.188 & \underline{3.937} & 4.121 & \textbf{3.927} & 4.102 & 4.594\\
    ResNet50 & 0.795 & 0.796 & 0.813 & 0.798 & 0.790 & 0.844 & 3.998 & 3.833 & 4.034 & \textbf{3.762} &\underline{3.789} & 4.409\\
    EfficientNet-V2-S & 0.794 & 0.794 & 0.812 & 0.795 & 0.812 & 0.856 & 3.927 & \underline{3.716} & 3.908 & \textbf{3.683} & \underline{3.888} & 4.427\\
    \bottomrule
    \end{tabular}
    \caption{Visual Acuity Prediction. Coverage Rate and Average Interval Width. In terms of average width, CP outperforms others. PAC-based algorithms tend to conservative (wider interval) for the training-conditional guarantee (PAC-style guarantee), and "O-[]" variatnts' group-conditional coverage guarantees.}
    \label{tab:res_va_general}
\end{table}

\begin{figure}
    \centering
    \begin{subfigure}{0.33\linewidth}
        \includegraphics[width=\linewidth]{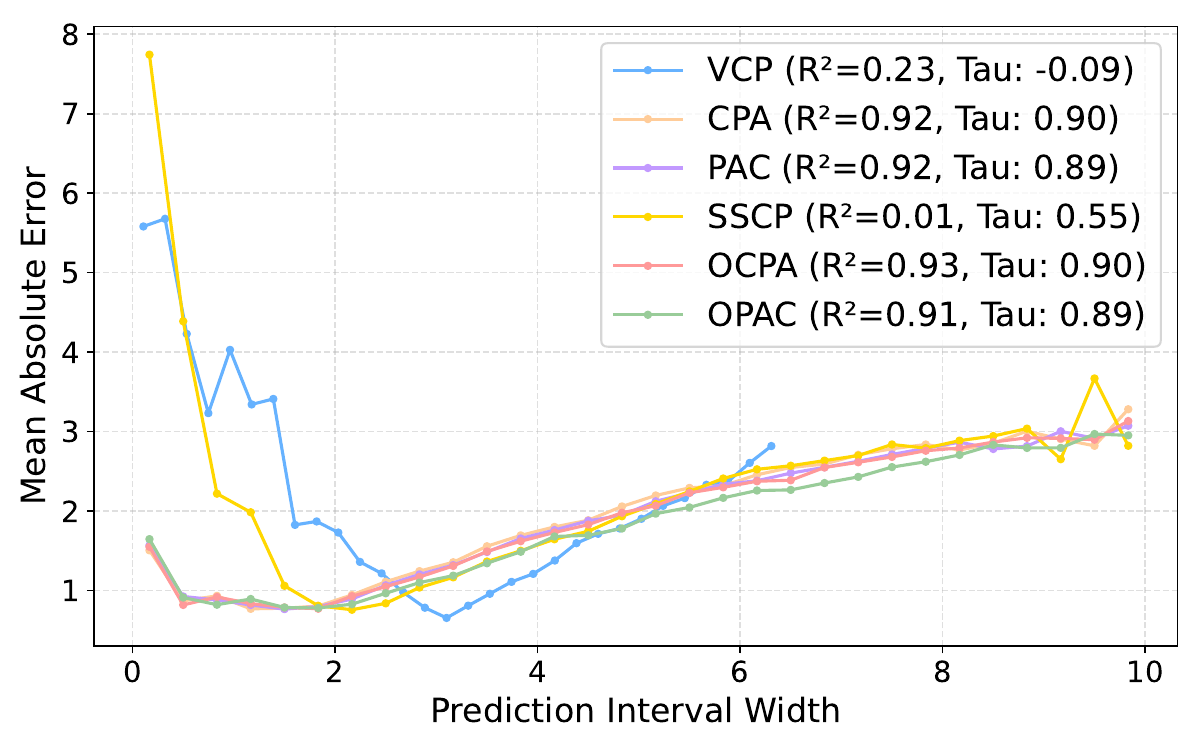}
        \caption{Simple-CNN, $\alpha=0.2$}
    \end{subfigure}
    \begin{subfigure}{0.33\linewidth}
        \includegraphics[width=\linewidth]{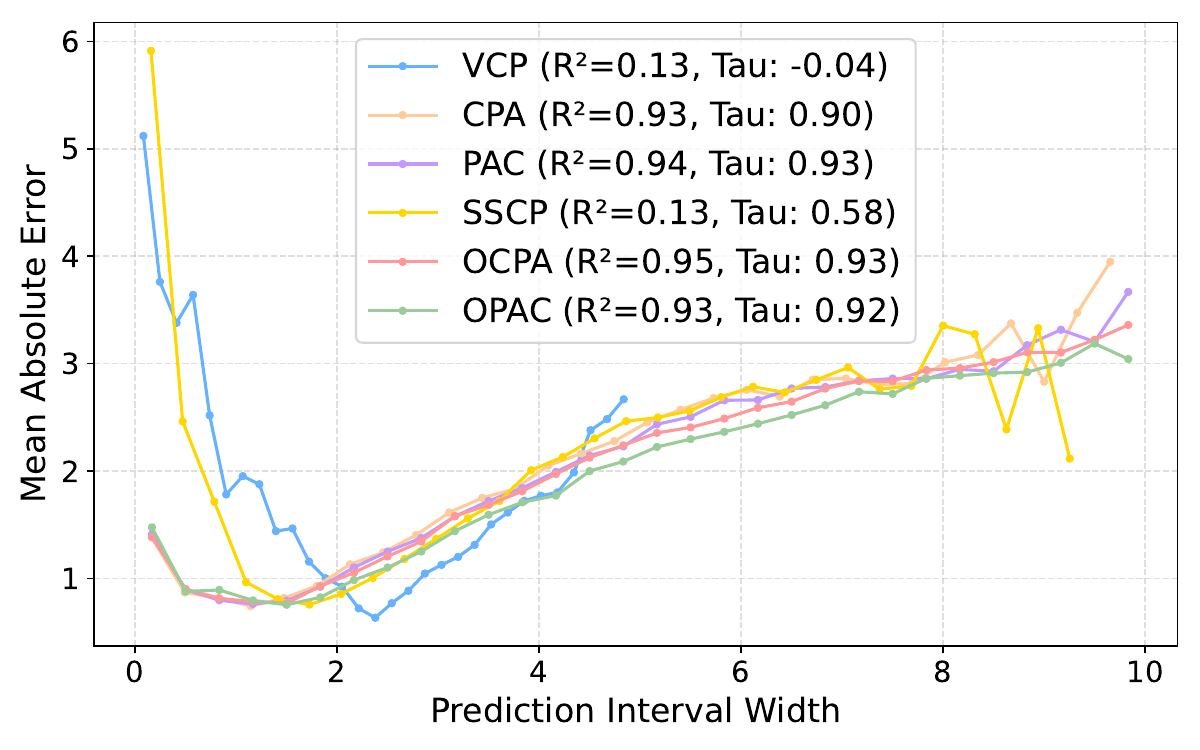}
        \caption{Simple-CNN, $\alpha=0.3$}
    \end{subfigure}
    \begin{subfigure}{0.33\linewidth}
        \includegraphics[width=\linewidth]{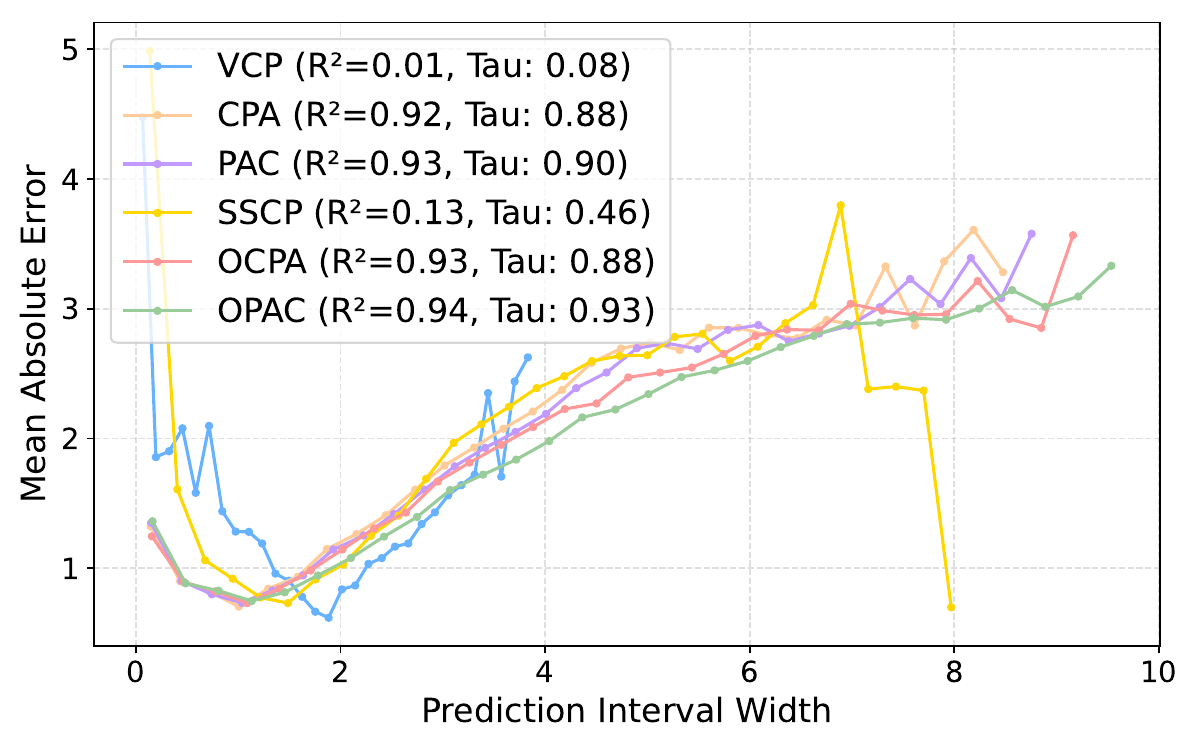}
        \caption{Simple-CNN, $\alpha=0.4$}
    \end{subfigure}
    \begin{subfigure}{0.33\linewidth}
        \includegraphics[width=\linewidth]{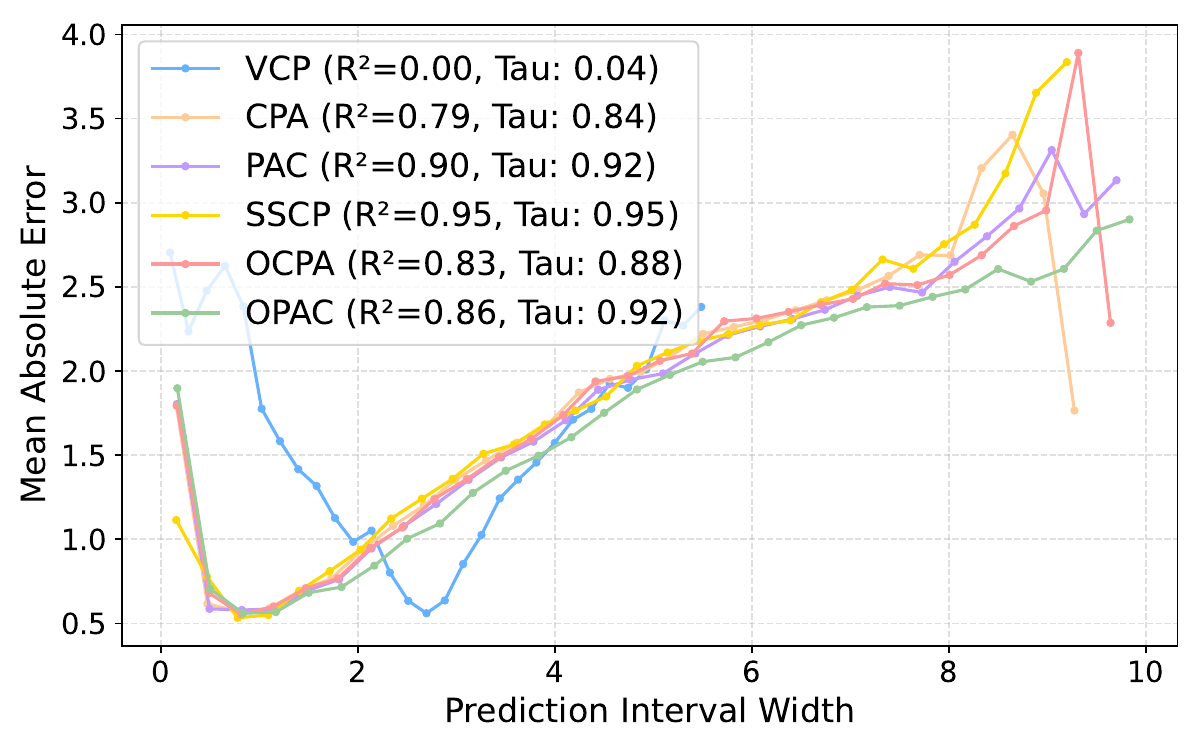}
        \caption{ResNet18, $\alpha=0.2$}
    \end{subfigure}
    \begin{subfigure}{0.33\linewidth}
        \includegraphics[width=\linewidth]{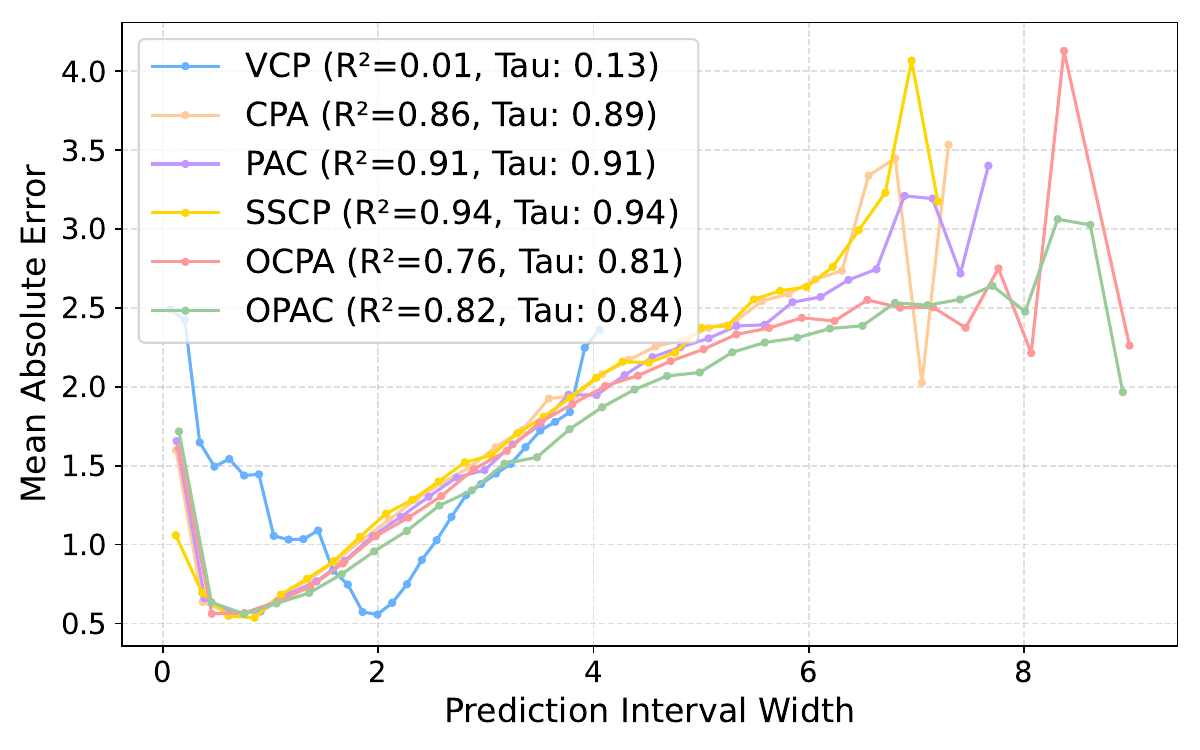}
        \caption{ResNet18, $\alpha=0.3$}
    \end{subfigure}
    \begin{subfigure}{0.33\linewidth}
        \includegraphics[width=\linewidth]{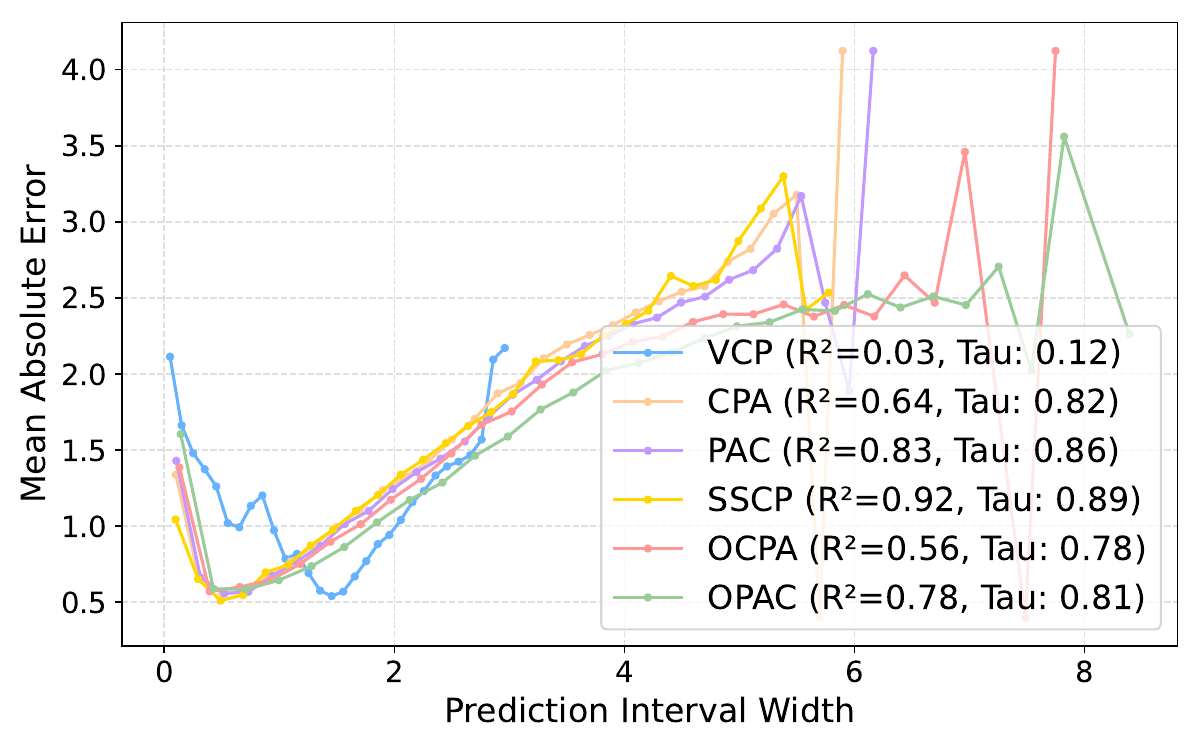}
        \caption{ResNet18, $\alpha=0.4$}
    \end{subfigure}
    \begin{subfigure}{0.33\linewidth}
        \includegraphics[width=\linewidth]{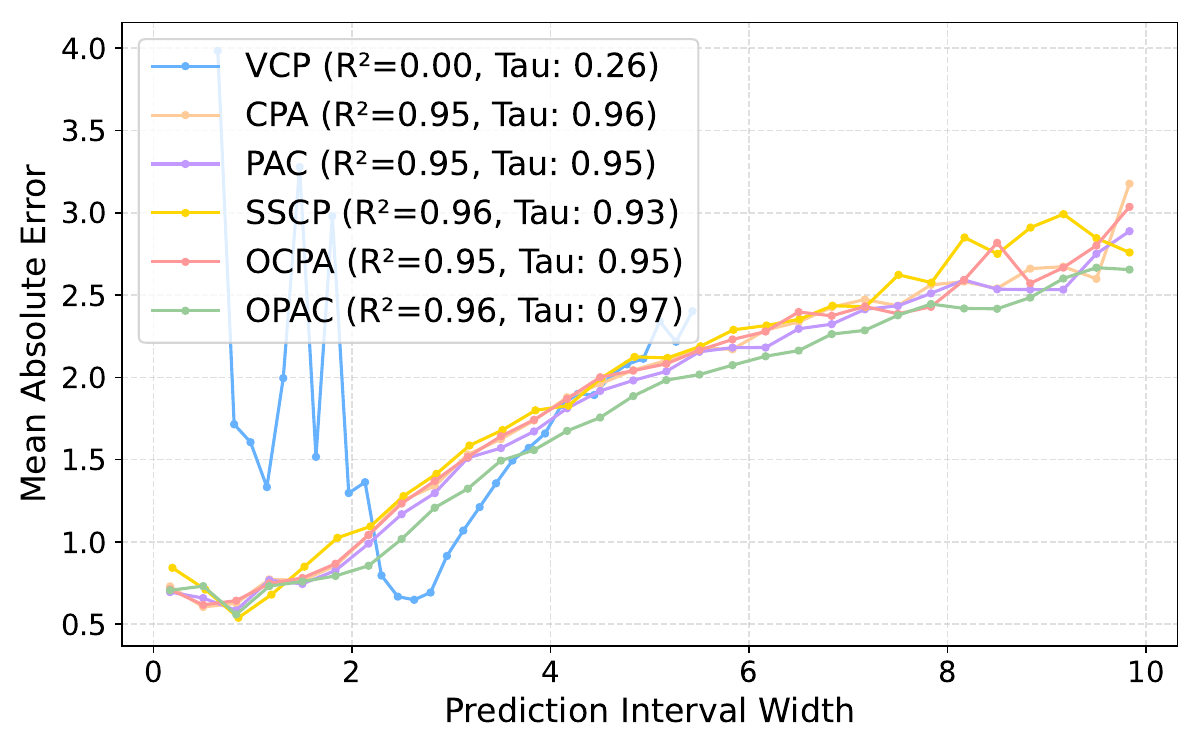}
        \caption{ResNet50, $\alpha=0.2$}
    \end{subfigure}
    \begin{subfigure}{0.33\linewidth}
        \includegraphics[width=\linewidth]{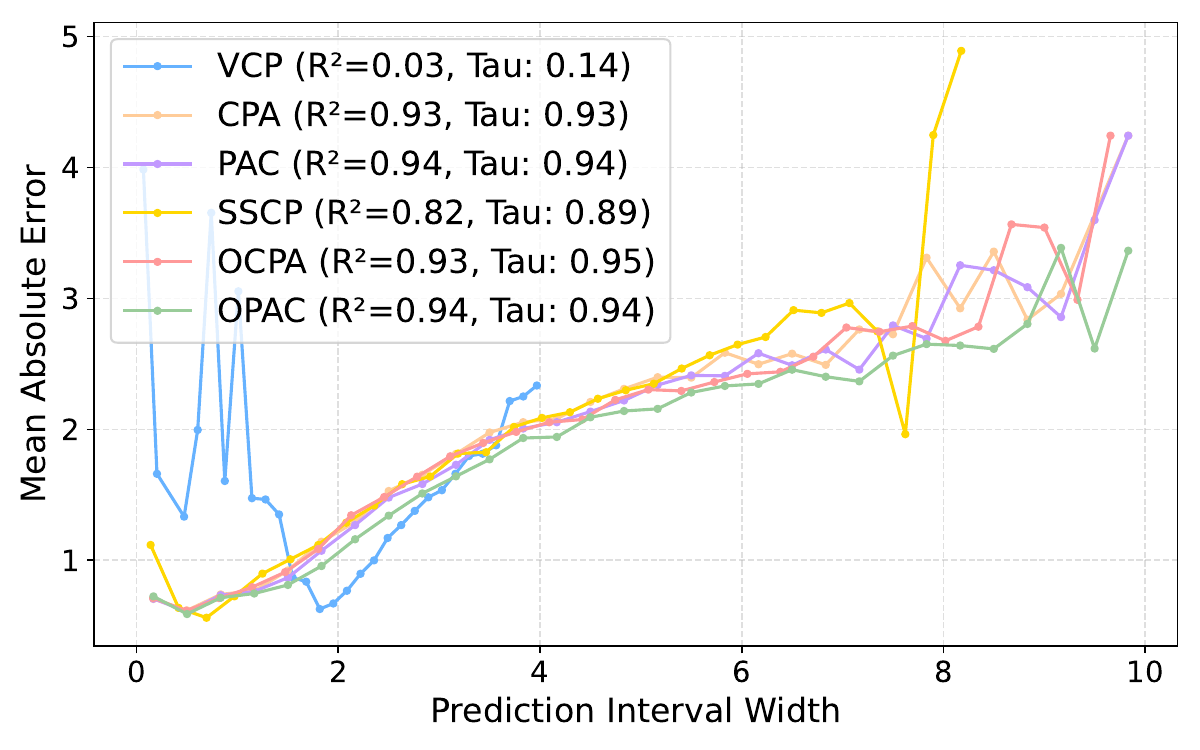}
        \caption{ResNet50, $\alpha=0.3$}
    \end{subfigure}
    \begin{subfigure}{0.33\linewidth}
        \includegraphics[width=\linewidth]{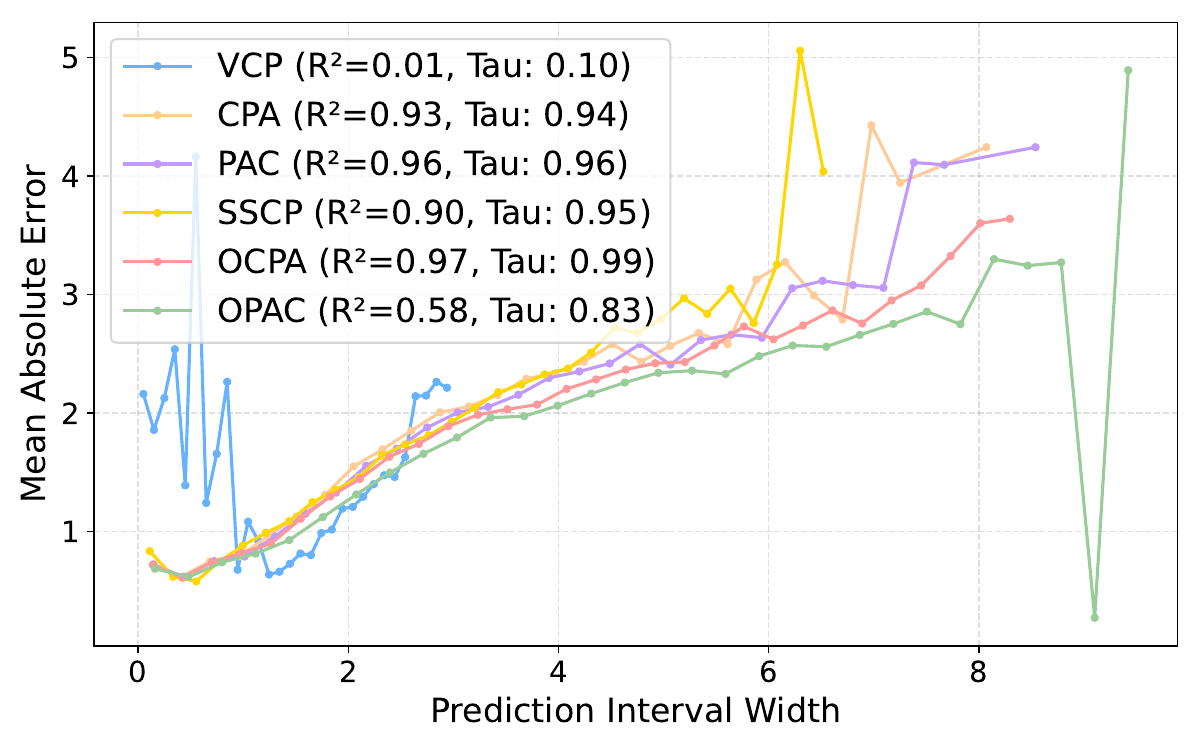}
        \caption{ResNet50, $\alpha=0.4$}
    \end{subfigure}
        \begin{subfigure}{0.33\linewidth}
        \includegraphics[width=\linewidth]{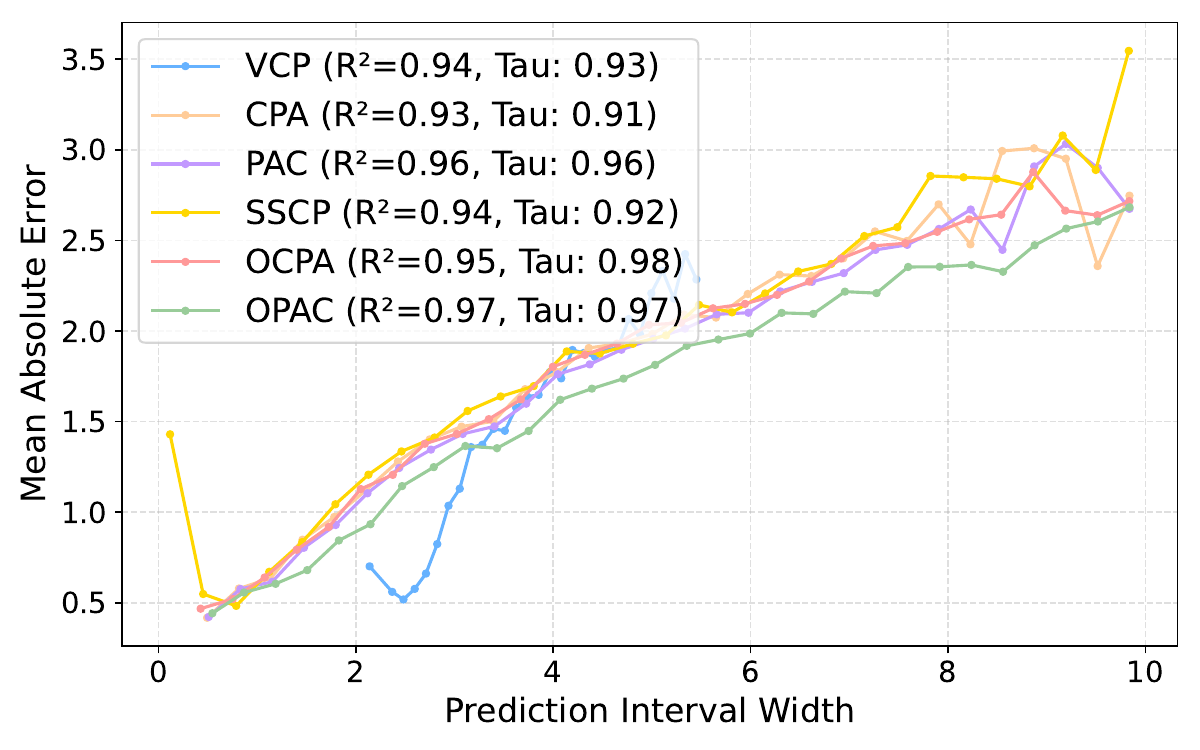}
        \caption{EfficientNet-V2-S, $\alpha=0.2$}
    \end{subfigure}
    \begin{subfigure}{0.33\linewidth}
        \includegraphics[width=\linewidth]{figures/va/rel/res_rel_efficientnetv2s-pd_b30_alpha0.3.pdf}
        \caption{EfficientNet-V2-S, $\alpha=0.3$}
    \end{subfigure}
    \begin{subfigure}{0.33\linewidth}
        \includegraphics[width=\linewidth]{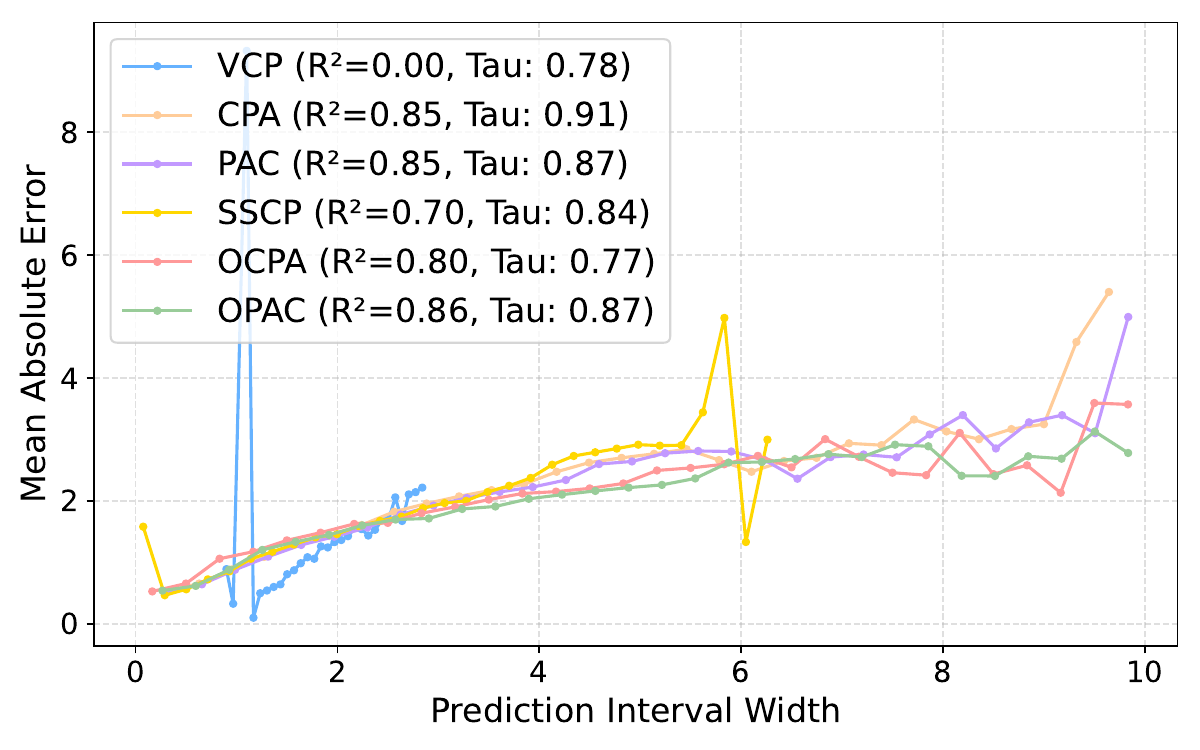}
        \caption{EfficientNet-V2-S, $\alpha=0.4$}
    \end{subfigure}
    \caption{Relation diagram between prediction interval width and prediction error for all base models.}
    \label{fig:va_rel_all_models}
\end{figure}

\FloatBarrier

\section{Computing infrastructure}
We conduct experiments using the two systems. 

\begin{table}[!hb]
    \centering
    \begin{tabular}{ccc}
    \toprule
    System & Component & Specification\\
    \midrule
    \multirow{6}{*}{System 1} & CPU & Intel(R) Xeon(R) Gold 6248 CPU @ 2.50GHz 40 cores\\
    & Memory & 754GB\\
    & GPU & NVIDIA GeForce RTX 2080 TI 11GB GDDR6\\
    & OS & Ubuntu 22.04.5 LTS\\
    & Python & 3.8\\
    & PyTorch & 1.13.1\\
    \midrule
    \multirow{6}{*}{System 2} & CPU & Intel(R) Xeon(R) Gold 6338 CPU @ 2.00GHz 64 cores\\
    & Memory & 1TB\\
    & GPU & NVIDIA A100 PCIe 80GB HBM2e\\
    & OS & Debian 12\\
    & Python & 3.8\\
    & PyTorch & 2.0.1\\
    \bottomrule
    \end{tabular}
    \caption{System Specification.}
    \label{tab:system_spec}
\end{table}

\end{document}